\documentclass{article}

\pdfoutput=1
\usepackage[preprint]{neurips_2023}

\usepackage[utf8]{inputenc} % allow utf-8 input
\usepackage[T1]{fontenc}    % use 8-bit T1 fonts
\usepackage{hyperref}       % hyperlinks
\usepackage{url}            % simple URL typesetting
\usepackage{booktabs}       % professional-quality tables
\usepackage{amsfonts}       % blackboard math symbols
\usepackage{nicefrac}       % compact symbols for 1/2, etc.
\usepackage{microtype}      % microtypography
\usepackage[dvipsnames]{xcolor}         % colors
\usepackage{multicol}
\usepackage{multirow}
\usepackage{lipsum}
\usepackage{verbatim}
\usepackage{amsmath}
\usepackage[toc,page,header]{appendix}
\usepackage{minitoc}
\usepackage{graphicx}

\usepackage{floatrow}
\usepackage{array}
\usepackage{subcaption}
\usepackage{hyperref}
\hypersetup{
    colorlinks=true,
    linkcolor=blue,
    filecolor=magenta,      
    urlcolor=blue,
    pdftitle={Overleaf Example},
    pdfpagemode=FullScreen,
    }

\setcitestyle{numbers,open={[},close={]}} 

\title{Evaluating the Zero-shot Robustness of Instruction-tuned Language Models} 

\author{%
  Jiuding Sun \\
  Khoury College of Computer Sciences\\
  Northeastern University\\
  \texttt{sun.jiu@northeastern.edu} \\
  \And 
  Chantal Shaib \\
  Khoury College of Computer Sciences\\
  Northeastern University\\
  \texttt{shaib.c@northeastern.edu}
  \And 
  Byron C. Wallace \\
  Khoury College of Computer Sciences\\
  Northeastern University\\
  \texttt{b.wallace@northeastern.edu}
}

\begin{document}

% Table float box with bottom caption, box width adjusted to content

\doparttoc % Tell to minitoc to generate a toc for the parts
\faketableofcontents % Run a fake tableofcontents command for the partocs

\floatsetup[figure]{}

\maketitle

\begin{abstract}
  \emph{Instruction fine-tuning} has recently emerged as a promising approach for improving the zero-shot capabilities of Large Language Models (LLMs) on new tasks.  
This technique has shown particular strength in improving the performance of modestly sized LLMs, sometimes inducing performance competitive with much larger model variants.
In this paper we ask two questions: (1) How sensitive are instruction-tuned models to the particular phrasings of instructions, and, (2) How can we make them more robust to
 such natural language variation? 
To answer the former, we collect a set of 319 instructions manually written by NLP practitioners for over 80 unique tasks included in widely used benchmarks, and we evaluate the variance and average performance of these instructions as compared to instruction phrasings observed during instruction fine-tuning. 
We find that using novel (unobserved) but appropriate instruction phrasings consistently degrades model performance, sometimes substantially so. Further, such natural instructions yield a wide variance in downstream performance, despite their semantic equivalence. 
Put another way, instruction-tuned models are not especially robust to instruction re-phrasings. 
We propose a simple method to mitigate this issue by introducing ``soft prompt'' embedding parameters and optimizing these to maximize the similarity between representations of semantically equivalent instructions. We show that this method consistently improves the robustness of instruction-tuned models. \footnote{The code and instructions are publicly available at: \url{https://github.com/jiudingsun01/InstructionEval}}

\end{abstract}

\section{Introduction}

\begin{figure}[h]
    \centering
    \includegraphics[width=140mm]{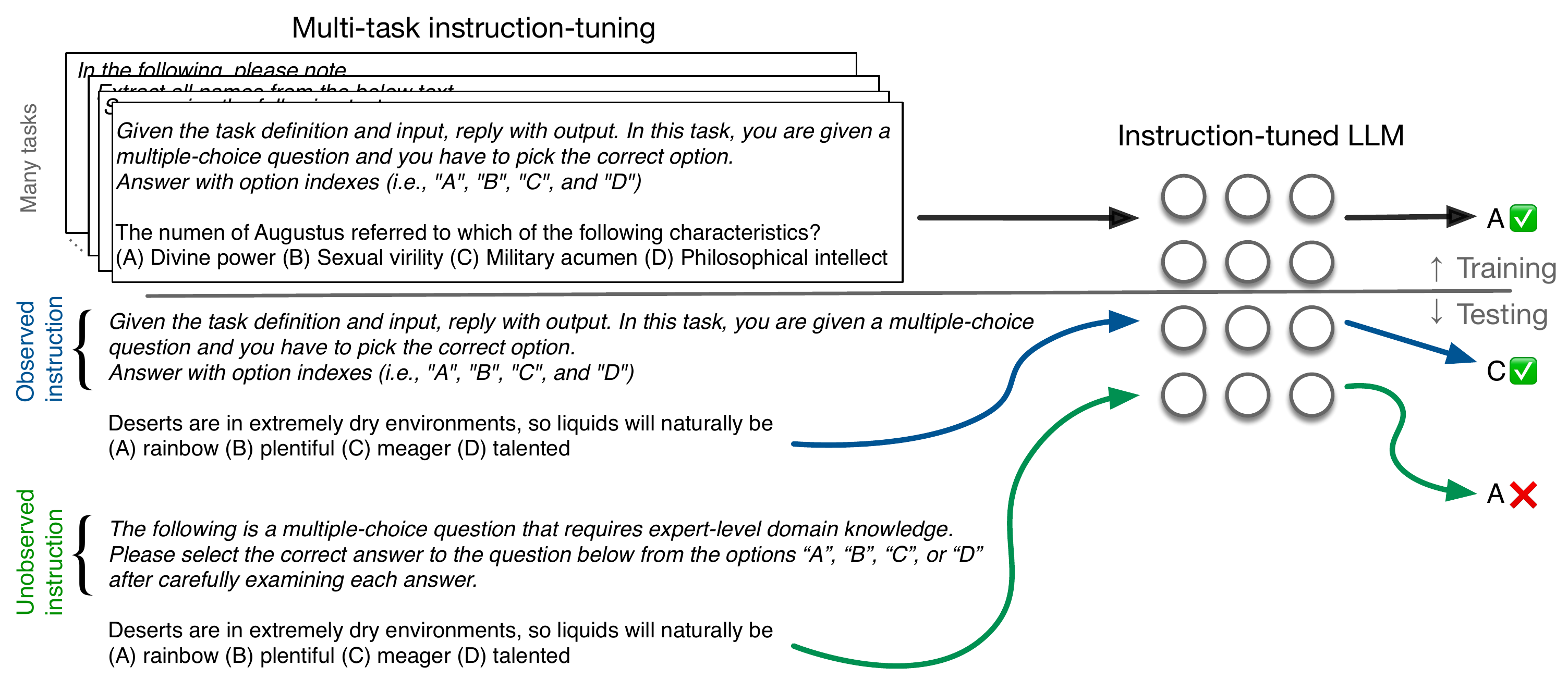}
    \caption{How well do models trained on instruction-tuning datasets generalize to novel instructions (unobserved in training)? Our analysis suggests that they do not do so very well. Above we show a case where pairing an example with an observed instruction yields the correct output, while providing a distinct but semantically equivalent instruction produces an incorrect response.
     We propose and evaluate a simple method that improves this.}
    %Flan \cite{chung2022scaling} is a meta-dataset comprising a total of 1.8k tasks. We ask: How well do models trained on this collection of instructions generalize to novel instructions and tasks (unobserved in training), and how can we improve their ability to do so?} %We study how far can the models tuned with such instruction collection be generalized into other unobserved instructions and tasks.}
    \label{fig:main_fig}
\end{figure}

Large Language Models (LLMs) have come to dominate NLP, in part because they enable zero- and few-shot adaptation to new tasks via \emph{prompting} \cite{brown2020language, chowdhery2022palm, hoffmann2022training, zeng2022glm}.
Recent work has demonstrated the promise of fine-tuning such models with natural language instructions. 
Such \emph{instruction-tuning} improves LLM performance in zero- and few-shot settings, sometimes dramatically, especially for ``mid-sized'' models \cite{chung2022scaling, ouyang2022training}. 
For example, on some benchmarks the instruction-tuned Flan-T5-XL (3B parameters) \cite{chung2022scaling} outperforms GPT-3 (175B), despite being dramatically smaller. 
Furthermore, LLaMa-7B \cite{touvron2023llama}---after being fine-tuned on large-scale corpora on the Alpaca \cite{alpaca} instruction set---outperforms GPT-3 across a range of NLP benchmarks.
%How does instruction-tuning work? Several efforts have investigated

These empirical successes have motivated efforts to curate instruction-augmented task collections for meta-learning \cite{wang2022benchmarking, wei2021finetuned, wei2021finetuned}, and research into improving instruction-tuning \cite{longpre2023flan, xu2022multiinstruct, sanh2021multitask}. %Nonetheless, it is not clear how far the ability gained from instruction tuning 
In this work we investigate how robust instruction-tuned models are.  
%can be generalized to domains, tasks, and instructions that are \textbf{\textit{unobserved}} during instruction fine-tuning. In this work, we investigate the \emph{robustness} of instructions. 
More specifically, we ask: How sensitive are instruction-tuned LMs to shifts in instruction phrasings at test time? 
This is particularly important given that the primary motivation of instruction tuning is to facilitate zero-shot adaptation via natural language instruction: If models are overly sensitive to the particular phrasing of a task instruction it may greatly limit their utility in practice. 

%the shifts of instruction and task distribution from the training stage?

Prior work---reviewed at length in Section \ref{section:related-work}---has established that LLMs do not seem to intuitively ``understand'' prompts \cite{webson2021prompt,jang2023can, zhang2023aligning}, but these efforts did not consider instruction-tuned models specifically.
Recent, contemporaneous work to ours \cite{gu2023robustness} investigated the robustness of instruction-tuned models, and found that instruction-tuned T5 \cite{raffel2020exploring} is robust to instruction perturbations in few-shot settings, but less so in zero-shot application. 
We contribute a more in-depth analysis of this phenomena across a much wider set of instruction-tuned models and benchmarks.
We also introduce and evaluate a method for improving the robustness of such models, with promising results.  
% by imposing an objective encouraging LLMs to induce similar representations for semantically equivalent instructions.

%To address this research question, 
More specifically, we collect a relatively large set of task instructions manually composed by NLP researchers; these are valid instructions but distinct from those found in the Flan collection.
%\textbf{\textit{unobserved}} during instruction fine-tuning. 
We then assess the performance of LLMs fine-tuned on the Flan collection instruction set when given these novel instructions on two benchmarks: \textsc{MMLU} \cite{hendrycks2020measuring} and \textsc{BBL} \cite{srivastava2022beyond}.
%We perform inferences on MMLU \cite{hendrycks2020measuring} and BBL \cite{srivastava2022beyond} with LMs that are instruction-tuned with the Flan collection. 
%It is observed that there is an 
We find that using novel instructions in zero-shot application degrades accuracy considerably (Figure \ref{fig:main_fig} illustrates this).
For example, comparing the performance of Flan-T5 XXL when using (a) instructions that were seen in training to (b) semantically equivalent but unobserved in training, we observe a 6.9 point drop in absolute performance on average across large benchmarks.  

Our {\bf main contributions} are summarized as follows. (1) We perform a comprehensive and in-depth analysis of the robustness of instruction-tuned LLMs across three ``families'' of such models (Flan-T5 \cite{wei2021finetuned}, Alpaca \cite{alpaca}, and T0 \cite{sanh2021multitask}) using large benchmarks \cite{hendrycks2020measuring,srivastava2022beyond}. 
For this we collect a large set of new task instructions manually composed by researchers in NLP; we will release this dataset to facilitate additional work on instruction robustness. We observe substantial performance degradation when using ``novel'' (unseen in training) instructions. 
(2) We propose a simple method to improve robustness by imposing an objective encouraging LLMs to induce similar representations for semantically equivalent instructions. 
We find that this consistently improves the performance realized when using novel but appropriate task instructions.

\section{Related Work}
\label{section:related-work}

\paragraph{Multitask learning and instruction-tuning}

%Before the instruction era, previous works of multi-task fine-tuning focused on bettering the model's NLU ability by unifying several downstream tasks into one and fine-tuning over the unified corpus 
%Prior to the introduction of \emph{instruction-tuning}, 
Training a single text-to-text model capable of providing responses to arbitrary queries has been an aspiration in NLP for at least half a decade. 
%pre-dates the current wave of LLMs and modern prompting and instructing strategies. 
%Prior to modern prompting and instructing strategies, 
For example, prior to modern prompting and instructing strategies, there were efforts to unify disparate tasks by reframing them as instances of general \emph{question answering} \cite{mccann2018natural, khashabi2020unifiedqa, keskar2019unifying}. 
%After that, many contemporary works carry out meta-dataset resources for multi-task tuning with instructions to unlock the hidden knowledge learned through large-scale unsupervised learning, the representative works are: Flan 2021 collection \cite{wei2021finetuned}, Natural Instructions \cite{mishra2021cross} and T0 \cite{sanh2021multitask}. 
More recent efforts have focussed on compiling and fine-tuning LLMs on corpora comprising diverse tasks with associated natural language instructions \cite{wei2021finetuned,mishra2021cross,sanh2021multitask}; we refer to this strategy as instruction-tuning. 
One example of this is {\tt Super-NaturalInstructions} \cite{wang2022benchmarking}, which compiles over 1600 tasks and enriches these with both instructions and negative examples. %to enrich the instruction collection. On top of that, t
Similarly, the recently released OPT-IML Bench \cite{iyer2022opt} comprises 2000 NLP tasks. %covering a range of categories.
The Flan 2022 task collection \cite{longpre2023flan} additionally features \emph{Chain-of-Thought} (CoT) style ``reasoning'' chains in instruction templates; the authors show that including these (as well as zero-shot examples and ``input inversions'') during instruction fine-tuning yields improvements on held-out tasks. 
%he new Flan collection further incorporates Chain-of-Thought (CoT) reasoning into the instruction templates and shows that it benefits instruction fine-tuning. 

These meta-resources---collections of instructions, tasks, and samples---have facilitated the training of instruction-tuned model families such as Flan-T5, Flan-PaLM \cite{chung2022scaling}, and OPT-IML \cite{iyer2022opt}.\footnote{Somewhat confusingly, in the case of FLAN and OPT, the corpora (i.e., benchmarks comprising tasks and instructions) and LLMs fine-tuned using them are both referred to with the associated acronym as prefix: For instance, Flan-T5 denotes a T5 \cite{raffel2020exploring} variant fine-tuned with the Flan collection.}
Results have been encouraging; fine-tuning LLMs to follow instructions provides clear and consistent gains across models, and, perhaps most exciting, enables relatively ``small'' ($\sim$10B) LLMs to achieve near SOTA performance comparable to massive ($\sim$175B) models \cite{alpaca}. 
This has motivated interest in characterizing how instructions help models, and developing techniques to further improve instruction-tuning; we review recent efforts related to these two research threads below. 

%Subsequently, with all these resources available many instruction-tuned models like Flan-T5, Flan-PaLM \cite{chung2022scaling}, and OPT-IML \cite{iyer2022opt} are trained and achieved state-of-the-art performance on many downstream tasks.
\paragraph{Evaluating prompting and instruction capabilities}
%Inherited from the prompt learning paradigm, many works aim to evaluate the hidden mechanism of instruction tuning and its potential flaws. 
Instructions may be seen as a special sort of model prompting, which a few recent efforts have critically evaluated. 
For example, Webson and Pavlick ask whether models meaningfully ``understand'' prompts \cite{webson2021prompt}, finding that they largely do not: %looks at the choices of prompt for inference over instruction fine-tuned model and found that the 
Performance is often unaffected when irrelevant and misleading prompts are provided. 
In follow up work, Jang \emph{et al.}
\cite{jang2023can} evaluates performance on negated prompts, observing an ``inverse-scaling'' phenomenon in which larger models perform worse in this case.
%studies the negated prompt and found that it inversely scaled with the size of the model. 

Other work has attempted to characterize how and when \emph{in-context learning} (ICL)---i.e., including a few examples in prompts---works \cite{min2022rethinking,wang2023large,dai2022can,akyurek2022learning,yu2022alert}. 
ICL is a form of prompting orthogonal to the present effort, as we are primarily interested in the zero-shot adaptability of instruction-tuned LLMs.

%With respect to instruction tuned models specifically, Yu \emph{et al.} \cite{yu2022alert} show that instruction tuning enhances zero-shot performance of the model at the cost of harming its ability to follow various instructions. 
%\cite{min2022rethinking} examine the performance of instruction-tuned models on classification tasks and discover that altering the label space in the instruction has a marginal impact on the performance.

In work contemporaneous to ours, Gu \emph{et al.} \cite{gu2023robustness} investigated how robust instruction-tuned models are to instruction perturbations (e.g., dropping words) and paraphrasings. 
They found that models are relatively robust when given examples (i.e., in few-shot settings), but quite sensitive when used zero-shot; this is qualitatively in line with our findings.
Our work differs in important way from this coincident research: (1) We provide a much more comprehensive analysis of robustness; Gu \emph{et al.} considered \emph{only} T5 instruction-tuned on a single instruction dataset, whereas we evaluate three LLMs (and different sizes of each) using five instruction tuning datasets, and we evaluate using over 80 test tasks in all (Gu \emph{et al.} considered only 12). (2) We propose and evaluate a new approach to \emph{improving} the robustness of instruction-tuned models; Gu \emph{et al.} offered no mechanism to improve robustness.

\paragraph{Improving instruction-tuning}
%Most of the works aim to improve the performance of instruction-tuned models are focusing on two threads. 
Past work has also sought to improve instruction-tuning in various ways.
One means to do so is to instruction tune based on human feedback \cite{ouyang2022training, glaese2022improving, bai2022training, nakano2021webgpt, zhang2023wisdom}. 
This tends to improve open-ended model responses but degrade performance on downstream tasks. %improves the model's open-ended task performance at the cost of NLP tasks performance degradation. 
Another strategy is to leverage existing resources to automatically generate instruction-tuning datasets at scale. 
For example, Wang \emph{et al.} \cite{wang2022self} use LLMs to generate instructions, inputs, and outputs and use these to improve their own instruction-following capabilities. 
%improves the instruction-following capabilities of LLMs by bootstrapping their own generations to train the model.
In a similarly meta vein, Zhou and colleagues \cite{zhou2022large}  propose using LLMs to engineer prompts. 
%regards instructions as program to perform text-to-structure generation with LLM. 
Finally, Ye \emph{et al.} \cite{ye2022guess} propose ``flipping'' the standard task by tasking LLMs with generating \emph{instructions}, given an input and label. 
%trains the LM to produce instructions given the input and labels. 
%\input{sections/03_instruction_robustness.tex}
\section{Instruction Datasets}

\subsection{Evaluation Benchmarks}

%We perform the evaluation of instruction-tuned models 
We evaluate a set of instruction-tuned models on two large benchmarks: \textsc{MMLU} \cite{hendrycks2020measuring} and \textsc{Big-Bench} \cite{srivastava2022beyond}. \textsc{MMLU} is a multiple-choice question-answering benchmark comprising 57 tasks that require expert knowledge.
\textsc{Big-Bench} is a collaboratively built benchmark containing 204 diverse tasks from various domains; here %we use the 18 task 
consider the \textsc{Big-Bench Lite} subset, and we include only QA, multi-class, and binary classification tasks, yielding 18 tasks from in all. %in \textsc{Big-Bench}.

%More specifically, We conduct our experiment over all 57 tasks on MMLU. An 18 tasks subset of \textsc{Big-Bench Lite} consists of all QA, multi-class, and binary classification tasks.

\subsection{Collecting New Instructions from NLP Researchers}
\label{section:new-instructions}

We aim to evaluate instruction-tuned models when they are provided instructions which are semantically equivalent to, but superficially different from, those with which they were trained.
To this end, we enlist NLP researchers (graduate students) to compose novel instructions for the tasks considered; these particular instruction phrasings were therefore \emph{unobserved} during instruction fine-tuning. 

%For each instruction-tuned language model we  we collect a set of instructions unobserved to the model by expert annotation, and we collect a set of instructions observed during training from the original instruction-tuning collection.

% TODO: to have some small tables here to show the stats

%\subsubsection{Unobserved Instruction}

%We perform large-scale crowd-sourcing from 
More specifically, we recruited 36 NLP graduate students working in NLP.
All had at least some experience with instruction-tuned models and the downstream tasks included in the evaluation benchmarks. 
For each of the 18 tasks in \textsc{BBL} and all tasks in \textsc{MMLU}, we asked 12 graduate students to write one (distinct) instruction they would use for zero-shot inference with an instruction-tuned model. 
%To ensure fairness, the information on models is omitted to avoid having priors to fit the pattern of the specific model. 
%The detailed instruction collection process can be seen in Appendix A.
We provide details on this instruction collection process in Appendix A. 
We will release all 319 instructions acquired for this work to ensure the reproducibility of this work and to facilitate further research on instruction-tuned model robustness. 
% We treat 57 tasks of MMLU as a whole (general QA template). Don't know how to make the word clearer

%\subsubsection{Observed Instruction}

% could be generally applied to one of ``multiple-choice QA'', ``binary label classification'', and ``m
%``multi-label classification" tasks. 

% TODO: to have some small tables here to show the stats

\section{Evaluating the Robustness of Instruction-tuned LLMs}

\subsection{Models and Data}

We conduct experiments with model variants trained over three instruction collections (these provide \emph{observed} task instructions): P3 \cite{sanh2021multitask}, Flan-2022 \cite{chung2022scaling}, and Alpaca \cite{alpaca}. 
%For each model and associated published instruction collection, we manually perused entire collection and pick the 
To facilitate our analyses, we manually identified all instructions that correspond to (a) multiple-choice question answering (QA), (b) binary classification (BC), or tasks that demand ``yes'' or ``no'' responses, and (c) multi-class classification (MC), which requires classifying inputs into a finite set of categories. 

To evaluate model robustness with respect to instruction phrasings we use two benchmarks: \textsc{MMLU} \cite{hendrycks2020measuring} and \textsc{Big-Bench Lite} (\textsc{BBL}) \cite{srivastava2022beyond} along with the acquired set of novel instructions described in Section \ref{section:new-instructions}.
%We evaluate all tasks with multi-choice grading with logit scores to keep a fair comparison. 
We include all 57 tasks from \textsc{MMLU}, and 14 of 24 tasks from \textsc{BBL}. 
From the latter we exclude two tasks that rely on generation metrics, four that use exact-match, and four that contain tokens unrecognized by the T5 and/or LLaMa tokenizer (e.g., inputs are emojis in one task).

%2 tasks with NLG metric, 4 tasks with exact-match metric, and 4 tasks containing T5/LLaMa unrecognized tokens were removed from evaluation.
%For unobserved instruction, we follow the procedure in section 3.2 to collect task-specific instructions for each task that we are evaluating. 
%For observed instruction, we classify all tasks into three categories: 
%We group observed instructions under three categories corresponding to task types (which imply particular output formats): (1) Multiple-choice QA, i.e., tasks that entail selecting an answer to a question from a finite set of options;
%that present themselves as questions with choices. 
%(2) Multi-class classification, which requires classifying inputs into a finite set of categories; and (3) Binary classification, or tasks that demand ``yes'' or ``no'' responses. 

\begin{table}[h]
\small
    \centering
    \begin{tabular}{c l}
    \toprule
    \multirow{3}{*}{\textsc{QA}} & In this task, you are given a multiple-choice question and you have to pick the\\
    & correct option. Answer with option indexes (i.e., "A", "B", "C", and "D").  \\
    & Q: \textcolor{ForestGreen}{\{question\}} A. \textcolor{MidnightBlue}{\{choiceA\}} B. \textcolor{MidnightBlue}{\{choiceB\}} C. \textcolor{MidnightBlue}{\{choiceC\}} D. \textcolor{MidnightBlue}{\{choiceD\}}\\
    \midrule
    \multirow{1}{*}{\textsc{MC}} & Pick one category for the following text. The options are - \textcolor{MidnightBlue}{\{options\}} \textcolor{ForestGreen}{\{text\}} % \\
    %& \textcolor{ForestGreen}{\{text\}} \\
    \\
    \midrule
    \multirow{2}{*}{\textsc{BC}} & \textcolor{ForestGreen}{\{paragraph\}} Choose your answer: According to the above paragraph, the \\
    & question "\textcolor{ForestGreen}{\{question\}}" is "\textcolor{MidnightBlue}{\{response\}}"? \\
    \bottomrule
    \end{tabular}
    \caption{Examples of observed instructions we collected for three general types of tasks.}
    \label{table:instruction-examples}
\end{table}

We use the same instructions for all tasks in the same category, taken from the published instruction tuning datasets associated with each model.
These instructions are general, e.g., in the case of classification they request that the model consider an example with respect to categorization criteria and label space provided by the instance, and select an appropriate category (examples in Table \ref{table:instruction-examples}). 
One can ``mix-and-match'' such instructions so long as they are appropriate for the task type.

%this means there will be cases where we use an ``incorrect'' instruction for a particular dataset, i.e., instructing the model to select an answer on a basis that is in fact irrelevant to the task being considered (but such that the elicited output format will be correct).
%This may degrade the performance of ``observed'' instructions, as compared to results which might be obtained if one manually aligned instructions to datasets within benchmarks. 
%We made this analysis decision to avoid biasing results by inflating the performance of ``observed'' instructions; we are interested in how robust instruction-tuned LLMs are

%3) Binary-classification: tasks that require the answer 'yes' or 'no'. 
%To avoid subjective bias, we use the same instructions for all tasks in the same category. They are collected from the published instruction tuning dataset for each model.

\begin{table}%[h]
    \centering
    \small
        \begin{tabular}[t]{l l c c c}
        \multicolumn{5}{c}{\textsc{Observed Instructions}} \\
        \toprule
        %& \textsc{MMLU} & \multicolumn{3}{c}{\textsc{BBL}} \\
        %& \multicolumn{3}{c}{\textsc{MMLU}} & \multicolumn{1}{c}{\textsc{BBL}} \\
        \emph{Instruction Type} & \multicolumn{2}{c}{QA} & MC & BC \\
        Flan & \multicolumn{2}{c}{50} & 35 & 18 \\
        Alpaca & \multicolumn{2}{c}{20} & 20 & 11 \\
        P3 & \multicolumn{2}{c}{13} & 8 & 7 \\
    \end{tabular}
    \quad
    \begin{tabular}[t]{l l|c|c|c}
        \multicolumn{5}{c}{\textsc{Unobserved Instructions}} \\
        \toprule
        Number of tasks & \multicolumn{2}{c}{1} & \multicolumn{2}{c}{14} \\
        Instructions per task & \multicolumn{2}{c}{20} & \multicolumn{2}{c}{10} \\
        \hline
        Total instructions & \multicolumn{2}{c}{20} & \multicolumn{2}{c}{140} \\
    \end{tabular}

    \caption{Counts of instruction phrasings (unobserved and observed) we use for evaluations.}
    \label{tab:data_stat}
\end{table}

\subsection{Results}
\label{section:main-analysis-results}

\begin{figure}[htbp]
  \centering
  \begin{subfigure}{0.475\textwidth}
    \includegraphics[width=\textwidth]{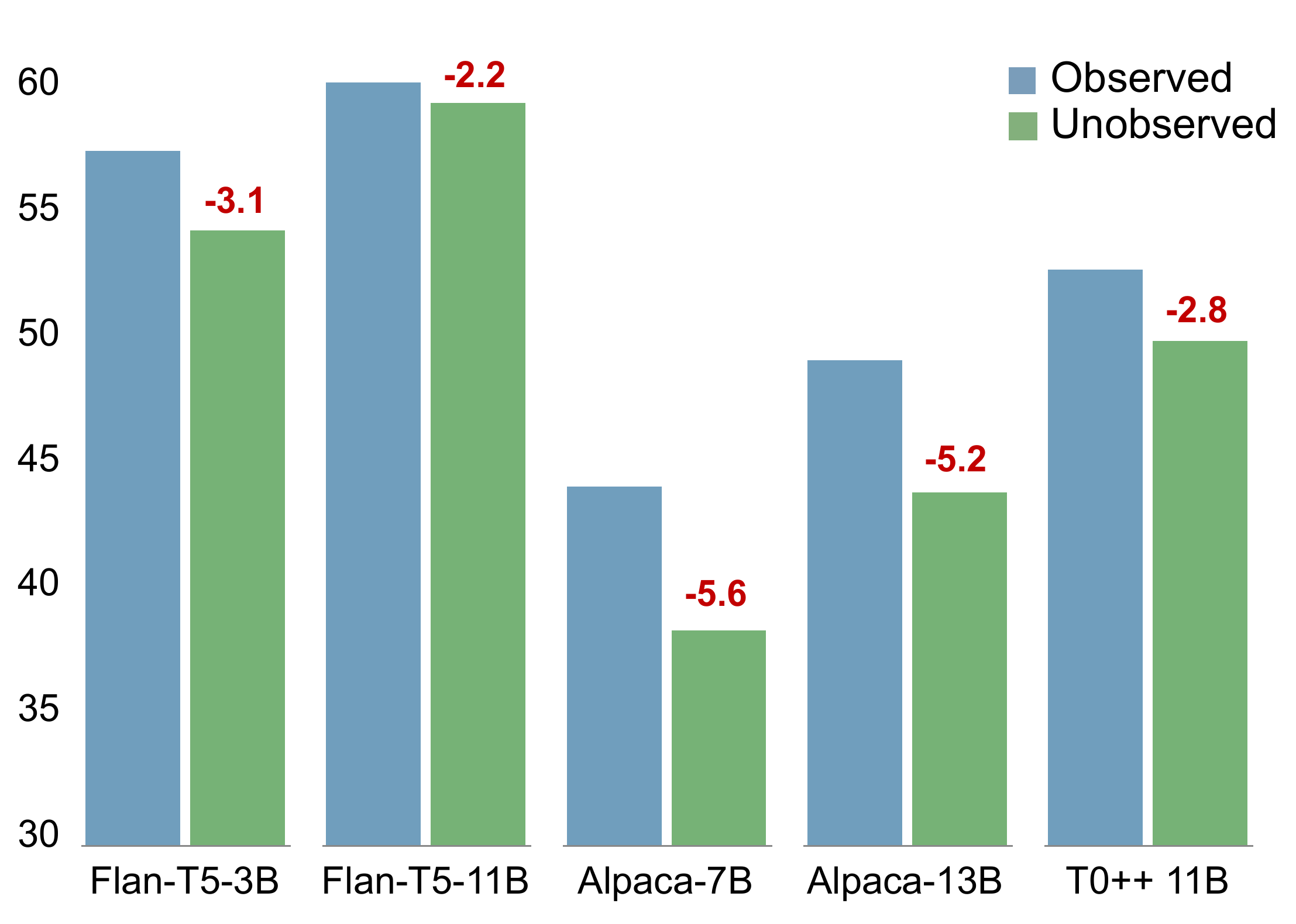}
    \caption{Average zero-shot performance over all tasks when using observed and unobserved instructions.}
    \label{fig:main-results-main_results}
  \end{subfigure}
  \hfill
  \begin{subfigure}{0.475\textwidth}
    \includegraphics[width=\textwidth]{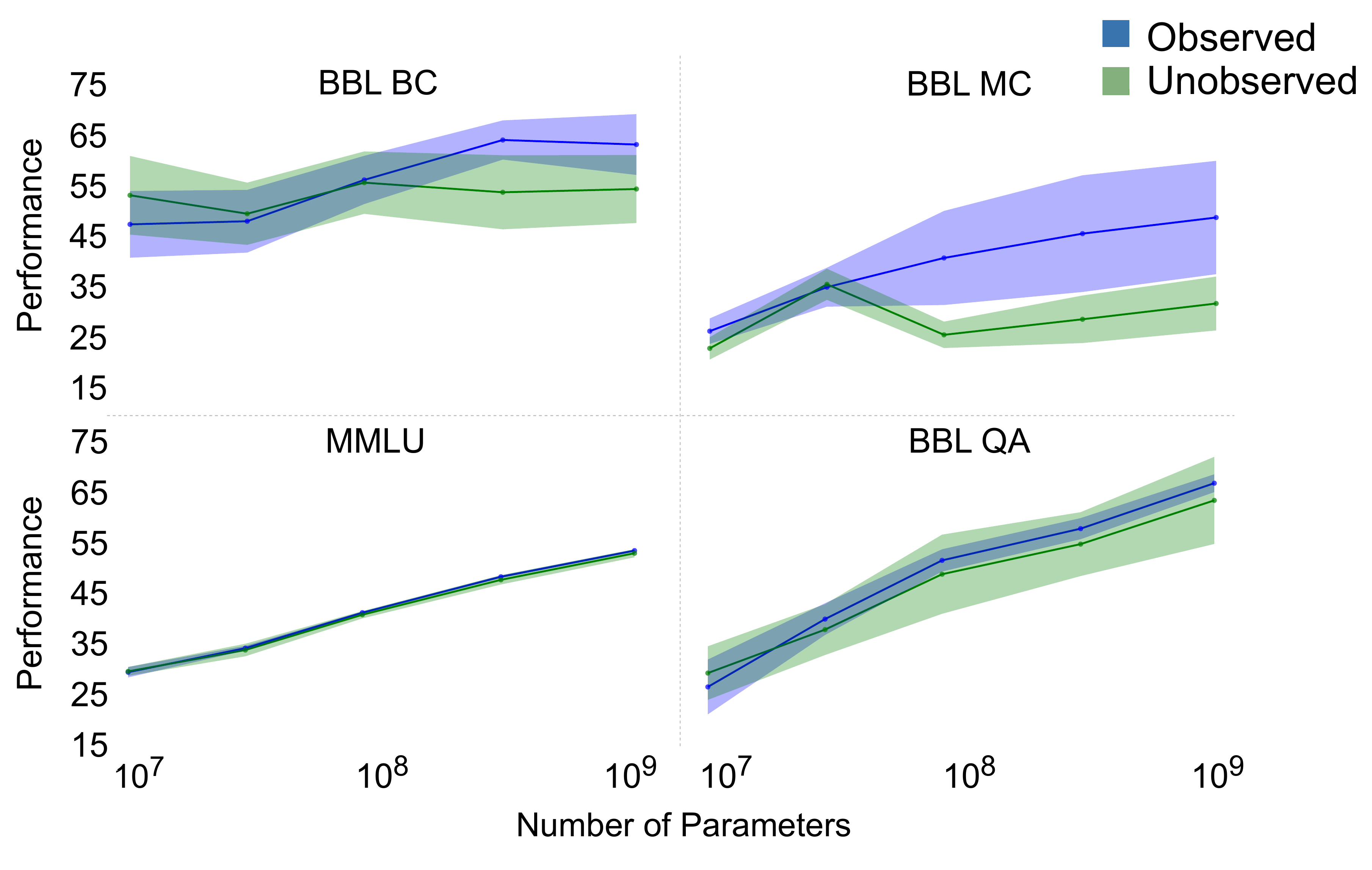}
    \caption{Performances of Flan-T5 using observed and unobserved instructions as a function of model size.}
    \label{fig:main_scaling_reesults}
  \end{subfigure}
  \caption{Using novel but valid instructions at test time (phrasings unobserved in training) consistently degrades the performance of instruction-tuned LLMs (a). Scale does not necessarily fix this (b).}
  \label{fig:main-results}
\end{figure}

We present the main aggregated analysis results in Figure \ref{fig:main-results} and Table \ref{tab:main_result}.
The take-away here is that using instructions unobserved in training---but manually composed for the task at hand and so semantically appropriate---leads to considerable degradation in performance: On average, unobserved instructions reduce accuracy by over five points across models considered. 
Table \ref{tab:main_result} reports results disaggregated by task type; we observe that classification tasks are most harmed by use of novel instructions. 
We provide additional, more granular (dataset-level) results in the Appendix.

\begin{comment}
\begin{figure}
    \centering
    \includegraphics[width=14.4cm]{images/main_results_v4.pdf}
    \caption{Average performance across tasks between observed and unobserved instructions.}
    \label{fig:main_results}
\end{figure}

\begin{figure}
    \centering
    \includegraphics[width=14.4cm]{images/main_scaling_results_v1.pdf}
    \caption{}
    \label{fig:main_scaling_results}
\end{figure}
\end{comment}

\begin{table}[h]
\small
    \centering
    \begin{tabular}{l c c c c c}
        \toprule
         \multirow{2}{*}{\textbf{Model}} & \textsc{MMLU} & \textsc{BBL-QA} & \textsc{BBL-BC} & \textsc{BBL-MC} & \textbf{Overall} \\ [0.5ex]
        & Avg. \ \ Std. & Avg. \ \ Std. & Avg. \ \ Std. & Avg. \ \ Std. & Avg. \ \ Std. \\
        \hline 
        \rule{0pt}{12pt} Flan-T5-3B  & & & &  \\
        \hspace{0.25cm} \textsc{Observed}   & $\textbf{48.1} \ \ (\pm 0.3)$ & $\textbf{59.0} \ \ (\pm 2.1)$ & $\textbf{66.5} \ \ (\pm 3.8)$ & $\textbf{55.6} \ \ (\pm 0.7)$ & $\textbf{57.3} \ \ (\pm 1.7)$ \\        
        \hspace{0.25cm} \textsc{Unobserved} & $47.5 \ \ (\pm 0.9)$ & $56.0 \ \ (\pm 7.3)$ & $61.1 \ \ (\pm 6.9)$ & $52.1 \ \ (\pm 5.4)$ & $54.2 \ \ (\pm 5.1)$ \\
        \hspace{0.25cm} \textbf{Performance $\Delta$} & \textcolor{red}{$\downarrow 0.6$} & \textcolor{red}{$\downarrow 3.0$} & \textcolor{red}{$\downarrow 5.5$} & \textcolor{red}{$\downarrow 3.5$} & \textcolor{red}{$\downarrow 3.1$} \\
        %\hline 
        \rule{0pt}{12pt} Alpaca-7B  & & & & \\
        \hspace{0.25cm} \textsc{Observed}   & $\textbf{41.9} \ \ (\pm 0.6)$ & $\textbf{48.6} \ \ (\pm 2.8)$ & $\textbf{53.8} \ \ (\pm 3.4)$ & $\textbf{32.1} \ \ (\pm 2.2)$ & $\textbf{44.1} \ \ (\pm 2.3)$  \\
        \hspace{0.25cm} \textsc{Unobserved} & $39.7 \ \ (\pm 2.2)$ & $45.3 \ \ (\pm 6.5)$ & $52.4 \ \ (\pm 6.5)$ & $16.4 \ \ (\pm 3.5)$ & $38.5 \ \ (\pm 4.7)$  \\
        \hspace{0.25cm} \textbf{Performance $\Delta$}  & \textcolor{red}{$\downarrow 2.2$} & \textcolor{red}{$\downarrow 3.3$} & \textcolor{red}{$\downarrow 1.4$} & \textcolor{red}{$\downarrow 15.7$} & \textcolor{red}{$\downarrow 5.6$} \\
        %\hline 
        \rule{0pt}{12pt} T0++ 11B & & & & \\
        \hspace{0.25cm} \textsc{Observed}   & $48.3 \ \ (\pm 0.9)$ & $54.1 \ \ (\pm 4.1)$  & $\textbf{66.1} \ \ (\pm 2.1)$ & $\textbf{42.0} \ \ (\pm 2.1)$ & $\textbf{52.6} \ \ (\pm 2.3)$ \\            
        \hspace{0.25cm} \textsc{Unobserved} & $\textbf{48.5} \ \ (\pm 0.9)$ & $\textbf{54.7} \ \ (\pm 3.7)$ & $54.7 \ \ (\pm 4.3)$ & $41.4 \ \ (\pm 2.4)$ & $49.8 \ \ (\pm 2.8)$ \\
        \hspace{0.25cm} \textbf{Performance $\Delta$} & \textcolor{ForestGreen}{$\uparrow 0.2$} &  \textcolor{ForestGreen}{$\uparrow 0.7$} & \textcolor{red}{$\downarrow 11.4$} & \textcolor{red}{$\downarrow 0.6$} & \textcolor{red}{$\downarrow 2.8$} \\
        %\hline 
        \rule{0pt}{12pt} Flan-T5-11B & & & & \\
        \hspace{0.25cm} \textsc{Observed}   & $\textbf{53.2} \ \ (\pm0.2)$ & $\textbf{67.9} \ \ (\pm1.8)$ & $\textbf{65.6} \ \ (\pm6.0)$ & $\textbf{58.7} \ \ (\pm0.5)$ & $\textbf{61.4} \ \ (\pm2.1)$ \\            
        \hspace{0.25cm} \textsc{Unobserved} & $52.7 \ \ (\pm0.8)$ & $64.6 \ \ (\pm8.5)$ &  $63.6 \ \ (\pm6.1)$ & $55.9 \ \ (\pm5.5)$ & $59.2 \ \ (\pm5.2)$ \\
        \hspace{0.25cm} \textbf{Performance $\Delta$} & \textcolor{red}{$\downarrow 0.5$} & \textcolor{red}{$\downarrow 3.4$} &  \textcolor{red}{$\downarrow 2.0$} & \textcolor{red}{$\downarrow 2.8$} & \textcolor{red}{$\downarrow 2.2$} \\
        %\hline
        \rule{0pt}{12pt} Alpaca-13B  & & & & \\
        \hspace{0.25cm} \textsc{Observed}   & $\textbf{47.8} \ \ (\pm 0.5)$ & $\textbf{53.9} \ \ (\pm 2.2)$ & $\textbf{57.9} \ \ (\pm 4.8)$ & $\textbf{36.7} \ \ (\pm 1.8)$ & $\textbf{49.1} \ \ (\pm 2.3)$ \\            
        \hspace{0.25cm} \textsc{Unobserved} & $47.0 \ \ (\pm 0.8)$ & $51.7 \ \ (\pm 5.7)$ & $54.1 \ \ (\pm 5.6)$ & $22.7 \ \ (\pm 7.5)$ & $43.9 \ \ (\pm 14.0)$ \\
        \hspace{0.25cm} \textbf{Performance $\Delta$} & \textcolor{red}{$\downarrow 0.9$} & \textcolor{red}{$\downarrow 2.2$} & \textcolor{red}{$\downarrow 3.8$} & \textcolor{red}{$\downarrow 14.0$} & \textcolor{red}{$\downarrow 5.2$} \\
        \bottomrule
    \end{tabular}
    \vspace{0.25cm}
    \caption{Results using observed and unobserved instructions across benchmark tasks (grouped by type). Performance degrades---sometimes by 10+ points---when one uses (\textsc{unobserved}) instructions, suggesting that instruction-tuned models are not particularly robust. BC, MC, and QA stand for binary classification, multi-class classification, and question answering, respectively.} %The overall performance of instruction-tuned models with observed and unobserved data. \textsc{MC} stands for multiple classification, and \textsc{BC} stands for binary classification.}
    \label{tab:main_result}
\end{table}

\subsection{A Closer Look at Instruction Robustness}
\label{section:closer-look}

\begin{figure}
    \centering
    \includegraphics[scale=0.335]{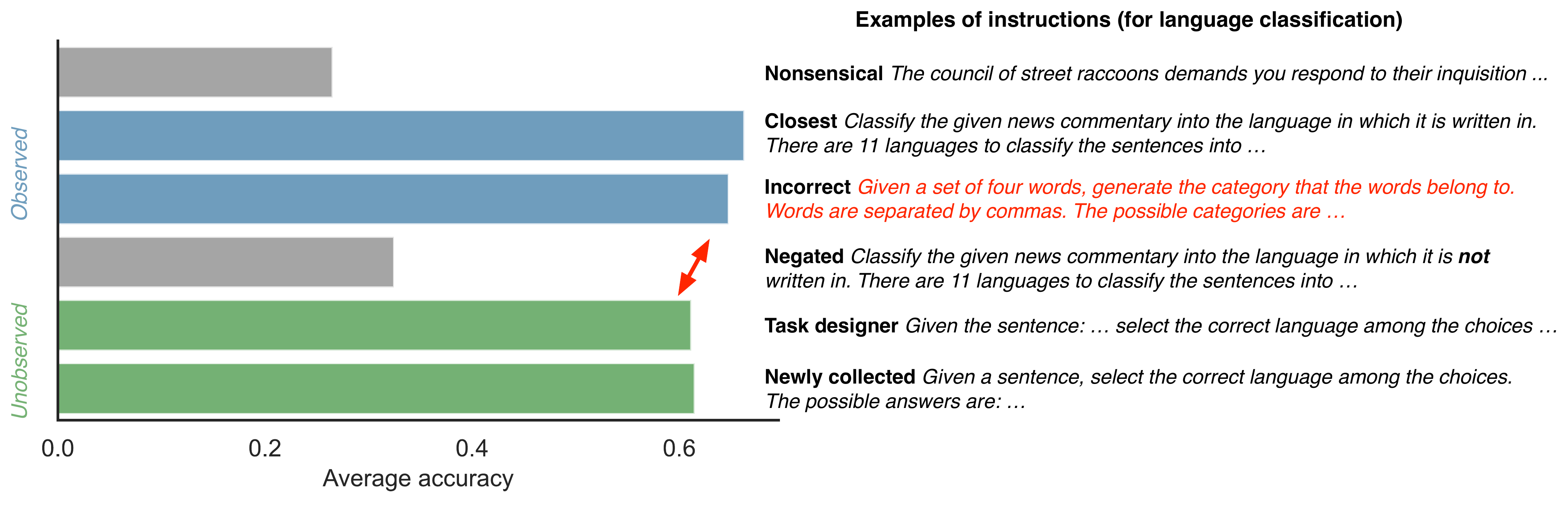}
    \caption{{\bf \emph{Incorrect} but observed instructions perform better on average than \emph{correct} but unobserved instructions}. We report averages over benchmarks, but show example instructions on the right for a specific, illustrative task. We provide all instructions in the Appendix.}
    \vspace{-0.5em}
    \label{fig:adversarial}
\end{figure}

Above we used general instructions requesting the model to perform tasks (Table \ref{table:instruction-examples}).
%to avoid data contamination due to researcher bias. 
Here we delve further into the performance degradation observed when using novel instructions. % that results from using unobserved instructions at inference time, 
We report a curious result highlighting the degree to which models rely on having previously observed instructions: Incorrect but observed instructions outperform appropriate but unobserved instructions (Figure \ref{fig:adversarial}).
%degree to which these models depend on having observed an instruction given at inference time. 

We come to this observation by evaluating the performance of Flan-T5-XXL (11B) using six instruction types over seven datasets from \textsc{Big-Bench}. %, following the same protocol as in \ref{section:main-analysis-results}, 
%we evaluate the performance of Flan-T5-XXL (11B) with six different instruction types. %under six different settings. 
In particular, this includes (variants of) two instructions \emph{observed} in training:
\textbf{Closest} is the instruction from the most similar task in the instruction-tuning set; \textbf{Incorrect} is an observed instruction for a \emph{completely different} and inappropriate task (but which has the same desired output format, e.g., classification)---intuitively these should not yield the desired behavior; \textbf{Negated} is the same as \textbf{closest}, but we negate the instruction to indicate that it should \emph{not} perform the task. 
%to reverse the instruction.

For \emph{unobserved} instructions, we consider:
\textbf{Task designer}, the instruction (task prefix) provided by the author of the task in \textsc{Big-Bench}, and; 
\textbf{Newly collected}, or the novel instructions collected from NLP graduate students, described above.
As a control for reference, we also consider \textbf{Nonsensical}, which is a random ``instruction'' completely irrelevant to any task.

\begin{comment}

\begin{figure}
    \centering
    \includegraphics[width=140mm]{images/adversarial_w_examples_cropped.pdf}
    \caption{Caption}
    \label{fig:adversarial}
\end{figure}
\end{comment}

Figure \ref{fig:adversarial} reports average results for these variants. 
Consistent with our findings, using instructions unobserved in training degrades performance. 
Strikingly, here we also find that using an \emph{inappropriate but observed} instruction outperforms using \emph{appropriate but unobserved} instructions.
This indicates that instruction-tuned models---or at least modestly sized ones we have evaluated here---may in some way overrely on having observed instructions in training, and do not generalize to new instructions and phrasings as we might hope. We provide all the instructions and results in the Appendix.
%generalize somewhat poorly to new instructions, compared to their performance when used with instructions seen in training.
%That observed but incorrect instructions fare better than unobserved but correct instructions illustrates this point. 

%The result shows that in most of cases, the observed instructions outperform the unobserved ones by a large margin. More surprisingly, the incorrect, observed instructions - being semantically distracting to the actual task - have comparable performance with the unobserved instruction, which is semantically accurate but not trained before.

% Specifically, we ... % TODO
%Here, to provide a closer look at the gap between the performance of observed and unobserved instructions, we conduct a more specific experiment that tests the  

\begin{comment}
\begin{table}[h]
    \centering
    \begin{tabular}{l c}
    \toprule
        Settings & \textbf{Avg. Acc.} \\ [0.5ex]
        \hline
        Task Designer & 45.2 \\
        Unobserved & 44.1 \\
        \hline
        Observed - Correct & \textbf{50.5} \\
        Observed - Incorrect & 47.2 \\
        Observed - Negated & 43.7 \\
        \hline
        Random Text & 35.9 \\
        \bottomrule
    \end{tabular}
    \caption{Results on 7 Datasets from \textsc{Big-Bench} with different instruction settings.}
    \label{tab:my_label}
\end{table}
\end{comment}

\subsection{Scaling}

%Due to the limited accessibility of LLMs instruction-tuning collections, we are not able to evaluate the instruction robustness of 
Does instruction robustness begin to emerge as a function of scale? 
To attempt to answer this, we repeated all experiments from Table \ref{tab:main_result} with Flan-T5 model sizes ranging from small (80M parameters) to XXL (11B). 
We observe in Figure \ref{fig:main_scaling_reesults} that the disparity between results achieved with observed versus unobserved instructions \textbf{does not} seem to decrease with model scale, at least up to this point.
That said, massive models (175B+) may offer greater robustness. 
However, we reiterate that much of the excitement about instruction tuning is the possibility that this technique appears to allow much smaller models to achieve results competitive with massive alternatives. 
%the scalability of the issue that we have discovered, we took Flan-T5, the instruction-tuned model with the largest size variation, by repeating all the experiments in \ref{tab:main_result} from Flan-T5-Small (80M) to Flan-T5-XXL (11B). The performance gap on both \textsc{MMLU} and \textsc{BBL} is not decreased as the model scales exponentially.

\subsection{Robustness with Semantic Distance}
\label{section:mmlu_variance}
%Results Not So Variable on MMLU?}

One observation in \ref{section:main-analysis-results} is that performance on \textsc{MMLU} is less affected by using unobserved instructions.
%seems to suffer less from the O.O.D. instructions. 
\textsc{MMLU} is a benchmark with 57 QA tasks about different knowledge domains; these tasks all share a similar form of input-output (question, four choices $\rightarrow$ answer).
During instruction collection, we treated all tasks in \textsc{MMLU} as a general QA task and asked NLP researchers to write general QA instructions. 
%to give instructions that can apply to all the questions. Hence, these "meta" 
As a result, we hypothesize that these instructions are comparatively similar to the observed instructions, and this in turn explains the relative robustness in this case. 
% that we collected for evaluation. Therefore, the degradation of the performance is relatively light.

\begin{figure}[h]
    \centering
    \includegraphics[scale=0.255]{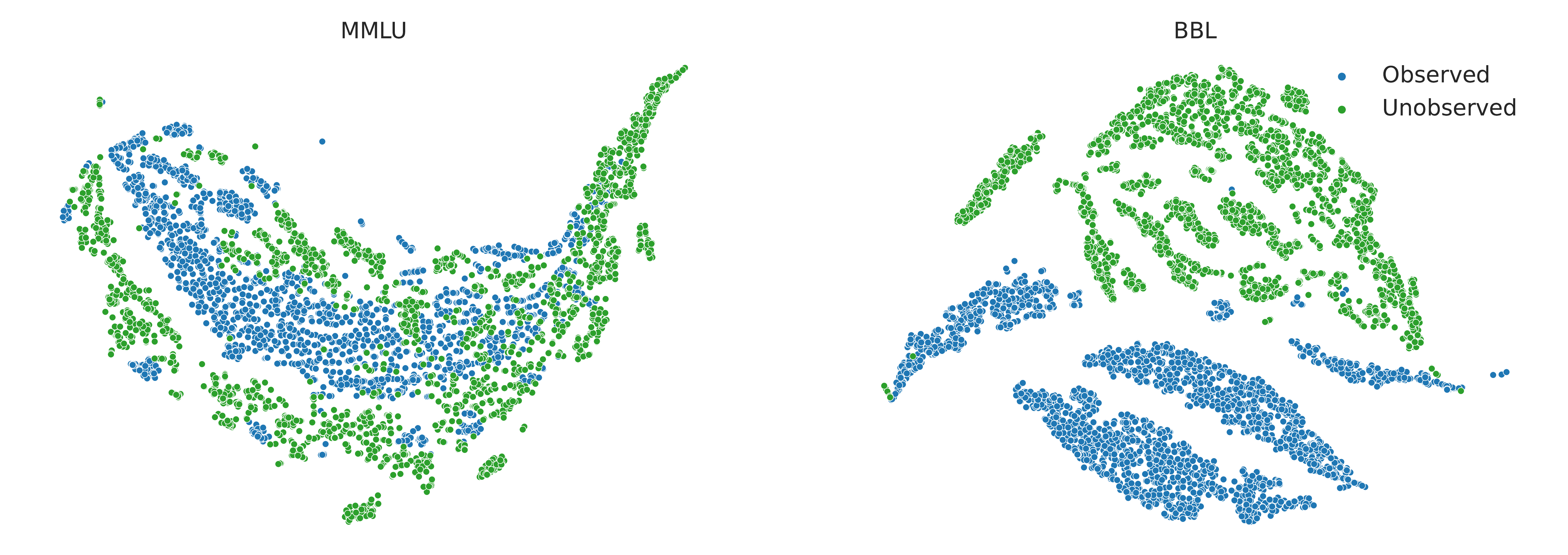}
    \caption{tSNE plots of representations for the first decoded tokens of 300 randomly sampled examples from \textsc{MMLU} and \textsc{BBL} with Flan-T5 (XXL). Embeddings of observed and unobserved instructions for \textsc{MMLU} are similar, while for \textsc{BBL} they are quite different. This result holds across most but not all models considered: See the \ref{section:embeddings} for visualizations over all models.}%The pattern of the tSNE plot is not an absolute metric and varies from model to model based on the embedding size. However, most of the examples we visualize show similar patterns as demonstrated, and we report all of the visualizations in the Appendix.}
    \label{fig:embeddings}
\end{figure}

\begin{figure}[h]
\begin{floatrow}
\floatbox{figure}[.5\textwidth][\FBheight][t]{
    \centering
    \includegraphics[scale=0.45]{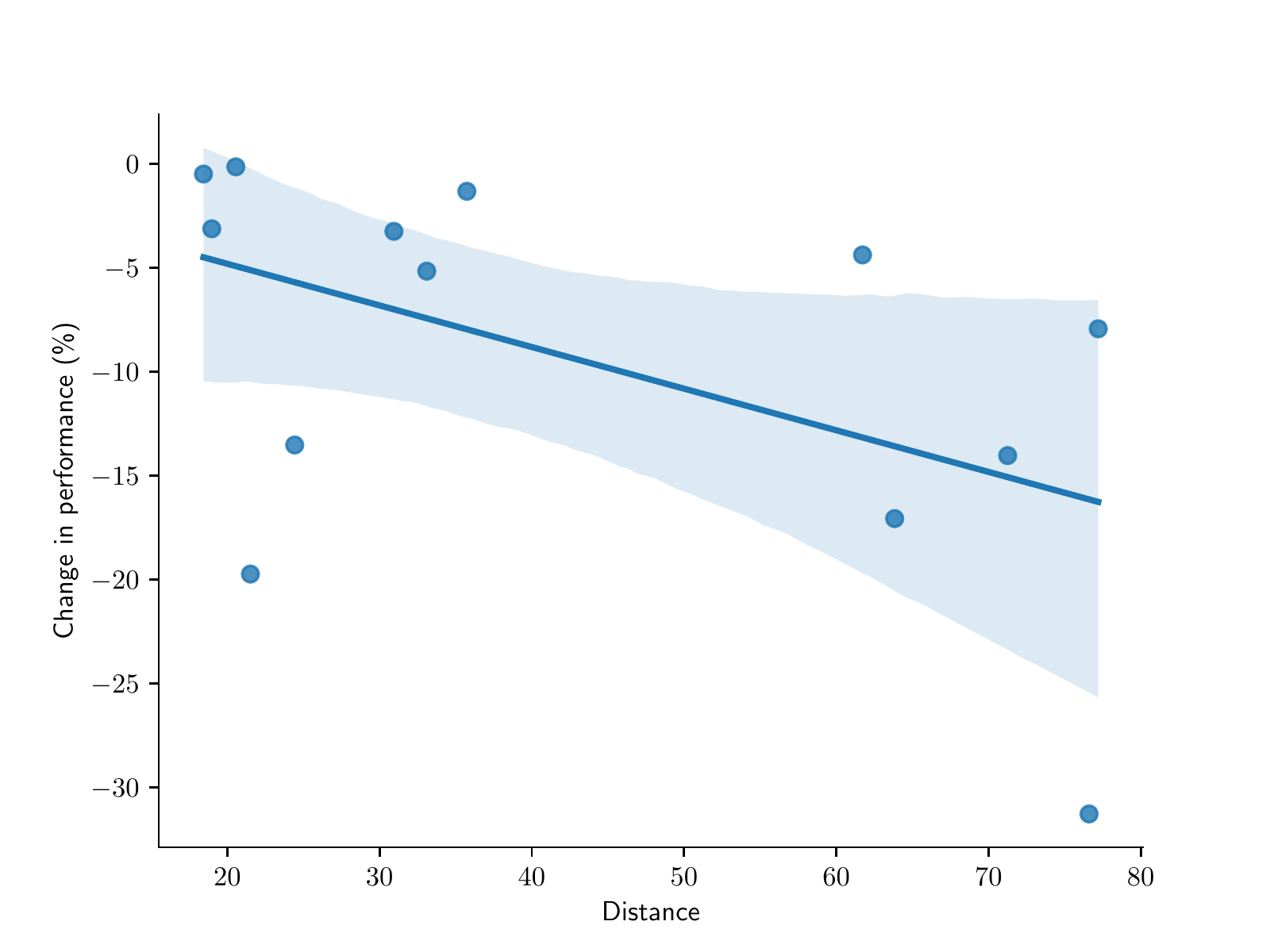}
}
{
\caption{Plots of average degradations in performance versus the semantic distance while using unobserved instructions.}
\label{fig:perf-dist-reg}
}
% \label{fig:perf-dist-reg}
\floatbox{table}[.5\textwidth][\FBheight][t]{
    \vspace{-30pt}
    \begin{tabular}{l l l}
    \toprule
        \textbf{Dataset} & \textbf{Avg.} $\Delta\ell$2 & \textbf{Avg. $\Delta$ Acc.} \\
        \midrule
        \textsc{MMLU} & \textbf{19.8} & -\textbf{0.5} \\
        \midrule
        \textsc{BBL-QA} & 37.9 & -3.4 \\
        \textsc{BBL-BC} & 25.3 & -2.0 \\
        \textsc{BBL-MC} & 26.1 & -2.8 \\
        \bottomrule
    \end{tabular}
}
{
\caption{Average degradations in performance for four categories. It could be seen that \textsc{MMLU} has minimal average distance, which indicates a smaller distribution shift, and hence leads to the smallest degradation}
\label{table:distances}
}
\end{floatrow}
\end{figure}

We empirically verify this in Figure \ref{fig:embeddings} and Table \ref{table:distances}. For each instance (instruction plus example), we extract the representation at the penultimate layer for the first decoded token. %To visualize the distribution of instances with observed and unobserved instructions, w
We use tSNE \cite{van2008visualizing} to visualize these representations of observed and unobserved instructions over instances in \textsc{MMLU} and \textsc{BBL}. 
Figure \ref{fig:embeddings} shows that in the case of \textsc{MMLU} the unobserved instructions we collected are quite similar to the observed, while there is a greater separation between unobserved and observed instructions in \textsc{BBL}. 
We also provide a numerical measurement of this phenomonen in Table \ref{table:distances}.
We report the average $\ell$2 distance between representations of unobserved instructions and those of their nearest observed counterparts.
We see that \textsc{MMLU} unobserved instructions are, on average, closer to the nearest observed instruction; this correlates with the lower observed performance drop.
These findings are in line with the hypothesis that the unobserved instructions for \textsc{MMLU} are more similar to the observed instructions for this dataset, and this likely explains the apparent robustness in this case. %We report analogous results for all datasets in the Appendix.
%we provide a numerical measurement of what the plot is indicating. For each unobserved instruction, we match it with the closest observed instruction by the average L2 distance across all the examples that we sampled. We compute the average distances and accuracy degradation with all the observed and unobserved instructions pairs.
%, we show that the instructions we collected for MMLU and BBL have an astonishing difference. In MMLU, the instances with observed instruction and unobserved instruction are less distinguishable, whereas in BBL, the two clusters are clearly separated in the latent space.
 
%To verify our hypothesis, propose a method to measure the distributional difference between observed and unobserved instructions. 

\begin{comment}
\begin{table}[h]
\small
    \centering
    \begin{tabular}{l l l}
    \toprule
        \textbf{Dataset} & \textbf{Avg. $\ell$2} ($\ell$2) & \textbf{Avg. $\Delta$ Accuracy} (\%) \\
        \midrule
        \textsc{MMLU} & \textbf{19.8} & -\textbf{1.5}\% \\
        \midrule
        \textsc{Novel Concepts} & 22.0 & -3.1\% \\
        \textsc{StrangeQA} & 55.3 & -5.5\% \\
        \textsc{Language Identification} & 59.0 & -11.3\% \\
        \bottomrule
    \end{tabular}
    \caption{Average distances and accuracy degradations (as \%) on three datasets in \textsc{BBL}.}
    \label{table:distances}
\end{table}
\end{comment}

%The result shows that the accuracy degradation for using unobserved instructions is highly correlated with the similarity between the instructions used and the instruction trained. 

We plot mean performance degradation (as \%) as a function of average similarity between the similarity of the first decoded tokens (following \emph{unobserved} instructions) and the same for the \emph{most similar} \emph{observed} instruction. 
The negative slope implies the intuitive relationship: Instructions that are dissimilar (in terms of model representations) tend to result in poorer performance.  However, the relationship is relatively weak, yielding an intercept estimate of -0.8 and a slope of -0.2 ($p=$0.08).

\begin{comment}
    \caption{Average degradations in performance observed when using instructions unobserved training as a function of the similarity between (a) the representation induced by the model for a given instruction, and, (b) the same for the \emph{nearest} observed instruction. This is for Flan-T5 (XXL). We use representations for the first token following the instruction, extracted from the penultimate layer in the network.}
\end{comment}

    \vspace{-0.5em}
\subsection{Robustness Under In-Context Learning (ICL)}

Previous study \cite{gu2023robustness} has shown that the LLMs are less sensitive to prompt / instruction variation when few-shot examples are provided in context. 
While we are focused on zero-shot capabilities, for completeness, we re-ran all experiments in a few-shot setting. 
%Although this is not the focus of our work, we have conducted all of our previous experiments equally under few-shot settings. 
We report these results in the \ref{section:icl_robustness}. The main finding is that while some discrepancy remains, in general ICL \textbf{slightly} decreases the sensitivity of models to the use of unobserved instructions.
This is intuitive, given that the examples themselves likely imply the desired task and may affect the distribution.

\begin{figure}
    \centering
    \includegraphics[scale=0.17]{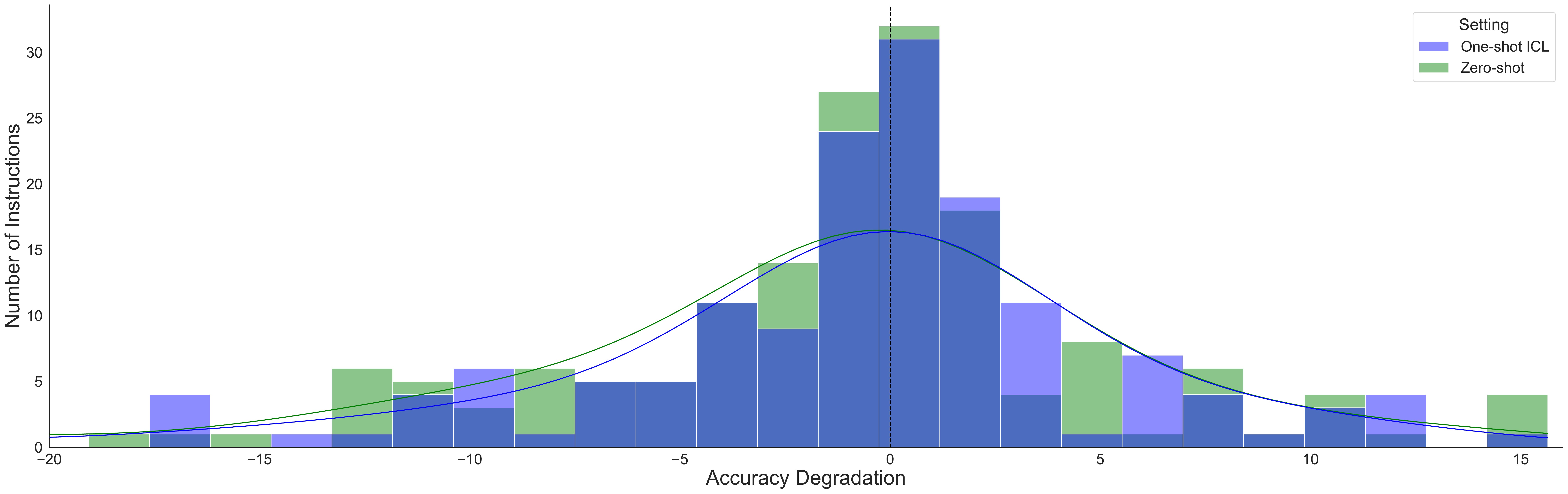}
    \caption{The performance degradation when using unobserved instruction at \textsc{BBL} and \textsc{MMLU} with Flan-T5-XXL. We plot the accuracy degradation of all the unobserved instructions compared with the average accuracy of the observed ones. It could be seen that under one-shot in-context learning, the model is slightly more robust as the performance difference converges closer to 0}
    \label{fig:icl}
\end{figure}

\begin{comment}
\begin{figure}[h]
\begin{floatrow}
\floatbox{figure}[.5\textwidth][\FBheight][t]{
    \centering
    \includegraphics[scale=0.16]{images/icl.pdf}
    \label{fig:perf-dist-reg}
}
{
\caption{Plots of average degradations in performance versus the semantic distance while using unobserved instructions.}
}
% \label{fig:perf-dist-reg}
\floatbox{table}[.5\textwidth][\FBheight][t]{
    \vspace{-10pt}
    \begin{tabular}{l c c c}
    \toprule
        \textbf{Dataset} & \textbf{0-shot} & \textbf{1-shot} &  $\Delta$ \textbf{Acc.}  \\
        \midrule
        \textsc{Small (80M)} & 22.0 & -3.1 & - \\
        \textsc{Small (80M)} & 55.3 & -5.5 & - \\
        \textsc{Small (80M)} & 59.0 & -11.3 & -\\
        \textsc{Small (80M)} & 59.0 & -11.3 & -\\
        \textsc{Small (80M)} & 59.0 & -11.3 & -\\
        \bottomrule
    \end{tabular}
    \label{table:icl_improvement}
}
{
\caption{Average degradations in performance for four categories. It could be seen that \textsc{MMLU} has the minimal distance and hence yields the least degradation}
}
\end{floatrow}
\end{figure}
\end{comment}

\section{Aligning Equivalent Instructions} 
    \vspace{-0.5em}
\label{section:methods}

%As an attempt to address the issues discussed above, 
We now introduce a simple, lightweight, but effective method to improve the robustness of instruction-tuned LLMs. 
The intuition is to introduce a term in the objective which explicitly encourages the model to yield similar predictions (and hence similar representations) for the same input when provided distinct but semantically equivalent instructions.

%\subsection{Aligning Instructions}

\begin{figure}[h]
        \centering
        \begin{subfigure}[b]{0.6\textwidth}
            \centering
            {\includegraphics[scale=0.365]{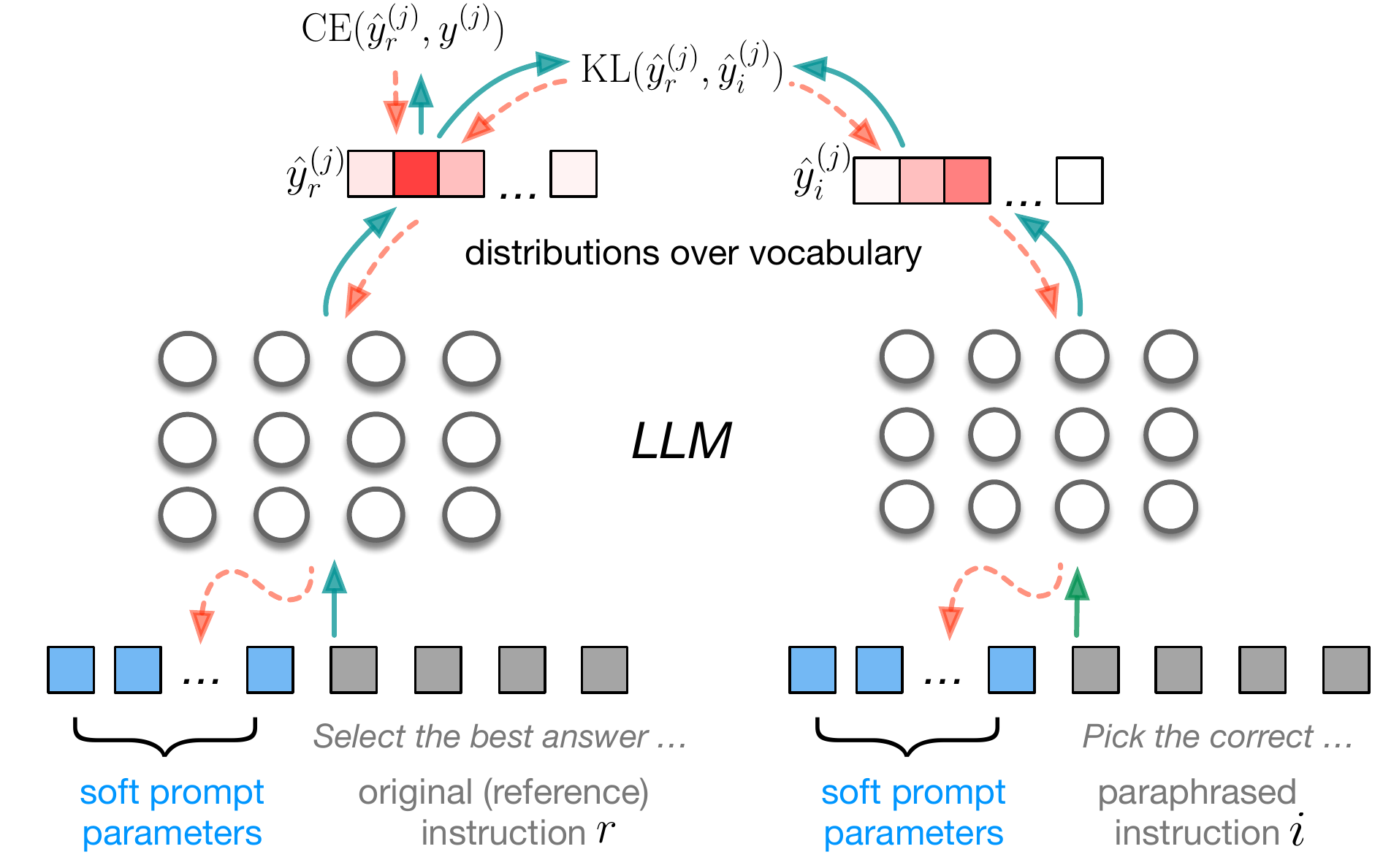}}
        \end{subfigure}
        \hfill
        \begin{subfigure}[b]{0.125\textwidth}
            \begin{align*}
                &\mathcal{L} = (1-\lambda)\mathcal{L}_{\text{CE}} + \lambda\mathcal{L}_{\text{KL}} \\
                &\mathcal{L}_{\text{KL}} = \frac{1}{N-1}\sum_{i\neq r}^{N}\sum_j \text{KL}(\hat{y}_{i}^{(j)} ||\hat{y}_r^{(j)}) \\
                &\hat{y}_{i}^{(j)} = \text{Softmax}(p_{i}^{(j)})\text{, }p_{i}^{(j)}\in\mathbb{R}^{|\mathcal{V}|} \\
            \end{align*}
        \end{subfigure}
        \caption{Schematic depiction of the proposed instruction alignment method (left) and associated loss terms (right). Dotted (red) lines indicate backpropagation; we update only the soft prompt parameters, which we show yields performance superior to fine-tuning all model parameters.
        \label{fig:method-schematic}}
   \end{figure}

%For an autoregressive Language Model with
More specifically, we aim to align semantically equivalent instructions in the space induced by the model.
To this end we introduce soft embedding parameters with dimensions $\mathbb{R}^{d \times n}$; this is equivalent to adding $n$ novel tokens (with embedding dimension $d$) as prefixes to inputs (preceding instructions). 
The intuition is to push the representations for semantically equivalent tasks close together. 
To this end, we add additional term to the loss: The KL-divergence $\mathcal{L}_{\text{KL}}$ of the output probabilities between a reference instruction for a given task and paraphrased (semantically equivalent) version of the same. 
We combine this with the standard cross-entropy loss, and fine-tune \emph{only} the introduced soft prompt parameters under this objective (Figure \ref{fig:method-schematic}). 
Here $\lambda$ is a loss-weighting hyper-parameter, $\hat{y}^{(j)}_i$ and $\hat{y}_r^{(j)}$ are the distributions over the vocabulary $\mathcal{V}$ induced by the model with paraphrased instruction $i$ and the reference instruction $r$ at token position $j$.\footnote{We pad instances such that the lengths in a given batch are effectively equal; the sum is therefore from 1 to the length associated with the current batch, we omit this for simplicity.}

Optimizing for the above objective requires paraphrased instructions $i$ for each task in the training data; we generate these automatically as follows. 
For instruction-tuning dataset, we sample a small amount of training data to use for alignment. %to serve as the instances for soft prompt alignment. 
 We paraphrase these reference instructions using GPT-4. For the Alpaca collection, we randomly sampled 1000 tasks and paraphrased them with three prompts, and collected the top three candidates under temperature 0.5. For the Flan collection, we randomly sampled 986 instances from the mixture with 3 prompts with greedy decoding. % Table \ref{tab:data_stat} reports counts yielded for different tasks we assembled in this way. 

For fine-tuning, we then create instances for each example by pairing them with every distinct instruction available for the corresponding task.  
We then form batches by including one instance featuring the original instruction and the rest comprising paraphrased instructions. For the implementation of the prefix, we follow the setting of \cite{li2021prefix}, which freezes the model parameters and just trains the prefix embeddings with the MLP layers.
%For each instruction (1 origin instruction + paraphrases) of the dataset, we combine it with all the instances. 
%Hence, for each instance with the original instruction $I$, we have N counterparts with paraphrased instruction $I'_{n}$ for $n\in \{1...N\}$.  

\section{Results}

We experiment with the proposed method using two representative instruction-tuned LLMs: Flan-XL (3B) and Alpaca (7B). 
We compare the canonical versions of these models trained in the usual way (the same evaluated in Table \ref{tab:main_result}) to variants fine-tuned using our proposed approach. 
We ablate components of our method to tease out the contributions of data and objectives.
%components of the proposed alignment strategy. 
%That is, we ablate the components of the proposed approach and report corresponding results to tease out the contributions of data and objectives. 

Specifically, we consider variants where we: Fine-tune all model parameters on the additional, automatically generated instruction paraphrases (FT); impose the new KL loss term (again fine-tuning all model parameters; FT+KL); introduce the additional soft prompt parameters and fine-tune on the paraphrase instances, but without KL (PT); and then the full proposed strategy, which introduces the soft prompt parameters and optimizes them for the loss augmented with the KL term ({\bf PT+KL}). 

%We compare variants of these models fine-tuned on the corresponding instruction datasets in the usual way to results achieved 
%For fine-tuning the baseline, we compare the performance of fine-tuning and so. More implementation details can be seen in Appendix ?

\begin{table}[h]
\centering
\small
\begin{tabular}{l c c c c c c }
\toprule
  & \multicolumn{3}{c}{\textsc{MMLU}} & \multicolumn{3}{c}{\textsc{BBL}} \\ [0.5ex]
\midrule
\textbf{Model} & \textsc{Obs.} & \textsc{Unobs.} & Avg. & \textsc{Obs.} & \textsc{Unobs.} & Avg. \\
    \textsc{Flan-T5-3B}   & 48.1 & 47.5 & 47.8 & \textbf{56.1} & 51.9 & 54.0 \\
    FT & 39.4 \textcolor{red}{\textbf{(-8.7)}} & 40.1 \textcolor{red}{\textbf{(-7.4)}} & 39.8 \textcolor{red}{\textbf{(-8.0)}} & 48.2 \textcolor{red}{\textbf{(-7.9)}} & 42.3 \textcolor{red}{\textbf{(-9.2)}} & 45.3 \textcolor{red}{\textbf{(-8.7)}} \\
    FT+KL    & 41.8 \textcolor{red}{\textbf{(-6.3)}} & 43.6 \textcolor{red}{\textbf{(-3.9)}} & 45.9 \textcolor{red}{\textbf{(-1.9)}} & 47.7 \textcolor{red}{\textbf{(-8.4)}} & 43.1 \textcolor{red}{\textbf{(-8.8)}} & 45.4 \textcolor{red}{\textbf{(-8.6)}} \\
     PT & 48.1 \textcolor{MidnightBlue}{\textbf{(+0.0)}} & 47.6 \textcolor{ForestGreen}{\textbf{(+0.1)}} & 47.9 \textcolor{ForestGreen}{\textbf{(+0.1)}} & 55.9 \textcolor{red}{\textbf{(-0.2)}} & 52.1 \textcolor{ForestGreen}{\textbf{(+0.2)}}  & 54.0 \textcolor{MidnightBlue}{\textbf{(+0.0)}} \\
     \textbf{PT+KL}   & \textbf{48.1} \textcolor{ForestGreen}{\textbf{(+0.1)}} & \textbf{47.9} \textcolor{ForestGreen}{\textbf{(+0.4)}} & \textbf{48.0} \textcolor{ForestGreen}{\textbf{(+0.2)}} & 55.9 \textcolor{red}{\textbf{(-0.2)}} & \textbf{53.7} \textcolor{ForestGreen}{\textbf{(+1.8)}} & \textbf{54.8} \textcolor{ForestGreen}{\textbf{(+0.8)}}\\
     \midrule
     \textsc{Alpaca-7B}   & 41.9 & 39.7 & 40.8 & 47.6 & 42.9 & 45.3 \\
     FT                 & 40.3 \textcolor{red}{\textbf{(-1.6)}} & 39.1 \textcolor{red}{\textbf{(-0.6)}} & 39.7 \textcolor{red}{\textbf{(-1.1)}} & 44.4 \textcolor{red}{\textbf{(-3.2)}} & 42.1 \textcolor{red}{\textbf{(-0.8)}} & 43.4 \textcolor{red}{\textbf{(-2.0)}} \\
     FT+KL              & 39.7 \textcolor{red}{\textbf{(-2.2)}} & 40.2 \textcolor{ForestGreen}{\textbf{(+0.5)}} & 40.0 \textcolor{red}{\textbf{(-0.8)}} & 45.6 \textcolor{red}{\textbf{(-2.0)}} & 42.8 \textcolor{red}{\textbf{(-0.1)}} & 44.2 \textcolor{red}{\textbf{(-1.1)}} \\
     PT                 & 42.1 \textcolor{ForestGreen}{\textbf{(+0.2)}} & 40.0 \textcolor{ForestGreen}{\textbf{(+0.3)}} & 41.1 \textcolor{ForestGreen}{\textbf{(+0.3)}} & 47.5 \textcolor{red}{\textbf{(-0.1)}} & 43.0 \textcolor{ForestGreen}{\textbf{(+0.1)}} & 45.3 \textcolor{MidnightBlue}{\textbf{(+0.0)}} \\
     \textbf{PT+KL}     & \textbf{42.4} \textcolor{ForestGreen}{\textbf{(+0.5)}} & \textbf{41.8} \textcolor{ForestGreen}{\textbf{(+2.1)}} & \textbf{42.1} \textcolor{ForestGreen}{\textbf{(+1.3)}} & \textbf{47.9} \textcolor{ForestGreen}{\textbf{(+0.3)}} & \textbf{46.6} \textcolor{ForestGreen}{\textbf{(+3.7)}} & \textbf{47.3} \textcolor{ForestGreen}{\textbf{(+2.0)}}\\
\bottomrule
\end{tabular}
\caption{Results and ablations of the proposed soft prompt alignment method. All ablated versions use the augmented set with automatically paraphrased instructions. FT refers to simply fine-tuning (with teacher-forcing) on this additional data; PT denotes prefix tuning (i.e., introducing soft prompt parameters); KL refers to the alignment objective that we proposed above. Using all of these components together yields the best performance, especially on unobserved instructions.}
\label{tab:alignment-results}
\end{table}

We report results in Table \ref{tab:alignment-results}. 
Two observations: (1) The proposed soft prompt alignment strategy ({\bf PT+KL}) yields consistent improvements across the tasks and models considered and especially improves performance on unobserved instructions, as anticipated. (2) The full benefit of the approach is realized only when all components---the additional automatically paraphrased training instructions, soft prompt parameters, and additional KL loss term---are in place.

\begin{table}[h]
\small
    \centering
    \begin{tabular}{l l l l}
    \toprule
        \textbf{Dataset} & \textbf{Closest Distance Before} & \textbf{Closest Distance After} & \textbf{$\Delta$ Acc. Improvement} (\%)\\
        \midrule
        \textsc{MMLU} & 22.2 & 21.3 & + 0.3\%\\
        \midrule
        \textsc{BBL QA} & 22.4 & 23.0 & + 0.4\% \\
        \textsc{BBL BC} & 30.1 & \textbf{27.9} & \textbf{+ 4.2\%} \\
        \textsc{BBL MC} & 26.0 & 24.6 & + 0.3\% \\
        \bottomrule
    \end{tabular}
    \caption{Average distances before and after soft prompt alignment with Flan-T5-XL.} %Ten observed and unobserved instructions with 50 instances for each dataset.} %For each unobserved instruction, its closest observed instruction is found by the average $\ell$2 distances across instances. We report the result of all the datasets in the Appendix}
    \vspace{-0.5em}
    \label{tab:improvement_distance}
\end{table}

Following our approach in \ref{section:mmlu_variance}, we take the average distance between observed and unobserved instructions before and after alignment. 
Table \ref{tab:improvement_distance} shows that our method brings observed and unobserved instruction representations closer together. 
The similarity is most increased in the case of the biggest accuracy gain, further suggesting the mechanism of improvement provided by soft prompt alignment. 
%Further, a large reduction of distance is observed in most of the binary classification datasets, which is also where the improvement is the most significant.

%\subsection{Discussion}

\section{Conclusions}
\label{section:conclusions}
\vspace{-.35em}

Instruction-tuned LLMs have emerged as a promising means of achieving zero-shot performance with smaller models that is competitive to, and sometimes even better than, that observed using much larger LLMs \cite{longpre2023flan,alpaca}. 
In this work we empirically characterized the \emph{robustness} of such models with respect to instruction rephrasings. 
In particular, we collected manually composed instructions from 36 graduate students in NLP across 75 tasks, and we evaluated different families of instruction-tuned LLMs (Flan, Alpaca, and T0) when provided observed and unobserved instructions (seen in training and not, respectively). %(i.e., seen in training) and unobserved (manual instructions we collected).
We found that using the latter consistently degrades model performance, indicating that models are unduly sensitive to instruction phrasings.

We then proposed a simple mechanism intended to improve the robustness of instruction-tuned LLMs.
This approach entails introducing an additional loss term that penalizes the model for inducing dissimilar distributions over output tokens when using (a) paraphrased instructions as opposed to (b) a reference instruction for the same task.
We found that training under this objective consistently (though modestly) improves results, and in particular mitigates the degradation observed when previously unobserved instructions are used. 

\section{Limitations}
\label{section:limitations}
\vspace{-.35em}

This work has important limitations: For example we only evaluated ``mid-sized'' models (<20B parameters), it is unclear if our findings would generalize to much larger instruction-tuned models. (However, we note that instruction tuning has been most promising for smaller models.) 
We also restricted our evaluation to three task types: QA and multi-class and binary classification.

\vspace{0.1em}
\noindent{\bf Ethics} This work does not have an explicit ethical dimension, but we acknowledge that all LLMs are likely to encode problematic biases; it is unclear how instruction-tuning might interact with these. 

\section{Acknowledgments}

This work was supported by the National Science Foundation (NSF) grant 1901117.

We thank Jay DeYoung and Alberto Mario Ceballos Arroyo for their advice and feedback on the paper. 
We also thank Alberto Mario Ceballos Arroyo, Arnab Sen Sharma, Bowen Zhao, Eric Todd, Hanming Li, Hiba Ahsan, Hye Sun Yun, Shulin Cao, Jay DeYoung, Jered McInerney, Ji Qi, Jifan Yu, Jize Jiang, Kaisheng Zeng, Koyena Pal, Kundan Krishna, Linxiao Nie, Hailong Jin, Jinxin Matthew Liu, Millicent Li, Monica Munnangi, Nikhil Prakash, Pouya Pezeshpour, Sanjana Ramprasad, Sarthak Jain, Shangqing Tu, Somin Wadhwa, Tingjian Zhang, Hao Wesley Peng, Xiaozhi Wang, Xingyu Lu, Xin Lv, Zijun Yao for providing manually written instructions.

\bibliographystyle{plain}
\bibliography{neurips_2023}

\appendix
\cleardoublepage
\addcontentsline{toc}{section}{Appendix} % Add the appendix text to the document TOC
\part{Appendix} % Start the appendix part
\parttoc % Insert the appendix TOC
\newpage
\section{Experimental Setup Details}
To ensure reproducibility, we %here 
provide all %the 
details regarding our evaluation of %to evaluate the 
%instruction 
the robustness of instruction-tuned LLMs. %different models.

\subsection{Evaluation Protocols}

%Occasionally, 
LLMs %(Especially for models tuned with general instructions like Alpaca) tend to 
sometimes generate %correct output in diverse forms 
outputs that are \emph{correct} but different from a (natural language) target. 
%when given O.O.D. instructions. Hence, 
Therefore, %to ensure fairness, 
we predict %the answer 
answers according to %by
``multiple-choice'' grading suggested by \textsc{Big-Bench}, by which we take the logits score and argmax over all the possible choices to obtain the prediction. 
In most cases, this approach yields the same accuracy as using exact match for evaluation. Here are the configurations for all the models we evaluated.

\begin{table}[hp]
    \centering
    \begin{tabular}{l c c c c c}
    \toprule
        \textbf{Models} & Node Type & Precision & Batch Size & Hours & $CO_2$ emission (KG) \\
        \midrule
        \textbf{Inference} & \\ [0.5ex]
        Flan-T5-Small & V100-SXM2-32G & FP16 & 128 & 64 & 4.0 \\ [0.1ex]
        Flan-T5-Base & V100-SXM2-32G & FP16 & 128 & 128 & 8.1 \\ [0.1ex]
        Flan-T5-Large & V100-SXM2-32G & FP16 & 32 & 256 & 16.2 \\ [0.1ex]
        Flan-T5-XL & V100-SXM2-32G & FP16 & 32 & 512 & 32.3 \\ [0.1ex]
        Flan-T5-XXL & RTX-A6000-46G & BF16 & 8 & 600 & 37.8 \\ [0.1ex]
        T0++ & RTX-A6000-46G & BF16 & 2 & 128 & 8.1 \\ [0.1ex]
        Alpaca-7B & A100-SXM4-80G & BF16 & 16 & 160 & 13.4 \\ [0.1ex]
        Alpaca-13B & A100-SXM4-80G & BF16 & 8 & 192 & 16.1 \\ [0.1ex]
        \midrule
        \textbf{Training} & \\ [0.5ex]
        Flan-T5-XL & A100-SXM4-80G & BF16 & 256 & 256 & 21.5 \\ [0.1ex]
        Alpaca-7B & A100-SXM4-80G & BF16 & 128 & 80 & 6.7 \\ [0.1ex]
        \midrule
        \multicolumn{3}{l}{\textbf{Estimated Total $CO_2$ Emission (KG)}} & \multicolumn{3}{c}{164.2} \\ 
        \bottomrule
    \end{tabular}
    \caption{The configurations for evaluating different instruction-tuned LMs. The $CO_2$ emission is estimated by \cite{lacoste2019quantifying}. The total emission is estimated to be equivalent to 679 Km driven by an average ICE car.}
    \label{tab:my_label}
\end{table}

\subsection{Hyperparameters}

We conduct all our training and ablation studies on 8 A100s with 80GB memory. We kept the KL-Loss weight  to $0.8$. We train both Flan-T5-XL and Alpaca-7B with a batch size of 4. The weight decay is set to be $1e-5$. The learning rate is $5e-4$ for the experiment.

\newpage
\section{Disaggregated Results}

\subsection{Main Results and Scaling Results}

In the main paper we reported aggregated results over benchmark corpora. 
Here we report %the result over all the
results on individual datasets for \textsc{BBL}. 
For \textsc{MMLU}, we evaluate all 57 datasets together, because these are all QA tasks (and we would want a QA model to be capable of answering questions across a diverse set of domains). %a challenging QA task requires knowledge in various domains. 
%The accuracy and standard deviation of the accuracies over all the instructions are reported. 
We report means and stadard deviations of the accuracies achieved over all instructions in Table \ref{tab:granular_mmlu}.
The numbers on the left of the setting suggest the number of instructions used. 
%To see the performance of each instruction, 
We also share even more granular results---reporting the performance for each instruction---in CSV files provided in the supplemental material.

\begin{table}[h!]
    \centering
    \setlength{\extrarowheight}{1mm}
    \begin{tabular}{l c c c c c}
    \toprule
        \multicolumn{6}{c}{\textsc{MMLU}} \\
        \hline
        \textbf{Model} & Flan-T5-XL & Flan-T5-XXL & T0pp-11B & Alpaca-7B & Alpaca-13B \\
        \midrule
        \textbf{MMLU} & \\
        \textsc{Observed}  & \textbf{48.1 ($\pm$ 0.3)} & \textbf{53.2 ($\pm$ 0.2)} & 48.3 ($\pm$ 0.9) & \textbf{41.9 ($\pm$ 0.6)} & \textbf{47.8 ($\pm$ 0.5)} \\
        \textsc{Unobserved} & 47.5 ($\pm$ 0.9) & 52.7 ($\pm$ 0.8) & \textbf{48.5 ($\pm$ 0.9)} & 39.7 ($\pm$ 2.2) & 47.0 ($\pm$ 0.8) \\
        \bottomrule
    \end{tabular}
    \caption{Granular results for Table \ref{tab:main_result} on each dataset of \textsc{MMLU} We treated all tasks in MMLU equally as general QA and computed the overall accuracy.}
    \label{tab:granular_mmlu}
\end{table}

\begin{table}[h!]
    \centering
    \setlength{\extrarowheight}{1mm}
    \begin{tabular}{l c c c c c}
    \toprule
        \multicolumn{6}{c}{\textsc{MMLU}} \\
        \hline
        \textbf{Size Variance} & Small (80M) & Base (250M) & Large (780M) & XL (3B) & XXL (11B) \\
        \midrule
        \textbf{MMLU} & \\
        \textsc{Observed} & 29.4 ($\pm$ 1.0) & \textbf{34.1 ($\pm$ 0.4)} & \textbf{41.1 ($\pm$ 0.2)} & \textbf{48.1 ($\pm$ 0.3)} & \textbf{53.2 ($\pm$ 0.2)} \\
        \textsc{Unobserved} & \textbf{29.6 ($\pm$ 0.9)} & 33.8 ($\pm$ 1.2) & 40.7 ($\pm$ 0.7) & 47.5 ($\pm$ 0.9) & 52.7 ($\pm$ 0.8) \\
        \bottomrule
    \end{tabular}
    \caption{Granular results for Figure \ref{fig:main_scaling_reesults} on each dataset of \textsc{MMLU} We treated all tasks in MMLU equally as general QA and computed the overall accuracy.}
    \label{tab:granular_mmlu}
\end{table}

\begin{table}[h!]
    \centering
    \setlength{\extrarowheight}{1mm}
    \begin{tabular}{l c c c c c }
    \toprule
        \multicolumn{6}{c}{\textsc{BBL-QA}} \\
        \hline
        \textbf{Model} & Flan-T5-XL & Flan-T5-XXL & T0pp-11B & Alpaca-7B & Alpaca-13B \\
        \midrule
        \textbf{BBQ Lite} & \\
        \textsc{Observed} & 66.5 ($\pm$ 1.5) & \textbf{77.4 ($\pm$ 2.4)} & \textbf{51.8 ($\pm$ 5.3)} & 32.6 ($\pm$ 1.0) & 43.5 ($\pm$ 1.4)  \\
        \textsc{Unobserved} & \textbf{67.0 ($\pm$ 7.0)} & 73.7 ($\pm$ 11.4) & 51.6 ($\pm$ 3.0) & \textbf{33.1 ($\pm$ 1.3)} & \textbf{45.5 ($\pm$ 2.9)} \\
        \midrule
        \textbf{Code Desc.} & \\
        \textsc{Observed} & \textbf{73.6 ($\pm$ 3.4)} & \textbf{83.6 ($\pm$ 1.7)} & 70.3 ($\pm$ 3.0) & \textbf{70.2 ($\pm$ 2.5)} & \textbf{85.2 ($\pm$ 2.4)}  \\
        \textsc{Unobserved} & 69.7 ($\pm$ 12.4) & 72.9 ($\pm$ 22.2) & \textbf{70.5 ($\pm$ 3.7)} & 67.5 ($\pm$ 11.3) & 82.2 ($\pm$ 8.5) \\
        \midrule
        \textbf{Hindu Know.} & \\
        \textsc{Observed} & \textbf{52.4 ($\pm$ 1.6)} & 53.9 ($\pm$ 1.8) & \textbf{57.1 ($\pm$ 2.5)} & 50.9 \textbf{($\pm$ 2.1)} & 63.8 ($\pm$ 0.7) \\
        \textsc{Unobserved} & 47.1 ($\pm$ 5.4) & \textbf{56.5 ($\pm$ 3.5)} & 53.2 ($\pm$ 3.0) & 49.8 ($\pm$ 5.1) & \textbf{63.9 ($\pm$ 1.1)} \\
        \midrule
        \textbf{Known Unk.} & \\
        \textsc{Observed} & \textbf{79.3 ($\pm$ 2.5)} & \textbf{84.7 ($\pm$ 2.1)} & 70.9 ($\pm$ 10.2) & \textbf{75.2 ($\pm$ 4.7)} & \textbf{81.9 ($\pm$ 4.3)}  \\
        \textsc{Unobserved}  & 69.0 ($\pm$ 6.7) & 80.6 ($\pm$ 8.1) & \textbf{76.1 ($\pm$ 5.9)} & 60.9 ($\pm$ 11.2) & 71.1 ($\pm$ 16.3) \\
        \midrule
        \textbf{Logical Ded.} & \\
        \textsc{Observed} & \textbf{52.5 ($\pm$ 1.0)} & \textbf{58.0 ($\pm$ 0.7)} & \textbf{45.5 ($\pm$ 0.8)} & \textbf{25.5 ($\pm$ 1.1)} & \textbf{29.2 ($\pm$ 1.3)} \\
        \textsc{Unobserved} & 52.1 ($\pm$ 1.1) & 57.8 ($\pm$ 0.6) & 45.3 ($\pm$ 1.2) & 24.5 ($\pm$ 2.3) & 28.0 ($\pm$ 1.6) \\
        \midrule
        \textbf{Novel Conc.} & \\
        \textsc{Observed} & 29.8 ($\pm$ 2.4) & \textbf{50.1 ($\pm$ 1.9)} & 28.8 ($\pm$ 2.9) & \textbf{37.2 ($\pm$ 5.2)} & \textbf{20.0 ($\pm$ 3.1)} \\
        \textsc{Unobserved} & \textbf{31.2 ($\pm$ 5.0)} & 46.0 ($\pm$ 5.4) & \textbf{31.5 ($\pm$ 5.1)} & 36.1 ($\pm$ 7.6) & 19.6 ($\pm$ 4.0) \\
        \midrule
        \textbf{Logic Grid} & \\
        \textsc{Observed} & \textbf{41.8 ($\pm$ 1.1)} & \textbf{43.2 ($\pm$ 1.8)} & \textbf{37.6 ($\pm$ 1.7)} & 24.4 ($\pm$ 1.9) & \textbf{29.3 ($\pm$ 0.8)} \\
        \textsc{Unobserved} & 38.6 ($\pm$ 5.4) & 39.6 ($\pm$ 4.4) & 36.4 ($\pm$ 3.6) & \textbf{25.4 ($\pm$ 1.1)} & 28.7 ($\pm$ 1.2) \\
        \bottomrule
        \textbf{Conc. Com.} & \\
        \textsc{Observed} & \textbf{75.9 ($\pm$ 1.8)} & \textbf{75.0 ($\pm$ 2.6)} & 73.3 ($\pm$ 2.3) & \textbf{58.7 ($\pm$ 4.0)} & \textbf{63.0 ($\pm$ 2.2)}\\
        \textsc{Unobserved} & 75.0 ($\pm$ 1.9) & 73.6 ($\pm$ 4.8) & \textbf{74.2 ($\pm$ 2.6)} & 55.9 ($\pm$ 6.2) & 61.1 ($\pm$ 3.3) \\
        \bottomrule
    \end{tabular}
    \caption{Granular results for Table \ref{tab:main_result} on each dataset of category \textsc{BBL-QA}}
    \label{tab:granular_qa}
\end{table}

\begin{table}[h!]
    \centering
    \setlength{\extrarowheight}{1mm}
    \begin{tabular}{l c c c c c }
    \toprule
        \multicolumn{6}{c}{\textsc{BBL-QA}} \\
        \hline
        \textbf{Size Variance} & Small (80M) & Base (250M) & Large (780M) & XL (3B) & XXL (11B) \\
        \midrule
        \textbf{BBQ Lite} & \\
        \textsc{Observed} & 28.3 ($\pm$ 1.3) & \textbf{51.5 ($\pm$ 1.4)} & 56.6 ($\pm$ 2.0) & 66.5 ($\pm$ 1.5) & \textbf{77.4 ($\pm$ 2.4)} \\
        \textsc{Unobserved} & \textbf{28.6 ($\pm$ 4.3)} & 50.5 ($\pm$ 4.1) & \textbf{56.7 ($\pm$ 4.7)} & \textbf{67.0 ($\pm$ 7.0)} & 73.7 ($\pm$ 11.4) \\
        \midrule
        \textbf{Code Desc.} & \\
        \textsc{Observed} & 22.0 ($\pm$ 4.0) & \textbf{55.7 ($\pm$ 3.3)} & \textbf{72.4 ($\pm$ 3.2)} & \textbf{73.6 ($\pm$ 3.4)} & \textbf{83.6 ($\pm$ 1.7)} \\
        \textsc{Unobserved} & \textbf{32.1 ($\pm$ 7.2)} & 48.6 ($\pm$ 7.2) & 63.3 ($\pm$ 14.2) & 69.7 ($\pm$ 12.4) & 72.9 ($\pm$ 22.2) \\
        \midrule
        \textbf{Hindu Know.} & \\
        \textsc{Observed} & 25.1 ($\pm$ 15.2) & \textbf{30.7 ($\pm$ 2.6)} & 34.9 ($\pm$ 0.9) & \textbf{52.4 ($\pm$ 1.6)} & 53.9 ($\pm$ 1.8) \\
        \textsc{Unobserved} & \textbf{31.6 ($\pm$ 10.7)} & 26.9 ($\pm$ 4.4) & \textbf{37.5 ($\pm$ 7.0)} & 47.1 ($\pm$ 5.4) & \textbf{56.5 ($\pm$ 3.5)} \\
        \midrule
        \textbf{Known Unk.} & \\
        \textsc{Observed} & 49.9 ($\pm$ 1.9) & \textbf{66.9 ($\pm$ 4.7)} & \textbf{76.2 ($\pm$ 3.5)} & \textbf{79.3 ($\pm$ 2.5)} & \textbf{84.7 ($\pm$ 2.1)}  \\
        \textsc{Unobserved}  & \textbf{52.8 ($\pm$ 5.2)} & 63.8 ($\pm$ 7.3) & 68.4 ($\pm$ 11.1) & 69.0 ($\pm$ 6.7) & 80.6 ($\pm$ 8.1) \\
        \midrule
        \textbf{Logical Ded.} & \\
        \textsc{Observed} & 19.8 ($\pm$ 0.7) & 27.1 ($\pm$ 1.3) & 45.9 ($\pm$ 1.1) & \textbf{52.5 ($\pm$ 1.0)} & \textbf{58.0 ($\pm$ 0.7)} \\
        \textsc{Unobserved} & \textbf{19.9 ($\pm$ 0.4)} & \textbf{28.9 ($\pm$ 2.8)} & \textbf{46.4 ($\pm$ 2.6)} & 52.1 ($\pm$ 1.1) & 57.8 ($\pm$ 0.6) \\
        \midrule
        \textbf{Novel Conc.} & \\
        \textsc{Observed} & \textbf{22.9 ($\pm$ 9.2)} & 15.9 ($\pm$ 5.4) & \textbf{31.0 ($\pm$ 2.3)} & 29.8 ($\pm$ 2.4) & \textbf{50.1 ($\pm$ 1.9)} \\
        \textsc{Unobserved} & 19.3 ($\pm$ 3.6) & \textbf{16.8 ($\pm$ 4.0)} & 28.4 ($\pm$ 6.9) & \textbf{31.2 ($\pm$ 5.0)} & 46.0 ($\pm$ 5.4) \\
        \midrule
        \textbf{Logic Grid} & \\
        \textsc{Observed} & 22.3 ($\pm$ 4.0) & \textbf{31.7 ($\pm$ 0.8)} & 32.6 ($\pm$ 2.1) & \textbf{41.8 ($\pm$ 1.1)} & \textbf{43.2 ($\pm$ 1.8)} \\
        \textsc{Unobserved} & \textbf{28.8 ($\pm$ 3.1)} & 29.4 ($\pm$ 5.1) & \textbf{34.1 ($\pm$ 2.8)} & 38.6 ($\pm$ 5.4) & 39.6 ($\pm$ 4.4) \\
        \bottomrule
        \textbf{Conc. Com.} & \\
        \textsc{Observed} & 30.4 ($\pm$ 10.6) & \textbf{55.6 ($\pm$ 5.1)} & \textbf{64.2 ($\pm$ 1.9)} & \textbf{75.9 ($\pm$ 1.8)} & \textbf{75.0 ($\pm$ 2.6)} \\
        \textsc{Unobserved} & \textbf{32.2 ($\pm$ 16.7)} & 54.5 ($\pm$ 9.4) & 58.1 ($\pm$ 11.6) & 75.0 ($\pm$ 1.9) & 73.6 ($\pm$ 4.8) \\
        \bottomrule
    \end{tabular}
    \caption{Granular results for Figure \ref{fig:main_scaling_reesults} on each dataset of category \textsc{BBL-QA}}
    \label{tab:granular_qa}
\end{table}

\begin{table}[h!]
    \centering
    \setlength{\extrarowheight}{1mm}
    \begin{tabular}{l c c c c c}
    \toprule
        \multicolumn{6}{c}{\textsc{BBL-BC}} \\
        \hline
        \textbf{Model} & Flan-T5-XL & Flan-T5-XXL & T0pp-11B & Alpaca-7B & Alpaca-13B \\
        \midrule
        \textbf{Play Dialog} & \\
        \textsc{Observed} & \textbf{61.6 ($\pm$ 5.8)} & 51.8 ($\pm$ 9.5) & \textbf{62.7 ($\pm$ 0.4)} & \textbf{45.0 ($\pm$ 2.0)} & \textbf{53.4 ($\pm$ 5.8)} \\
        \textsc{Unobserved} & 53.0 ($\pm$ 6.9) & \textbf{58.1 ($\pm$ 4.4)} & 55.2 ($\pm$ 8.1) & 42.9 ($\pm$ 7.9) & 42.9 ($\pm$ 8.8) \\
        \midrule
        \textbf{Strat. QA} & \\
        \textsc{Observed} & 58.7 ($\pm$ 3.3) & \textbf{64.2 ($\pm$ 3.0)} & 51.0 ($\pm$ 1.8) & 53.0 ($\pm$ 2.1) & 56.7 ($\pm$ 3.8) \\
        \textsc{Unobserved} & \textbf{60.7 ($\pm$ 7.5)} & 59.3 ($\pm$ 6.1) & \textbf{54.5 ($\pm$ 0.9)} & \textbf{53.3 ($\pm$ 4.1)} & \textbf{61.0 ($\pm$ 1.9)} \\
        \midrule
        \textbf{Strange St.} & \\
        \textsc{Observed} & 69.3 ($\pm$ 4.4) & 71.0 ($\pm$ 7.3) & \textbf{51.2 ($\pm$ 5.1)} & \textbf{67.0 ($\pm$ 4.7)} & \textbf{69.8 ($\pm$ 5.0)}\\
        \textsc{Unobserved} & \textbf{70.5 ($\pm$ 7.0)} & \textbf{77.4 ($\pm$ 6.1)} & 48.4 ($\pm$ 3.1) & 59.9 ($\pm$ 9.4) & 57.5 ($\pm$ 5.6)\\
        \midrule
        \textbf{Winowhy} & \\
        \textsc{Observed} & \textbf{76.5 ($\pm$ 1.9)} & \textbf{75.6 ($\pm$ 4.0)} & \textbf{99.6 ($\pm$ 1.0)} & 50.1 ($\pm$ 4.8) & 51.9 ($\pm$ 4.6)\\
        \textsc{Unobserved} & 60.2 ($\pm$ 6.2) & 59.7 ($\pm$ 7.7) & 60.9 ($\pm$ 5.1) & \textbf{53.4 ($\pm$ 4.6)} & \textbf{55.2 ($\pm$ 6.0)} \\
        \bottomrule
    \end{tabular}
    \caption{Granular results for Table \ref{tab:main_result} on each dataset of category \textsc{BBL-BC}}
    \label{tab:granular_bc_main}
\end{table}

\begin{table}[h!]
    \centering
    \setlength{\extrarowheight}{1mm}
    \begin{tabular}{l c c c c c}
    \toprule
        \multicolumn{6}{c}{\textsc{BBL-BC}} \\
        \hline
        \textbf{Size Variance} & Small (80M) & Base (250M) & Large (780M) & XL (3B) & XXL (11B) \\
        \midrule
        \textbf{Play Dialog} & \\
        \textsc{Observed} & 51.6 ($\pm$ 13.3) & 54.6 ($\pm$ 10.7) & \textbf{59.0 ($\pm$ 6.7)} & \textbf{61.6 ($\pm$ 5.8)} & 51.8 ($\pm$ 9.5) \\
        \textsc{Unobserved} & \textbf{61.6 ($\pm$ 4.6)} & \textbf{56.3 ($\pm$ 10.9)} & 57.3 ($\pm$ 7.8) & 53.0 ($\pm$ 6.9) & \textbf{58.1 ($\pm$ 4.4)} \\
        \midrule
        \textbf{Strat. QA} & \\
        \textsc{Observed} & \textbf{52.3 ($\pm$ 1.0)} & 48.9 ($\pm$ 2.1) & \textbf{60.9 ($\pm$ 1.3)} & 58.7 ($\pm$ 3.3) & \textbf{64.2 ($\pm$ 3.0)} \\
        \textsc{Unobserved} & 51.5 ($\pm$ 2.7) & \textbf{52.9 ($\pm$ 1.3)} & 53.9 ($\pm$ 3.8) & \textbf{60.7 ($\pm$ 7.5)} & 59.3 ($\pm$ 6.1) \\
        \midrule
        \textbf{Strange St.} & \\
        \textsc{Observed} & 41.3 ($\pm$ 10.3) & \textbf{43.1 ($\pm$ 4.2)} & 54.4 ($\pm$ 1.2) & 69.3 ($\pm$ 4.4) & 71.0 ($\pm$ 7.3) \\
        \textsc{Unobserved} & \textbf{55.9 ($\pm$ 18.5)} & 42.0 ($\pm$ 5.5) & \textbf{67.9 ($\pm$ 8.0)} & \textbf{70.5 ($\pm$ 7.0)} & \textbf{77.4 ($\pm$ 6.1)} \\
        \midrule
        \textbf{Winowhy} & \\
        \textsc{Observed} & \textbf{54.8 ($\pm$ 1.6)} & 55.9 ($\pm$ 7.6) & \textbf{60.4 ($\pm$ 9.8)} & \textbf{76.5 ($\pm$ 1.9)} & \textbf{75.6 ($\pm$ 4.0)} \\
        \textsc{Unobserved} & 53.7 ($\pm$ 5.1) & \textbf{57.1 ($\pm$ 6.7)} & 53.5 ($\pm$ 4.9) & 60.2 ($\pm$ 6.2) & 59.7 ($\pm$ 7.7) \\
        \bottomrule
    \end{tabular}
    \caption{Granular results for Figure \ref{fig:main_scaling_reesults} on each dataset of category \textsc{BBL-BC}}
    \label{tab:granular_bc_scale}
\end{table}

\begin{table}[h!]
    \centering
    \setlength{\extrarowheight}{1mm}
    \begin{tabular}{l c c c c c }
    \toprule
        \multicolumn{6}{c}{\textsc{BBL-MC}} \\
        \hline
        \textbf{Model} & Flan-T5-XL & Flan-T5-XXL & T0pp-11B & Alpaca-7B & Alpaca-13B \\
        \textbf{Language ID} & \\
        \textsc{Observed} & \textbf{32.6 ($\pm$ 0.2)} & \textbf{38.9 ($\pm$ 0.3)} & \textbf{15.7 ($\pm$ 3.0)} & 12.9 ($\pm$ 0.7) & 18.5 ($\pm$ 0.7)\\
        \textsc{Unobserved} & 25.5 ($\pm$ 7.3) & 31.6 ($\pm$ 9.4) & 14.3 ($\pm$ 2.4) & \textbf{14.7 ($\pm$ 1.7)} & \textbf{21.7 ($\pm$ 0.7)} \\
        \midrule
        \textbf{Vitamin C} & \\
        \textsc{Observed} & 78.6 ($\pm$ 1.1) & 78.5 ($\pm$ 0.7) & 68.3 ($\pm$ 1.1) & \textbf{51.4 ($\pm$ 3.6)} & \textbf{54.9 ($\pm$ 2.9)}\\
        \textsc{Unobserved} & 78.6 ($\pm$ 3.6) & \textbf{80.2 ($\pm$ 1.6)} & \textbf{68.5 ($\pm$ 2.4)} & 18.1 ($\pm$ 5.3) & 23.6 ($\pm$ 14.2)\\
        \bottomrule
    \end{tabular}
    \caption{Granular results for Table \ref{tab:main_result} on each dataset of category \textsc{BBL-MC}}
    \label{tab:granular_mc_main}
\end{table}

\begin{table}[h!]
    \centering
    \setlength{\extrarowheight}{1mm}
    \begin{tabular}{l c c c c c }
    \toprule
        \multicolumn{6}{c}{\textsc{BBL-MC}} \\
        \hline
        \textbf{Size Variance} & Small (80M) & Base (250M) & Large (780M) & XL (3B) & XXL (11B) \\
        \textbf{Language ID} & \\
        \textsc{Observed} & \textbf{11.9 ($\pm$ 0.2)} & \textbf{17.0 ($\pm$ 0.3)} & \textbf{25.8 ($\pm$ 0.3)} & \textbf{32.6 ($\pm$ 0.2)} & \textbf{38.9 ($\pm$ 0.3)} \\
        \textsc{Unobserved} & 9.5 ($\pm$ 0.2) & 12.4 ($\pm$ 1.5) & 19.2 ($\pm$ 4.4) & 25.5 ($\pm$ 7.3) & 31.6 ($\pm$ 9.4) \\
        \midrule
        \textbf{Vitamin C} & \\
        \textsc{Observed} & \textbf{46.6 ($\pm$ 4.0)} & 60.7 ($\pm$ 5.6) & \textbf{72.6 ($\pm$ 1.5)} & 78.6 ($\pm$ 1.1) & 78.5 ($\pm$ 0.7) \\
        \textsc{Unobserved} & 40.8 ($\pm$ 4.2) & \textbf{63.0 ($\pm$ 4.6)} & 36.4 ($\pm$ 0.8) & 78.6 ($\pm$ 3.6) & \textbf{80.2 ($\pm$ 1.6)} \\
        \bottomrule
    \end{tabular}
    \caption{Granular results for Figure \ref{fig:main_scaling_reesults} on each dataset of category \textsc{BBL-MC}}
    \label{tab:granular_mc_scale}
\end{table}

\clearpage
\subsection{"Closer Look" Experiment Results}

Here, we provide the detailed results that we reported in \ref{fig:adversarial}
\begin{table}[h!]
    \centering
    \begin{tabular}{c c c c c c c}
    \toprule
        \multirow{2}{*}{\textbf{Dataset}} & \multicolumn{2}{c}{\textbf{Observed}} & \multicolumn{2}{c}{\textbf{Unobserved}} & \multicolumn{2}{c}{\textbf{Control}} \\
         & Closest & Incorrect & Collected & Task Designer & Negated & Nonsensical \\
        \midrule
        Intent & 93.6 & 93.1 & 94.1 & \textbf{94.66} & 28.0 & 40.7 \\
        Recognition & $\pm 0.3$ & $\pm 1.0$ & $\pm 0.6$ & - & $\pm 6.5$ & $\pm 7.2$ \\
        \midrule
        Empirical & 39.2 & \textbf{41.62} & 37.6 & 37.4 & 28.1 & 30.9 \\
        Judgments & $\pm 0.8$ & $\pm 6.3$ & $\pm 1.7$ & - & $\pm 3.5$ & $\pm 2.5$ \\
        \midrule
        Conceptual & 78.0 & \textbf{78.92} & 75.3 & 58.3 & 11.2 & 63.7 \\
        Combinations & $\pm 1.6$ & $\pm 0.5$ & $\pm 3.3$ & - & $\pm 2.1$ & $\pm 2.6$ \\
        \midrule
        Language & \textbf{38.94} & 29.3 & 28.8 & 27.6 & 36.9 & 12.4 \\
        Identification & $\pm 0.3$ & $\pm 5.0$ & $\pm 5.8$ & - & $\pm 0.5$ & $\pm 0.5$ \\
        \midrule
        Logical & \textbf{56.92} & 49.4 & 52.8 & 53.8 & 11.8 & 34.4 \\
        Sequence & $\pm 6.6$ & $\pm 6.1$ & $\pm 5.3$ & - & $\pm 6.9$ & $\pm 5.9$ \\
        \midrule 
        Crash & 53.6 & 50.0 & 50.5 & \textbf{63.16} & 28.6 & 43.7 \\
        Blossom & $\pm 2.8$ & $\pm 5.3$ & $\pm 2.2$ & - & $\pm 6.2$ & $\pm 1.4$ \\
        \midrule 
        Epistemic & 62.8 & 59.3 & 58.1 & 60.2 & \textbf{65.49} & 49.5 \\
        Reasoning & $\pm 2.9$ & $\pm 3.4$ & $\pm 1.7$ & - & $\pm 4.6$ & $\pm 1.3$ \\
        \midrule \midrule
        \multirow{2}{*}{Overall} & \textbf{60.45} & 57.4 & 56.8 & 56.4 & 30.0 & 39.3 \\
         & $\pm 2.1$ & $\pm 2.2$ & $\pm 1.8$ & - & $\pm 2.2$ & $\pm 2.3$ \\
         \bottomrule
    \end{tabular}
    \caption{The detailed results of "A Closer Look" experiment. We provide all the instructions picked and their sources in Section \ref{section:granular_instructions}. In could be seen that the "Incorrect" but observed instruction, in most cases, outperform the correct but unobserved instructions ("Collected" and "Task Designer").}
    \label{tab:closer_look}
\end{table}
\newpage
\section{Instruction Robustness with In-context Learning}
\label{section:icl_robustness}

We have been focused on zero-shot settings, but here we also report results achieved under In-context Learning (ICL). 
We consider one-shot ICL for Flan-T5 because its context window limit precludes providing additional shots in context.
%We consider the effect of
%Due to the %window size 
%context window limit of Flan-T5, w
We repeat the experiments from the main paper with the one-shot ICL below. 
%We remove the first data from every dataset as the in-context example. 
We (arbitrarily) take the first instance from every dataset to use as the shot. It could be seen from the result that the performance gap between observed and unobserved instructions is significantly narrowed down with the in-context example provided.
%Flan-T5 is explicitly instruction-tuned with few-shots instruction, we think that it possess the basic in-context learning ability.

\begin{table}[h!]
    \centering
    \setlength{\extrarowheight}{1mm}
    \begin{tabular}{l c c c c c}
    \toprule
        \multicolumn{6}{c}{\textsc{MMLU} \textbf{(One-shot)}} \\
        \hline
        \textbf{Flan-T5} & Small (80M) & Base (250M) & Large (780M) & XL (3B) & XXL (11B) \\
        \midrule
        \textbf{MMLU} & \\
        \textsc{Observed} & 29.3 ($\pm$ 0.9) & \textbf{33.9 ($\pm$ 0.5)} & \textbf{40.7 ($\pm$ 0.2)} & \textbf{47.5 ($\pm$ 0.2)} & 52.7 ($\pm$ 0.2)\\
        \textsc{Unobserved} & \textbf{29.6 ($\pm$ 0.6)} & 33.8 ($\pm$ 1.0) & 40.4 ($\pm$ 0.9) & 47.5 ($\pm$ 0.7) & \textbf{52.8 ($\pm$ 1.0)} \\
        \bottomrule
    \end{tabular}
    \caption{Granular results on each dataset of \textsc{MMLU} for Flan-T5 models and one-shot ICL. We group all QA tasks in \textsc{MMLU} and report overall accuracy on these.}%We treated all tasks in MMLU equally as general QA and computed the overall accuracy.}
    \label{tab:mmlu_icl}
\end{table}

\begin{table}[h!]
    \centering
    \setlength{\extrarowheight}{1mm}
    \begin{tabular}{l c c c c c }
    \toprule
        \multicolumn{6}{c}{\textsc{BBL-QA} \textbf{(One-shot)}} \\
        \hline
        \textbf{Flan-T5} & Small (80M) & Base (250M) & Large (780M) & XL (3B) & XXL (11B) \\
        \midrule
        \textbf{BBQ Lite} & \\
        \textsc{Observed} & \textbf{29.6 ($\pm$ 2.2)} & 50.5 ($\pm$ 1.7) & 57.0 ($\pm$ 2.0) & 66.7 ($\pm$ 1.6) & 77.2 ($\pm$ 2.7) \\
        \textsc{Unobserved} & 28.9 ($\pm$ 1.4) & \textbf{51.0 ($\pm$ 3.6)} & \textbf{58.0 ($\pm$ 2.7)} & \textbf{69.0 ($\pm$ 5.9)} & \textbf{77.6 ($\pm$ 5.5)} \\
        \midrule
        \textbf{Code Desc.} & \\
        \textsc{Observed} & 20.5 ($\pm$ 3.1) & \textbf{56.9 ($\pm$ 5.1)} & \textbf{76.0 ($\pm$ 2.2)} & \textbf{73.6 ($\pm$ 1.8)} & \textbf{85.3 ($\pm$ 1.6)} \\
        \textsc{Unobserved} & \textbf{25.6 ($\pm$ 8.7)} & 43.6 ($\pm$ 8.4) & 53.5 ($\pm$ 16.0) & 65.8 ($\pm$ 15.3) & 75.0 ($\pm$ 12.3) \\
        \midrule
        \textbf{Hindu Know.} & \\
        \textsc{Observed} & \textbf{24.5 ($\pm$ 12.7)} & \textbf{30.5 ($\pm$ 3.6)} & 36.2 ($\pm$ 1.7) & \textbf{50.9 ($\pm$ 1.3)} & 52.9 ($\pm$ 1.9) \\
        \textsc{Unobserved} & 23.0 ($\pm$ 5.8) & 22.4 ($\pm$ 5.8) & \textbf{37.7 ($\pm$ 4.7)} & 50.2 ($\pm$ 5.8) & \textbf{54.6 ($\pm$ 3.2)} \\
        \midrule
        \textbf{Known Unk.} & \\
        \textsc{Observed} & \textbf{49.2 ($\pm$ 3.8)} & \textbf{66.7 ($\pm$ 8.3)} & \textbf{73.6 ($\pm$ 2.7)} & \textbf{76.3 ($\pm$ 2.0)} & \textbf{84.7 ($\pm$ 3.8)} \\
        \textsc{Unobserved}  & 49.1 ($\pm$ 4.1) & 60.2 ($\pm$ 7.3) & 67.7 ($\pm$ 12.9) & 63.7 ($\pm$ 9.8) & 76.0 ($\pm$ 12.7) \\
        \midrule
        \textbf{Logical Ded.} & \\
        \textsc{Observed} & 20.2 ($\pm$ 0.4) & 26.8 ($\pm$ 1.0) & \textbf{45.9 ($\pm$ 1.1)} & \textbf{53.0 ($\pm$ 0.7)} & \textbf{58.2 ($\pm$ 0.5)} \\
        \textsc{Unobserved} & \textbf{20.2 ($\pm$ 0.8)} & \textbf{27.8 ($\pm$ 3.5)} & 44.3 ($\pm$ 8.6) & 48.6 ($\pm$ 10.1) & 55.0 ($\pm$ 10.5) \\
        \midrule
        \textbf{Novel Conc.} & \\
        \textsc{Observed} & \textbf{21.3 ($\pm$ 4.3)} & \textbf{14.8 ($\pm$ 4.9)} & \textbf{29.1 ($\pm$ 4.0)} & 31.2 ($\pm$ 2.0) & \textbf{47.7 ($\pm$ 3.1)} \\
        \textsc{Unobserved} & 17.2 ($\pm$ 6.8) & 14.7 ($\pm$ 9.0) & 24.7 ($\pm$ 6.2) & \textbf{35.1 ($\pm$ 5.2)} & 42.5 ($\pm$ 9.0) \\
        \bottomrule
        \textbf{Conc. Com.} & \\
        \textsc{Observed} & 28.0 ($\pm$ 3.6) & \textbf{38.5 ($\pm$ 2.3)} & \textbf{65.0 ($\pm$ 1.8)} & \textbf{77.7 ($\pm$ 1.6)} & \textbf{77.1 ($\pm$ 2.2)} \\
        \textsc{Unobserved} & \textbf{28.6 ($\pm$ 4.5)} & 36.3 ($\pm$ 6.4) & 58.2 ($\pm$ 10.7) & 74.8 ($\pm$ 11.9) & 75.2 ($\pm$ 2.8) \\
        \bottomrule
    \end{tabular}
    \caption{Granular results on each dataset of category \textsc{BBL-QA} with one-shot in-context learning.}
    \label{tab:qa_icl}
\end{table}

\begin{table}[h!]
    \centering
    \setlength{\extrarowheight}{1mm}
    \begin{tabular}{l c c c c c}
    \toprule
        \multicolumn{6}{c}{\textsc{BBL-BC} \textbf{(One-shot)}} \\
        \hline
        \textbf{Flan-T5} & Small (80M) & Base (250M) & Large (780M) & XL (3B) & XXL (11B) \\
        \midrule
        \textbf{Play Dialog} & \\
        \textsc{Observed} & 49.8 ($\pm$ 11.9) & \textbf{54.4 ($\pm$ 10.3)} & \textbf{58.1 ($\pm$ 5.7)} & \textbf{57.8 ($\pm$ 3.3)} & 43.9 ($\pm$ 4.2) \\
        \textsc{Unobserved} & \textbf{55.8 ($\pm$ 10.2)} & 52.5 ($\pm$ 10.7) & 48.6 ($\pm$ 8.7) & 46.3 ($\pm$ 3.9) & \textbf{51.3 ($\pm$ 3.3)} \\
        \midrule
        \textbf{Strat. QA} & \\
        \textsc{Observed} & 52.5 ($\pm$ 0.8) & 49.3 ($\pm$ 3.1) & \textbf{60.6 ($\pm$ 1.5)} & 60.6 ($\pm$ 6.2) & \textbf{66.2 ($\pm$ 2.6)} \\
        \textsc{Unobserved} & \textbf{53.2 ($\pm$ 0.0)} & \textbf{53.3 ($\pm$ 0.8)} & 55.9 ($\pm$ 4.3) & \textbf{61.2 ($\pm$ 6.0)} & 62.2 ($\pm$ 5.5) \\
        \midrule
        \textbf{Strange St.} & \\
        \textsc{Observed} & 40.7 ($\pm$ 12.3) & \textbf{41.8 ($\pm$ 2.3)} & 51.7 ($\pm$ 1.6) & 74.4 ($\pm$ 3.6) & 78.5 ($\pm$ 2.1)\\
        \textsc{Unobserved} & \textbf{46.7 ($\pm$ 5.1)} & 37.9 ($\pm$ 8.6) & \textbf{56.3 ($\pm$ 3.0)} & \textbf{78.7 ($\pm$ 3.2)} & \textbf{83.2 ($\pm$ 7.6)} \\
        \midrule
        \textbf{Winowhy} & \\
        \textsc{Observed} & \textbf{52.7 ($\pm$ 2.8)} & 57.3 ($\pm$ 6.2) & \textbf{62.1 ($\pm$ 5.9)} & \textbf{77.2 ($\pm$ 0.6)} & \textbf{76.7 ($\pm$ 1.0)} \\
        \textsc{Unobserved} & 48.1 ($\pm$ 4.2) & \textbf{58.3 ($\pm$ 8.1)} & 57.1 ($\pm$ 8.8) & 66.3 ($\pm$ 8.8) & 65.5 ($\pm$ 9.9) \\
        \bottomrule
    \end{tabular}
    \caption{Granular results on each dataset of category \textsc{BBL-BC} with one-shot in-context learning.}
    \label{tab:bc_icl}
\end{table}

\begin{table}[h!]
    \centering
    \setlength{\extrarowheight}{1mm}
    \begin{tabular}{l c c c c c }
    \toprule
        \multicolumn{6}{c}{\textsc{BBL-MC} \textbf{(One-shot)}} \\
        \hline
        \textbf{Flan-T5} & Small (80M) & Base (250M) & Large (780M) & XL (3B) & XXL (11B) \\
        \textbf{Language ID} & \\
        \textsc{Observed} & \textbf{11.7 ($\pm$ 0.2)} & \textbf{13.5 ($\pm$ 2.8)} & \textbf{25.6 ($\pm$ 0.4)} & \textbf{31.8 ($\pm$ 0.3)} & \textbf{38.7 ($\pm$ 0.5)}\\
        \textsc{Unobserved} & 9.7 ($\pm$ 0.9) & 11.6 ($\pm$ 1.5) & 16.9 ($\pm$ 5.4) & 20.3 ($\pm$ 7.5) & 28.4 ($\pm$ 10.3) \\
        \midrule
        \textbf{Vitamin C} & \\
        \textsc{Observed} & \textbf{50.9 ($\pm$ 1.6)} & 60.9 ($\pm$ 5.4) & \textbf{73.3 ($\pm$ 0.8)} & \textbf{78.6 ($\pm$ 1.4)} & 80.4 ($\pm$ 0.5) \\
        \textsc{Unobserved} & 50.1 ($\pm$ 0.9) & \textbf{64.5 ($\pm$ 1.8)} & 73.2 ($\pm$ 1.5) & 77.5 ($\pm$ 9.3) & \textbf{80.7 ($\pm$ 3.6)} \\
        \bottomrule
    \end{tabular}
    \caption{Granular results on each dataset of category \textsc{BBL-MC} with one-shot in-context learning.}
    \label{tab:granular_mc_scale}
\end{table}
\newpage
\section{Representational Similarity and Model Performance}
\label{section:embeddings}

We re-generate the visualization in Figure \ref{fig:embeddings} for all T5 model sizes in Figure \label{fig:tSNE-appendix}. The qualitative result---representations of tokens following observed instructions are generally dissimilar from those following unobserved instructions---remains largely consistent, although is less pronounced, e.g., for XL (in particular here, the \textsc{MMLU} samples are not as entangled as we might expect).

\begin{figure}
    \centering
    \includegraphics[scale=0.25]{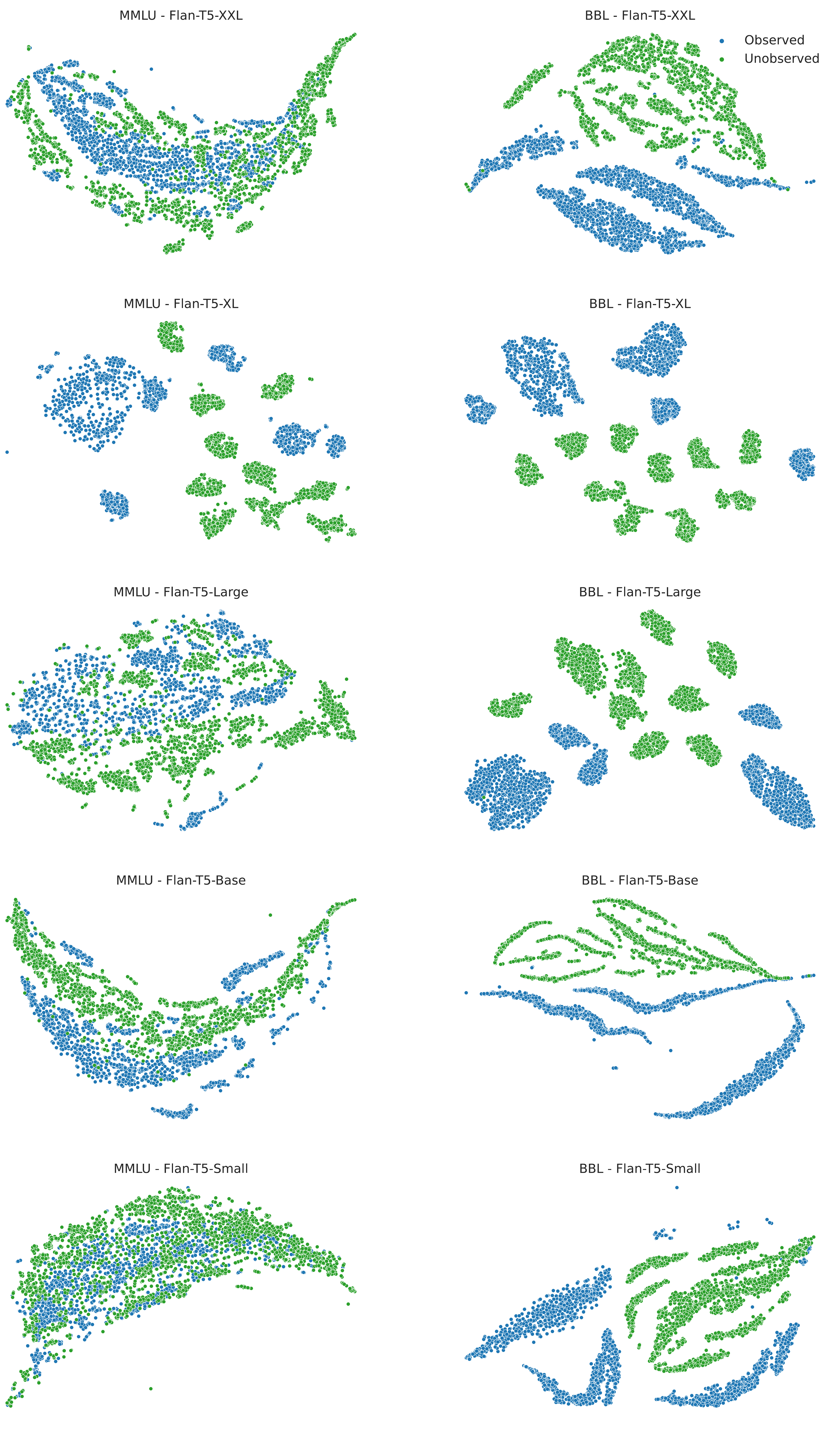}
    \caption{We reproduce Figure \ref{fig:embeddings} from the main paper over all T5 model sizes. }
    \label{fig:tSNE-appendix}
\end{figure}

\clearpage
\section{Instruction Collection}

In this section, we report in detail how we collected the instructions we used to evaluate the out of domain (O.O.D.) robustness of instruction-tuned LLMs.

\subsection{Observed Instructions} 

We manually review the instruction-tuning collections that were used to train the three instruction-tuned LLMs that we evaluated: Flan \cite{chung2022scaling}, Alpaca \cite{alpaca}, and P3 \cite{sanh2021multitask}. %are evaluating. 
%For %three types of tasks 
%task types (QA, Binary Classification, Multi-Class Classification), we collect %the 
%instructions in the collection that are general enough to ``mix-and-match'' for a given one of these tasks. %these tasks. 
We consider instructions that are sufficiently general to be able to ``mix-and-match'' for one of the three task types: QA, Binary Classification, Multiclass Classification. 
%with the same type of tasks that we are evaluating. 
%We collect observed instructions over three major instruction-tuning dataset: Flan \cite{chung2022scaling}, Alpaca \cite{alpaca}, and P3 \cite{sanh2021multitask}. 
%Here, 

Below we provide instruction templates %of 
for all of the (observed) instructions %we collected 
we aggregated, and their source in the collections:

\subsubsection*{Flan}

For simplicity purposes, we provide the task and indices for the observed instructions templates we aggregated from the Flan collection. The exact code we used to process the data may be found in the publically available Flan repository \footnote{https://github.com/google-research/FLAN} \\

\paragraph{QA - 01} Source: NIV2 - Task 73 - Template 1 \\ [1ex]
\textbf{Input}: \textcolor{MidnightBlue}{\{question\}}, \textcolor{ForestGreen}{\{options\}} \\ [0.5ex]
\textbf{Template}: \\ [0.5ex]
You are given a question and some answer options (associated with "A", "B", "C", "D"). You should choose the correct answer based on commonsense knowledge. Avoid answering questions based on associations, the set of answers are chosen deliberately to capture common sense beyond associations. Do not generate anything else apart from one of the following characters: \textcolor{ForestGreen}{\{options letter\}} and only give one answer for each question. \\  \\ \textcolor{MidnightBlue}{\{question\}} \textcolor{ForestGreen}{\{options\}}

\paragraph{QA - 02} Source: NIV2 - Task 73 - Template 2 \\ [1ex]
\textbf{Input}: \textcolor{MidnightBlue}{\{question\}}, \textcolor{ForestGreen}{\{options\}} \\ [0.5ex]
\textbf{Template}: \\ [0.5ex]
You will be given a definition of a task first, then some input of the task. \\ You are given a question and some answer options (associated with "A", "B", "C", "D"). You should choose the correct answer based on commonsense knowledge. Avoid answering questions based on associations, the set of answers are chosen deliberately to capture common sense beyond associations. Do not generate anything else apart from one of the following characters: \textcolor{ForestGreen}{\{options letter\}} and only give one answer for each question. \\  \\ \textcolor{MidnightBlue}{\{question\}} \textcolor{ForestGreen}{\{options\}} \\ Output:

\paragraph{QA - 03} Source: NIV2 - Task 73 - Template 3 \\ [1ex]
\textbf{Input}: \textcolor{MidnightBlue}{\{question\}}, \textcolor{ForestGreen}{\{options\}} \\ [0.5ex]
\textbf{Template}: \\ [0.5ex]
Definition: You are given a question and some answer options (associated with "A", "B", "C", "D"). You should choose the correct answer based on commonsense knowledge. Avoid answering questions based on associations, the set of answers are chosen deliberately to capture common sense beyond associations. Do not generate anything else apart from one of the following characters: \textcolor{ForestGreen}{\{options letter\}} and only give one answer for each question. \\ Input: \textcolor{MidnightBlue}{\{question\}} \textcolor{ForestGreen}{\{options\}} \\ Output:

\paragraph{QA - 04} Source: NIV2 - Task 73 - Template 4 \\ [1ex]
\textbf{Input}: \textcolor{MidnightBlue}{\{question\}}, \textcolor{ForestGreen}{\{options\}} \\ [0.5ex]
\textbf{Template}: \\ [0.5ex]
Instructions: You are given a question and some answer options (associated with "A", "B", "C", "D"). You should choose the correct answer based on commonsense knowledge. Avoid answering questions based on associations, the set of answers are chosen deliberately to capture common sense beyond associations. Do not generate anything else apart from one of the following characters: \textcolor{ForestGreen}{\{options letter\}} and only give one answer for each question. \\ Input: \textcolor{MidnightBlue}{\{question\}} \textcolor{ForestGreen}{\{options\}} \\ Output:

\paragraph{QA - 05} Source: NIV2 - Task 73 - Template 5 \\ [1ex]
\textbf{Input}: \textcolor{MidnightBlue}{\{question\}}, \textcolor{ForestGreen}{\{options\}} \\ [0.5ex]
\textbf{Template}: \\ [0.5ex]
You are given a question and some answer options (associated with "A", "B", "C", "D"). You should choose the correct answer based on commonsense knowledge. Avoid answering questions based on associations, the set of answers are chosen deliberately to capture common sense beyond associations. Do not generate anything else apart from one of the following characters: \textcolor{ForestGreen}{\{options letter\}} and only give one answer for each question. \\ Q: \textcolor{MidnightBlue}{\{question\}} \textcolor{ForestGreen}{\{options\}} \\ A:

\paragraph{QA - 06} Source: NIV2 - Task 73 - Template 6 \\ [1ex]
\textbf{Input}: \textcolor{MidnightBlue}{\{question\}}, \textcolor{ForestGreen}{\{options\}} \\ [0.5ex]
\textbf{Template}: \\ [0.5ex]
Given the task definition and input, reply with output. You are given a question and some answer options (associated with "A", "B", "C", "D"). You should choose the correct answer based on commonsense knowledge. Avoid answering questions based on associations, the set of answers are chosen deliberately to capture common sense beyond associations. Do not generate anything else apart from one of the following characters: \textcolor{ForestGreen}{\{options letter\}} and only give one answer for each question. \\  \\ \textcolor{MidnightBlue}{\{question\}} \textcolor{ForestGreen}{\{options\}} \\

\paragraph{QA - 07} Source: NIV2 - Task 73 - Template 7 \\ [1ex]
\textbf{Input}: \textcolor{MidnightBlue}{\{question\}}, \textcolor{ForestGreen}{\{options\}} \\ [0.5ex]
\textbf{Template}: \\ [0.5ex]
Teacher: You are given a question and some answer options (associated with "A", "B", "C", "D"). You should choose the correct answer based on commonsense knowledge. Avoid answering questions based on associations, the set of answers are chosen deliberately to capture common sense beyond associations. Do not generate anything else apart from one of the following characters: \textcolor{ForestGreen}{\{options letter\}} and only give one answer for each question. \\ Teacher: Now, understand the problem? Solve this instance: \textcolor{MidnightBlue}{\{question\}} \textcolor{ForestGreen}{\{options\}} \\ Student:

\paragraph{QA - 08} Source: NIV2 - Task 73 - Template 8 \\ [1ex]
\textbf{Input}: \textcolor{MidnightBlue}{\{question\}}, \textcolor{ForestGreen}{\{options\}} \\ [0.5ex]
\textbf{Template}: \\ [0.5ex]
Q: You are given a question and some answer options (associated with "A", "B", "C", "D"). You should choose the correct answer based on commonsense knowledge. Avoid answering questions based on associations, the set of answers are chosen deliberately to capture common sense beyond associations. Do not generate anything else apart from one of the following characters: \textcolor{ForestGreen}{\{options letter\}} and only give one answer for each question. \\ \textcolor{MidnightBlue}{\{question\}} \textcolor{ForestGreen}{\{options\}} \\ A:

\paragraph{QA - 09} Source: NIV2 - Task 73 - Template 9 \\ [1ex]
\textbf{Input}: \textcolor{MidnightBlue}{\{question\}}, \textcolor{ForestGreen}{\{options\}} \\ [0.5ex]
\textbf{Template}: \\ [0.5ex]
Detailed Instructions: You are given a question and some answer options (associated with "A", "B", "C", "D"). You should choose the correct answer based on commonsense knowledge. Avoid answering questions based on associations, the set of answers are chosen deliberately to capture common sense beyond associations. Do not generate anything else apart from one of the following characters: \textcolor{ForestGreen}{\{options letter\}} and only give one answer for each question. \\ Problem:\textcolor{MidnightBlue}{\{question\}} \textcolor{ForestGreen}{\{options\}} \\ Solution:

\paragraph{QA - 10} Source: NIV2 - Task 73 - Template 10 \\ [1ex]
\textbf{Input}: \textcolor{MidnightBlue}{\{question\}}, \textcolor{ForestGreen}{\{options\}} \\ [0.5ex]
\textbf{Template}: \\ [0.5ex]
Detailed Instructions: You are given a question and some answer options (associated with "A", "B", "C", "D"). You should choose the correct answer based on commonsense knowledge. Avoid answering questions based on associations, the set of answers are chosen deliberately to capture common sense beyond associations. Do not generate anything else apart from one of the following characters: \textcolor{ForestGreen}{\{options letter\}} and only give one answer for each question. \\ Q: \textcolor{MidnightBlue}{\{question\}} \textcolor{ForestGreen}{\{options\}} \\ A:

\paragraph{QA - 11} Source: NIV2 - Task 1420 - Template 1 \\ [1ex]
\textbf{Input}: \textcolor{MidnightBlue}{\{question\}}, \textcolor{ForestGreen}{\{options\}} \\ [0.5ex]
\textbf{Template}: \\ [0.5ex]
In this task, you need to provide the correct option for a given problem from the provided options. \\  \\ Problem:\textcolor{MidnightBlue}{\{question\}}  \\ \textcolor{ForestGreen}{\{options\}}

\paragraph{QA - 12} Source: NIV2 - Task 1420 - Template 2 \\ [1ex]
\textbf{Input}: \textcolor{MidnightBlue}{\{question\}}, \textcolor{ForestGreen}{\{options\}} \\ [0.5ex]
\textbf{Template}: \\ [0.5ex]
You will be given a definition of a task first, then some input of the task. \\ In this task, you need to provide the correct option for a given problem from the provided options. \\  \\ Problem:\textcolor{MidnightBlue}{\{question\}}  \\ \textcolor{ForestGreen}{\{options\}} \\ Output:

\paragraph{QA - 13} Source: NIV2 - Task 1420 - Template 3 \\ [1ex]
\textbf{Input}: \textcolor{MidnightBlue}{\{question\}}, \textcolor{ForestGreen}{\{options\}} \\ [0.5ex]
\textbf{Template}: \\ [0.5ex]
Definition: In this task, you need to provide the correct option for a given problem from the provided options. \\ Input: Problem:\textcolor{MidnightBlue}{\{question\}}  \\ \textcolor{ForestGreen}{\{options\}} \\ Output:

\paragraph{QA - 14} Source: NIV2 - Task 1420 - Template 4 \\ [1ex]
\textbf{Input}: \textcolor{MidnightBlue}{\{question\}}, \textcolor{ForestGreen}{\{options\}} \\ [0.5ex]
\textbf{Template}: \\ [0.5ex]
Instructions: In this task, you need to provide the correct option for a given problem from the provided options. \\ Input: Problem:\textcolor{MidnightBlue}{\{question\}}  \\ \textcolor{ForestGreen}{\{options\}} \\ Output:

\paragraph{QA - 15} Source: NIV2 - Task 1420 - Template 5 \\ [1ex]
\textbf{Input}: \textcolor{MidnightBlue}{\{question\}}, \textcolor{ForestGreen}{\{options\}} \\ [0.5ex]
\textbf{Template}: \\ [0.5ex]
In this task, you need to provide the correct option for a given problem from the provided options. \\ Q: Problem:\textcolor{MidnightBlue}{\{question\}}  \\ \textcolor{ForestGreen}{\{options\}} \\ A:

\paragraph{QA - 16} Source: NIV2 - Task 1420 - Template 6 \\ [1ex]
\textbf{Input}: \textcolor{MidnightBlue}{\{question\}}, \textcolor{ForestGreen}{\{options\}} \\ [0.5ex]
\textbf{Template}: \\ [0.5ex]
Given the task definition and input, reply with output. In this task, you need to provide the correct option for a given problem from the provided options. \\  \\ Problem:\textcolor{MidnightBlue}{\{question\}}  \\ \textcolor{ForestGreen}{\{options\}} \\

\paragraph{QA - 17} Source: NIV2 - Task 1420 - Template 7 \\ [1ex]
\textbf{Input}: \textcolor{MidnightBlue}{\{question\}}, \textcolor{ForestGreen}{\{options\}} \\ [0.5ex]
\textbf{Template}: \\ [0.5ex]
Teacher: In this task, you need to provide the correct option for a given problem from the provided options. \\ Teacher: Now, understand the problem? Solve this instance: Problem:\textcolor{MidnightBlue}{\{question\}}  \\ \textcolor{ForestGreen}{\{options\}} \\ Student:

\paragraph{QA - 18} Source: NIV2 - Task 1420 - Template 8 \\ [1ex]
\textbf{Input}: \textcolor{MidnightBlue}{\{question\}}, \textcolor{ForestGreen}{\{options\}} \\ [0.5ex]
\textbf{Template}: \\ [0.5ex]
Q: In this task, you need to provide the correct option for a given problem from the provided options. \\ Problem:\textcolor{MidnightBlue}{\{question\}}  \\ \textcolor{ForestGreen}{\{options\}} \\ A:

\paragraph{QA - 19} Source: NIV2 - Task 1420 - Template 9 \\ [1ex]
\textbf{Input}: \textcolor{MidnightBlue}{\{question\}}, \textcolor{ForestGreen}{\{options\}} \\ [0.5ex]
\textbf{Template}: \\ [0.5ex]
Detailed Instructions: In this task, you need to provide the correct option for a given problem from the provided options. \\ Problem:Problem:\textcolor{MidnightBlue}{\{question\}}  \\ \textcolor{ForestGreen}{\{options\}} \\ Solution:

\paragraph{QA - 20} Source: NIV2 - Task 1420 - Template 10 \\ [1ex]
\textbf{Input}: \textcolor{MidnightBlue}{\{question\}}, \textcolor{ForestGreen}{\{options\}} \\ [0.5ex]
\textbf{Template}: \\ [0.5ex]
Detailed Instructions: In this task, you need to provide the correct option for a given problem from the provided options. \\ Q: Problem:\textcolor{MidnightBlue}{\{question\}}  \\ \textcolor{ForestGreen}{\{options\}} \\ A:

\paragraph{QA - 21} Source: NIV2 - Task 1286 - Template 1 \\ [1ex]
\textbf{Input}: \textcolor{MidnightBlue}{\{question\}}, \textcolor{ForestGreen}{\{options\}} \\ [0.5ex]
\textbf{Template}: \\ [0.5ex]
In this task, you are given a multiple-choice question and you have to pick the incorrect option. Answer with option indexes (i.e., \textcolor{ForestGreen}{\{options letter\}}). \\  \\ \textcolor{MidnightBlue}{\{question\}} \textcolor{ForestGreen}{\{options\}}

\paragraph{QA - 22} Source: NIV2 - Task 1286 - Template 2 \\ [1ex]
\textbf{Input}: \textcolor{MidnightBlue}{\{question\}}, \textcolor{ForestGreen}{\{options\}} \\ [0.5ex]
\textbf{Template}: \\ [0.5ex]
You will be given a definition of a task first, then some input of the task. \\ In this task, you are given a multiple-choice question and you have to pick the incorrect option. Answer with option indexes (i.e., \textcolor{ForestGreen}{\{options letter\}}). \\  \\ \textcolor{MidnightBlue}{\{question\}} \textcolor{ForestGreen}{\{options\}} \\ Output:

\paragraph{QA - 23} Source: NIV2 - Task 1286 - Template 3 \\ [1ex]
\textbf{Input}: \textcolor{MidnightBlue}{\{question\}}, \textcolor{ForestGreen}{\{options\}} \\ [0.5ex]
\textbf{Template}: \\ [0.5ex]
Definition: In this task, you are given a multiple-choice question and you have to pick the incorrect option. Answer with option indexes (i.e., \textcolor{ForestGreen}{\{options letter\}}). \\ Input: \textcolor{MidnightBlue}{\{question\}} \textcolor{ForestGreen}{\{options\}} \\ Output:

\paragraph{QA - 24} Source: NIV2 - Task 1286 - Template 4 \\ [1ex]
\textbf{Input}: \textcolor{MidnightBlue}{\{question\}}, \textcolor{ForestGreen}{\{options\}} \\ [0.5ex]
\textbf{Template}: \\ [0.5ex]
Instructions: In this task, you are given a multiple-choice question and you have to pick the incorrect option. Answer with option indexes (i.e., \textcolor{ForestGreen}{\{options letter\}}). \\ Input: \textcolor{MidnightBlue}{\{question\}} \textcolor{ForestGreen}{\{options\}} \\ Output:

\paragraph{QA - 25} Source: NIV2 - Task 1286 - Template 5 \\ [1ex]
\textbf{Input}: \textcolor{MidnightBlue}{\{question\}}, \textcolor{ForestGreen}{\{options\}} \\ [0.5ex]
\textbf{Template}: \\ [0.5ex]
In this task, you are given a multiple-choice question and you have to pick the incorrect option. Answer with option indexes (i.e., \textcolor{ForestGreen}{\{options letter\}}). \\ Q: \textcolor{MidnightBlue}{\{question\}} \textcolor{ForestGreen}{\{options\}} \\ A: 

\paragraph{QA - 26} Source: NIV2 - Task 1286 - Template 6 \\ [1ex]
\textbf{Input}: \textcolor{MidnightBlue}{\{question\}}, \textcolor{ForestGreen}{\{options\}} \\ [0.5ex]
\textbf{Template}: \\ [0.5ex]
Given the task definition and input, reply with output. In this task, you are given a multiple-choice question and you have to pick the incorrect option. Answer with option indexes (i.e., \textcolor{ForestGreen}{\{options letter\}}). \\  \\ \textcolor{MidnightBlue}{\{question\}} \textcolor{ForestGreen}{\{options\}} \\

\paragraph{QA - 27} Source: NIV2 - Task 1286 - Template 7 \\ [1ex]
\textbf{Input}: \textcolor{MidnightBlue}{\{question\}}, \textcolor{ForestGreen}{\{options\}} \\ [0.5ex]
\textbf{Template}: \\ [0.5ex]
Teacher: In this task, you are given a multiple-choice question and you have to pick the incorrect option. Answer with option indexes (i.e., \textcolor{ForestGreen}{\{options letter\}}). \\ Teacher: Now, understand the problem? Solve this instance: \textcolor{MidnightBlue}{\{question\}} \textcolor{ForestGreen}{\{options\}} \\ Student:

\paragraph{QA - 28} Source: NIV2 - Task 1286 - Template 8 \\ [1ex]
\textbf{Input}: \textcolor{MidnightBlue}{\{question\}}, \textcolor{ForestGreen}{\{options\}} \\ [0.5ex]
\textbf{Template}: \\ [0.5ex]
Q: In this task, you are given a multiple-choice question and you have to pick the incorrect option. Answer with option indexes (i.e., \textcolor{ForestGreen}{\{options letter\}}). \\ \textcolor{MidnightBlue}{\{question\}} \textcolor{ForestGreen}{\{options\}} \\ A:

\paragraph{QA - 29} Source: NIV2 - Task 1286 - Template 9 \\ [1ex]
\textbf{Input}: \textcolor{MidnightBlue}{\{question\}}, \textcolor{ForestGreen}{\{options\}} \\ [0.5ex]
\textbf{Template}: \\ [0.5ex]
Detailed Instructions: In this task, you are given a multiple-choice question and you have to pick the incorrect option. Answer with option indexes (i.e., \textcolor{ForestGreen}{\{options letter\}}). \\ Problem:\textcolor{MidnightBlue}{\{question\}} \textcolor{ForestGreen}{\{options\}} \\ Solution:

\paragraph{QA - 30} Source: NIV2 - Task 1286 - Template 10 \\ [1ex]
\textbf{Input}: \textcolor{MidnightBlue}{\{question\}}, \textcolor{ForestGreen}{\{options\}} \\ [0.5ex]
\textbf{Template}: \\ [0.5ex]
Detailed Instructions: In this task, you are given a multiple-choice question and you have to pick the incorrect option. Answer with option indexes (i.e., \textcolor{ForestGreen}{\{options letter\}}). \\ Q: \textcolor{MidnightBlue}{\{question\}} \textcolor{ForestGreen}{\{options\}} \\ A:

\paragraph{QA - 31} Source: NIV2 - Task 1565 - Template 1 \\ [1ex]
\textbf{Input}: \textcolor{MidnightBlue}{\{question\}}, \textcolor{ForestGreen}{\{options\}} \\ [0.5ex]
\textbf{Template}: \\ [0.5ex]
This task involves asking a question, providing a set of \textcolor{ForestGreen}{\{options length\}} options. You are expected to choose the best answer to the question. The output will be in the form of \textcolor{ForestGreen}{\{options letter\}}, corresponding to which option is chosen. \\  \\ \textcolor{MidnightBlue}{\{question\}}, Options: [\textcolor{ForestGreen}{\{options\}}]

\paragraph{QA - 32} Source: NIV2 - Task 1565 - Template 2 \\ [1ex]
\textbf{Input}: \textcolor{MidnightBlue}{\{question\}}, \textcolor{ForestGreen}{\{options\}} \\ [0.5ex]
\textbf{Template}: \\ [0.5ex]
You will be given a definition of a task first, then some input of the task. \\ This task involves asking a question, providing a set of \textcolor{ForestGreen}{\{options length\}} options. You are expected to choose the best answer to the question. The output will be in the form of \textcolor{ForestGreen}{\{options letter\}}, corresponding to which option is chosen. \\  \\ \textcolor{MidnightBlue}{\{question\}}, Options: [\textcolor{ForestGreen}{\{options\}}] \\ Output:

\paragraph{QA - 33} Source: NIV2 - Task 1565 - Template 3 \\ [1ex]
\textbf{Input}: \textcolor{MidnightBlue}{\{question\}}, \textcolor{ForestGreen}{\{options\}} \\ [0.5ex]
\textbf{Template}: \\ [0.5ex]
Definition: This task involves asking a question, providing a set of \textcolor{ForestGreen}{\{options length\}} options. You are expected to choose the best answer to the question. The output will be in the form of \textcolor{ForestGreen}{\{options letter\}}, corresponding to which option is chosen. \\ Input: \textcolor{MidnightBlue}{\{question\}}, Options: [\textcolor{ForestGreen}{\{options\}}] \\ Output:

\paragraph{QA - 34} Source: NIV2 - Task 1565 - Template 4 \\ [1ex]
\textbf{Input}: \textcolor{MidnightBlue}{\{question\}}, \textcolor{ForestGreen}{\{options\}} \\ [0.5ex]
\textbf{Template}: \\ [0.5ex]
Instructions: This task involves asking a question, providing a set of \textcolor{ForestGreen}{\{options length\}} options. You are expected to choose the best answer to the question. The output will be in the form of \textcolor{ForestGreen}{\{options letter\}}, corresponding to which option is chosen. \\ Input: \textcolor{MidnightBlue}{\{question\}}, Options: [\textcolor{ForestGreen}{\{options\}}] \\ Output:

\paragraph{QA - 35} Source: NIV2 - Task 1565 - Template 5 \\ [1ex]
\textbf{Input}: \textcolor{MidnightBlue}{\{question\}}, \textcolor{ForestGreen}{\{options\}} \\ [0.5ex]
\textbf{Template}: \\ [0.5ex]
This task involves asking a question, providing a set of \textcolor{ForestGreen}{\{options length\}} options. You are expected to choose the best answer to the question. The output will be in the form of \textcolor{ForestGreen}{\{options letter\}}, corresponding to which option is chosen. \\ Q: \textcolor{MidnightBlue}{\{question\}}, Options: [\textcolor{ForestGreen}{\{options\}}] \\ A: 

\paragraph{QA - 36} Source: NIV2 - Task 1565 - Template 6 \\ [1ex]
\textbf{Input}: \textcolor{MidnightBlue}{\{question\}}, \textcolor{ForestGreen}{\{options\}} \\ [0.5ex]
\textbf{Template}: \\ [0.5ex]
Given the task definition and input, reply with output. This task involves asking a question, providing a set of \textcolor{ForestGreen}{\{options length\}} options. You are expected to choose the best answer to the question. The output will be in the form of \textcolor{ForestGreen}{\{options letter\}}, corresponding to which option is chosen. \\  \\ \textcolor{MidnightBlue}{\{question\}}, Options: [\textcolor{ForestGreen}{\{options\}}] \\ 

\paragraph{QA - 37} Source: NIV2 - Task 1565 - Template 7 \\ [1ex]
\textbf{Input}: \textcolor{MidnightBlue}{\{question\}}, \textcolor{ForestGreen}{\{options\}} \\ [0.5ex]
\textbf{Template}: \\ [0.5ex]
Teacher: This task involves asking a question, providing a set of \textcolor{ForestGreen}{\{options length\}} options. You are expected to choose the best answer to the question. The output will be in the form of \textcolor{ForestGreen}{\{options letter\}}, corresponding to which option is chosen. \\ Teacher: Now, understand the problem? Solve this instance: \textcolor{MidnightBlue}{\{question\}}, Options: [\textcolor{ForestGreen}{\{options\}}] \\ Student:

\paragraph{QA - 38} Source: NIV2 - Task 1565 - Template 8 \\ [1ex]
\textbf{Input}: \textcolor{MidnightBlue}{\{question\}}, \textcolor{ForestGreen}{\{options\}} \\ [0.5ex]
\textbf{Template}: \\ [0.5ex]
Q: This task involves asking a question, providing a set of \textcolor{ForestGreen}{\{options length\}} options. You are expected to choose the best answer to the question. The output will be in the form of \textcolor{ForestGreen}{\{options letter\}}, corresponding to which option is chosen. \\ \textcolor{MidnightBlue}{\{question\}}, Options: [\textcolor{ForestGreen}{\{options\}}] \\ A:

\paragraph{QA - 39} Source: NIV2 - Task 1565 - Template 9 \\ [1ex]
\textbf{Input}: \textcolor{MidnightBlue}{\{question\}}, \textcolor{ForestGreen}{\{options\}} \\ [0.5ex]
\textbf{Template}: \\ [0.5ex]
Detailed Instructions: This task involves asking a question, providing a set of \textcolor{ForestGreen}{\{options length\}} options. You are expected to choose the best answer to the question. The output will be in the form of \textcolor{ForestGreen}{\{options letter\}}, corresponding to which option is chosen. \\ Problem:\textcolor{MidnightBlue}{\{question\}}, Options: [\textcolor{ForestGreen}{\{options\}}] \\ Solution:

\paragraph{QA - 40} Source: NIV2 - Task 1565 - Template 10 \\ [1ex]
\textbf{Input}: \textcolor{MidnightBlue}{\{question\}}, \textcolor{ForestGreen}{\{options\}} \\ [0.5ex]
\textbf{Template}: \\ [0.5ex]
Detailed Instructions: This task involves asking a question, providing a set of \textcolor{ForestGreen}{\{options length\}} options. You are expected to choose the best answer to the question. The output will be in the form of \textcolor{ForestGreen}{\{options letter\}}, corresponding to which option is chosen. \\ Q: \textcolor{MidnightBlue}{\{question\}}, Options: [\textcolor{ForestGreen}{\{options\}}] \\ A:

\paragraph{QA - 41} Source: NIV2 - Task 229 - Template 1 \\ [1ex]
\textbf{Input}: \textcolor{MidnightBlue}{\{question\}}, \textcolor{ForestGreen}{\{options\}} \\ [0.5ex]
\textbf{Template}: \\ [0.5ex]
You are given a science question (hard-level) and \textcolor{ForestGreen}{\{option length\}} answer options (associated with \textcolor{ForestGreen}{\{option letter\}}). Your task is to find the correct answer based on scientific facts, knowledge, and reasoning. Do not generate anything else apart from one of the following characters: \textcolor{ForestGreen}{\{option letter\}}. There is only one correct answer for each question. \\  \\ \textcolor{MidnightBlue}{\{question\}} \textcolor{ForestGreen}{\{options\}}

\paragraph{QA - 42} Source: NIV2 - Task 229 - Template 2 \\ [1ex]
\textbf{Input}: \textcolor{MidnightBlue}{\{question\}}, \textcolor{ForestGreen}{\{options\}} \\ [0.5ex]
\textbf{Template}: \\ [0.5ex]
You will be given a definition of a task first, then some input of the task. \\ You are given a science question (hard-level) and \textcolor{ForestGreen}{\{option length\}} answer options (associated with \textcolor{ForestGreen}{\{option letter\}}). Your task is to find the correct answer based on scientific facts, knowledge, and reasoning. Do not generate anything else apart from one of the following characters: \textcolor{ForestGreen}{\{option letter\}}. There is only one correct answer for each question. \\  \\ \textcolor{MidnightBlue}{\{question\}} \textcolor{ForestGreen}{\{options\}} \\ Output:

\paragraph{QA - 43} Source: NIV2 - Task 229 - Template 3 \\ [1ex]
\textbf{Input}: \textcolor{MidnightBlue}{\{question\}}, \textcolor{ForestGreen}{\{options\}} \\ [0.5ex]
\textbf{Template}: \\ [0.5ex]
Definition: You are given a science question (hard-level) and \textcolor{ForestGreen}{\{option length\}} answer options (associated with \textcolor{ForestGreen}{\{option letter\}}). Your task is to find the correct answer based on scientific facts, knowledge, and reasoning. Do not generate anything else apart from one of the following characters: \textcolor{ForestGreen}{\{option letter\}}. There is only one correct answer for each question. \\ Input: \textcolor{MidnightBlue}{\{question\}} \textcolor{ForestGreen}{\{options\}} \\ Output:

\paragraph{QA - 44} Source: NIV2 - Task 229 - Template 4 \\ [1ex]
\textbf{Input}: \textcolor{MidnightBlue}{\{question\}}, \textcolor{ForestGreen}{\{options\}} \\ [0.5ex]
\textbf{Template}: \\ [0.5ex]
Instructions: You are given a science question (hard-level) and \textcolor{ForestGreen}{\{option length\}} answer options (associated with \textcolor{ForestGreen}{\{option letter\}}). Your task is to find the correct answer based on scientific facts, knowledge, and reasoning. Do not generate anything else apart from one of the following characters: \textcolor{ForestGreen}{\{option letter\}}. There is only one correct answer for each question. \\ Input: \textcolor{MidnightBlue}{\{question\}} \textcolor{ForestGreen}{\{options\}} \\ Output:

\paragraph{QA - 45} Source: NIV2 - Task 229 - Template 5 \\ [1ex]
\textbf{Input}: \textcolor{MidnightBlue}{\{question\}}, \textcolor{ForestGreen}{\{options\}} \\ [0.5ex]
\textbf{Template}: \\ [0.5ex]
You are given a science question (hard-level) and \textcolor{ForestGreen}{\{option length\}} answer options (associated with \textcolor{ForestGreen}{\{option letter\}}). Your task is to find the correct answer based on scientific facts, knowledge, and reasoning. Do not generate anything else apart from one of the following characters: \textcolor{ForestGreen}{\{option letter\}}. There is only one correct answer for each question. \\ Q: \textcolor{MidnightBlue}{\{question\}} \textcolor{ForestGreen}{\{options\}} \\ A:

\paragraph{QA - 46} Source: NIV2 - Task 229 - Template 6 \\ [1ex]
\textbf{Input}: \textcolor{MidnightBlue}{\{question\}}, \textcolor{ForestGreen}{\{options\}} \\ [0.5ex]
\textbf{Template}: \\ [0.5ex]
Given the task definition and input, reply with output. You are given a science question (hard-level) and \textcolor{ForestGreen}{\{option length\}} answer options (associated with \textcolor{ForestGreen}{\{option letter\}}). Your task is to find the correct answer based on scientific facts, knowledge, and reasoning. Do not generate anything else apart from one of the following characters: \textcolor{ForestGreen}{\{option letter\}}. There is only one correct answer for each question. \\  \\ \textcolor{MidnightBlue}{\{question\}} \textcolor{ForestGreen}{\{options\}} \\

\paragraph{QA - 47} Source: NIV2 - Task 229 - Template 7 \\ [1ex]
\textbf{Input}: \textcolor{MidnightBlue}{\{question\}}, \textcolor{ForestGreen}{\{options\}} \\ [0.5ex]
\textbf{Template}: \\ [0.5ex]
Teacher: You are given a science question (hard-level) and \textcolor{ForestGreen}{\{option length\}} answer options (associated with \textcolor{ForestGreen}{\{option letter\}}). Your task is to find the correct answer based on scientific facts, knowledge, and reasoning. Do not generate anything else apart from one of the following characters: \textcolor{ForestGreen}{\{option letter\}}. There is only one correct answer for each question. \\ Teacher: Now, understand the problem? Solve this instance: \textcolor{MidnightBlue}{\{question\}} \textcolor{ForestGreen}{\{options\}} \\ Student:

\paragraph{QA - 48} Source: NIV2 - Task 229 - Template 8 \\ [1ex]
\textbf{Input}: \textcolor{MidnightBlue}{\{question\}}, \textcolor{ForestGreen}{\{options\}} \\ [0.5ex]
\textbf{Template}: \\ [0.5ex]
Q: You are given a science question (hard-level) and \textcolor{ForestGreen}{\{option length\}} answer options (associated with \textcolor{ForestGreen}{\{option letter\}}). Your task is to find the correct answer based on scientific facts, knowledge, and reasoning. Do not generate anything else apart from one of the following characters: \textcolor{ForestGreen}{\{option letter\}}. There is only one correct answer for each question. \\ \textcolor{MidnightBlue}{\{question\}} \textcolor{ForestGreen}{\{options\}} \\ A:

\paragraph{QA - 49} Source: NIV2 - Task 229 - Template 9 \\ [1ex]
\textbf{Input}: \textcolor{MidnightBlue}{\{question\}}, \textcolor{ForestGreen}{\{options\}} \\ [0.5ex]
\textbf{Template}: \\ [0.5ex]
Detailed Instructions: You are given a science question (hard-level) and \textcolor{ForestGreen}{\{option length\}} answer options (associated with \textcolor{ForestGreen}{\{option letter\}}). Your task is to find the correct answer based on scientific facts, knowledge, and reasoning. Do not generate anything else apart from one of the following characters: \textcolor{ForestGreen}{\{option letter\}}. There is only one correct answer for each question. \\ Problem:\textcolor{MidnightBlue}{\{question\}} \textcolor{ForestGreen}{\{options\}} \\ Solution:

\paragraph{QA - 50} Source: NIV2 - Task 229 - Template 10 \\ [1ex]
\textbf{Input}: \textcolor{MidnightBlue}{\{question\}}, \textcolor{ForestGreen}{\{options\}} \\ [0.5ex]
\textbf{Template}: \\ [0.5ex]
Detailed Instructions: You are given a science question (hard-level) and \textcolor{ForestGreen}{\{option length\}} answer options (associated with \textcolor{ForestGreen}{\{option letter\}}). Your task is to find the correct answer based on scientific facts, knowledge, and reasoning. Do not generate anything else apart from one of the following characters: \textcolor{ForestGreen}{\{option letter\}}. There is only one correct answer for each question. \\ Q: \textcolor{MidnightBlue}{\{question\}} \textcolor{ForestGreen}{\{options\}} \\ A:

\paragraph{MC - 01} Source: NIV2 - Task 1135 - Template 1 \\ [1ex]
\textbf{Input}: \textcolor{MidnightBlue}{\{question\}}, \textcolor{ForestGreen}{\{options\}} \\ [0.5ex]
\textbf{Template}: \\ [0.5ex]
In this task, you will be presented with a question that has multiple possible answers. You should choose the most suitable option out of \textcolor{ForestGreen}{\{options letter\}}, based on your commonsense knowledge. \\  \\ \textcolor{MidnightBlue}{\{question\}} Options: \textcolor{ForestGreen}{\{options\}}

\paragraph{MC - 02} Source: NIV2 - Task 1135 - Template 2 \\ [1ex]
\textbf{Input}: \textcolor{MidnightBlue}{\{question\}}, \textcolor{ForestGreen}{\{options\}} \\ [0.5ex]
\textbf{Template}: \\ [0.5ex]
You will be given a definition of a task first, then some input of the task. \\ In this task, you will be presented with a question that has multiple possible answers. You should choose the most suitable option out of \textcolor{ForestGreen}{\{options letter\}}, based on your commonsense knowledge. \\  \\ \textcolor{MidnightBlue}{\{question\}} Options: \textcolor{ForestGreen}{\{options\}} \\ Output:

\paragraph{MC - 03} Source: NIV2 - Task 1135 - Template 3 \\ [1ex]
\textbf{Input}: \textcolor{MidnightBlue}{\{question\}}, \textcolor{ForestGreen}{\{options\}} \\ [0.5ex]
\textbf{Template}: \\ [0.5ex]
Definition: In this task, you will be presented with a question that has multiple possible answers. You should choose the most suitable option out of \textcolor{ForestGreen}{\{options letter\}}, based on your commonsense knowledge. \\ Input: \textcolor{MidnightBlue}{\{question\}} Options: \textcolor{ForestGreen}{\{options\}} \\ Output:

\paragraph{MC - 04} Source: NIV2 - Task 1135 - Template 4 \\ [1ex]
\textbf{Input}: \textcolor{MidnightBlue}{\{question\}}, \textcolor{ForestGreen}{\{options\}} \\ [0.5ex]
\textbf{Template}: \\ [0.5ex]
Instructions: In this task, you will be presented with a question that has multiple possible answers. You should choose the most suitable option out of \textcolor{ForestGreen}{\{options letter\}}, based on your commonsense knowledge. \\ Input: \textcolor{MidnightBlue}{\{question\}} Options: \textcolor{ForestGreen}{\{options\}} \\ Output:

\paragraph{MC - 05} Source: NIV2 - Task 1135 - Template 5 \\ [1ex]
\textbf{Input}: \textcolor{MidnightBlue}{\{question\}}, \textcolor{ForestGreen}{\{options\}} \\ [0.5ex]
\textbf{Template}: \\ [0.5ex]
In this task, you will be presented with a question that has multiple possible answers. You should choose the most suitable option out of \textcolor{ForestGreen}{\{options letter\}}, based on your commonsense knowledge. \\ Q: \textcolor{MidnightBlue}{\{question\}} Options: \textcolor{ForestGreen}{\{options\}} \\ A: 

\paragraph{MC - 06} Source: NIV2 - Task 1135 - Template 6 \\ [1ex]
\textbf{Input}: \textcolor{MidnightBlue}{\{question\}}, \textcolor{ForestGreen}{\{options\}} \\ [0.5ex]
\textbf{Template}: \\ [0.5ex]
Given the task definition and input, reply with output. In this task, you will be presented with a question that has multiple possible answers. You should choose the most suitable option out of \textcolor{ForestGreen}{\{options letter\}}, based on your commonsense knowledge. \\  \\ \textcolor{MidnightBlue}{\{question\}} Options: \textcolor{ForestGreen}{\{options\}} \\ 

\paragraph{MC - 07} Source: NIV2 - Task 1135 - Template 7 \\ [1ex]
\textbf{Input}: \textcolor{MidnightBlue}{\{question\}}, \textcolor{ForestGreen}{\{options\}} \\ [0.5ex]
\textbf{Template}: \\ [0.5ex]
Teacher: In this task, you will be presented with a question that has multiple possible answers. You should choose the most suitable option out of \textcolor{ForestGreen}{\{options letter\}}, based on your commonsense knowledge. \\ Teacher: Now, understand the problem? Solve this instance: \textcolor{MidnightBlue}{\{question\}} Options: \textcolor{ForestGreen}{\{options\}} \\ Student:

\paragraph{MC - 08} Source: NIV2 - Task 1135 - Template 8 \\ [1ex]
\textbf{Input}: \textcolor{MidnightBlue}{\{question\}}, \textcolor{ForestGreen}{\{options\}} \\ [0.5ex]
\textbf{Template}: \\ [0.5ex]
Q: In this task, you will be presented with a question that has multiple possible answers. You should choose the most suitable option out of \textcolor{ForestGreen}{\{options letter\}}, based on your commonsense knowledge. \\ \textcolor{MidnightBlue}{\{question\}} Options: \textcolor{ForestGreen}{\{options\}} \\ A:

\paragraph{MC - 09} Source: NIV2 - Task 1135 - Template 9 \\ [1ex]
\textbf{Input}: \textcolor{MidnightBlue}{\{question\}}, \textcolor{ForestGreen}{\{options\}} \\ [0.5ex]
\textbf{Template}: \\ [0.5ex]
Detailed Instructions: In this task, you will be presented with a question that has multiple possible answers. You should choose the most suitable option out of \textcolor{ForestGreen}{\{options letter\}}, based on your commonsense knowledge. \\ Problem:\textcolor{MidnightBlue}{\{question\}} Options: \textcolor{ForestGreen}{\{options\}} \\ Solution:

\paragraph{MC - 10} Source: NIV2 - Task 1135 - Template 10 \\ [1ex]
\textbf{Input}: \textcolor{MidnightBlue}{\{question\}}, \textcolor{ForestGreen}{\{options\}} \\ [0.5ex]
\textbf{Template}: \\ [0.5ex]
Detailed Instructions: In this task, you will be presented with a question that has multiple possible answers. You should choose the most suitable option out of \textcolor{ForestGreen}{\{options letter\}}, based on your commonsense knowledge. \\ Q: \textcolor{MidnightBlue}{\{question\}} Options: \textcolor{ForestGreen}{\{options\}} \\ A:

\paragraph{MC - 11} Source: NIV2 - Task 900 - Template 1 \\ [1ex]
\textbf{Input}: \textcolor{MidnightBlue}{\{text\}}, \textcolor{ForestGreen}{\{options\}} \\ [0.5ex]
\textbf{Template}: \\ [0.5ex]
Given a trivia question, classify broad topical category from this list: \textcolor{ForestGreen}{\{options\}}. \\  \\ \textcolor{MidnightBlue}{\{question\}}

\paragraph{MC - 12} Source: NIV2 - Task 900 - Template 2 \\ [1ex]
\textbf{Input}: \textcolor{MidnightBlue}{\{text\}}, \textcolor{ForestGreen}{\{options\}} \\ [0.5ex]
\textbf{Template}: \\ [0.5ex]
You will be given a definition of a task first, then some input of the task. \\ Given a trivia question, classify broad topical category from this list: \textcolor{ForestGreen}{\{options\}}. \\  \\ \textcolor{MidnightBlue}{\{question\}} \\ Output:

\paragraph{MC - 13} Source: NIV2 - Task 900 - Template 3 \\ [1ex]
\textbf{Input}: \textcolor{MidnightBlue}{\{text\}}, \textcolor{ForestGreen}{\{options\}} \\ [0.5ex]
\textbf{Template}: \\ [0.5ex]
Definition: Given a trivia question, classify broad topical category from this list: \textcolor{ForestGreen}{\{options\}}. \\ Input: \textcolor{MidnightBlue}{\{question\}} \\ Output:

\paragraph{MC - 14} Source: NIV2 - Task 900 - Template 4 \\ [1ex]
\textbf{Input}: \textcolor{MidnightBlue}{\{text\}}, \textcolor{ForestGreen}{\{options\}} \\ [0.5ex]
\textbf{Template}: \\ [0.5ex]
Instructions: Given a trivia question, classify broad topical category from this list: \textcolor{ForestGreen}{\{options\}}. \\ Input: \textcolor{MidnightBlue}{\{question\}} \\ Output:

\paragraph{MC - 15} Source: NIV2 - Task 900 - Template 5 \\ [1ex]
\textbf{Input}: \textcolor{MidnightBlue}{\{text\}}, \textcolor{ForestGreen}{\{options\}} \\ [0.5ex]
\textbf{Template}: \\ [0.5ex]
Given a trivia question, classify broad topical category from this list: \textcolor{ForestGreen}{\{options\}}. \\ Q: \textcolor{MidnightBlue}{\{question\}} \\ A: 

\paragraph{MC - 16} Source: NIV2 - Task 900 - Template 6 \\ [1ex]
\textbf{Input}: \textcolor{MidnightBlue}{\{text\}}, \textcolor{ForestGreen}{\{options\}} \\ [0.5ex]
\textbf{Template}: \\ [0.5ex]
Given the task definition and input, reply with output. Given a trivia question, classify broad topical category from this list: \textcolor{ForestGreen}{\{options\}}. \\  \\ \textcolor{MidnightBlue}{\{question\}} \\ 

\paragraph{MC - 17} Source: NIV2 - Task 900 - Template 7 \\ [1ex]
\textbf{Input}: \textcolor{MidnightBlue}{\{text\}}, \textcolor{ForestGreen}{\{options\}} \\ [0.5ex]
\textbf{Template}: \\ [0.5ex]
Teacher: Given a trivia question, classify broad topical category from this list: \textcolor{ForestGreen}{\{options\}}. \\ Teacher: Now, understand the problem? Solve this instance: \textcolor{MidnightBlue}{\{question\}} \\ Student:

\paragraph{MC - 18} Source: NIV2 - Task 900 - Template 8 \\ [1ex]
\textbf{Input}: \textcolor{MidnightBlue}{\{text\}}, \textcolor{ForestGreen}{\{options\}} \\ [0.5ex]
\textbf{Template}: \\ [0.5ex]
Q: Given a trivia question, classify broad topical category from this list: \textcolor{ForestGreen}{\{options\}}. \\ \textcolor{MidnightBlue}{\{question\}} \\ A:

\paragraph{MC - 19} Source: NIV2 - Task 900 - Template 9 \\ [1ex]
\textbf{Input}: \textcolor{MidnightBlue}{\{text\}}, \textcolor{ForestGreen}{\{options\}} \\ [0.5ex]
\textbf{Template}: \\ [0.5ex]
Detailed Instructions: Given a trivia question, classify broad topical category from this list: \textcolor{ForestGreen}{\{options\}}. \\ Problem:\textcolor{MidnightBlue}{\{question\}} \\ Solution:

\paragraph{MC - 20} Source: NIV2 - Task 900 - Template 10 \\ [1ex]
\textbf{Input}: \textcolor{MidnightBlue}{\{text\}}, \textcolor{ForestGreen}{\{options\}} \\ [0.5ex]
\textbf{Template}: \\ [0.5ex]
Detailed Instructions: Given a trivia question, classify broad topical category from this list: \textcolor{ForestGreen}{\{options\}}. \\ Q: \textcolor{MidnightBlue}{\{question\}} \\ A:

\paragraph{MC - 21} Source: Flan2021 - ARC - Template 1 \\ [1ex]
\textbf{Input}: \textcolor{MidnightBlue}{\{question\}}, \textcolor{ForestGreen}{\{options\}} \\ [0.5ex]
\textbf{Template}: \\ [0.5ex]
\textcolor{MidnightBlue}{\{text\}} \\OPTIONS: \\ \textcolor{ForestGreen}{\{options\}}

\paragraph{MC - 22} Source: Flan2021 - ARC - Template 2 \\ [1ex]
\textbf{Input}: \textcolor{MidnightBlue}{\{question\}}, \textcolor{ForestGreen}{\{options\}} \\ [0.5ex]
\textbf{Template}: \\ [0.5ex]
Question: \textcolor{MidnightBlue}{\{text\}}? \\ OPTIONS:\textcolor{ForestGreen}{\{options\}} \\ Answer:

\paragraph{MC - 23} Source: Flan2021 - ARC - Template 3 \\ [1ex]
\textbf{Input}: \textcolor{MidnightBlue}{\{question\}}, \textcolor{ForestGreen}{\{options\}} \\ [0.5ex]
\textbf{Template}: \\ [0.5ex]
Question: \textcolor{MidnightBlue}{\{text\}} \\  \\ What is the correct answer to the question from the following choices? \\ OPTIONS:\textcolor{ForestGreen}{\{options\}}

\paragraph{MC - 24} Source: Flan2021 - ARC - Template 4 \\ [1ex]
\textbf{Input}: \textcolor{MidnightBlue}{\{question\}}, \textcolor{ForestGreen}{\{options\}} \\ [0.5ex]
\textbf{Template}: \\ [0.5ex]
Question: \textcolor{MidnightBlue}{\{text\}} \\ What is the correct answer to this question? \\ OPTIONS:\textcolor{ForestGreen}{\{options\}}...A: 

\paragraph{MC - 25} Source: Flan2021 - ARC - Template 5 \\ [1ex]
\textbf{Input}: \textcolor{MidnightBlue}{\{question\}}, \textcolor{ForestGreen}{\{options\}} \\ [0.5ex]
\textbf{Template}: \\ [0.5ex]
Choose your answer? \\  \\ \textcolor{MidnightBlue}{\{text\}} \\  \\ OPTIONS:\textcolor{ForestGreen}{\{options\}}

\paragraph{MC - 26} Source: Flan2021 - ARC - Template 6 \\ [1ex]
\textbf{Input}: \textcolor{MidnightBlue}{\{question\}}, \textcolor{ForestGreen}{\{options\}} \\ [0.5ex]
\textbf{Template}: \\ [0.5ex]
Answer the question \\  \\ \textcolor{MidnightBlue}{\{text\}} \\ OPTIONS:\textcolor{ForestGreen}{\{options\}}

\paragraph{MC - 27} Source: Flan2021 - ARC - Template 7 \\ [1ex]
\textbf{Input}: \textcolor{MidnightBlue}{\{question\}}, \textcolor{ForestGreen}{\{options\}} \\ [0.5ex]
\textbf{Template}: \\ [0.5ex]
\textcolor{MidnightBlue}{\{text\}} \\  \\ Pick the answer from these options \\  \\ OPTIONS:\textcolor{ForestGreen}{\{options\}}

\paragraph{MC - 28} Source: Flan2021 - CosmosQA - Template 1 \\ [1ex]
\textbf{Input}: \textcolor{MidnightBlue}{\{context\}}, \textcolor{MidnightBlue}{\{question\}}, \textcolor{ForestGreen}{\{options\}} \\ [0.5ex]
\textbf{Template}: \\ [0.5ex]
\textcolor{MidnightBlue}{\{context\}} \\  \\ Question with options to choose from: \textcolor{MidnightBlue}{\{question\}} \\ OPTIONS:\textcolor{ForestGreen}{\{options\}}

\paragraph{MC - 29} Source: Flan2021 - CosmosQA - Template 2 \\ [1ex]
\textbf{Input}: \textcolor{MidnightBlue}{\{context\}}, \textcolor{MidnightBlue}{\{question\}}, \textcolor{ForestGreen}{\{options\}} \\ [0.5ex]
\textbf{Template}: \\ [0.5ex]
\textcolor{MidnightBlue}{\{context\}} \\  \\ OPTIONS: \textcolor{ForestGreen}{\{options\}} \\ Q: \textcolor{MidnightBlue}{\{question\}}

\paragraph{MC - 30} Source: Flan2021 - CosmosQA - Template 3 \\ [1ex]
\textbf{Input}: \textcolor{MidnightBlue}{\{context\}}, \textcolor{MidnightBlue}{\{question\}}, \textcolor{ForestGreen}{\{options\}} \\ [0.5ex]
\textbf{Template}: \\ [0.5ex]
\textcolor{MidnightBlue}{\{context\}} \\  \\ OPTIONS: \textcolor{ForestGreen}{\{options\}} \\ Answer the following question: \textcolor{MidnightBlue}{\{question\}} \\ 

\paragraph{MC - 31} Source: Flan2021 - CosmosQA - Template 4 \\ [1ex]
\textbf{Input}: \textcolor{MidnightBlue}{\{context\}}, \textcolor{MidnightBlue}{\{question\}}, \textcolor{ForestGreen}{\{options\}} \\ [0.5ex]
\textbf{Template}: \\ [0.5ex]
\textcolor{MidnightBlue}{\{context\}} \\  \\ Based on the preceding passage, choose your answer for question \textcolor{MidnightBlue}{\{question\}} \\ OPTIONS: \textcolor{ForestGreen}{\{options\}}

\paragraph{MC - 32} Source: Flan2021 - CosmosQA - Template 5 \\ [1ex]
\textbf{Input}: \textcolor{MidnightBlue}{\{context\}}, \textcolor{MidnightBlue}{\{question\}}, \textcolor{ForestGreen}{\{options\}} \\ [0.5ex]
\textbf{Template}: \\ [0.5ex]
\textcolor{MidnightBlue}{\{context\}} \\  \\ Q with options: Give answer the following question using evidence from the above passage: \textcolor{MidnightBlue}{\{question\}} \\ OPTIONS:\textcolor{ForestGreen}{\{options\}}

\paragraph{MC - 33} Source: Flan2021 - CosmosQA - Template 6 \\ [1ex]
\textbf{Input}: \textcolor{MidnightBlue}{\{context\}}, \textcolor{MidnightBlue}{\{question\}}, \textcolor{ForestGreen}{\{options\}} \\ [0.5ex]
\textbf{Template}: \\ [0.5ex]
Context: \textcolor{MidnightBlue}{\{context\}} \\ Question \textcolor{MidnightBlue}{\{question\}} \\ Possible answers: \\ \textcolor{ForestGreen}{\{options\}} \\ The answer:

\paragraph{MC - 34} Source: Flan2021 - CosmosQA - Template 7 \\ [1ex]
\textbf{Input}: \textcolor{MidnightBlue}{\{context\}}, \textcolor{MidnightBlue}{\{question\}}, \textcolor{ForestGreen}{\{options\}} \\ [0.5ex]
\textbf{Template}: \\ [0.5ex]
Read the following article and answer the question by choosing from the options. \\  \\ \textcolor{MidnightBlue}{\{context\}} \\  \\ \textcolor{MidnightBlue}{\{question\}} \\OPTIONS: \textcolor{ForestGreen}{\{options\}}...A:

\paragraph{MC - 35} Source: Flan2021 - CosmosQA - Template 8 \\ [1ex]
\textbf{Input}: \textcolor{MidnightBlue}{\{context\}}, \textcolor{MidnightBlue}{\{question\}}, \textcolor{ForestGreen}{\{options\}} \\ [0.5ex]
\textbf{Template}: \\ [0.5ex]
This question has options. Answer the question about text: \\  \\ \textcolor{MidnightBlue}{\{context\}} \\  \\ \textcolor{MidnightBlue}{\{question\}} \\ OPTIONS:\textcolor{ForestGreen}{\{options\}}

\paragraph{BC - 01} Source: NIV2 - Task 56 - Template 1 \\ [1ex]
\textbf{Input}: \textcolor{MidnightBlue}{\{paragraph\}}, \textcolor{MidnightBlue}{\{question\}}, \textcolor{ForestGreen}{\{correct answer\}} \\ [0.5ex]
\textbf{Template}: \\ [0.5ex]
In this task, your goal is to judge a correct answer to a given question based on an associated paragraph and decide if it is a good correct answer or not. A good correct answer is one that correctly and completely answers the question. A bad correct answer addresses the question only partially or incorrectly. If you think the given correct answer is good, indicate it by responding "Yes". Otherwise, respond "No". There are only two types of responses possible: "Yes" and "No". \\  \\ Paragraph- \textcolor{MidnightBlue}{\{paragraph\}} Question: \textcolor{MidnightBlue}{\{question\}} Correct Answer: \textcolor{ForestGreen}{\{correct answer\}}

\paragraph{BC - 02} Source: NIV2 - Task 56 - Template 2 \\ [1ex]
\textbf{Input}: \textcolor{MidnightBlue}{\{paragraph\}}, \textcolor{MidnightBlue}{\{question\}}, \textcolor{ForestGreen}{\{correct answer\}} \\ [0.5ex]
\textbf{Template}: \\ [0.5ex]
You will be given a definition of a task first, then some input of the task. \\ In this task, your goal is to judge a correct answer to a given question based on an associated paragraph and decide if it is a good correct answer or not. A good correct answer is one that correctly and completely answers the question. A bad correct answer addresses the question only partially or incorrectly. If you think the given correct answer is good, indicate it by responding "Yes". Otherwise, respond "No". There are only two types of responses possible: "Yes" and "No". \\  \\ Paragraph- \textcolor{MidnightBlue}{\{paragraph\}} Question: \textcolor{MidnightBlue}{\{question\}} Correct Answer: \textcolor{ForestGreen}{\{correct answer\}} \\ Output:

\paragraph{BC - 03} Source: NIV2 - Task 56 - Template 3 \\ [1ex]
\textbf{Input}: \textcolor{MidnightBlue}{\{paragraph\}}, \textcolor{MidnightBlue}{\{question\}}, \textcolor{ForestGreen}{\{correct answer\}} \\ [0.5ex]
\textbf{Template}: \\ [0.5ex]
Definition: In this task, your goal is to judge a correct answer to a given question based on an associated paragraph and decide if it is a good correct answer or not. A good correct answer is one that correctly and completely answers the question. A bad correct answer addresses the question only partially or incorrectly. If you think the given correct answer is good, indicate it by responding "Yes". Otherwise, respond "No". There are only two types of responses possible: "Yes" and "No". \\ Input: Paragraph- \textcolor{MidnightBlue}{\{paragraph\}} Question: \textcolor{MidnightBlue}{\{question\}} Correct Answer: \textcolor{ForestGreen}{\{correct answer\}} \\ Output:

\paragraph{BC - 04} Source: NIV2 - Task 56 - Template 4 \\ [1ex]
\textbf{Input}: \textcolor{MidnightBlue}{\{paragraph\}}, \textcolor{MidnightBlue}{\{question\}}, \textcolor{ForestGreen}{\{correct answer\}} \\ [0.5ex]
\textbf{Template}: \\ [0.5ex]
Instructions: In this task, your goal is to judge a correct answer to a given question based on an associated paragraph and decide if it is a good correct answer or not. A good correct answer is one that correctly and completely answers the question. A bad correct answer addresses the question only partially or incorrectly. If you think the given correct answer is good, indicate it by responding "Yes". Otherwise, respond "No". There are only two types of responses possible: "Yes" and "No". \\ Input: Paragraph- \textcolor{MidnightBlue}{\{paragraph\}} Question: \textcolor{MidnightBlue}{\{question\}} Correct Answer: \textcolor{ForestGreen}{\{correct answer\}} \\ Output:

\paragraph{BC - 05} Source: NIV2 - Task 56 - Template 5 \\ [1ex]
\textbf{Input}: \textcolor{MidnightBlue}{\{paragraph\}}, \textcolor{MidnightBlue}{\{question\}}, \textcolor{ForestGreen}{\{correct answer\}} \\ [0.5ex]
\textbf{Template}: \\ [0.5ex]
In this task, your goal is to judge a correct answer to a given question based on an associated paragraph and decide if it is a good correct answer or not. A good correct answer is one that correctly and completely answers the question. A bad correct answer addresses the question only partially or incorrectly. If you think the given correct answer is good, indicate it by responding "Yes". Otherwise, respond "No". There are only two types of responses possible: "Yes" and "No". \\ Q: Paragraph- \textcolor{MidnightBlue}{\{paragraph\}} Question: \textcolor{MidnightBlue}{\{question\}} Correct Answer: \textcolor{ForestGreen}{\{correct answer\}} \\ A:

\paragraph{BC - 06} Source: NIV2 - Task 56 - Template 6 \\ [1ex]
\textbf{Input}: \textcolor{MidnightBlue}{\{paragraph\}}, \textcolor{MidnightBlue}{\{question\}}, \textcolor{ForestGreen}{\{correct answer\}} \\ [0.5ex]
\textbf{Template}: \\ [0.5ex]
Given the task definition and input, reply with output. In this task, your goal is to judge a correct answer to a given question based on an associated paragraph and decide if it is a good correct answer or not. A good correct answer is one that correctly and completely answers the question. A bad correct answer addresses the question only partially or incorrectly. If you think the given correct answer is good, indicate it by responding "Yes". Otherwise, respond "No". There are only two types of responses possible: "Yes" and "No". \\  \\ Paragraph- \textcolor{MidnightBlue}{\{paragraph\}} Question: \textcolor{MidnightBlue}{\{question\}} Correct Answer: \textcolor{ForestGreen}{\{correct answer\}} \\

\paragraph{BC - 07} Source: NIV2 - Task 56 - Template 7 \\ [1ex]
\textbf{Input}: \textcolor{MidnightBlue}{\{paragraph\}}, \textcolor{MidnightBlue}{\{question\}}, \textcolor{ForestGreen}{\{correct answer\}} \\ [0.5ex]
\textbf{Template}: \\ [0.5ex]
Teacher: In this task, your goal is to judge a correct answer to a given question based on an associated paragraph and decide if it is a good correct answer or not. A good correct answer is one that correctly and completely answers the question. A bad correct answer addresses the question only partially or incorrectly. If you think the given correct answer is good, indicate it by responding "Yes". Otherwise, respond "No". There are only two types of responses possible: "Yes" and "No". \\ Teacher: Now, understand the problem? Solve this instance: Paragraph- \textcolor{MidnightBlue}{\{paragraph\}} Question: \textcolor{MidnightBlue}{\{question\}} Correct Answer: \textcolor{ForestGreen}{\{correct answer\}} \\ Student:

\paragraph{BC - 08} Source: NIV2 - Task 56 - Template 8 \\ [1ex]
\textbf{Input}: \textcolor{MidnightBlue}{\{paragraph\}}, \textcolor{MidnightBlue}{\{question\}}, \textcolor{ForestGreen}{\{correct answer\}} \\ [0.5ex]
\textbf{Template}: \\ [0.5ex]
Q: In this task, your goal is to judge a correct answer to a given question based on an associated paragraph and decide if it is a good correct answer or not. A good correct answer is one that correctly and completely answers the question. A bad correct answer addresses the question only partially or incorrectly. If you think the given correct answer is good, indicate it by responding "Yes". Otherwise, respond "No". There are only two types of responses possible: "Yes" and "No". \\ Paragraph- \textcolor{MidnightBlue}{\{paragraph\}} Question: \textcolor{MidnightBlue}{\{question\}} Correct Answer: \textcolor{ForestGreen}{\{correct answer\}} \\ A:

\paragraph{BC - 09} Source: NIV2 - Task 56 - Template 9 \\ [1ex]
\textbf{Input}: \textcolor{MidnightBlue}{\{paragraph\}}, \textcolor{MidnightBlue}{\{question\}}, \textcolor{ForestGreen}{\{correct answer\}} \\ [0.5ex]
\textbf{Template}: \\ [0.5ex]
Detailed Instructions: In this task, your goal is to judge a correct answer to a given question based on an associated paragraph and decide if it is a good correct answer or not. A good correct answer is one that correctly and completely answers the question. A bad correct answer addresses the question only partially or incorrectly. If you think the given correct answer is good, indicate it by responding "Yes". Otherwise, respond "No". There are only two types of responses possible: "Yes" and "No". \\ Problem:Paragraph- \textcolor{MidnightBlue}{\{paragraph\}} Question: \textcolor{MidnightBlue}{\{question\}} Correct Answer: \textcolor{ForestGreen}{\{correct answer\}} \\ Solution:

\paragraph{BC - 10} Source: NIV2 - Task 56 - Template 10 \\ [1ex]
\textbf{Input}: \textcolor{MidnightBlue}{\{paragraph\}}, \textcolor{MidnightBlue}{\{question\}}, \textcolor{ForestGreen}{\{correct answer\}} \\ [0.5ex]
\textbf{Template}: \\ [0.5ex]
Detailed Instructions: In this task, your goal is to judge a correct answer to a given question based on an associated paragraph and decide if it is a good correct answer or not. A good correct answer is one that correctly and completely answers the question. A bad correct answer addresses the question only partially or incorrectly. If you think the given correct answer is good, indicate it by responding "Yes". Otherwise, respond "No". There are only two types of responses possible: "Yes" and "No". \\ Q: Paragraph- \textcolor{MidnightBlue}{\{paragraph\}} Question: \textcolor{MidnightBlue}{\{question\}} Correct Answer: \textcolor{ForestGreen}{\{correct answer\}} \\ A:

\paragraph{BC - 11} Source: Flan2021 - MultiRC - Template 1 \\ [1ex]
\textbf{Input}: \textcolor{MidnightBlue}{\{paragraph\}}, \textcolor{MidnightBlue}{\{question\}}, \textcolor{ForestGreen}{\{response\}} \\ [0.5ex]
\textbf{Template}: \\ [0.5ex]
\textcolor{MidnightBlue}{\{paragraph\}} \\  \\ Question: "\textcolor{MidnightBlue}{\{question\}}" \\  \\ Response: "\textcolor{ForestGreen}{\{response\}}" \\  \\ Does the response correctly answer the question? \\  \\

\paragraph{BC - 12} Source: Flan2021 - MultiRC - Template 2 \\ [1ex]
\textbf{Input}: \textcolor{MidnightBlue}{\{paragraph\}}, \textcolor{MidnightBlue}{\{question\}}, \textcolor{ForestGreen}{\{response\}} \\ [0.5ex]
\textbf{Template}: \\ [0.5ex]
\textcolor{MidnightBlue}{\{paragraph\}} \\  \\ Question: "\textcolor{MidnightBlue}{\{question\}}" \\  \\ Response: "\textcolor{ForestGreen}{\{response\}}" \\  \\ Based on the paragraph, is the response to the question is factually correct? \\  \\ 

\paragraph{BC - 13} Source: Flan2021 - MultiRC - Template 3 \\ [1ex]
\textbf{Input}: \textcolor{MidnightBlue}{\{paragraph\}}, \textcolor{MidnightBlue}{\{question\}}, \textcolor{ForestGreen}{\{response\}} \\ [0.5ex]
\textbf{Template}: \\ [0.5ex]
\textcolor{MidnightBlue}{\{paragraph\}} \\  \\ Question: "\textcolor{MidnightBlue}{\{question\}}" \\  \\ Answer: "\textcolor{ForestGreen}{\{response\}}" \\  \\ Is this answer correct? \\  \\ ...I think the answer is

\paragraph{BC - 14} Source: Flan2021 - MultiRC - Template 4 \\ [1ex]
\textbf{Input}: \textcolor{MidnightBlue}{\{paragraph\}}, \textcolor{MidnightBlue}{\{question\}}, \textcolor{ForestGreen}{\{response\}} \\ [0.5ex]
\textbf{Template}: \\ [0.5ex]
Paragraph: \textcolor{MidnightBlue}{\{paragraph\}} \\  \\ Question: "\textcolor{MidnightBlue}{\{question\}}" \\  \\ Answer: "\textcolor{ForestGreen}{\{response\}}" \\  \\ Based on the paragraph, choose if the answer is correct: \\  \\

\paragraph{BC - 15} Source: Flan2021 - MultiRC - Template 5 \\ [1ex]
\textbf{Input}: \textcolor{MidnightBlue}{\{paragraph\}}, \textcolor{MidnightBlue}{\{question\}}, \textcolor{ForestGreen}{\{response\}} \\ [0.5ex]
\textbf{Template}: \\ [0.5ex]
\textcolor{MidnightBlue}{\{paragraph\}} \\  \\ Choose from options: Based on the paragraph, does the response "\textcolor{ForestGreen}{\{response\}}" correctly answer the question "\textcolor{MidnightBlue}{\{question\}}"? \\  \\

\paragraph{BC - 16} Source: Flan2021 - MultiRC - Template 6 \\ [1ex]
\textbf{Input}: \textcolor{MidnightBlue}{\{paragraph\}}, \textcolor{MidnightBlue}{\{question\}}, \textcolor{ForestGreen}{\{response\}} \\ [0.5ex]
\textbf{Template}: \\ [0.5ex]
\textcolor{MidnightBlue}{\{paragraph\}} \\  \\ Choose your answer: According to the above paragraph, the correct answer to the question "\textcolor{MidnightBlue}{\{question\}}" is "\textcolor{ForestGreen}{\{response\}}"? \\  \\ 

\paragraph{BC - 17} Source: Flan2021 - MultiRC - Template 7 \\ [1ex]
\textbf{Input}: \textcolor{MidnightBlue}{\{paragraph\}}, \textcolor{MidnightBlue}{\{question\}}, \textcolor{ForestGreen}{\{response\}} \\ [0.5ex]
\textbf{Template}: \\ [0.5ex]
\textcolor{MidnightBlue}{\{paragraph\}} \\  \\ After reading the above, is "\textcolor{ForestGreen}{\{response\}}" the correct answer to the question "\textcolor{MidnightBlue}{\{question\}}"? \\  \\ 

\paragraph{BC - 18} Source: Flan2021 - MultiRC - Template 8 \\ [1ex]
\textbf{Input}: \textcolor{MidnightBlue}{\{paragraph\}}, \textcolor{MidnightBlue}{\{question\}}, \textcolor{ForestGreen}{\{response\}} \\ [0.5ex]
\textbf{Template}: \\ [0.5ex]
\textcolor{MidnightBlue}{\{paragraph\}} \\  \\ Question: "\textcolor{MidnightBlue}{\{question\}}" \\  \\ Answer: "\textcolor{ForestGreen}{\{response\}}" \\  \\ Is this answer to the question correct? \\ 

\subsubsection*{Alpaca}

\paragraph{QA/MC - 01} Source: Alpaca Tasks Collection \\ [1ex]
\textbf{Input}: \textcolor{MidnightBlue}{\{question\}}, \textcolor{ForestGreen}{\{options\}} \\ [0.5ex]
\textbf{Template}: \\ [0.5ex]
Below is an instruction that describes a task, paired with an input that provides further context. Write a response that appropriately completes the request. \\ \\
\#\#\# Instruction: \\
Select the correct letter in the parentheses. 
\\ \\
\#\#\# Input: \\
Question: \textcolor{MidnightBlue}{\{question\}} \\
\textcolor{ForestGreen}{\{options\}} \\ \\
\#\#\# Response: 

\paragraph{QA/MC - 02} Source: Alpaca Tasks Collection \\ [1ex]
\textbf{Input}: \textcolor{MidnightBlue}{\{question\}}, \textcolor{ForestGreen}{\{options\}} \\ [0.5ex]
\textbf{Template}: \\ [0.5ex]
Below is an instruction that describes a task, paired with an input that provides further context. Write a response that appropriately completes the request. \\ \\
\#\#\# Instruction: \\
Select the correct option from the following choices.
\\ \\
\#\#\# Input: \\
Question: \textcolor{MidnightBlue}{\{question\}} \\
\textcolor{ForestGreen}{\{options\}} \\ \\
\#\#\# Response:

\paragraph{QA/MC - 03} Source: Alpaca Tasks Collection \\ [1ex]
\textbf{Input}: \textcolor{MidnightBlue}{\{question\}}, \textcolor{ForestGreen}{\{options\}} \\ [0.5ex]
\textbf{Template}: \\ [0.5ex]
Below is an instruction that describes a task, paired with an input that provides further context. Write a response that appropriately completes the request. \\ \\
\#\#\# Instruction: \\
Answer this multiple choice question.
\\ \\
\#\#\# Input: \\
Question: \textcolor{MidnightBlue}{\{question\}} \\
\textcolor{ForestGreen}{\{options\}} \\ \\
\#\#\# Response:

\paragraph{QA/MC - 04} Source: Alpaca Tasks Collection \\ [1ex]
\textbf{Input}: \textcolor{MidnightBlue}{\{question\}}, \textcolor{ForestGreen}{\{options\}} \\ [0.5ex]
\textbf{Template}: \\ [0.5ex]
Below is an instruction that describes a task, paired with an input that provides further context. Write a response that appropriately completes the request. \\ \\
\#\#\# Instruction: \\
Read the answer choices and select the correct one.
\\ \\
\#\#\# Input: \\
Question: \textcolor{MidnightBlue}{\{question\}} \\
\textcolor{ForestGreen}{\{options\}} \\ \\
\#\#\# Response:

\paragraph{QA/MC - 05} Source: Alpaca Tasks Collection \\ [1ex]
\textbf{Input}: \textcolor{MidnightBlue}{\{question\}}, \textcolor{ForestGreen}{\{options\}} \\ [0.5ex]
\textbf{Template}: \\ [0.5ex]
Below is an instruction that describes a task, paired with an input that provides further context. Write a response that appropriately completes the request. \\ \\
\#\#\# Instruction: \\
Identify the correct answer from the choices below.
\\ \\
\#\#\# Input: \\
Question: \textcolor{MidnightBlue}{\{question\}} \\
\textcolor{ForestGreen}{\{options\}} \\ \\
\#\#\# Response:

\paragraph{QA/MC - 06} Source: Alpaca Tasks Collection \\ [1ex]
\textbf{Input}: \textcolor{MidnightBlue}{\{question\}}, \textcolor{ForestGreen}{\{options\}} \\ [0.5ex]
\textbf{Template}: \\ [0.5ex]
Below is an instruction that describes a task, paired with an input that provides further context. Write a response that appropriately completes the request. \\ \\
\#\#\# Instruction: \\
Determine which choice is correct and output it.
\\ \\
\#\#\# Input: \\
Question: \textcolor{MidnightBlue}{\{question\}} \\
\textcolor{ForestGreen}{\{options\}} \\ \\
\#\#\# Response:

\paragraph{QA/MC - 07} Source: Alpaca Tasks Collection \\ [1ex]
\textbf{Input}: \textcolor{MidnightBlue}{\{question\}}, \textcolor{ForestGreen}{\{options\}} \\ [0.5ex]
\textbf{Template}: \\ [0.5ex]
Below is an instruction that describes a task, paired with an input that provides further context. Write a response that appropriately completes the request. \\ \\
\#\#\# Instruction: \\
Refer to the given input and identify the correct answer.
\\ \\
\#\#\# Input: \\
Question: \textcolor{MidnightBlue}{\{question\}} \\
\textcolor{ForestGreen}{\{options\}} \\ \\
\#\#\# Response:

\paragraph{QA/MC - 08} Source: Alpaca Tasks Collection \\ [1ex]
\textbf{Input}: \textcolor{MidnightBlue}{\{question\}}, \textcolor{ForestGreen}{\{options\}} \\ [0.5ex]
\textbf{Template}: \\ [0.5ex]
Below is an instruction that describes a task, paired with an input that provides further context. Write a response that appropriately completes the request. \\ \\
\#\#\# Instruction: \\
From the given \textcolor{ForestGreen}{\{options length\}} options, select the one most relevant to the given input.
\\ \\
\#\#\# Input: \\
Question: \textcolor{MidnightBlue}{\{question\}} \\
\textcolor{ForestGreen}{\{options\}} \\ \\
\#\#\# Response: 

\paragraph{QA/MC - 09} Source: Alpaca Tasks Collection \\ [1ex]
\textbf{Input}: \textcolor{MidnightBlue}{\{question\}}, \textcolor{ForestGreen}{\{options\}} \\ [0.5ex]
\textbf{Template}: \\ [0.5ex]
Below is an instruction that describes a task, paired with an input that provides further context. Write a response that appropriately completes the request. \\ \\
\#\#\# Instruction: \\
Select the most optimal response.
\\ \\
\#\#\# Input: \\
Question: \textcolor{MidnightBlue}{\{question\}} \\
\textcolor{ForestGreen}{\{options\}} \\ \\
\#\#\# Response:

\paragraph{QA/MC - 10} Source: Alpaca Tasks Collection \\ [1ex]
\textbf{Input}: \textcolor{MidnightBlue}{\{question\}}, \textcolor{ForestGreen}{\{options\}} \\ [0.5ex]
\textbf{Template}: \\ [0.5ex]
Below is an instruction that describes a task, paired with an input that provides further context. Write a response that appropriately completes the request. \\ \\
\#\#\# Instruction: \\
Read the answer choices and select the correct one.
\\ \\
\#\#\# Input: \\
Question: \textcolor{MidnightBlue}{\{question\}} \\
\textcolor{ForestGreen}{\{options\}} \\ \\
\#\#\# Response: 

\paragraph{QA/MC - 11} Source: Alpaca Tasks Collection \\ [1ex]
\textbf{Input}: \textcolor{MidnightBlue}{\{question\}}, \textcolor{ForestGreen}{\{options\}} \\ [0.5ex]
\textbf{Template}: \\ [0.5ex]
Below is an instruction that describes a task, paired with an input that provides further context. Write a response that appropriately completes the request. \\ \\
\#\#\# Instruction: \\
Select the best answer out of given options.
\\ \\
\#\#\# Input: \\
Question: \textcolor{MidnightBlue}{\{question\}} \\
\textcolor{ForestGreen}{\{options\}} \\ \\
\#\#\# Response:

\paragraph{QA/MC - 12} Source: Alpaca Tasks Collection \\ [1ex]
\textbf{Input}: \textcolor{MidnightBlue}{\{question\}}, \textcolor{ForestGreen}{\{options\}} \\ [0.5ex]
\textbf{Template}: \\ [0.5ex]
Below is an instruction that describes a task, paired with an input that provides further context. Write a response that appropriately completes the request. \\ \\
\#\#\# Instruction: \\
Determine which of these options is the correct answer.
\\ \\
\#\#\# Input: \\
Question: \textcolor{MidnightBlue}{\{question\}} \\
\textcolor{ForestGreen}{\{options\}} \\ \\
\#\#\# Response:

\paragraph{QA/MC - 13} Source: Alpaca Tasks Collection \\ [1ex]
\textbf{Input}: \textcolor{MidnightBlue}{\{question\}}, \textcolor{ForestGreen}{\{options\}} \\ [0.5ex]
\textbf{Template}: \\ [0.5ex]
Below is an instruction that describes a task, paired with an input that provides further context. Write a response that appropriately completes the request. \\ \\
\#\#\# Instruction: \\
Choose the best option.
\\ \\
\#\#\# Input: \\
Question: \textcolor{MidnightBlue}{\{question\}} \\
\textcolor{ForestGreen}{\{options\}} \\ \\
\#\#\# Response:

\paragraph{QA/MC - 14} Source: Alpaca Tasks Collection \\ [1ex]
\textbf{Input}: \textcolor{MidnightBlue}{\{question\}}, \textcolor{ForestGreen}{\{options\}} \\ [0.5ex]
\textbf{Template}: \\ [0.5ex]
Below is an instruction that describes a task, paired with an input that provides further context. Write a response that appropriately completes the request. \\ \\
\#\#\# Instruction: \\
Choose the best option.
\\ \\
\#\#\# Input: \\
Question: \textcolor{MidnightBlue}{\{question\}} \\
\textcolor{ForestGreen}{\{options\}} \\ \\
\#\#\# Response:

\paragraph{QA/MC - 15} Source: Alpaca Tasks Collection \\ [1ex]
\textbf{Input}: \textcolor{MidnightBlue}{\{question\}}, \textcolor{ForestGreen}{\{options\}} \\ [0.5ex]
\textbf{Template}: \\ [0.5ex]
Below is an instruction that describes a task, paired with an input that provides further context. Write a response that appropriately completes the request. \\ \\
\#\#\# Instruction: \\
Select the best answer.
\\ \\
\#\#\# Input: \\
Question: \textcolor{MidnightBlue}{\{question\}} \\
\textcolor{ForestGreen}{\{options\}} \\ \\
\#\#\# Response:

\paragraph{QA/MC - 16} Source: Alpaca Tasks Collection \\ [1ex]
\textbf{Input}: \textcolor{MidnightBlue}{\{question\}}, \textcolor{ForestGreen}{\{options\}} \\ [0.5ex]
\textbf{Template}: \\ [0.5ex]
Below is an instruction that describes a task, paired with an input that provides further context. Write a response that appropriately completes the request. \\ \\
\#\#\# Instruction: \\
Choose the correct answer.
\\ \\
\#\#\# Input: \\
Question: \textcolor{MidnightBlue}{\{question\}} \\
\textcolor{ForestGreen}{\{options\}} \\ \\
\#\#\# Response: 

\paragraph{QA/MC - 17} Source: Alpaca Tasks Collection \\ [1ex]
\textbf{Input}: \textcolor{MidnightBlue}{\{question\}}, \textcolor{ForestGreen}{\{options\}} \\ [0.5ex]
\textbf{Template}: \\ [0.5ex]
Below is an instruction that describes a task, paired with an input that provides further context. Write a response that appropriately completes the request. \\ \\
\#\#\# Instruction: \\
Select the correct answer from a list.
\\ \\
\#\#\# Input: \\
Question: \textcolor{MidnightBlue}{\{question\}} \\
\textcolor{ForestGreen}{\{options\}} \\ \\
\#\#\# Response:

\paragraph{QA/MC - 18} Source: Alpaca Tasks Collection \\ [1ex]
\textbf{Input}: \textcolor{MidnightBlue}{\{question\}}, \textcolor{ForestGreen}{\{options\}} \\ [0.5ex]
\textbf{Template}: \\ [0.5ex]
Below is an instruction that describes a task, paired with an input that provides further context. Write a response that appropriately completes the request. \\ \\
\#\#\# Instruction: \\
Choose the best answer.
\\ \\
\#\#\# Input: \\
Question: \textcolor{MidnightBlue}{\{question\}} \\
\textcolor{ForestGreen}{\{options\}} \\ \\
\#\#\# Response:

\paragraph{QA/MC - 19} Source: Alpaca Tasks Collection \\ [1ex]
\textbf{Input}: \textcolor{MidnightBlue}{\{question\}}, \textcolor{ForestGreen}{\{options\}} \\ [0.5ex]
\textbf{Template}: \\ [0.5ex]
Below is an instruction that describes a task, paired with an input that provides further context. Write a response that appropriately completes the request. \\ \\
\#\#\# Instruction: \\
Choose the statement that best suits the given context.
\\ \\
\#\#\# Input: \\
Question: \textcolor{MidnightBlue}{\{question\}} \\
\textcolor{ForestGreen}{\{options\}} \\ \\
\#\#\# Response:

\paragraph{QA/MC - 20} Source: Alpaca Tasks Collection \\ [1ex]
\textbf{Input}: \textcolor{MidnightBlue}{\{question\}}, \textcolor{ForestGreen}{\{options\}} \\ [0.5ex]
\textbf{Template}: \\ [0.5ex]
Below is an instruction that describes a task, paired with an input that provides further context. Write a response that appropriately completes the request. \\ \\
\#\#\# Instruction: \\
Answer the question based on common sense and your knowledge.
\\ \\
\#\#\# Input: \\
Question: \textcolor{MidnightBlue}{\{question\}} \\
\textcolor{ForestGreen}{\{options\}} \\ \\
\#\#\# Response:

\paragraph{BC - 01} Source: Alpaca Tasks Collection \\ [1ex]
\textbf{Input}: \textcolor{MidnightBlue}{\{claim\}} \\ [0.5ex]
\textbf{Template}: \\ [0.5ex]
Below is an instruction that describes a task, paired with an input that provides further context. Write a response that appropriately completes the request. \\ \\
\#\#\# Instruction: \\
Determine if this claim is true or false:
\\ \\
\#\#\# Input: \\
Claim: \textcolor{MidnightBlue}{\{claim\}} \\ \\
\#\#\# Response:

\paragraph{BC - 02} Source: Alpaca Tasks Collection \\ [1ex]
\textbf{Input}: \textcolor{MidnightBlue}{\{sentence\}} \\ [0.5ex]
\textbf{Template}: \\ [0.5ex]
Below is an instruction that describes a task, paired with an input that provides further context. Write a response that appropriately completes the request. \\ \\
\#\#\# Instruction: \\
Is the following sentence true or false?
\\ \\
\#\#\# Input: \\
\textcolor{MidnightBlue}{\{sentence\}} \\ \\
\#\#\# Response:

\paragraph{BC - 03} Source: Alpaca Tasks Collection \\ [1ex]
\textbf{Input}: \textcolor{MidnightBlue}{\{statement\}} \\ [0.5ex]
\textbf{Template}: \\ [0.5ex]
Below is an instruction that describes a task, paired with an input that provides further context. Write a response that appropriately completes the request. \\ \\
\#\#\# Instruction: \\
Identify whether the following phrase is a true or false statement
\\ \\
\#\#\# Input: \\
\textcolor{MidnightBlue}{\{statement\}} \\ \\
\#\#\# Response:

\paragraph{BC - 04} Source: Alpaca Tasks Collection \\ [1ex]
\textbf{Input}: \textcolor{MidnightBlue}{\{statement\}} \\ [0.5ex]
\textbf{Template}: \\ [0.5ex]
Below is an instruction that describes a task, paired with an input that provides further context. Write a response that appropriately completes the request. \\ \\
\#\#\# Instruction: \\
Check if the following statement is true or false:
\\ \\
\#\#\# Input: \\
\textcolor{MidnightBlue}{\{statement\}} \\ \\
\#\#\# Response:

\paragraph{BC - 05} Source: Alpaca Tasks Collection \\ [1ex]
\textbf{Input}: \textcolor{MidnightBlue}{\{statement\}} \\ [0.5ex]
\textbf{Template}: \\ [0.5ex]
Below is an instruction that describes a task, paired with an input that provides further context. Write a response that appropriately completes the request. \\ \\
\#\#\# Instruction: \\
Classify the following statement as true or false:
\\ \\
\#\#\# Input: \\
\textcolor{MidnightBlue}{\{statement\}} \\ \\
\#\#\# Response:

\paragraph{BC - 06} Source: Alpaca Tasks Collection \\ [1ex]
\textbf{Input}: \textcolor{MidnightBlue}{\{statement\}} \\ [0.5ex]
\textbf{Template}: \\ [0.5ex]
Below is an instruction that describes a task, paired with an input that provides further context. Write a response that appropriately completes the request. \\ \\
\#\#\# Instruction: \\
Classify the following statement as true or false:
\\ \\
\#\#\# Input: \\
\textcolor{MidnightBlue}{\{statement\}} \\ \\
\#\#\# Response:

\paragraph{BC - 07} Source: Alpaca Tasks Collection \\ [1ex]
\textbf{Input}: \textcolor{MidnightBlue}{\{statement\}} \\ [0.5ex]
\textbf{Template}: \\ [0.5ex]
Below is an instruction that describes a task, paired with an input that provides further context. Write a response that appropriately completes the request. \\ \\
\#\#\# Instruction: \\
Do a fact check to confirm the accuracy of the statement and output true or false.
\\ \\
\#\#\# Input: \\
\textcolor{MidnightBlue}{\{statement\}} \\ \\
\#\#\# Response:

\paragraph{BC - 08} Source: Alpaca Tasks Collection \\ [1ex]
\textbf{Input}: \textcolor{MidnightBlue}{\{sentence\}} \\ [0.5ex]
\textbf{Template}: \\ [0.5ex]
Below is an instruction that describes a task, paired with an input that provides further context. Write a response that appropriately completes the request. \\ \\
\#\#\# Instruction: \\
Label whether an input sentence is true or false.
\\ \\
\#\#\# Input: \\
\textcolor{MidnightBlue}{\{sentence\}} \\ \\
\#\#\# Response:

\paragraph{BC - 09} Source: Alpaca Tasks Collection \\ [1ex]
\textbf{Input}: \textcolor{MidnightBlue}{\{sentence\}} \\ [0.5ex]
\textbf{Template}: \\ [0.5ex]
Below is an instruction that describes a task, paired with an input that provides further context. Write a response that appropriately completes the request. \\ \\
\#\#\# Instruction: \\
Indicate a yes or no answer to the given statement..
\\ \\
\#\#\# Input: \\
\textcolor{MidnightBlue}{\{sentence\}} \\ \\
\#\#\# Response:

\paragraph{BC - 10} Source: Alpaca Tasks Collection \\ [1ex]
\textbf{Input}: \textcolor{MidnightBlue}{\{sentence\}} \\ [0.5ex]
\textbf{Template}: \\ [0.5ex]
Below is an instruction that describes a task, paired with an input that provides further context. Write a response that appropriately completes the request. \\ \\
\#\#\# Instruction: \\
Evaluate the following proposal as a yes or no response.
\\ \\
\#\#\# Input: \\
\textcolor{MidnightBlue}{\{sentence\}} \\ \\
\#\#\# Response:

\paragraph{BC - 11} Source: Alpaca Tasks Collection \\ [1ex]
\textbf{Input}: \textcolor{MidnightBlue}{\{statement\}} \\ [0.5ex]
\textbf{Template}: \\ [0.5ex]
Below is an instruction that describes a task, paired with an input that provides further context. Write a response that appropriately completes the request. \\ \\
\#\#\# Instruction: \\
Respond to the following statement with a yes or no.
\\ \\
\#\#\# Input: \\
\textcolor{MidnightBlue}{\{statement\}} \\ \\
\#\#\# Response:

\subsubsection*{P3 (T0)}

\paragraph{QA - 01} Source: T0 Arc Challenge - Template 1 \\ [1ex]
\textbf{Input}: \textcolor{MidnightBlue}{\{question\}}, \textcolor{ForestGreen}{\{options\}} \\ [0.5ex]
\textbf{Template}: \\ [0.5ex]
Here's a problem to solve: \textcolor{MidnightBlue}{\{question\}} \\  \\ Among the 4 following options, which is the correct answer? \textcolor{ForestGreen}{\{options\}}

\paragraph{QA - 02} Source: T0 Arc Challenge - Template 2 \\ [1ex]
\textbf{Input}: \textcolor{MidnightBlue}{\{question\}}, \textcolor{ForestGreen}{\{options\}} \\ [0.5ex]
\textbf{Template}: \\ [0.5ex]
\textcolor{MidnightBlue}{\{question\}} \\  \\ Options:\textcolor{ForestGreen}{\{options\}}

\paragraph{QA - 03} Source: T0 Arc Challenge - Template 3 \\ [1ex]
\textbf{Input}: \textcolor{MidnightBlue}{\{question\}}, \textcolor{ForestGreen}{\{options\}} \\ [0.5ex]
\textbf{Template}: \\ [0.5ex]
I am hesitating between 4 options to answer the following question, which option should I choose? \\ Question: \textcolor{MidnightBlue}{\{question\}} \\ Possibilities:\textcolor{ForestGreen}{\{options\}}

\paragraph{QA - 04} Source: T0 Arc Challenge - Template 4 \\ [1ex]
\textbf{Input}: \textcolor{MidnightBlue}{\{question\}}, \textcolor{ForestGreen}{\{options\}} \\ [0.5ex]
\textbf{Template}: \\ [0.5ex]
I gave my students this multiple choice question: \textcolor{MidnightBlue}{\{question\}} \\  \\ Only one answer is correct among these 4 choices:\textcolor{ForestGreen}{\{options\}} \\  \\ Could you tell me which one is correct?

\paragraph{QA - 05} Source: T0 Arc Challenge - Template 5 \\ [1ex]
\textbf{Input}: \textcolor{MidnightBlue}{\{question\}}, \textcolor{ForestGreen}{\{options\}} \\ [0.5ex]
\textbf{Template}: \\ [0.5ex]
Pick the most correct option to answer the following question. \\  \\ \textcolor{MidnightBlue}{\{question\}} \\  \\ Options:\textcolor{ForestGreen}{\{options\}}

\paragraph{QA - 06} Source: T0 Cos e - Template 1 \\ [1ex]
\textbf{Input}: \textcolor{MidnightBlue}{\{question\}}, \textcolor{ForestGreen}{\{options\}} \\ [0.5ex]
\textbf{Template}: \\ [0.5ex]
\textcolor{MidnightBlue}{\{question\}} \\ Choose the most suitable option to answer the above question. \\ Options:\textcolor{ForestGreen}{\{options\}}

\paragraph{QA - 07} Source: T0 Cos e - Template 2 \\ [1ex]
\textbf{Input}: \textcolor{MidnightBlue}{\{question\}}, \textcolor{ForestGreen}{\{options\}} \\ [0.5ex]
\textbf{Template}: \\ [0.5ex]
\textcolor{MidnightBlue}{\{question\}} \\ Choose the most suitable option to answer the above question. \\ Options\textcolor{ForestGreen}{\{options\}}

\paragraph{QA - 08} Source: T0 Cos e - Template 3 \\ [1ex]
\textbf{Input}: \textcolor{MidnightBlue}{\{question\}}, \textcolor{ForestGreen}{\{options\}} \\ [0.5ex]
\textbf{Template}: \\ [0.5ex]
\textcolor{MidnightBlue}{\{question\}}\textcolor{ForestGreen}{\{options\}} \\ The best answer is:

\paragraph{QA - 09} Source: T0 Cos e - Template 4 \\ [1ex]
\textbf{Input}: \textcolor{MidnightBlue}{\{question\}}, \textcolor{ForestGreen}{\{options\}} \\ [0.5ex]
\textbf{Template}: \\ [0.5ex]
Pick the option in line with common sense to answer the question. \\ Question: \textcolor{MidnightBlue}{\{question\}} \\ Options:\textcolor{ForestGreen}{\{options\}} \\ The best answer is:

\paragraph{QA - 10} Source: T0 Cos e - Template 5 \\ [1ex]
\textbf{Input}: \textcolor{MidnightBlue}{\{question\}}, \textcolor{ForestGreen}{\{options\}} \\ [0.5ex]
\textbf{Template}: \\ [0.5ex]
Pick the option in line with common sense to answer the question. \\ Question: \textcolor{MidnightBlue}{\{question\}} \\ Options:\textcolor{ForestGreen}{\{options\}} \\

\paragraph{QA - 11} Source: T0 Cos e - Template 6 \\ [1ex]
\textbf{Input}: \textcolor{MidnightBlue}{\{question\}}, \textcolor{ForestGreen}{\{options\}} \\ [0.5ex]
\textbf{Template}: \\ [0.5ex]
Pick the option in line with common sense to answer the question. \\ Questions: \textcolor{MidnightBlue}{\{question\}} \\ Options:\textcolor{ForestGreen}{\{options\}}

\paragraph{QA - 12} Source: T0 OpenbookQA - Template 1 \\ [1ex]
\textbf{Input}: \textcolor{MidnightBlue}{\{question\}}, \textcolor{ForestGreen}{\{options\}} \\ [0.5ex]
\textbf{Template}: \\ [0.5ex]
\textcolor{MidnightBlue}{\{question\}} \\  \\ Choose an answer from this list:\textcolor{ForestGreen}{\{options\}}

\paragraph{QA - 13} Source: T0 OpenbookQA - Template 2 \\ [1ex]
\textbf{Input}: \textcolor{MidnightBlue}{\{question\}}, \textcolor{ForestGreen}{\{options\}} \\ [0.5ex]
\textbf{Template}: \\ [0.5ex]
\textcolor{MidnightBlue}{\{question\}} \\  \\ Which is the correct answer?\textcolor{ForestGreen}{\{options\}}

\paragraph{QA - 14} Source: T0 OpenbookQA - Template 3 \\ [1ex]
\textbf{Input}: \textcolor{MidnightBlue}{\{question\}}, \textcolor{ForestGreen}{\{options\}} \\ [0.5ex]
\textbf{Template}: \\ [0.5ex]
\textcolor{MidnightBlue}{\{question\}}\textcolor{ForestGreen}{\{options\}} \\ Is the right answer "\textcolor{ForestGreen}{\{options letter\}}"

\paragraph{QA - 15} Source: T0 OpenbookQA - Template 4 \\ [1ex]
\textbf{Input}: \textcolor{MidnightBlue}{\{question\}}, \textcolor{ForestGreen}{\{options\}} \\ [0.5ex]
\textbf{Template}: \\ [0.5ex]
\textcolor{MidnightBlue}{\{question\}} \\  \\ Choices:\textcolor{ForestGreen}{\{options\}}

\paragraph{QA - 16} Source: T0 OpenbookQA - Template 5 \\ [1ex]
\textbf{Input}: \textcolor{MidnightBlue}{\{question\}}, \textcolor{ForestGreen}{\{options\}} \\ [0.5ex]
\textbf{Template}: \\ [0.5ex]
\textcolor{MidnightBlue}{\{question\}}\textcolor{ForestGreen}{\{options\}}

\paragraph{QA - 17} Source: T0 OpenbookQA - Template 6 \\ [1ex]
\textbf{Input}: \textcolor{MidnightBlue}{\{question\}}, \textcolor{ForestGreen}{\{options\}} \\ [0.5ex]
\textbf{Template}: \\ [0.5ex]
\textcolor{MidnightBlue}{\{question\}}\textcolor{ForestGreen}{\{options\}} \\  \\ Which is the correct answer?

\paragraph{BC - 01} Source: T0 MultiRC - Template 1 \\ [1ex]
\textbf{Input}: \textcolor{MidnightBlue}{\{paragraph\}}, \textcolor{MidnightBlue}{\{question\}}, \textcolor{ForestGreen}{\{answer\}} \\ [0.5ex]
\textbf{Template}: \\ [0.5ex]
\textcolor{MidnightBlue}{\{paragraph\}} \\  \\ Question: \textcolor{MidnightBlue}{\{question\}} \\ I found this answer "\textcolor{ForestGreen}{\{answer\}}". Is that correct? Yes or no?

\paragraph{BC - 02} Source: T0 MultiRC - Template 2 \\ [1ex]
\textbf{Input}: \textcolor{MidnightBlue}{\{paragraph\}}, \textcolor{MidnightBlue}{\{question\}}, \textcolor{ForestGreen}{\{answer\}} \\ [0.5ex]
\textbf{Template}: \\ [0.5ex]
\textcolor{MidnightBlue}{\{paragraph\}} \\ Based on the previous passage, \textcolor{MidnightBlue}{\{question\}} \\ Is "\textcolor{ForestGreen}{\{answer\}}" a correct answer?

\paragraph{BC - 03} Source: T0 MultiRC - Template 3 \\ [1ex]
\textbf{Input}: \textcolor{MidnightBlue}{\{paragraph\}}, \textcolor{MidnightBlue}{\{question\}}, \textcolor{ForestGreen}{\{answer\}} \\ [0.5ex]
\textbf{Template}: \\ [0.5ex]
\textcolor{MidnightBlue}{\{paragraph\}} \\ Question: \textcolor{MidnightBlue}{\{question\}} \\  \\ I am grading my students' exercises. Is the answer "\textcolor{ForestGreen}{\{answer\}}" correct?

\paragraph{BC - 04} Source: T0 MultiRC - Template 4 \\ [1ex]
\textbf{Input}: \textcolor{MidnightBlue}{\{paragraph\}}, \textcolor{MidnightBlue}{\{question\}}, \textcolor{ForestGreen}{\{answer\}} \\ [0.5ex]
\textbf{Template}: \\ [0.5ex]
\textcolor{MidnightBlue}{\{paragraph\}} \\ \textcolor{MidnightBlue}{\{question\}} \\ Would it be good to answer "\textcolor{ForestGreen}{\{answer\}}"?

\paragraph{BC - 05} Source: T0 MultiRC - Template 5 \\ [1ex]
\textbf{Input}: \textcolor{MidnightBlue}{\{paragraph\}}, \textcolor{MidnightBlue}{\{question\}}, \textcolor{ForestGreen}{\{answer\}} \\ [0.5ex]
\textbf{Template}: \\ [0.5ex]
\textcolor{MidnightBlue}{\{paragraph\}} \\ Question: \textcolor{MidnightBlue}{\{question\}} \\ Is it "\textcolor{ForestGreen}{\{answer\}}"?

\paragraph{BC - 06} Source: T0 MultiRC - Template 6 \\ [1ex]
\textbf{Input}: \textcolor{MidnightBlue}{\{paragraph\}}, \textcolor{MidnightBlue}{\{question\}}, \textcolor{ForestGreen}{\{answer\}} \\ [0.5ex]
\textbf{Template}: \\ [0.5ex]
\textcolor{MidnightBlue}{\{paragraph\}} \\  \\ Decide whether"\textcolor{ForestGreen}{\{answer\}}" is a valid answer to the following question: \\ \textcolor{MidnightBlue}{\{question\}} \\ Answer yes or no.

\paragraph{BC - 07} Source: T0 MultiRC - Template 7 \\ [1ex]
\textbf{Input}: \textcolor{MidnightBlue}{\{paragraph\}}, \textcolor{MidnightBlue}{\{question\}}, \textcolor{ForestGreen}{\{answer\}} \\ [0.5ex]
\textbf{Template}: \\ [0.5ex]
\textcolor{MidnightBlue}{\{paragraph\}} \\ Question: \textcolor{MidnightBlue}{\{question\}} \\ Is the correct answer "\textcolor{ForestGreen}{\{answer\}}"?

\paragraph{BC - 08} Source: T0 MultiRC - Template 8 \\ [1ex]
\textbf{Input}: \textcolor{MidnightBlue}{\{paragraph\}}, \textcolor{MidnightBlue}{\{question\}}, \textcolor{ForestGreen}{\{answer\}} \\ [0.5ex]
\textbf{Template}: \\ [0.5ex]
Is "\textcolor{ForestGreen}{\{answer\}}" a correct answer to the following question? \\ Question: \textcolor{MidnightBlue}{\{question\}} \\  \\ Rely on the following text: \textcolor{MidnightBlue}{\{paragraph\}}

\paragraph{BC - 09} Source: T0 MultiRC - Template 9 \\ [1ex]
\textbf{Input}: \textcolor{MidnightBlue}{\{paragraph\}}, \textcolor{MidnightBlue}{\{question\}}, \textcolor{ForestGreen}{\{answer\}} \\ [0.5ex]
\textbf{Template}: \\ [0.5ex]
\textcolor{MidnightBlue}{\{paragraph\}} \\  \\ Question: \textcolor{MidnightBlue}{\{question\}} \\ I think "\textcolor{ForestGreen}{\{answer\}}" is a valid answer. Could you confirm? Yes or no?

\paragraph{BC - 10} Source: T0 MultiRC - Template 10 \\ [1ex]
\textbf{Input}: \textcolor{MidnightBlue}{\{paragraph\}}, \textcolor{MidnightBlue}{\{question\}}, \textcolor{ForestGreen}{\{answer\}} \\ [0.5ex]
\textbf{Template}: \\ [0.5ex]
\textcolor{MidnightBlue}{\{paragraph\}} \\ \textcolor{MidnightBlue}{\{question\}} \\ I was going to say "\textcolor{ForestGreen}{\{answer\}}". Does that sound right?

\paragraph{MC - 1} Source: T0 DBPedia - Template 1 \\ [1ex]
\textbf{Input}: \textcolor{MidnightBlue}{\{question\}}, \textcolor{ForestGreen}{\{categories\}} \\ [0.5ex]
\textbf{Template}: \\ [0.5ex]
\textcolor{MidnightBlue}{\{question\}} Given a list of categories: \textcolor{ForestGreen}{\{categories\}}, what category does the paragraph belong to?

\paragraph{MC - 2} Source: T0 DBPedia - Template 2 \\ [1ex]
\textbf{Input}: \textcolor{MidnightBlue}{\{question\}}, \textcolor{ForestGreen}{\{categories\}} \\ [0.5ex]
\textbf{Template}: \\ [0.5ex]
Pick one category for the following text. The options are - \textcolor{ForestGreen}{\{categories\}}. \textcolor{MidnightBlue}{\{question\}}

\paragraph{MC - 3} Source: T0 DBPedia - Template 3 \\ [1ex]
\textbf{Input}: \textcolor{MidnightBlue}{\{question\}}, \textcolor{ForestGreen}{\{categories\}} \\ [0.5ex]
\textbf{Template}: \\ [0.5ex]
\textcolor{MidnightBlue}{\{question\}} Given a choice of categories \textcolor{ForestGreen}{\{categories\}}, the text refers to which one?

\paragraph{MC - 4} Source: T0 TREC - Template 1 \\ [1ex]
\textbf{Input}: \textcolor{MidnightBlue}{\{question\}}, \textcolor{ForestGreen}{\{categories\}} \\ [0.5ex]
\textbf{Template}: \\ [0.5ex]
Categories: \textcolor{ForestGreen}{\{categories\}} \\  \\ What category best describes: \textcolor{MidnightBlue}{\{question\}} \\ Answer:

\paragraph{MC - 5} Source: T0 TREC - Template 2 \\ [1ex]
\textbf{Input}: \textcolor{MidnightBlue}{\{question\}}, \textcolor{ForestGreen}{\{categories\}} \\ [0.5ex]
\textbf{Template}: \\ [0.5ex]
Question: \textcolor{MidnightBlue}{\{question\}} \\  \\ Descriptors: \textcolor{ForestGreen}{\{categories\}} \\  \\ Best Descriptor?

\paragraph{MC - 6} Source: T0 TREC - Template 3 \\ [1ex]
\textbf{Input}: \textcolor{MidnightBlue}{\{question\}}, \textcolor{ForestGreen}{\{categories\}} \\ [0.5ex]
\textbf{Template}: \\ [0.5ex]
Which category best describes the following question: \textcolor{MidnightBlue}{\{question\}} \\  \\ Choose from the following list: \\ \textcolor{ForestGreen}{\{categories\}}

\paragraph{MC - 7} Source: T0 TREC - Template 4 \\ [1ex]
\textbf{Input}: \textcolor{MidnightBlue}{\{question\}}, \textcolor{ForestGreen}{\{categories\}} \\ [0.5ex]
\textbf{Template}: \\ [0.5ex]
\textcolor{MidnightBlue}{\{question\}}Is this asking about \textcolor{ForestGreen}{\{categories\}}?

\paragraph{MC - 8} Source: T0 TREC - Template 1 \\ [1ex]
\textbf{Input}: \textcolor{MidnightBlue}{\{question\}}, \textcolor{ForestGreen}{\{categories\}} \\ [0.5ex]
\textbf{Template}: \\ [0.5ex]
Is the following question asking about  \textcolor{ForestGreen}{\{categories\}}? \\  \\ \textcolor{MidnightBlue}{\{question\}}

\subsection{Unobserved Instructions}

\begin{figure}[h]
    \centering
    \includegraphics[width=140mm]{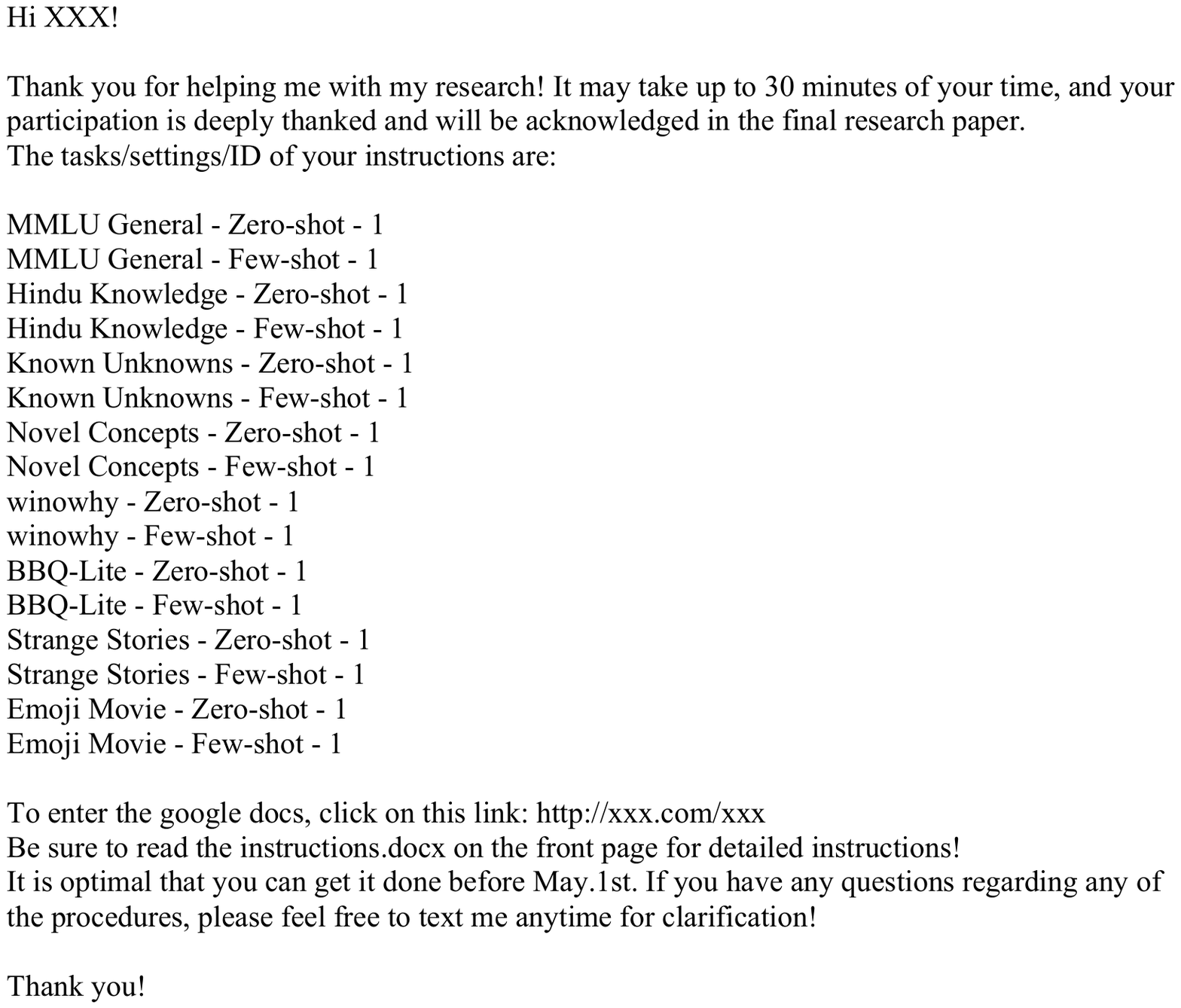}
    \caption{Invitation note send to participant}
    \label{fig:invitation}
\end{figure}

\begin{figure}[h]
    \centering
    \includegraphics[width=140mm]{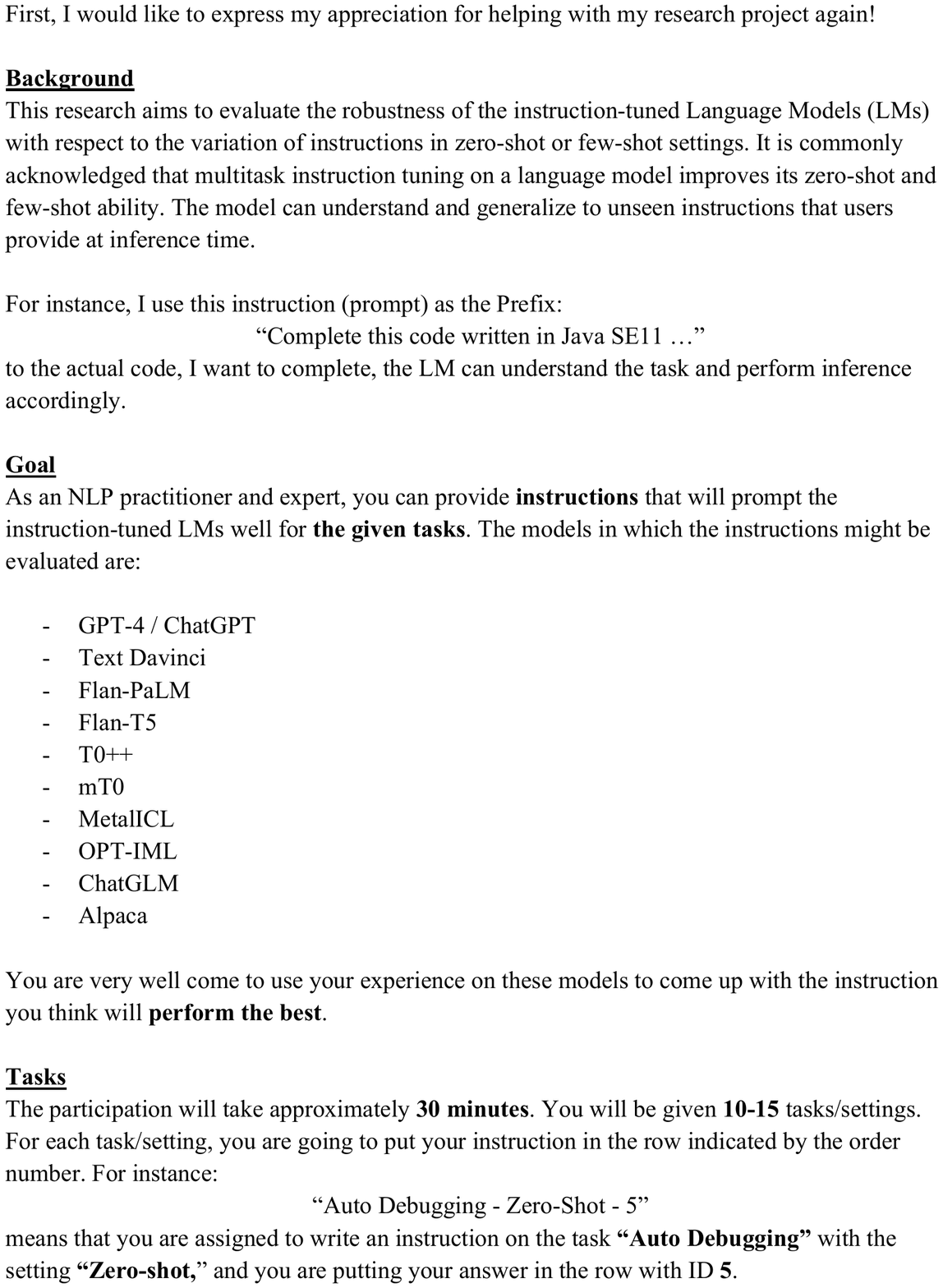}
    \caption{The first page of the instruction given to the annotator}
    \label{fig:collection_instruction_p1}
\end{figure}

\begin{figure}[h]
    \centering
    \includegraphics[width=140mm]{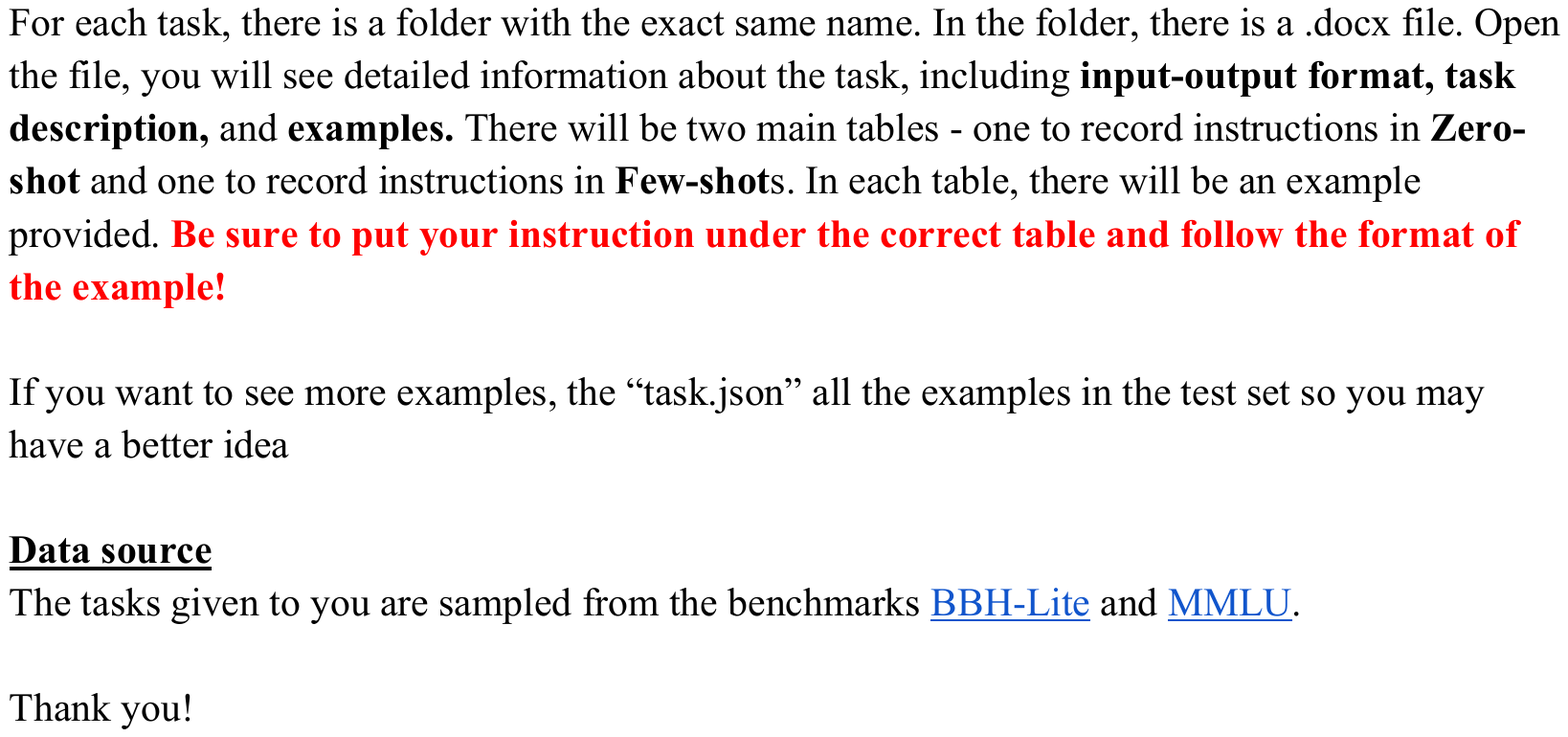}
    \caption{The second page of the instruction given to the annotator}
    \label{fig:collection_instruction_p2}
\end{figure}

\begin{figure}
    \centering
    \includegraphics[width=140mm]{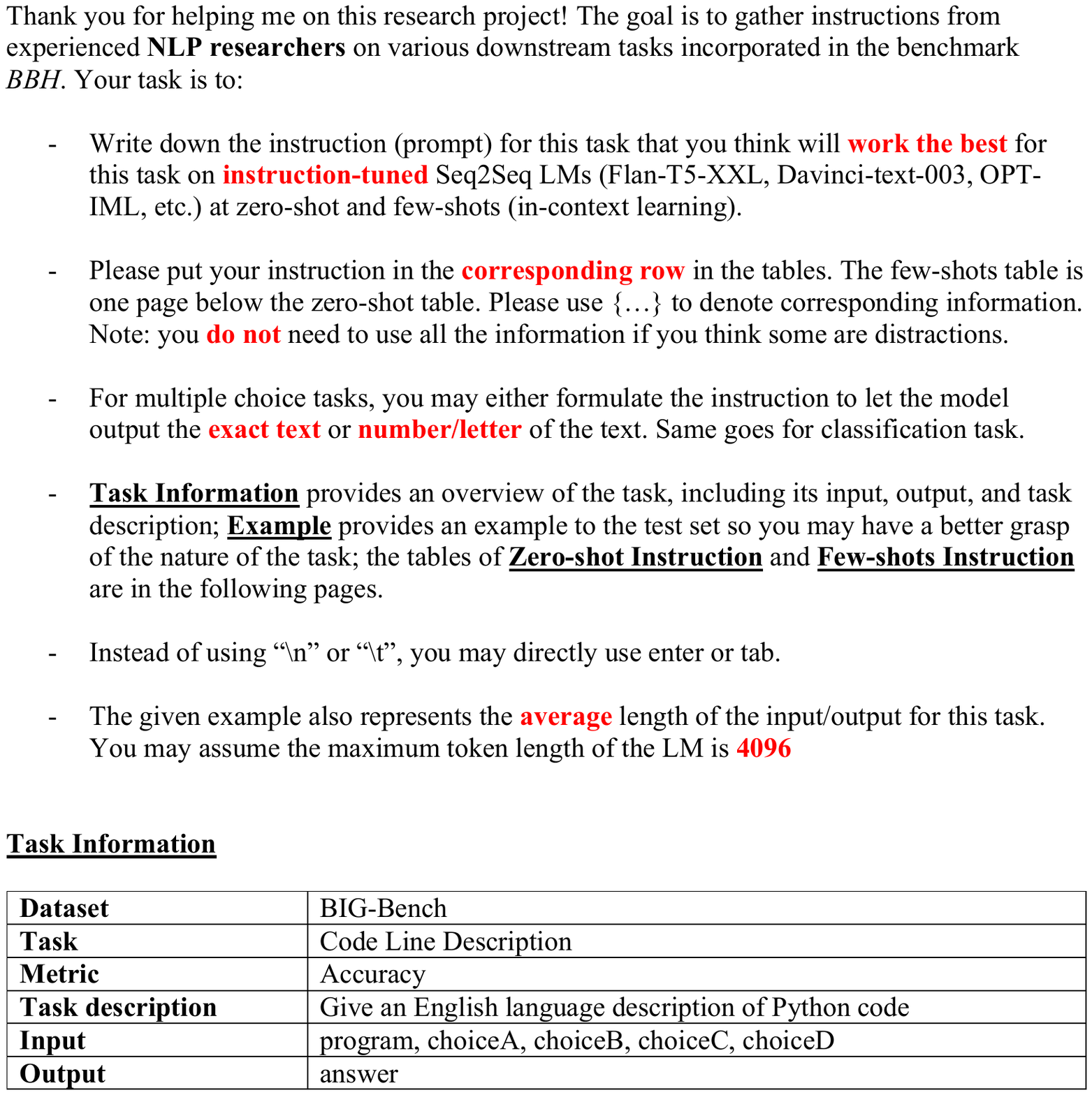}
    \caption{The first page of the dataset information}
    \label{fig:task_pg1}
\end{figure}

\begin{figure}
    \centering
    \includegraphics[width=140mm]{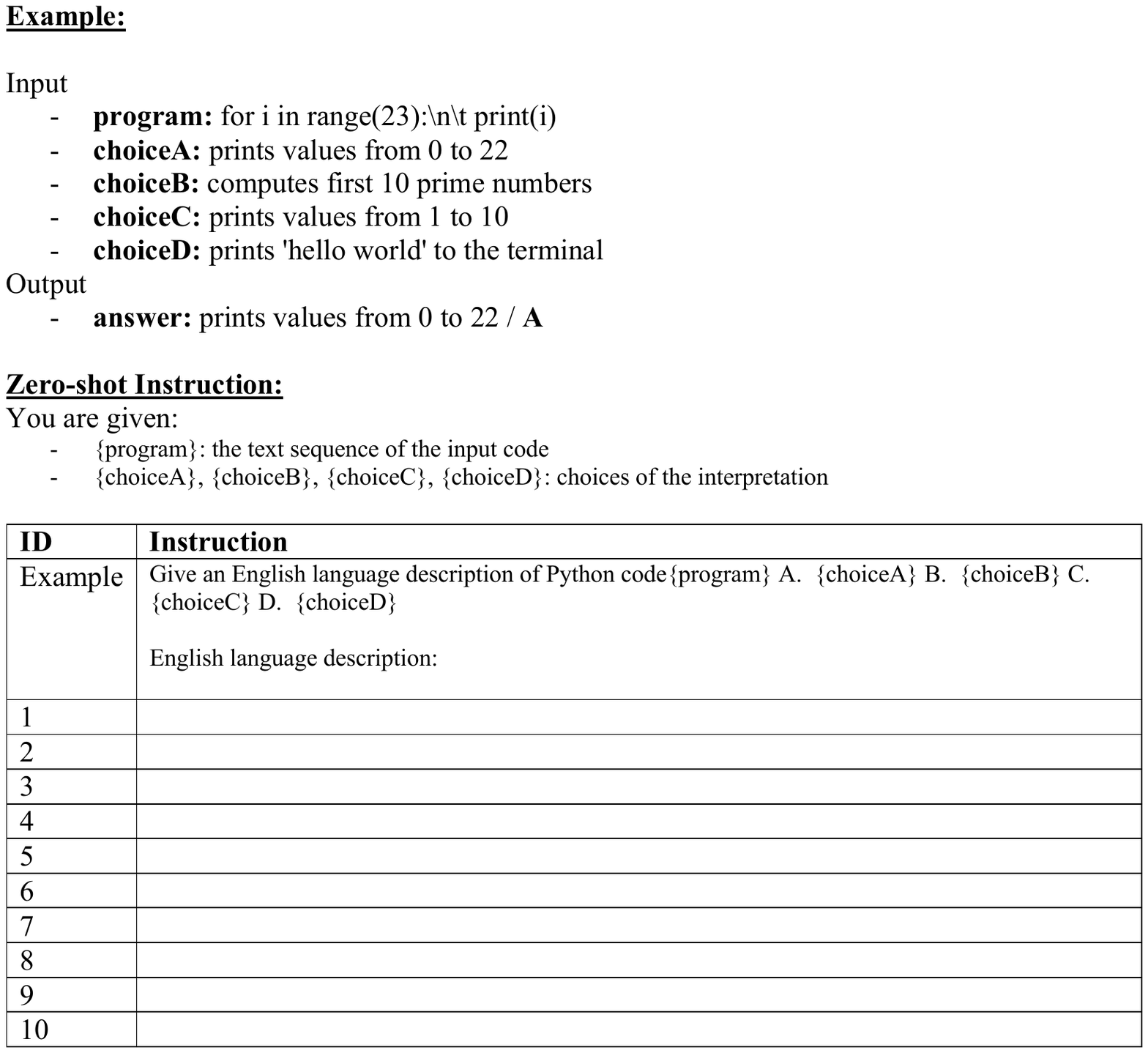}
    \caption{The second page of the dataset information}
    \label{fig:task_pg2}
\end{figure}

%With the agreement of participation, we allocate each annotator to provide one 
We collected novel, unobserved instructions---i.e.,, not seen in training---by enlisting researchers in NLP to write instructions for tasks \emph{de novo}.
To facilitate this we showed each annotator one zero-shot instruction and its few-shot form for \textsc{MMLU} and 12 datasets in \textsc{BBL} based on their field of expertise. We sent out an invitation message %through Email/Slack,  
to prospective participants, which contained a brief introduction to the goal of the research; we reproduce the full invitation in Figure \ref{fig:invitation}. 
%By clicking the link, the 
The in-line link redirected annotators to a designated Google Drive folder which included a detailed description of the procedure (see Figure \ref{fig:collection_instruction_p1} and Figure \ref{fig:collection_instruction_p2}). 
For each dataset, we provided detailed information about the task including the description, input-output format, demonstration instruction, and some examples (Shown by Figure \ref{fig:task_pg1} and Figure \ref{fig:task_pg2}). 
We asked participants %are asked 
to provide a prompt and its few-shot form for this task in the corresponding row of the table.\footnote{To evaluate Alpaca, we matched collected  instructions to corresponding templates that Alapca with which Alpaca was trained.} 
%To ensure independence, the prompt filled in by other participants are invisible. 
Participants were not shown prompts written by others, to preserve independence. 
Below we %provide 
reproduce all unobserved instructions that we collected for each benchmark task.

\subsubsection*{MMLU}

\paragraph{Unobserved - 01} Source: Annotator \\ [1ex]
\textbf{Input}: \textcolor{MidnightBlue}{\{question\}}, \textcolor{ForestGreen}{\{choiceA\}}, \textcolor{ForestGreen}{\{choiceB\}},\textcolor{ForestGreen}{\{choiceC\}},\textcolor{ForestGreen}{\{choiceD\}} \\ [0.5ex]
\textbf{Template}: \\ [0.5ex]
Please act as a domain expert to choose the most suitable answer from the given choices to the question below. Question: \textcolor{MidnightBlue}{\{question\}}. Choices: A. \textcolor{ForestGreen}{\{choiceA\}}  B. \textcolor{ForestGreen}{\{choiceB\}} C. \textcolor{ForestGreen}{\{choiceC\}} D. \textcolor{ForestGreen}{\{choiceD\}} \\ Please answer the question with your choice only without any other words.

\paragraph{Unobserved - 02} Source: Annotator \\ [1ex]
\textbf{Input}: \textcolor{MidnightBlue}{\{question\}}, \textcolor{ForestGreen}{\{choiceA\}}, \textcolor{ForestGreen}{\{choiceB\}},\textcolor{ForestGreen}{\{choiceC\}},\textcolor{ForestGreen}{\{choiceD\}} \\ [0.5ex]
\textbf{Template}: \\ [0.5ex]
Solve the question with professional knowledge and output the best option for the question from "A", "B", "C", "D" without other words: \\ Question: \textcolor{MidnightBlue}{\{question\}} \\ Options: \\ A: \textcolor{ForestGreen}{\{choiceA\}} \\ B: \textcolor{ForestGreen}{\{choiceB\}} \\ C: \textcolor{ForestGreen}{\{choiceC\}} \\ D: \textcolor{ForestGreen}{\{choiceD\}} \\ Answer:

\paragraph{Unobserved - 03} Source: Annotator \\ [1ex]
\textbf{Input}: \textcolor{MidnightBlue}{\{question\}}, \textcolor{ForestGreen}{\{choiceA\}}, \textcolor{ForestGreen}{\{choiceB\}},\textcolor{ForestGreen}{\{choiceC\}},\textcolor{ForestGreen}{\{choiceD\}} \\ [0.5ex]
\textbf{Template}: \\ [0.5ex]
Solve the question which requires deep understanding to the field. \textcolor{MidnightBlue}{\{question\}} \\ Choose from: \\ A: \textcolor{ForestGreen}{\{choiceA\}} \\ B: \textcolor{ForestGreen}{\{choiceB\}} \\ C: \textcolor{ForestGreen}{\{choiceC\}} \\ D: \textcolor{ForestGreen}{\{choiceD\}} \\ Answer:

\paragraph{Unobserved - 04} Source: Annotator \\ [1ex]
\textbf{Input}: \textcolor{MidnightBlue}{\{question\}}, \textcolor{ForestGreen}{\{choiceA\}}, \textcolor{ForestGreen}{\{choiceB\}},\textcolor{ForestGreen}{\{choiceC\}},\textcolor{ForestGreen}{\{choiceD\}} \\ [0.5ex]
\textbf{Template}: \\ [0.5ex]
\textcolor{MidnightBlue}{\{question\}} (A) \textcolor{ForestGreen}{\{choiceA\}} (B) \textcolor{ForestGreen}{\{choiceB\}} (C) \textcolor{ForestGreen}{\{choiceC\}} (D) \textcolor{ForestGreen}{\{choiceD\}} \\ The correct answer to this question is (

\paragraph{Unobserved - 05} Source: Annotator \\ [1ex]
\textbf{Input}: \textcolor{MidnightBlue}{\{question\}}, \textcolor{ForestGreen}{\{choiceA\}}, \textcolor{ForestGreen}{\{choiceB\}},\textcolor{ForestGreen}{\{choiceC\}},\textcolor{ForestGreen}{\{choiceD\}} \\ [0.5ex]
\textbf{Template}: \\ [0.5ex]
\textcolor{MidnightBlue}{\{question\}} \\  \\ A. \textcolor{ForestGreen}{\{choiceA\}} B. \textcolor{ForestGreen}{\{choiceB\}} C. \textcolor{ForestGreen}{\{choiceC\}} D. \textcolor{ForestGreen}{\{choiceD\}} \\ I know exactly the answer to this question! The correct choice is

\paragraph{Unobserved - 06} Source: Annotator \\ [1ex]
\textbf{Input}: \textcolor{MidnightBlue}{\{question\}}, \textcolor{ForestGreen}{\{choiceA\}}, \textcolor{ForestGreen}{\{choiceB\}},\textcolor{ForestGreen}{\{choiceC\}},\textcolor{ForestGreen}{\{choiceD\}} \\ [0.5ex]
\textbf{Template}: \\ [0.5ex]
You are given multiple-choice questions from a variety of domains. For each question, please select an answer from A, B, C, and D, and explain your reasoning. \\  \\ Question: \textcolor{MidnightBlue}{\{question\}} \\ The options are: \\ A: \textcolor{ForestGreen}{\{choiceA\}} \\ B: \textcolor{ForestGreen}{\{choiceB\}} \\ C: \textcolor{ForestGreen}{\{choiceC\}} \\ D: \textcolor{ForestGreen}{\{choiceD\}} \\  \\ Answer:

\paragraph{Unobserved - 07} Source: Annotator \\ [1ex]
\textbf{Input}: \textcolor{MidnightBlue}{\{question\}}, \textcolor{ForestGreen}{\{choiceA\}}, \textcolor{ForestGreen}{\{choiceB\}},\textcolor{ForestGreen}{\{choiceC\}},\textcolor{ForestGreen}{\{choiceD\}} \\ [0.5ex]
\textbf{Template}: \\ [0.5ex]
Please provide the correct answer to the following question, which requires expert level knowledge by choosing one of the options below and outputting it as your answer: \\  \\ Question: \textcolor{MidnightBlue}{\{question\}} \\ Options \\ A: \textcolor{ForestGreen}{\{choiceA\}} \\ B: \textcolor{ForestGreen}{\{choiceB\}} \\ C: \textcolor{ForestGreen}{\{choiceC\}} \\ D: \textcolor{ForestGreen}{\{choiceD\}} \\ Your answer:

\paragraph{Unobserved - 08} Source: Annotator \\ [1ex]
\textbf{Input}: \textcolor{MidnightBlue}{\{question\}}, \textcolor{ForestGreen}{\{choiceA\}}, \textcolor{ForestGreen}{\{choiceB\}},\textcolor{ForestGreen}{\{choiceC\}},\textcolor{ForestGreen}{\{choiceD\}} \\ [0.5ex]
\textbf{Template}: \\ [0.5ex]
\textcolor{MidnightBlue}{\{question\}} \\ Options: \\ 	- A \textcolor{ForestGreen}{\{choiceA\}} \\ 	- B \textcolor{ForestGreen}{\{choiceB\}} \\ 	- C \textcolor{ForestGreen}{\{choiceC\}} \\ 	- D \textcolor{ForestGreen}{\{choiceD\}} \\ Which option is correct?:

\paragraph{Unobserved - 09} Source: Annotator \\ [1ex]
\textbf{Input}: \textcolor{MidnightBlue}{\{question\}}, \textcolor{ForestGreen}{\{choiceA\}}, \textcolor{ForestGreen}{\{choiceB\}},\textcolor{ForestGreen}{\{choiceC\}},\textcolor{ForestGreen}{\{choiceD\}} \\ [0.5ex]
\textbf{Template}: \\ [0.5ex]
Given the question: \textcolor{MidnightBlue}{\{question\}}, and the choices for the answer are A. \textcolor{ForestGreen}{\{choiceA\}}, B. \textcolor{ForestGreen}{\{choiceB\}}, C. \textcolor{ForestGreen}{\{choiceC\}}, D. \textcolor{ForestGreen}{\{choiceD\}}. Output one of A, B, C, and D to indicate the correct choice. The correct choice is:

\paragraph{Unobserved - 10} Source: Annotator \\ [1ex]
\textbf{Input}: \textcolor{MidnightBlue}{\{question\}}, \textcolor{ForestGreen}{\{choiceA\}}, \textcolor{ForestGreen}{\{choiceB\}},\textcolor{ForestGreen}{\{choiceC\}},\textcolor{ForestGreen}{\{choiceD\}} \\ [0.5ex]
\textbf{Template}: \\ [0.5ex]
What is the answer to the question: \textcolor{MidnightBlue}{\{question\}} A. \textcolor{ForestGreen}{\{choiceA\}}, B. \textcolor{ForestGreen}{\{choiceB\}}, C. \textcolor{ForestGreen}{\{choiceC\}}, D. \textcolor{ForestGreen}{\{choiceD\}}

\paragraph{Unobserved - 11} Source: Annotator \\ [1ex]
\textbf{Input}: \textcolor{MidnightBlue}{\{question\}}, \textcolor{ForestGreen}{\{choiceA\}}, \textcolor{ForestGreen}{\{choiceB\}},\textcolor{ForestGreen}{\{choiceC\}},\textcolor{ForestGreen}{\{choiceD\}} \\ [0.5ex]
\textbf{Template}: \\ [0.5ex]
You are given a question that requires knowledge from a specific domain. Question: \textcolor{MidnightBlue}{\{question\}}. Select tha answer from A. '\textcolor{ForestGreen}{\{choiceA\}}', B. '\textcolor{ForestGreen}{\{choiceB\}}', C. '\textcolor{ForestGreen}{\{choiceC\}}', D. and '\textcolor{ForestGreen}{\{choiceD\}}'. Answer:

\paragraph{Unobserved - 12} Source: Annotator \\ [1ex]
\textbf{Input}: \textcolor{MidnightBlue}{\{question\}}, \textcolor{ForestGreen}{\{choiceA\}}, \textcolor{ForestGreen}{\{choiceB\}},\textcolor{ForestGreen}{\{choiceC\}},\textcolor{ForestGreen}{\{choiceD\}} \\ [0.5ex]
\textbf{Template}: \\ [0.5ex]
I want to know the answer to this question: \textcolor{MidnightBlue}{\{question\}}. Please select from the following: A. \textcolor{ForestGreen}{\{choiceA\}}, B. \textcolor{ForestGreen}{\{choiceB\}}, C. \textcolor{ForestGreen}{\{choiceC\}}, D. \textcolor{ForestGreen}{\{choiceD\}}. Indicate your choice with the letter.

\paragraph{Unobserved - 13} Source: Annotator \\ [1ex]
\textbf{Input}: \textcolor{MidnightBlue}{\{question\}}, \textcolor{ForestGreen}{\{choiceA\}}, \textcolor{ForestGreen}{\{choiceB\}},\textcolor{ForestGreen}{\{choiceC\}},\textcolor{ForestGreen}{\{choiceD\}} \\ [0.5ex]
\textbf{Template}: \\ [0.5ex]
Question: \textcolor{MidnightBlue}{\{question\}}. Choices: A. \textcolor{ForestGreen}{\{choiceA\}}, B. \textcolor{ForestGreen}{\{choiceB\}}, C. \textcolor{ForestGreen}{\{choiceC\}}, D. \textcolor{ForestGreen}{\{choiceD\}}. Answer:

\paragraph{Unobserved - 14} Source: Annotator \\ [1ex]
\textbf{Input}: \textcolor{MidnightBlue}{\{question\}}, \textcolor{ForestGreen}{\{choiceA\}}, \textcolor{ForestGreen}{\{choiceB\}},\textcolor{ForestGreen}{\{choiceC\}},\textcolor{ForestGreen}{\{choiceD\}} \\ [0.5ex]
\textbf{Template}: \\ [0.5ex]
Task: Multiple-choice question answering. \\ Question: \textcolor{MidnightBlue}{\{question\}} \\ Choices: (A) \textcolor{ForestGreen}{\{choiceA\}} (B) \textcolor{ForestGreen}{\{choiceB\}} (C) \textcolor{ForestGreen}{\{choiceC\}} (D) \textcolor{ForestGreen}{\{choiceD\}} \\ Answer: (

\paragraph{Unobserved - 15} Source: Annotator \\ [1ex]
\textbf{Input}: \textcolor{MidnightBlue}{\{question\}}, \textcolor{ForestGreen}{\{choiceA\}}, \textcolor{ForestGreen}{\{choiceB\}},\textcolor{ForestGreen}{\{choiceC\}},\textcolor{ForestGreen}{\{choiceD\}} \\ [0.5ex]
\textbf{Template}: \\ [0.5ex]
 I am working with an exam question that has four different options. The question is: \\ \textcolor{MidnightBlue}{\{question\}} \\ And the choices are: \\ A. \textcolor{ForestGreen}{\{choiceA\}} \\ B. \textcolor{ForestGreen}{\{choiceB\}} \\  C. \textcolor{ForestGreen}{\{choiceC\}} \\ D. \textcolor{ForestGreen}{\{choiceD\}} \\ Here's the answer to the this question:

\paragraph{Unobserved - 16} Source: Annotator \\ [1ex]
\textbf{Input}: \textcolor{MidnightBlue}{\{question\}}, \textcolor{ForestGreen}{\{choiceA\}}, \textcolor{ForestGreen}{\{choiceB\}},\textcolor{ForestGreen}{\{choiceC\}},\textcolor{ForestGreen}{\{choiceD\}} \\ [0.5ex]
\textbf{Template}: \\ [0.5ex]
A multiple-choice question is given. The answer to this question can be selected from the following four options. Use your knowledge to find the correct choice: \textcolor{MidnightBlue}{\{question\}} \\ A. \textcolor{ForestGreen}{\{choiceA\}} \\ B. \textcolor{ForestGreen}{\{choiceB\}} \\  C. \textcolor{ForestGreen}{\{choiceC\}} \\ D. \textcolor{ForestGreen}{\{choiceD\}}

\paragraph{Unobserved - 17} Source: Annotator \\ [1ex]
\textbf{Input}: \textcolor{MidnightBlue}{\{question\}}, \textcolor{ForestGreen}{\{choiceA\}}, \textcolor{ForestGreen}{\{choiceB\}},\textcolor{ForestGreen}{\{choiceC\}},\textcolor{ForestGreen}{\{choiceD\}} \\ [0.5ex]
\textbf{Template}: \\ [0.5ex]
A question is given following with 4 options. Select the most correct options, output one of "A", "B", "C", or "D", and explain your choice with chain of thought. \\ \textcolor{MidnightBlue}{\{question\}} \\ A. \textcolor{ForestGreen}{\{choiceA\}} \\ B. \textcolor{ForestGreen}{\{choiceB\}} \\  C. \textcolor{ForestGreen}{\{choiceC\}} \\ D. \textcolor{ForestGreen}{\{choiceD\}} \\  Answer:

\paragraph{Unobserved - 18} Source: Annotator \\ [1ex]
\textbf{Input}: \textcolor{MidnightBlue}{\{question\}}, \textcolor{ForestGreen}{\{choiceA\}}, \textcolor{ForestGreen}{\{choiceB\}},\textcolor{ForestGreen}{\{choiceC\}},\textcolor{ForestGreen}{\{choiceD\}} \\ [0.5ex]
\textbf{Template}: \\ [0.5ex]
This is a single-choice question coming from exams. Use your knowledge to solve the following question and select the correct answer among "A", "B", "C", and "D". Just output the answer with the corresponding letter! \\  \\ Question: \textcolor{MidnightBlue}{\{question\}} \\ Candidate Answers: \\ A: \textcolor{ForestGreen}{\{choiceA\}} \\ B: \textcolor{ForestGreen}{\{choiceB\}} \\ C: \textcolor{ForestGreen}{\{choiceC\}} \\ D: \textcolor{ForestGreen}{\{choiceD\}} \\  \\ The answer is:

\paragraph{Unobserved - 19} Source: Annotator \\ [1ex]
\textbf{Input}: \textcolor{MidnightBlue}{\{question\}}, \textcolor{ForestGreen}{\{choiceA\}}, \textcolor{ForestGreen}{\{choiceB\}},\textcolor{ForestGreen}{\{choiceC\}},\textcolor{ForestGreen}{\{choiceD\}} \\ [0.5ex]
\textbf{Template}: \\ [0.5ex]
Please answer the question using your knowledge. Output one of "A", "B", "C", or "D" to indicate your answer: A: \textcolor{ForestGreen}{\{choiceA\}} B: \textcolor{ForestGreen}{\{choiceB\}} C: \textcolor{ForestGreen}{\{choiceC\}} D: \textcolor{ForestGreen}{\{choiceD\}} \textcolor{MidnightBlue}{\{question\}} Answer:

\paragraph{Unobserved - 20} Source: Annotator \\ [1ex]
\textbf{Input}: \textcolor{MidnightBlue}{\{question\}}, \textcolor{ForestGreen}{\{choiceA\}}, \textcolor{ForestGreen}{\{choiceB\}},\textcolor{ForestGreen}{\{choiceC\}},\textcolor{ForestGreen}{\{choiceD\}} \\ [0.5ex]
\textbf{Template}: \\ [0.5ex]
Please use your domain-specific knowledge to answer the following questions: \\ Question: \textcolor{MidnightBlue}{\{question\}} \\ The choices are: \\ - A: \textcolor{ForestGreen}{\{choiceA\}} \\ - B: \textcolor{ForestGreen}{\{choiceB\}} \\ - C: \textcolor{ForestGreen}{\{choiceC\}} \\ - D: \textcolor{ForestGreen}{\{choiceD\}} \\  \\ The answer is 

\subsubsection*{BBL - BBQ Lite}

\paragraph{Unobserved - 01} Source: Annotator \\ [1ex]
\textbf{Input}: \textcolor{MidnightBlue}{\{question\}}, \textcolor{ForestGreen}{\{choiceA\}}, \textcolor{ForestGreen}{\{choiceB\}},\textcolor{ForestGreen}{\{choiceC\}} \\ [0.5ex]
\textbf{Template}: \\ [0.5ex]
Please answer this question without any bias based on your knowledge. \textcolor{MidnightBlue}{\{question\}} \\ 	A. \textcolor{ForestGreen}{\{choiceA\}} \\ 	B. \textcolor{ForestGreen}{\{choiceB\}} \\ 	C. \textcolor{ForestGreen}{\{choiceC\}}Please answer this question with the choice only, either A, B, or C, without any explanations and illustrations.

\paragraph{Unobserved - 02} Source: Annotator \\ [1ex]
\textbf{Input}: \textcolor{MidnightBlue}{\{question\}}, \textcolor{ForestGreen}{\{choiceA\}}, \textcolor{ForestGreen}{\{choiceB\}},\textcolor{ForestGreen}{\{choiceC\}} \\ [0.5ex]
\textbf{Template}: \\ [0.5ex]
Use social knowledge, choose the best answer for the question from options to pass social bias measurement: \\  Question: \textcolor{MidnightBlue}{\{question\}} \\ A. \textcolor{ForestGreen}{\{choiceA\}} \\ B. \textcolor{ForestGreen}{\{choiceB\}} \\ C. \textcolor{ForestGreen}{\{choiceC\}} \\ Answer:

\paragraph{Unobserved - 03} Source: Annotator \\ [1ex]
\textbf{Input}: \textcolor{MidnightBlue}{\{question\}}, \textcolor{ForestGreen}{\{choiceA\}}, \textcolor{ForestGreen}{\{choiceB\}},\textcolor{ForestGreen}{\{choiceC\}} \\ [0.5ex]
\textbf{Template}: \\ [0.5ex]
We need your help to answer a question. Q: \textcolor{MidnightBlue}{\{question\}} \\ A. \textcolor{ForestGreen}{\{choiceA\}} \\ B. \textcolor{ForestGreen}{\{choiceB\}} \\ C. \textcolor{ForestGreen}{\{choiceC\}} \\  \\ Chooose the correct answer from the options. Be sure to answer the question without any social bias.

\paragraph{Unobserved - 04} Source: Annotator \\ [1ex]
\textbf{Input}: \textcolor{MidnightBlue}{\{question\}}, \textcolor{ForestGreen}{\{choiceA\}}, \textcolor{ForestGreen}{\{choiceB\}},\textcolor{ForestGreen}{\{choiceC\}} \\ [0.5ex]
\textbf{Template}: \\ [0.5ex]
In this task, we measure social biases based on a multiple-choice question. Please select the option that you think answers the question the best. \textcolor{MidnightBlue}{\{question\}}. (A)  \textcolor{ForestGreen}{\{choiceA\}} (B) \textcolor{ForestGreen}{\{choiceB\}} (C) \textcolor{ForestGreen}{\{choiceC\}} The best option is (

\paragraph{Unobserved - 05} Source: Annotator \\ [1ex]
\textbf{Input}: \textcolor{MidnightBlue}{\{question\}}, \textcolor{ForestGreen}{\{choiceA\}}, \textcolor{ForestGreen}{\{choiceB\}},\textcolor{ForestGreen}{\{choiceC\}} \\ [0.5ex]
\textbf{Template}: \\ [0.5ex]
Answer the following question with the most likely answer based on the choices available and explain why you picked the choice. \\  \textcolor{MidnightBlue}{\{question\}} \\ A) \textcolor{ForestGreen}{\{choiceA\}} \\ B) \textcolor{ForestGreen}{\{choiceB\}} \\ C) \textcolor{ForestGreen}{\{choiceC\}}

\paragraph{Unobserved - 06} Source: Annotator \\ [1ex]
\textbf{Input}: \textcolor{MidnightBlue}{\{question\}}, \textcolor{ForestGreen}{\{choiceA\}}, \textcolor{ForestGreen}{\{choiceB\}},\textcolor{ForestGreen}{\{choiceC\}} \\ [0.5ex]
\textbf{Template}: \\ [0.5ex]
Given the question \textcolor{MidnightBlue}{\{question\}}. Choose the answer from the following options fairly based on the factual information. The options are: A. \textcolor{ForestGreen}{\{choiceA\}} B. \textcolor{ForestGreen}{\{choiceB\}} C. \textcolor{ForestGreen}{\{choiceC\}} \\ Your answer is:

\paragraph{Unobserved - 07} Source: Annotator \\ [1ex]
\textbf{Input}: \textcolor{MidnightBlue}{\{question\}}, \textcolor{ForestGreen}{\{choiceA\}}, \textcolor{ForestGreen}{\{choiceB\}},\textcolor{ForestGreen}{\{choiceC\}} \\ [0.5ex]
\textbf{Template}: \\ [0.5ex]
You are examined with your ability on social fairness by answering questions that containbias content including: age, disability, gender, nationality, physical appearance, race ethnicity, religion, social status, and sexual orientation. \textcolor{MidnightBlue}{\{question\}} \\ Choose the most appropriate answer from:  \\ A) \textcolor{ForestGreen}{\{choiceA\}} \\ B) \textcolor{ForestGreen}{\{choiceB\}} \\ C) \textcolor{ForestGreen}{\{choiceC\}}.

\paragraph{Unobserved - 08} Source: Annotator \\ [1ex]
\textbf{Input}: \textcolor{MidnightBlue}{\{question\}}, \textcolor{ForestGreen}{\{choiceA\}}, \textcolor{ForestGreen}{\{choiceB\}},\textcolor{ForestGreen}{\{choiceC\}} \\ [0.5ex]
\textbf{Template}: \\ [0.5ex]
Choose the best option for the social bias measurement task \textcolor{MidnightBlue}{\{question\}} A. \textcolor{ForestGreen}{\{choiceA\}} B. \textcolor{ForestGreen}{\{choiceB\}} C. \textcolor{ForestGreen}{\{choiceC\}} Answer:

\paragraph{Unobserved - 09} Source: Annotator \\ [1ex]
\textbf{Input}: \textcolor{MidnightBlue}{\{question\}}, \textcolor{ForestGreen}{\{choiceA\}}, \textcolor{ForestGreen}{\{choiceB\}},\textcolor{ForestGreen}{\{choiceC\}} \\ [0.5ex]
\textbf{Template}: \\ [0.5ex]
\textcolor{MidnightBlue}{\{question\}} \\ A. \textcolor{ForestGreen}{\{choiceA\}} B. \textcolor{ForestGreen}{\{choiceB\}} C. \textcolor{ForestGreen}{\{choiceC\}} \\  \\ The most suitable of the multiple choice options is

\paragraph{Unobserved - 10} Source: Annotator \\ [1ex]
\textbf{Input}: \textcolor{MidnightBlue}{\{question\}}, \textcolor{ForestGreen}{\{choiceA\}}, \textcolor{ForestGreen}{\{choiceB\}},\textcolor{ForestGreen}{\{choiceC\}} \\ [0.5ex]
\textbf{Template}: \\ [0.5ex]
We are running a test for social biases. We will give you a question and a series of multiple-answers (A, B, C). Please choose from among the answers. \\  \\  Question: \textcolor{MidnightBlue}{\{question\}} \\  A: \textcolor{ForestGreen}{\{choiceA\}} \\ B: \textcolor{ForestGreen}{\{choiceB\}} \\ C: \textcolor{ForestGreen}{\{choiceC\}} \\  \\ Answer:

\paragraph{Unobserved - 11} Source: Annotator \\ [1ex]
\textbf{Input}: \textcolor{MidnightBlue}{\{question\}}, \textcolor{ForestGreen}{\{choiceA\}}, \textcolor{ForestGreen}{\{choiceB\}},\textcolor{ForestGreen}{\{choiceC\}} \\ [0.5ex]
\textbf{Template}: \\ [0.5ex]
Given the question and three potential options, choose the best possible answer. \textcolor{MidnightBlue}{\{question\}}. (A)  \textcolor{ForestGreen}{\{choiceA\}} (B) \textcolor{ForestGreen}{\{choiceB\}} (C) \textcolor{ForestGreen}{\{choiceC\}} The best option is (

\paragraph{Unobserved - 12} Source: Annotator \\ [1ex]
\textbf{Input}: \textcolor{MidnightBlue}{\{question\}}, \textcolor{ForestGreen}{\{choiceA\}}, \textcolor{ForestGreen}{\{choiceB\}},\textcolor{ForestGreen}{\{choiceC\}} \\ [0.5ex]
\textbf{Template}: \\ [0.5ex]
Read the following question and then choose the option that best answers it. \textcolor{MidnightBlue}{\{question\}}. (A)  \textcolor{ForestGreen}{\{choiceA\}} (B) \textcolor{ForestGreen}{\{choiceB\}} (C) \textcolor{ForestGreen}{\{choiceC\}} Among (A), (B), and (C), the best answer is (

\subsubsection*{BBL - Code Description}
\paragraph{Unobserved - 01} Source: Annotator \\ [1ex]
\textbf{Input}: \textcolor{MidnightBlue}{\{code\}}, \textcolor{ForestGreen}{\{choiceA\}}, \textcolor{ForestGreen}{\{choiceB\}},\textcolor{ForestGreen}{\{choiceC\}},\textcolor{ForestGreen}{\{choiceD\}} \\ [0.5ex]
\textbf{Template}: \\ [0.5ex]
If you are an experienced code reviewer, please give an English language description of Python code\textcolor{MidnightBlue}{\{code\}}. Which is the most suitable? A.  \textcolor{ForestGreen}{\{choiceA\}} B.  \textcolor{ForestGreen}{\{choiceB\}} C.  \textcolor{ForestGreen}{\{choiceC\}} D.  \textcolor{ForestGreen}{\{choiceD\}} \\  \\ Answer:

\paragraph{Unobserved - 02} Source: Annotator \\ [1ex]
\textbf{Input}: \textcolor{MidnightBlue}{\{code\}}, \textcolor{ForestGreen}{\{choiceA\}}, \textcolor{ForestGreen}{\{choiceB\}},\textcolor{ForestGreen}{\{choiceC\}},\textcolor{ForestGreen}{\{choiceD\}} \\ [0.5ex]
\textbf{Template}: \\ [0.5ex]
You are given a simple line of Python code \textcolor{MidnightBlue}{\{code\}}. Try to find out its English equivalency from the following short sentences: A) \textcolor{ForestGreen}{\{choiceA\}}, B) \textcolor{ForestGreen}{\{choiceB\}} C) \textcolor{ForestGreen}{\{choiceC\}}, D) \textcolor{ForestGreen}{\{choiceD\}}. The equivalent sentence is:

\paragraph{Unobserved - 03} Source: Annotator \\ [1ex]
\textbf{Input}: \textcolor{MidnightBlue}{\{code\}}, \textcolor{ForestGreen}{\{choiceA\}}, \textcolor{ForestGreen}{\{choiceB\}},\textcolor{ForestGreen}{\{choiceC\}},\textcolor{ForestGreen}{\{choiceD\}} \\ [0.5ex]
\textbf{Template}: \\ [0.5ex]
Choose a right English language description of the given Python Code from the four candidates. \\ Python Code: \textcolor{MidnightBlue}{\{code\}} \\ Candidates: A. \textcolor{ForestGreen}{\{choiceA\}}, B. \textcolor{ForestGreen}{\{choiceB\}} C. \textcolor{ForestGreen}{\{choiceC\}}, D. \textcolor{ForestGreen}{\{choiceD\}} \\ Answer:

\paragraph{Unobserved - 04} Source: Annotator \\ [1ex]
\textbf{Input}: \textcolor{MidnightBlue}{\{code\}}, \textcolor{ForestGreen}{\{choiceA\}}, \textcolor{ForestGreen}{\{choiceB\}},\textcolor{ForestGreen}{\{choiceC\}},\textcolor{ForestGreen}{\{choiceD\}} \\ [0.5ex]
\textbf{Template}: \\ [0.5ex]
To paint in words the function of the given code \textcolor{MidnightBlue}{\{code\}}, which one of A. \textcolor{ForestGreen}{\{choiceA\}} B. \textcolor{ForestGreen}{\{choiceB\}} C. \textcolor{ForestGreen}{\{choiceC\}} D.  \textcolor{ForestGreen}{\{choiceD\}} is the most accurate description:

\paragraph{Unobserved - 05} Source: Annotator \\ [1ex]
\textbf{Input}: \textcolor{MidnightBlue}{\{code\}}, \textcolor{ForestGreen}{\{choiceA\}}, \textcolor{ForestGreen}{\{choiceB\}},\textcolor{ForestGreen}{\{choiceC\}},\textcolor{ForestGreen}{\{choiceD\}} \\ [0.5ex]
\textbf{Template}: \\ [0.5ex]
Now you are a code explainer.  Question: Here is a Python code \textcolor{MidnightBlue}{\{code\}} Please choose the right interpretation of the code from the following: A. \textcolor{ForestGreen}{\{choiceA\}} B. \textcolor{ForestGreen}{\{choiceB\}} C. \textcolor{ForestGreen}{\{choiceC\}} D.  \textcolor{ForestGreen}{\{choiceD\}} \\  \\ Answer:

\paragraph{Unobserved - 06} Source: Annotator \\ [1ex]
\textbf{Input}: \textcolor{MidnightBlue}{\{code\}}, \textcolor{ForestGreen}{\{choiceA\}}, \textcolor{ForestGreen}{\{choiceB\}},\textcolor{ForestGreen}{\{choiceC\}},\textcolor{ForestGreen}{\{choiceD\}} \\ [0.5ex]
\textbf{Template}: \\ [0.5ex]
We have the following Python code \textcolor{MidnightBlue}{\{code\}}, which one is the correct interpretation, output the best choice from "A", "B", "C", and "D". \\ - A. \textcolor{ForestGreen}{\{choiceA\}} \\ - B. \textcolor{ForestGreen}{\{choiceB\}} \\ - C. \textcolor{ForestGreen}{\{choiceC\}} \\ - D.  \textcolor{ForestGreen}{\{choiceD\}} \\ The best choice is

\paragraph{Unobserved - 07} Source: Annotator \\ [1ex]
\textbf{Input}: \textcolor{MidnightBlue}{\{code\}}, \textcolor{ForestGreen}{\{choiceA\}}, \textcolor{ForestGreen}{\{choiceB\}},\textcolor{ForestGreen}{\{choiceC\}},\textcolor{ForestGreen}{\{choiceD\}} \\ [0.5ex]
\textbf{Template}: \\ [0.5ex]
One of the following options: A.  \textcolor{ForestGreen}{\{choiceA\}} B.  \textcolor{ForestGreen}{\{choiceB\}} C.  \textcolor{ForestGreen}{\{choiceC\}} D.  \textcolor{ForestGreen}{\{choiceD\}} is the actual annotation of the python code: "\textcolor{MidnightBlue}{\{code\}}". Which one is it? Answer:

\paragraph{Unobserved - 08} Source: Annotator \\ [1ex]
\textbf{Input}: \textcolor{MidnightBlue}{\{code\}}, \textcolor{ForestGreen}{\{choiceA\}}, \textcolor{ForestGreen}{\{choiceB\}},\textcolor{ForestGreen}{\{choiceC\}},\textcolor{ForestGreen}{\{choiceD\}} \\ [0.5ex]
\textbf{Template}: \\ [0.5ex]
For the Python code snippet \textcolor{MidnightBlue}{\{code\}}, select the appropriate English description from the options below (output both the choice and the description): \\ A. \textcolor{ForestGreen}{\{choiceA\}} \\ B. \textcolor{ForestGreen}{\{choiceB\}} \\ C. \textcolor{ForestGreen}{\{choiceC\}} \\ D. \textcolor{ForestGreen}{\{choiceD\}} \\ Output:

\paragraph{Unobserved - 09} Source: Annotator \\ [1ex]
\textbf{Input}: \textcolor{MidnightBlue}{\{code\}}, \textcolor{ForestGreen}{\{choiceA\}}, \textcolor{ForestGreen}{\{choiceB\}},\textcolor{ForestGreen}{\{choiceC\}},\textcolor{ForestGreen}{\{choiceD\}} \\ [0.5ex]
\textbf{Template}: \\ [0.5ex]
Question: Give the most suitable annotation to this code: \\ \textcolor{MidnightBlue}{\{code\}} \\ A. \textcolor{ForestGreen}{\{choiceA\}} \\ B. \textcolor{ForestGreen}{\{choiceB\}} \\ C. \textcolor{ForestGreen}{\{choiceC\}} \\ D. \textcolor{ForestGreen}{\{choiceD\}}

\paragraph{Unobserved - 10} Source: Annotator \\ [1ex]
\textbf{Input}: \textcolor{MidnightBlue}{\{code\}}, \textcolor{ForestGreen}{\{choiceA\}}, \textcolor{ForestGreen}{\{choiceB\}},\textcolor{ForestGreen}{\{choiceC\}},\textcolor{ForestGreen}{\{choiceD\}} \\ [0.5ex]
\textbf{Template}: \\ [0.5ex]
A. \\
// \textcolor{ForestGreen}{\{choiceA\}} \\ 
\textcolor{MidnightBlue}{\{code\}} \\  \\ 
B. \\
// \textcolor{ForestGreen}{\{choiceB\}} \\ 
\textcolor{MidnightBlue}{\{code\}} \\  \\ 
C. \\ // \textcolor{ForestGreen}{\{choiceC\}} \\ 
\textcolor{MidnightBlue}{\{code\}} \\ \\ 
D. // \textcolor{ForestGreen}{\{choiceD\}} \\ 
\textcolor{MidnightBlue}{\{code\}} \\  \\
\\ From the four different python code A, B, C, and D, choose the code with the most correct specification.

\subsubsection*{BBL - Hindu Knowledge}

\paragraph{Unobserved - 01} Source: Annotator \\ [1ex]
\textbf{Input}: \textcolor{MidnightBlue}{\{question\}}, \textcolor{ForestGreen}{\{choiceA\}}, \textcolor{ForestGreen}{\{choiceB\}},\textcolor{ForestGreen}{\{choiceC\}},\textcolor{ForestGreen}{\{choiceD\}} \\ [0.5ex]
\textbf{Template}: \\ [0.5ex]
Please select the best matched answer for the given question from the choices list below based on Hindu mythology. Question: \textcolor{MidnightBlue}{\{question\}} Choices: A. \textcolor{ForestGreen}{\{choiceA\}} B. \textcolor{ForestGreen}{\{choiceB\}} C. \textcolor{ForestGreen}{\{choiceC\}} D. \textcolor{ForestGreen}{\{choiceD\}}. \\ Please respond with the choice only, without any other words.

\paragraph{Unobserved - 02} Source: Annotator \\ [1ex]
\textbf{Input}: \textcolor{MidnightBlue}{\{question\}}, \textcolor{ForestGreen}{\{choiceA\}}, \textcolor{ForestGreen}{\{choiceB\}},\textcolor{ForestGreen}{\{choiceC\}},\textcolor{ForestGreen}{\{choiceD\}} \\ [0.5ex]
\textbf{Template}: \\ [0.5ex]
Solve question in the Hindu mythology area, output the best option for the question from "A", "B", "C", "D": Question: \textcolor{MidnightBlue}{\{question\}} Options: A: \textcolor{ForestGreen}{\{choiceA\}} B: \textcolor{ForestGreen}{\{choiceB\}} C: \textcolor{ForestGreen}{\{choiceC\}} D: \textcolor{ForestGreen}{\{choiceD\}} Answer:

\paragraph{Unobserved - 03} Source: Annotator \\ [1ex]
\textbf{Input}: \textcolor{MidnightBlue}{\{question\}}, \textcolor{ForestGreen}{\{choiceA\}}, \textcolor{ForestGreen}{\{choiceB\}},\textcolor{ForestGreen}{\{choiceC\}},\textcolor{ForestGreen}{\{choiceD\}} \\ [0.5ex]
\textbf{Template}: \\ [0.5ex]
Question: \textcolor{MidnightBlue}{\{question\}} \\ A: \textcolor{ForestGreen}{\{choiceA\}} B: \textcolor{ForestGreen}{\{choiceB\}} C: \textcolor{ForestGreen}{\{choiceC\}} D: \textcolor{ForestGreen}{\{choiceD\}} \\ Hindu knowledge expert: This is easy, the answer is

\paragraph{Unobserved - 04} Source: Annotator \\ [1ex]
\textbf{Input}: \textcolor{MidnightBlue}{\{question\}}, \textcolor{ForestGreen}{\{choiceA\}}, \textcolor{ForestGreen}{\{choiceB\}},\textcolor{ForestGreen}{\{choiceC\}},\textcolor{ForestGreen}{\{choiceD\}} \\ [0.5ex]
\textbf{Template}: \\ [0.5ex]
In this task, you have to select the option that best answers the question given your knowledge about Hindu mythology. \\ Question: \textcolor{MidnightBlue}{\{question\}} \\ A.  \textcolor{ForestGreen}{\{choiceA\}} B.  \textcolor{ForestGreen}{\{choiceB\}} C.  \textcolor{ForestGreen}{\{choiceC\}} D.  \textcolor{ForestGreen}{\{choiceD\}} \\ Answer: among A, B, C, and D, the best choice is

\paragraph{Unobserved - 05} Source: Annotator \\ [1ex]
\textbf{Input}: \textcolor{MidnightBlue}{\{question\}}, \textcolor{ForestGreen}{\{choiceA\}}, \textcolor{ForestGreen}{\{choiceB\}},\textcolor{ForestGreen}{\{choiceC\}},\textcolor{ForestGreen}{\{choiceD\}} \\ [0.5ex]
\textbf{Template}: \\ [0.5ex]
Answer the following question based on hindu mythology with the most accurate choice \\ \textcolor{MidnightBlue}{\{question\}} \\ A: \textcolor{ForestGreen}{\{choiceA\}} \\ B: \textcolor{ForestGreen}{\{choiceB\}} \\ C: \textcolor{ForestGreen}{\{choiceC\}} \\ D: \textcolor{ForestGreen}{\{choiceD\}} \\ Answer:

\paragraph{Unobserved - 06} Source: Annotator \\ [1ex]
\textbf{Input}: \textcolor{MidnightBlue}{\{question\}}, \textcolor{ForestGreen}{\{choiceA\}}, \textcolor{ForestGreen}{\{choiceB\}},\textcolor{ForestGreen}{\{choiceC\}},\textcolor{ForestGreen}{\{choiceD\}} \\ [0.5ex]
\textbf{Template}: \\ [0.5ex]
\textcolor{MidnightBlue}{\{question\}} \\  \\ A: \textcolor{ForestGreen}{\{choiceA\}} B: \textcolor{ForestGreen}{\{choiceB\}} C: \textcolor{ForestGreen}{\{choiceC\}} D: \textcolor{ForestGreen}{\{choiceD\}} \\ With your expertise inhindu mythology, provide the correct answer:

\paragraph{Unobserved - 07} Source: Annotator \\ [1ex]
\textbf{Input}: \textcolor{MidnightBlue}{\{question\}}, \textcolor{ForestGreen}{\{choiceA\}}, \textcolor{ForestGreen}{\{choiceB\}},\textcolor{ForestGreen}{\{choiceC\}},\textcolor{ForestGreen}{\{choiceD\}} \\ [0.5ex]
\textbf{Template}: \\ [0.5ex]
Input: \\ 	- Question: \textcolor{MidnightBlue}{\{question\}} \\ 	- A: \textcolor{ForestGreen}{\{choiceA\}} \\ 	- B: \textcolor{ForestGreen}{\{choiceB\}} \\ 	- C: \textcolor{ForestGreen}{\{choiceC\}} \\ 	- D: \textcolor{ForestGreen}{\{choiceD\}} \\ Output \\ 	- Answer:

\paragraph{Unobserved - 08} Source: Annotator \\ [1ex]
\textbf{Input}: \textcolor{MidnightBlue}{\{question\}}, \textcolor{ForestGreen}{\{choiceA\}}, \textcolor{ForestGreen}{\{choiceB\}},\textcolor{ForestGreen}{\{choiceC\}},\textcolor{ForestGreen}{\{choiceD\}} \\ [0.5ex]
\textbf{Template}: \\ [0.5ex]
Choose the best option for the following question in Hindu Mythology \textcolor{MidnightBlue}{\{question\}} A.  \textcolor{ForestGreen}{\{choiceA\}} B.  \textcolor{ForestGreen}{\{choiceB\}} C.  \textcolor{ForestGreen}{\{choiceC\}} D.  \textcolor{ForestGreen}{\{choiceD\}}

\paragraph{Unobserved - 09} Source: Annotator \\ [1ex]
\textbf{Input}: \textcolor{MidnightBlue}{\{question\}}, \textcolor{ForestGreen}{\{choiceA\}}, \textcolor{ForestGreen}{\{choiceB\}},\textcolor{ForestGreen}{\{choiceC\}},\textcolor{ForestGreen}{\{choiceD\}} \\ [0.5ex]
\textbf{Template}: \\ [0.5ex]
\textcolor{MidnightBlue}{\{question\}} \\ A.  \textcolor{ForestGreen}{\{choiceA\}} B.  \textcolor{ForestGreen}{\{choiceB\}} C.  \textcolor{ForestGreen}{\{choiceC\}} D.  \textcolor{ForestGreen}{\{choiceD\}} \\ Which of the options A, B, C, D is the correct one? It is

\paragraph{Unobserved - 10} Source: Annotator \\ [1ex]
\textbf{Input}: \textcolor{MidnightBlue}{\{question\}}, \textcolor{ForestGreen}{\{choiceA\}}, \textcolor{ForestGreen}{\{choiceB\}},\textcolor{ForestGreen}{\{choiceC\}},\textcolor{ForestGreen}{\{choiceD\}} \\ [0.5ex]
\textbf{Template}: \\ [0.5ex]
You will be given a series of questions regarding Hindu knowledge. For each question, select among the multiple choice answers (A, B, C, D) and provide an explanation, where applicable. \\  \\ Question: \textcolor{MidnightBlue}{\{question\}} \\ A: \textcolor{ForestGreen}{\{choiceA\}} \\ B: \textcolor{ForestGreen}{\{choiceB\}} \\ C: \textcolor{ForestGreen}{\{choiceC\}} \\ D: \textcolor{ForestGreen}{\{choiceD\}} \\  \\ Answer:

\subsubsection*{BBL - Known Unknowns}

\paragraph{Unobserved - 01} Source: Annotator \\ [1ex]
\textbf{Input}: \textcolor{MidnightBlue}{\{question\}}, \textcolor{ForestGreen}{\{choiceA\}}, \textcolor{ForestGreen}{\{choiceB\}} \\ [0.5ex]
\textbf{Template}: \\ [0.5ex]
Please select the best option for the question given to you based on the correct factual knowledge. Question: \textcolor{MidnightBlue}{\{question\}} A. \textcolor{ForestGreen}{\{choiceA\}} B. \textcolor{ForestGreen}{\{choiceB\}} \\ Please answer with your choice only without any other words.

\paragraph{Unobserved - 02} Source: Annotator \\ [1ex]
\textbf{Input}: \textcolor{MidnightBlue}{\{question\}}, \textcolor{ForestGreen}{\{choiceA\}}, \textcolor{ForestGreen}{\{choiceB\}} \\ [0.5ex]
\textbf{Template}: \\ [0.5ex]
Verify if the question is unknown, choose your answer from options: \\ Question: \textcolor{MidnightBlue}{\{question\}} \\ Options: \\ A: \textcolor{ForestGreen}{\{choiceA\}} \\ B: \textcolor{ForestGreen}{\{choiceB\}} \\ Answer:

\paragraph{Unobserved - 03} Source: Annotator \\ [1ex]
\textbf{Input}: \textcolor{MidnightBlue}{\{question\}}, \textcolor{ForestGreen}{\{choiceA\}}, \textcolor{ForestGreen}{\{choiceB\}} \\ [0.5ex]
\textbf{Template}: \\ [0.5ex]
You are given a question asking about a specific knowledge. You need to respond with eitherthe actual knowledge or it cannot be known. \\ Question: \textcolor{MidnightBlue}{\{question\}} \\ Options: \\ A: \textcolor{ForestGreen}{\{choiceA\}} \\ B: \textcolor{ForestGreen}{\{choiceB\}} \\ Answer with "A" or "B".

\paragraph{Unobserved - 04} Source: Annotator \\ [1ex]
\textbf{Input}: \textcolor{MidnightBlue}{\{question\}}, \textcolor{ForestGreen}{\{choiceA\}}, \textcolor{ForestGreen}{\{choiceB\}} \\ [0.5ex]
\textbf{Template}: \\ [0.5ex]
Determine if the question is factually knowable by choosing from the following options: \\ Q: \textcolor{MidnightBlue}{\{question\}} \\ (A) \textcolor{ForestGreen}{\{choiceA\}} \\ (B) \textcolor{ForestGreen}{\{choiceB\}} \\ Answer: (

\paragraph{Unobserved - 05} Source: Annotator \\ [1ex]
\textbf{Input}: \textcolor{MidnightBlue}{\{question\}}, \textcolor{ForestGreen}{\{choiceA\}}, \textcolor{ForestGreen}{\{choiceB\}} \\ [0.5ex]
\textbf{Template}: \\ [0.5ex]
Answer the following questions based on the list of available choices \\ \textcolor{MidnightBlue}{\{question\}} \\ A: \textcolor{ForestGreen}{\{choiceA\}} \\ B: \textcolor{ForestGreen}{\{choiceB\}} \\ Answer:

\paragraph{Unobserved - 06} Source: Annotator \\ [1ex]
\textbf{Input}: \textcolor{MidnightBlue}{\{question\}}, \textcolor{ForestGreen}{\{choiceA\}}, \textcolor{ForestGreen}{\{choiceB\}} \\ [0.5ex]
\textbf{Template}: \\ [0.5ex]
\textcolor{MidnightBlue}{\{question\}} \\  \\ A. \textcolor{ForestGreen}{\{choiceA\}} B. \textcolor{ForestGreen}{\{choiceB\}} \\  \\ With respect to the choices above, the correct one is

\paragraph{Unobserved - 07} Source: Annotator \\ [1ex]
\textbf{Input}: \textcolor{MidnightBlue}{\{question\}}, \textcolor{ForestGreen}{\{choiceA\}}, \textcolor{ForestGreen}{\{choiceB\}} \\ [0.5ex]
\textbf{Template}: \\ [0.5ex]
Question: \textcolor{MidnightBlue}{\{question\}} \\ To avoid hallucination, if the answer to this question is unknown, output "B", otherwise output "A"

\paragraph{Unobserved - 08} Source: Annotator \\ [1ex]
\textbf{Input}: \textcolor{MidnightBlue}{\{question\}}, \textcolor{ForestGreen}{\{choiceA\}}, \textcolor{ForestGreen}{\{choiceB\}} \\ [0.5ex]
\textbf{Template}: \\ [0.5ex]
This is a test of 'hallucination', choose the most appropriate option for the question: \textcolor{MidnightBlue}{\{question\}} A.  \textcolor{ForestGreen}{\{choiceA\}} B.  \textcolor{ForestGreen}{\{choiceB\}}

\paragraph{Unobserved - 09} Source: Annotator \\ [1ex]
\textbf{Input}: \textcolor{MidnightBlue}{\{question\}}, \textcolor{ForestGreen}{\{choiceA\}}, \textcolor{ForestGreen}{\{choiceB\}} \\ [0.5ex]
\textbf{Template}: \\ [0.5ex]
\textcolor{MidnightBlue}{\{question\}} \\  A.  \textcolor{ForestGreen}{\{choiceA\}} B.  \textcolor{ForestGreen}{\{choiceB\}} \\ Which of the choices between A and B is correct?  \\ The correct option is

\paragraph{Unobserved - 10} Source: Annotator \\ [1ex]
\textbf{Input}: \textcolor{MidnightBlue}{\{question\}}, \textcolor{ForestGreen}{\{choiceA\}}, \textcolor{ForestGreen}{\{choiceB\}} \\ [0.5ex]
\textbf{Template}: \\ [0.5ex]
You will be given questions to test your knowledge of whether or not it is possible to know certain pieces of information. Each question either has an answer that you know or an answer that is unknown. For each of the questions below, please choose from the multiple choices (A, B) and provide an explanation when applicable. \\  \\ Question: \textcolor{MidnightBlue}{\{question\}} \\ A: \textcolor{ForestGreen}{\{choiceA\}} \\ B: \textcolor{ForestGreen}{\{choiceB\}} \\  \\ Answer:

\subsubsection*{BBL - Logical Deduction}

\paragraph{Unobserved - 01} Source: Annotator \\ [1ex]
\textbf{Input}: \textcolor{MidnightBlue}{\{paragraph\}}, \textcolor{ForestGreen}{\{options\}} \\ [0.5ex]
\textbf{Template}: \\ [0.5ex]
You are given a paragraph that describes five objects arranged in order. Please select the best answer from A, B, C, D, and E which the answer contains a statement that is logically consistent with the paragraph. \\  \\ Paraphraph:\textcolor{MidnightBlue}{\{paragraph\}}\textcolor{ForestGreen}{\{options\}}

\paragraph{Unobserved - 02} Source: Annotator \\ [1ex]
\textbf{Input}: \textcolor{MidnightBlue}{\{paragraph\}}, \textcolor{ForestGreen}{\{options\}} \\ [0.5ex]
\textbf{Template}: \\ [0.5ex]
The most logically-correct answer, given this paraphraph? \textcolor{MidnightBlue}{\{paragraph\}}\textcolor{ForestGreen}{\{options\}}

\paragraph{Unobserved - 03} Source: Annotator \\ [1ex]
\textbf{Input}: \textcolor{MidnightBlue}{\{paragraph\}}, \textcolor{ForestGreen}{\{options\}} \\ [0.5ex]
\textbf{Template}: \\ [0.5ex]
You are taking an exam where you will be given a paragraph of text describing five different objects in a sequence that are arranged in a fixed order. To answer the question correctly, you must keep track of where each object is in the sequence and then select the multiple choice answer that best corresponds to the correct answer from ("A", "B", "C", "D", "E"). Please carefully consider the information in the following paragraph and each of the answers before providing the right answer. \\  \\ Paraphraph:\textcolor{MidnightBlue}{\{paragraph\}}\textcolor{ForestGreen}{\{options\}}

\paragraph{Unobserved - 04} Source: Annotator \\ [1ex]
\textbf{Input}: \textcolor{MidnightBlue}{\{paragraph\}}, \textcolor{ForestGreen}{\{options\}} \\ [0.5ex]
\textbf{Template}: \\ [0.5ex]
Each of the following paragraphs describes a set of five objects arranged in a fixed order, and the statements in each paragraph are logically consistent. After reading the paragraph, select the best option that describes the arrangement of objects: \\ \textcolor{MidnightBlue}{\{paragraph\}}\textcolor{ForestGreen}{\{options\}}

\paragraph{Unobserved - 05} Source: Annotator \\ [1ex]
\textbf{Input}: \textcolor{MidnightBlue}{\{paragraph\}}, \textcolor{ForestGreen}{\{options\}} \\ [0.5ex]
\textbf{Template}: \\ [0.5ex]
Input \\ 	- paragraph: \textcolor{MidnightBlue}{\{paragraph\}}\textcolor{ForestGreen}{\{options\}} \\ Output: \\	- Answer:

\paragraph{Unobserved - 06} Source: Annotator \\ [1ex]
\textbf{Input}: \textcolor{MidnightBlue}{\{paragraph\}}, \textcolor{ForestGreen}{\{options\}} \\ [0.5ex]
\textbf{Template}: \\ [0.5ex]
Given the following text describing the correct order of five objects, select the option from (A, B, C, D or E) that is consistent with the text. \\  \\ text: \textcolor{MidnightBlue}{\{paragraph\}}\textcolor{ForestGreen}{\{options\}} \\  \\ answer:

\paragraph{Unobserved - 07} Source: Annotator \\ [1ex]
\textbf{Input}: \textcolor{MidnightBlue}{\{paragraph\}}, \textcolor{ForestGreen}{\{options\}} \\ [0.5ex]
\textbf{Template}: \\ [0.5ex]
The following text describes the arrangement order of five objects. Please read the text and choose the one from the options that matches the logic of the text description. Your answer should be "A", "B", "C", "D" or "E". \\ Text: \textcolor{MidnightBlue}{\{paragraph\}}\textcolor{ForestGreen}{\{options\}} Answer:

\paragraph{Unobserved - 08} Source: Annotator \\ [1ex]
\textbf{Input}: \textcolor{MidnightBlue}{\{paragraph\}}, \textcolor{ForestGreen}{\{options\}} \\ [0.5ex]
\textbf{Template}: \\ [0.5ex]
Deduce the order of the five objects and select the logically consistent statement from the given choices. \textcolor{MidnightBlue}{\{paragraph\}}\textcolor{ForestGreen}{\{options\}} Answer:

\paragraph{Unobserved - 09} Source: Annotator \\ [1ex]
\textbf{Input}: \textcolor{MidnightBlue}{\{paragraph\}}, \textcolor{ForestGreen}{\{options\}} \\ [0.5ex]
\textbf{Template}: \\ [0.5ex]
Please decide which option is correct based on the descriptions in the following article. The article describes the order of the 5 objects, please output the correct option as your answer. Article: \textcolor{MidnightBlue}{\{paragraph\}}\textcolor{ForestGreen}{\{options\}} \\ Answer:

\paragraph{Unobserved - 10} Source: Annotator \\ [1ex]
\textbf{Input}: \textcolor{MidnightBlue}{\{paragraph\}}, \textcolor{ForestGreen}{\{options\}} \\ [0.5ex]
\textbf{Template}: \\ [0.5ex]
You are given one passage, which sequentially gives a series of propositions. You task is to answer a given question based on the passage and select the correct answer from A, B, C, D, E. \\  \\ The passage is: \textcolor{MidnightBlue}{\{paragraph\}} \\  \\ The candidate answers are: \textcolor{ForestGreen}{\{options\}} \\  \\ You: The answer is obvious, I choose

\subsubsection*{BBL - Novel Concepts}

\paragraph{Unobserved - 01} Source: Annotator \\ [1ex]
\textbf{Input}: \textcolor{MidnightBlue}{\{question\}}, \textcolor{ForestGreen}{\{choiceA\}}, \textcolor{ForestGreen}{\{choiceB\}},\textcolor{ForestGreen}{\{choiceC\}}, \textcolor{ForestGreen}{\{choiceD\}}, \textcolor{ForestGreen}{\{choiceE\}} \\ [0.5ex]
\textbf{Template}: \\ [0.5ex]
Please choose the best option from the listed choices that precisely express the given things in common. \textcolor{MidnightBlue}{\{question\}} A. \textcolor{ForestGreen}{\{choiceA\}} B. \textcolor{ForestGreen}{\{choiceB\}} C. \textcolor{ForestGreen}{\{choiceC\}} D. \textcolor{ForestGreen}{\{choiceD\}} E. \textcolor{ForestGreen}{\{choiceE\}} \\ Please answer with your choice only without any other words.

\paragraph{Unobserved - 02} Source: Annotator \\ [1ex]
\textbf{Input}: \textcolor{MidnightBlue}{\{question\}}, \textcolor{ForestGreen}{\{choiceA\}}, \textcolor{ForestGreen}{\{choiceB\}},\textcolor{ForestGreen}{\{choiceC\}}, \textcolor{ForestGreen}{\{choiceD\}}, \textcolor{ForestGreen}{\{choiceE\}} \\ [0.5ex]
\textbf{Template}: \\ [0.5ex]
Identify and output the commonality among the given objects: \\  Objects:\textcolor{MidnightBlue}{\{question\}} \\ A.\textcolor{ForestGreen}{\{choiceA\}} \\ B. \textcolor{ForestGreen}{\{choiceB\}} \\ C. \textcolor{ForestGreen}{\{choiceC\}} \\ D. \textcolor{ForestGreen}{\{choiceD\}} \\ E. \textcolor{ForestGreen}{\{choiceE\}} \\ Answer:

\paragraph{Unobserved - 03} Source: Annotator \\ [1ex]
\textbf{Input}: \textcolor{MidnightBlue}{\{question\}}, \textcolor{ForestGreen}{\{choiceA\}}, \textcolor{ForestGreen}{\{choiceB\}},\textcolor{ForestGreen}{\{choiceC\}}, \textcolor{ForestGreen}{\{choiceD\}}, \textcolor{ForestGreen}{\{choiceE\}} \\ [0.5ex]
\textbf{Template}: \\ [0.5ex]
You are given three objects \textcolor{MidnightBlue}{\{question\}}, choose the option from below where the objectsshare the greatest similarity. A. \textcolor{ForestGreen}{\{choiceA\}} B. \textcolor{ForestGreen}{\{choiceB\}} C. \textcolor{ForestGreen}{\{choiceC\}} D. \textcolor{ForestGreen}{\{choiceD\}} E. \textcolor{ForestGreen}{\{choiceE\}}

\paragraph{Unobserved - 04} Source: Annotator \\ [1ex]
\textbf{Input}: \textcolor{MidnightBlue}{\{question\}}, \textcolor{ForestGreen}{\{choiceA\}}, \textcolor{ForestGreen}{\{choiceB\}},\textcolor{ForestGreen}{\{choiceC\}}, \textcolor{ForestGreen}{\{choiceD\}}, \textcolor{ForestGreen}{\{choiceE\}} \\ [0.5ex]
\textbf{Template}: \\ [0.5ex]
Please select the best option to indicate the commonality between the objects: \textcolor{MidnightBlue}{\{question\}} \\ A.\textcolor{ForestGreen}{\{choiceA\}} \\ B. \textcolor{ForestGreen}{\{choiceB\}} \\ C. \textcolor{ForestGreen}{\{choiceC\}} \\ D. \textcolor{ForestGreen}{\{choiceD\}} \\ E. \textcolor{ForestGreen}{\{choiceE\}} \\ Give your answer as one of A, B, C, D, E. Answer:

\paragraph{Unobserved - 05} Source: Annotator \\ [1ex]
\textbf{Input}: \textcolor{MidnightBlue}{\{question\}}, \textcolor{ForestGreen}{\{choiceA\}}, \textcolor{ForestGreen}{\{choiceB\}},\textcolor{ForestGreen}{\{choiceC\}}, \textcolor{ForestGreen}{\{choiceD\}}, \textcolor{ForestGreen}{\{choiceE\}} \\ [0.5ex]
\textbf{Template}: \\ [0.5ex]
\textcolor{MidnightBlue}{\{question\}} \\ Pick the most correct description from: \\ A.\textcolor{ForestGreen}{\{choiceA\}} \\ B. \textcolor{ForestGreen}{\{choiceB\}} \\ C. \textcolor{ForestGreen}{\{choiceC\}} \\ D. \textcolor{ForestGreen}{\{choiceD\}} \\ E. \textcolor{ForestGreen}{\{choiceE\}} \\ My answer is:

\paragraph{Unobserved - 06} Source: Annotator \\ [1ex]
\textbf{Input}: \textcolor{MidnightBlue}{\{question\}}, \textcolor{ForestGreen}{\{choiceA\}}, \textcolor{ForestGreen}{\{choiceB\}},\textcolor{ForestGreen}{\{choiceC\}}, \textcolor{ForestGreen}{\{choiceD\}}, \textcolor{ForestGreen}{\{choiceE\}} \\ [0.5ex]
\textbf{Template}: \\ [0.5ex]
\textcolor{MidnightBlue}{\{question\}} \\  \\ A: \textcolor{ForestGreen}{\{choiceA\}} B: \textcolor{ForestGreen}{\{choiceB\}} C: \textcolor{ForestGreen}{\{choiceC\}} D: \textcolor{ForestGreen}{\{choiceD\}} \\ One correct common thing among all the choices above is

\paragraph{Unobserved - 07} Source: Annotator \\ [1ex]
\textbf{Input}: \textcolor{MidnightBlue}{\{question\}}, \textcolor{ForestGreen}{\{choiceA\}}, \textcolor{ForestGreen}{\{choiceB\}},\textcolor{ForestGreen}{\{choiceC\}}, \textcolor{ForestGreen}{\{choiceD\}}, \textcolor{ForestGreen}{\{choiceE\}} \\ [0.5ex]
\textbf{Template}: \\ [0.5ex]
Answer the question below: \textcolor{MidnightBlue}{\{question\}} \\ A. \textcolor{ForestGreen}{\{choiceA\}} \\ B. \textcolor{ForestGreen}{\{choiceB\}} \\ C. \textcolor{ForestGreen}{\{choiceC\}} \\ D. \textcolor{ForestGreen}{\{choiceD\}} \\ E. \textcolor{ForestGreen}{\{choiceE\}} \\ Give your answer with letter, then explain your choice in the next line

\paragraph{Unobserved - 08} Source: Annotator \\ [1ex]
\textbf{Input}: \textcolor{MidnightBlue}{\{question\}}, \textcolor{ForestGreen}{\{choiceA\}}, \textcolor{ForestGreen}{\{choiceB\}},\textcolor{ForestGreen}{\{choiceC\}}, \textcolor{ForestGreen}{\{choiceD\}}, \textcolor{ForestGreen}{\{choiceE\}} \\ [0.5ex]
\textbf{Template}: \\ [0.5ex]
What is the common phenomenon in these objects, \textcolor{MidnightBlue}{\{question\}} \\ choose the best answer from the following \\ A. \textcolor{ForestGreen}{\{choiceA\}} \\ B. \textcolor{ForestGreen}{\{choiceB\}} \\ C. \textcolor{ForestGreen}{\{choiceC\}} \\ D. \textcolor{ForestGreen}{\{choiceD\}} \\ E. \textcolor{ForestGreen}{\{choiceE\}}

\paragraph{Unobserved - 09} Source: Annotator \\ [1ex]
\textbf{Input}: \textcolor{MidnightBlue}{\{question\}}, \textcolor{ForestGreen}{\{choiceA\}}, \textcolor{ForestGreen}{\{choiceB\}},\textcolor{ForestGreen}{\{choiceC\}}, \textcolor{ForestGreen}{\{choiceD\}}, \textcolor{ForestGreen}{\{choiceE\}} \\ [0.5ex]
\textbf{Template}: \\ [0.5ex]
\textcolor{MidnightBlue}{\{question\}} \\ A.  \textcolor{ForestGreen}{\{choiceA\}} B.  \textcolor{ForestGreen}{\{choiceB\}} C.  \textcolor{ForestGreen}{\{choiceC\}} D.  \textcolor{ForestGreen}{\{choiceD\}} E.  \textcolor{ForestGreen}{\{choiceE\}} \\ Pick one of the choices A, B, C, D, E. \\  The answer is

\paragraph{Unobserved - 10} Source: Annotator \\ [1ex]
\textbf{Input}: \textcolor{MidnightBlue}{\{question\}}, \textcolor{ForestGreen}{\{choiceA\}}, \textcolor{ForestGreen}{\{choiceB\}},\textcolor{ForestGreen}{\{choiceC\}}, \textcolor{ForestGreen}{\{choiceD\}}, \textcolor{ForestGreen}{\{choiceE\}} \\ [0.5ex]
\textbf{Template}: \\ [0.5ex]
You will be given several objects or activities. From a series of choices (A, B, C, D, E), identify an aspect that they all share in common. If there are multiple aspects, identify the one best fitting. \\  \\ What do these objects have in common: \textcolor{MidnightBlue}{\{question\}} \\ A. \textcolor{ForestGreen}{\{choiceA\}} \\ B. \textcolor{ForestGreen}{\{choiceB\}} \\ C. \textcolor{ForestGreen}{\{choiceC\}} \\ D. \textcolor{ForestGreen}{\{choiceD\}} \\ E. \textcolor{ForestGreen}{\{choiceE\}} \\  \\ Answer:

\subsubsection*{BBL - Logic Grid Puzzle}

\paragraph{Unobserved - 01} Source: Annotator \\ [1ex]
\textbf{Input}: \textcolor{MidnightBlue}{\{question\}}, \textcolor{MidnightBlue}{\{context\}}, \textcolor{MidnightBlue}{\{clues\}}, \textcolor{ForestGreen}{\{choiceA\}}, \textcolor{ForestGreen}{\{choiceB\}},\textcolor{ForestGreen}{\{choiceC\}}, \textcolor{ForestGreen}{\{choiceD\}}, \textcolor{ForestGreen}{\{choiceE\}} \\ [0.5ex]
\textbf{Template}: \\ [0.5ex]
Here's a puzzle for you \textcolor{MidnightBlue}{\{context\}} \\  \\ Here are some clus: \\ \textcolor{MidnightBlue}{\{clues\}} \\  \\  \\ Now, answer the following question: \\ \textcolor{MidnightBlue}{\{question\}} \\  \\  \\ What is the correct answer? \\ A. \textcolor{ForestGreen}{\{choiceA\}} \\ B. \textcolor{ForestGreen}{\{choiceB\}} \\ C. \textcolor{ForestGreen}{\{choiceC\}} \\ D. \textcolor{ForestGreen}{\{choiceD\}} \\ E. \textcolor{ForestGreen}{\{choiceE\}} \\  \\  \\ The correct answer is: 

\paragraph{Unobserved - 02} Source: Annotator \\ [1ex]
\textbf{Input}: \textcolor{MidnightBlue}{\{question\}}, \textcolor{MidnightBlue}{\{context\}}, \textcolor{MidnightBlue}{\{clues\}}, \textcolor{ForestGreen}{\{choiceA\}}, \textcolor{ForestGreen}{\{choiceB\}},\textcolor{ForestGreen}{\{choiceC\}}, \textcolor{ForestGreen}{\{choiceD\}}, \textcolor{ForestGreen}{\{choiceE\}} \\ [0.5ex]
\textbf{Template}: \\ [0.5ex]
You are given a logic grid puzzle to test your sense of space and positions. You are given a context and some clues to pick the correct answer from the options to answer a question.Context: \textcolor{MidnightBlue}{\{context\}} \\ \textcolor{MidnightBlue}{\{clues\}} \\ Question: \textcolor{MidnightBlue}{\{question\}} \\ Options: \\ (A) \textcolor{ForestGreen}{\{choiceA\}} \\ (B) \textcolor{ForestGreen}{\{choiceB\}} \\ (C) \textcolor{ForestGreen}{\{choiceC\}} \\ (D) \textcolor{ForestGreen}{\{choiceD\}} \\ (E) \textcolor{ForestGreen}{\{choiceE\}} \\ Answer:

\paragraph{Unobserved - 03} Source: Annotator \\ [1ex]
\textbf{Input}: \textcolor{MidnightBlue}{\{question\}}, \textcolor{MidnightBlue}{\{context\}}, \textcolor{MidnightBlue}{\{clues\}}, \textcolor{ForestGreen}{\{choiceA\}}, \textcolor{ForestGreen}{\{choiceB\}},\textcolor{ForestGreen}{\{choiceC\}}, \textcolor{ForestGreen}{\{choiceD\}}, \textcolor{ForestGreen}{\{choiceE\}} \\ [0.5ex]
\textbf{Template}: \\ [0.5ex]
Question: \textcolor{MidnightBlue}{\{question\}}. Answer this question based on the following context and clues: \\  \\ Context: \textcolor{MidnightBlue}{\{context\}} \\ \textcolor{MidnightBlue}{\{clues\}} \\  \\ The answer is one of "1", "2", "3", "4", "5". The correct answer is:

\paragraph{Unobserved - 04} Source: Annotator \\ [1ex]
\textbf{Input}: \textcolor{MidnightBlue}{\{question\}}, \textcolor{MidnightBlue}{\{context\}}, \textcolor{MidnightBlue}{\{clues\}}, \textcolor{ForestGreen}{\{choiceA\}}, \textcolor{ForestGreen}{\{choiceB\}},\textcolor{ForestGreen}{\{choiceC\}}, \textcolor{ForestGreen}{\{choiceD\}}, \textcolor{ForestGreen}{\{choiceE\}} \\ [0.5ex]
\textbf{Template}: \\ [0.5ex]
You are a master at solving logic grid puzzles. Solve this: \textcolor{MidnightBlue}{\{context\}} \\  \\ \textcolor{MidnightBlue}{\{clues\}} \\  \\ \textcolor{MidnightBlue}{\{question\}}

\paragraph{Unobserved - 05} Source: Annotator \\ [1ex]
\textbf{Input}: \textcolor{MidnightBlue}{\{question\}}, \textcolor{MidnightBlue}{\{context\}}, \textcolor{MidnightBlue}{\{clues\}}, \textcolor{ForestGreen}{\{choiceA\}}, \textcolor{ForestGreen}{\{choiceB\}},\textcolor{ForestGreen}{\{choiceC\}}, \textcolor{ForestGreen}{\{choiceD\}}, \textcolor{ForestGreen}{\{choiceE\}} \\ [0.5ex]
\textbf{Template}: \\ [0.5ex]
\textcolor{MidnightBlue}{\{context\}} \\ \textcolor{MidnightBlue}{\{clues\}}Based on the puzzle provided above, \textcolor{MidnightBlue}{\{question\}} \\  \\ Options are: A: \textcolor{ForestGreen}{\{choiceA\}}, B: \textcolor{ForestGreen}{\{choiceB\}}, C: \textcolor{ForestGreen}{\{choiceC\}}, D: \textcolor{ForestGreen}{\{choiceD\}}, E: \textcolor{ForestGreen}{\{choiceE\}} \\  \\ Answer:

\paragraph{Unobserved - 06} Source: Annotator \\ [1ex]
\textbf{Input}: \textcolor{MidnightBlue}{\{question\}}, \textcolor{MidnightBlue}{\{context\}}, \textcolor{MidnightBlue}{\{clues\}}, \textcolor{ForestGreen}{\{choiceA\}}, \textcolor{ForestGreen}{\{choiceB\}},\textcolor{ForestGreen}{\{choiceC\}}, \textcolor{ForestGreen}{\{choiceD\}}, \textcolor{ForestGreen}{\{choiceE\}} \\ [0.5ex]
\textbf{Template}: \\ [0.5ex]
The task is to solve a logic grid puzzle. You will have the context to the problem and some clues to solve the puzzle. \\ Context: \textcolor{MidnightBlue}{\{context\}}\textcolor{MidnightBlue}{\{clues\}} \\ The question is: \textcolor{MidnightBlue}{\{question\}} \\ Options: \\ A. \textcolor{ForestGreen}{\{choiceA\}} \\ B. \textcolor{ForestGreen}{\{choiceB\}} \\ C. \textcolor{ForestGreen}{\{choiceC\}} \\ D. \textcolor{ForestGreen}{\{choiceD\}} \\ E. \textcolor{ForestGreen}{\{choiceE\}} \\ Output your answer as one of "A", "B", "C", "D", "E".

\paragraph{Unobserved - 07} Source: Annotator \\ [1ex]
\textbf{Input}: \textcolor{MidnightBlue}{\{question\}}, \textcolor{MidnightBlue}{\{context\}}, \textcolor{MidnightBlue}{\{clues\}}, \textcolor{ForestGreen}{\{choiceA\}}, \textcolor{ForestGreen}{\{choiceB\}},\textcolor{ForestGreen}{\{choiceC\}}, \textcolor{ForestGreen}{\{choiceD\}}, \textcolor{ForestGreen}{\{choiceE\}} \\ [0.5ex]
\textbf{Template}: \\ [0.5ex]
Here's a complex puzzle. Utitlize your skills to solve the puzzle by logic grid tables.\textcolor{MidnightBlue}{\{context\}} \\ \textcolor{MidnightBlue}{\{clues\}} \\ \textcolor{MidnightBlue}{\{question\}}

\paragraph{Unobserved - 08} Source: Annotator \\ [1ex]
\textbf{Input}: \textcolor{MidnightBlue}{\{question\}}, \textcolor{MidnightBlue}{\{context\}}, \textcolor{MidnightBlue}{\{clues\}}, \textcolor{ForestGreen}{\{choiceA\}}, \textcolor{ForestGreen}{\{choiceB\}},\textcolor{ForestGreen}{\{choiceC\}}, \textcolor{ForestGreen}{\{choiceD\}}, \textcolor{ForestGreen}{\{choiceE\}} \\ [0.5ex]
\textbf{Template}: \\ [0.5ex]
\textcolor{MidnightBlue}{\{question\}} \\ This question can only be answered if you fully understand the context: \textcolor{MidnightBlue}{\{context\}} \\ \textcolor{MidnightBlue}{\{clues\}}. Your choices are: \\ (A) \textcolor{ForestGreen}{\{choiceA\}} \\ (B) \textcolor{ForestGreen}{\{choiceB\}} \\ (C) \textcolor{ForestGreen}{\{choiceC\}} \\ (D) \textcolor{ForestGreen}{\{choiceD\}} \\ (E) \textcolor{ForestGreen}{\{choiceE\}} \\  \\ Answer:

\paragraph{Unobserved - 09} Source: Annotator \\ [1ex]
\textbf{Input}: \textcolor{MidnightBlue}{\{question\}}, \textcolor{MidnightBlue}{\{context\}}, \textcolor{MidnightBlue}{\{clues\}}, \textcolor{ForestGreen}{\{choiceA\}}, \textcolor{ForestGreen}{\{choiceB\}},\textcolor{ForestGreen}{\{choiceC\}}, \textcolor{ForestGreen}{\{choiceD\}}, \textcolor{ForestGreen}{\{choiceE\}} \\ [0.5ex]
\textbf{Template}: \\ [0.5ex]
You are tested for your ability to answer logic grid problems correctly. \textcolor{MidnightBlue}{\{question\}} \\ \textcolor{MidnightBlue}{\{clues\}} \\ \textcolor{MidnightBlue}{\{question\}} \\ The answer is always one of '1', '2', '3', '4', or '5'. Output your answer and give explanation

\paragraph{Unobserved - 10} Source: Annotator \\ [1ex]
\textbf{Input}: \textcolor{MidnightBlue}{\{question\}}, \textcolor{MidnightBlue}{\{context\}}, \textcolor{MidnightBlue}{\{clues\}}, \textcolor{ForestGreen}{\{choiceA\}}, \textcolor{ForestGreen}{\{choiceB\}},\textcolor{ForestGreen}{\{choiceC\}}, \textcolor{ForestGreen}{\{choiceD\}}, \textcolor{ForestGreen}{\{choiceE\}} \\ [0.5ex]
\textbf{Template}: \\ [0.5ex]
Context: \textcolor{MidnightBlue}{\{context\}} \\  \\ The following clues are always true: \textcolor{MidnightBlue}{\{clues\}} \\  \\ Now, infer the answer to this question: "\textcolor{MidnightBlue}{\{question\}}" and pick the correct answer from A) \textcolor{ForestGreen}{\{choiceA\}} B) \textcolor{ForestGreen}{\{choiceB\}} C) \textcolor{ForestGreen}{\{choiceC\}} D) \textcolor{ForestGreen}{\{choiceD\}} E) \textcolor{ForestGreen}{\{choiceE\}} \\  \\ Answer:

\subsubsection*{BBL - Conceptual Combinations}

\paragraph{Unobserved - 01} Source: Annotator \\ [1ex]
\textbf{Input}: \textcolor{MidnightBlue}{\{question\}}, \textcolor{MidnightBlue}{\{context\}}, \textcolor{ForestGreen}{\{choiceA\}}, \textcolor{ForestGreen}{\{choiceB\}},\textcolor{ForestGreen}{\{choiceC\}},\textcolor{ForestGreen}{\{choiceD\}} \\ [0.5ex]
\textbf{Template}: \\ [0.5ex]
\textcolor{MidnightBlue}{\{context\}} Question: \textcolor{MidnightBlue}{\{question\}} \\ The options are the following: \\ A. \textcolor{ForestGreen}{\{choiceA\}} \\ B. \textcolor{ForestGreen}{\{choiceB\}} \\ C. \textcolor{ForestGreen}{\{choiceC\}} \\ D. \textcolor{ForestGreen}{\{choiceD\}} \\ Use your common sense to output one of the letter "A", "B", "C", or "D" to indicate your answer.

\paragraph{Unobserved - 02} Source: Annotator \\ [1ex]
\textbf{Input}: \textcolor{MidnightBlue}{\{question\}}, \textcolor{MidnightBlue}{\{context\}}, \textcolor{ForestGreen}{\{choiceA\}}, \textcolor{ForestGreen}{\{choiceB\}},\textcolor{ForestGreen}{\{choiceC\}},\textcolor{ForestGreen}{\{choiceD\}} \\ [0.5ex]
\textbf{Template}: \\ [0.5ex]
You are given a concept or a factual context. Answer the multiple choice question based on the context by choosing from the choices provided. \\ Context: \textcolor{MidnightBlue}{\{context\}} \\ Question: \textcolor{MidnightBlue}{\{question\}} \\ Choices: \\ A. \textcolor{ForestGreen}{\{choiceA\}} \\ B. \textcolor{ForestGreen}{\{choiceB\}} \\ C. \textcolor{ForestGreen}{\{choiceC\}} \\ D. \textcolor{ForestGreen}{\{choiceD\}} \\ Answer:

\paragraph{Unobserved - 03} Source: Annotator \\ [1ex]
\textbf{Input}: \textcolor{MidnightBlue}{\{question\}}, \textcolor{MidnightBlue}{\{context\}}, \textcolor{ForestGreen}{\{choiceA\}}, \textcolor{ForestGreen}{\{choiceB\}},\textcolor{ForestGreen}{\{choiceC\}},\textcolor{ForestGreen}{\{choiceD\}} \\ [0.5ex]
\textbf{Template}: \\ [0.5ex]
You are a linguistic expert that knows most of the concepts and combinations of words. Now, answer the following question: \textcolor{MidnightBlue}{\{context\}} Question: \textcolor{MidnightBlue}{\{question\}} (A) \textcolor{ForestGreen}{\{choiceA\}} (B) \textcolor{ForestGreen}{\{choiceB\}} (C) \textcolor{ForestGreen}{\{choiceC\}} (D) \textcolor{ForestGreen}{\{choiceD\}} \\ Your answer is:

\paragraph{Unobserved - 04} Source: Annotator \\ [1ex]
\textbf{Input}: \textcolor{MidnightBlue}{\{question\}}, \textcolor{MidnightBlue}{\{context\}}, \textcolor{ForestGreen}{\{choiceA\}}, \textcolor{ForestGreen}{\{choiceB\}},\textcolor{ForestGreen}{\{choiceC\}},\textcolor{ForestGreen}{\{choiceD\}} \\ [0.5ex]
\textbf{Template}: \\ [0.5ex]
Answer the question about concepts combination. Specifically, you need to take contradictions, emergent properties, fanciful fictional combinations, homonyms, invented words, and surprising uncommon combinations into consideration. \textcolor{MidnightBlue}{\{context\}} Question: \textcolor{MidnightBlue}{\{question\}} (A) \textcolor{ForestGreen}{\{choiceA\}} (B) \textcolor{ForestGreen}{\{choiceB\}} (C) \textcolor{ForestGreen}{\{choiceC\}} (D) \textcolor{ForestGreen}{\{choiceD\}} \\ Your answer is:

\paragraph{Unobserved - 05} Source: Annotator \\ [1ex]
\textbf{Input}: \textcolor{MidnightBlue}{\{question\}}, \textcolor{MidnightBlue}{\{context\}}, \textcolor{ForestGreen}{\{choiceA\}}, \textcolor{ForestGreen}{\{choiceB\}},\textcolor{ForestGreen}{\{choiceC\}},\textcolor{ForestGreen}{\{choiceD\}} \\ [0.5ex]
\textbf{Template}: \\ [0.5ex]
Question: \textcolor{MidnightBlue}{\{question\}} \\ The options are:  \\ A. \textcolor{ForestGreen}{\{choiceA\}} \\ B. \textcolor{ForestGreen}{\{choiceB\}} \\ C. \textcolor{ForestGreen}{\{choiceC\}} \\ D. \textcolor{ForestGreen}{\{choiceD\}} \\ Here is a context to help you answer the question: \textcolor{MidnightBlue}{\{context\}}. Choose the best answer from "A", "B", "C", "D".

\paragraph{Unobserved - 06} Source: Annotator \\ [1ex]
\textbf{Input}: \textcolor{MidnightBlue}{\{question\}}, \textcolor{MidnightBlue}{\{context\}}, \textcolor{ForestGreen}{\{choiceA\}}, \textcolor{ForestGreen}{\{choiceB\}},\textcolor{ForestGreen}{\{choiceC\}},\textcolor{ForestGreen}{\{choiceD\}} \\ [0.5ex]
\textbf{Template}: \\ [0.5ex]
The following is a multiple-choice question answering problem about conceptual meaning of words. You should choose the answer that best answer the question based on the context. \textcolor{MidnightBlue}{\{context\}} Question: \textcolor{MidnightBlue}{\{question\}} \\ The options are: \\ A. \textcolor{ForestGreen}{\{choiceA\}} \\ B. \textcolor{ForestGreen}{\{choiceB\}} \\ C. \textcolor{ForestGreen}{\{choiceC\}} \\ D. \textcolor{ForestGreen}{\{choiceD\}} \\ Answer:

\paragraph{Unobserved - 07} Source: Annotator \\ [1ex]
\textbf{Input}: \textcolor{MidnightBlue}{\{question\}}, \textcolor{MidnightBlue}{\{context\}}, \textcolor{ForestGreen}{\{choiceA\}}, \textcolor{ForestGreen}{\{choiceB\}},\textcolor{ForestGreen}{\{choiceC\}},\textcolor{ForestGreen}{\{choiceD\}} \\ [0.5ex]
\textbf{Template}: \\ [0.5ex]
Context: \textcolor{MidnightBlue}{\{context\}}. Understand the context, and answer the following question: \textcolor{MidnightBlue}{\{question\}}Options: \\ (A) \textcolor{ForestGreen}{\{choiceA\}} \\ (B) \textcolor{ForestGreen}{\{choiceB\}} \\ (C) \textcolor{ForestGreen}{\{choiceC\}} \\ (D) \textcolor{ForestGreen}{\{choiceD\}} \\ Answer:

\paragraph{Unobserved - 08} Source: Annotator \\ [1ex]
\textbf{Input}: \textcolor{MidnightBlue}{\{question\}}, \textcolor{MidnightBlue}{\{context\}}, \textcolor{ForestGreen}{\{choiceA\}}, \textcolor{ForestGreen}{\{choiceB\}},\textcolor{ForestGreen}{\{choiceC\}},\textcolor{ForestGreen}{\{choiceD\}} \\ [0.5ex]
\textbf{Template}: \\ [0.5ex]
Linguistic Professor: \textcolor{MidnightBlue}{\{context\}} \textcolor{MidnightBlue}{\{question\}} \\ Student: can you provide the options? \\ Linguistic Professor: The choices are A) \textcolor{ForestGreen}{\{choiceA\}} B) \textcolor{ForestGreen}{\{choiceB\}} C) \textcolor{ForestGreen}{\{choiceC\}} D) \textcolor{ForestGreen}{\{choiceD\}} \\ Student: I got it. The answer is

\paragraph{Unobserved - 09} Source: Annotator \\ [1ex]
\textbf{Input}: \textcolor{MidnightBlue}{\{question\}}, \textcolor{MidnightBlue}{\{context\}}, \textcolor{ForestGreen}{\{choiceA\}}, \textcolor{ForestGreen}{\{choiceB\}},\textcolor{ForestGreen}{\{choiceC\}},\textcolor{ForestGreen}{\{choiceD\}} \\ [0.5ex]
\textbf{Template}: \\ [0.5ex]
The task is to answer the linguistic question about concepts combination. Context: \textcolor{MidnightBlue}{\{context\}} \\  \\ Question: \textcolor{MidnightBlue}{\{question\}} \\  \\ Options: \\ A. \textcolor{ForestGreen}{\{choiceA\}} \\ B. \textcolor{ForestGreen}{\{choiceB\}} \\ C. \textcolor{ForestGreen}{\{choiceC\}} \\ D. \textcolor{ForestGreen}{\{choiceD\}} \\  \\ Answer:

\paragraph{Unobserved - 10} Source: Annotator \\ [1ex]
\textbf{Input}: \textcolor{MidnightBlue}{\{question\}}, \textcolor{MidnightBlue}{\{context\}}, \textcolor{ForestGreen}{\{choiceA\}}, \textcolor{ForestGreen}{\{choiceB\}},\textcolor{ForestGreen}{\{choiceC\}},\textcolor{ForestGreen}{\{choiceD\}} \\ [0.5ex]
\textbf{Template}: \\ [0.5ex]
\textcolor{MidnightBlue}{\{question\}} \\ Options: A. \textcolor{ForestGreen}{\{choiceA\}}, B. \textcolor{ForestGreen}{\{choiceB\}}, C. \textcolor{ForestGreen}{\{choiceC\}}, or D. \textcolor{ForestGreen}{\{choiceD\}}. What is the correct answer to this conceptual combination question? Based on the context "\textcolor{MidnightBlue}{\{context\}}", I think the most accurate answer is

\subsubsection*{BBL - Play Dialog}

\paragraph{Unobserved - 01} Source: Annotator \\ [1ex]
\textbf{Input}: \textcolor{MidnightBlue}{\{play\}}, \textcolor{MidnightBlue}{\{line1\}}, \textcolor{MidnightBlue}{\{line2\}} \\ [0.5ex]
\textbf{Template}: \\ [0.5ex]
The following transcripts of dialogues have been taken from Shakespeare's plays, but the transcripts do not say who said what. Based on these contents and styles, your task is to identify whether the sentences in question were spoken by the same or different people. \\ \textcolor{MidnightBlue}{\{play\}} \\ From the above dialogue, are the lines \textcolor{MidnightBlue}{\{line1\}} and \textcolor{MidnightBlue}{\{line2\}} spoken by the same or different characters? \\  \\ Answer:

\paragraph{Unobserved - 02} Source: Annotator \\ [1ex]
\textbf{Input}: \textcolor{MidnightBlue}{\{play\}}, \textcolor{MidnightBlue}{\{line1\}}, \textcolor{MidnightBlue}{\{line2\}} \\ [0.5ex]
\textbf{Template}: \\ [0.5ex]
Below are transcripts of dialogues from Shakespeare plays. \\ \textcolor{MidnightBlue}{\{play\}} \\ Please identify whether the two scripts were spoken by the same people. Answer yes or no. \\ Script1: \textcolor{MidnightBlue}{\{line1\}} \\ Script2: \textcolor{MidnightBlue}{\{line2\}} \\ Answer:

\paragraph{Unobserved - 03} Source: Annotator \\ [1ex]
\textbf{Input}: \textcolor{MidnightBlue}{\{play\}}, \textcolor{MidnightBlue}{\{line1\}}, \textcolor{MidnightBlue}{\{line2\}} \\ [0.5ex]
\textbf{Template}: \\ [0.5ex]
You have read all the plays by Shakespeare. You surely recognized this dialogue:\textcolor{MidnightBlue}{\{play\}} \\ Now, are these two lines spoken by the same character or different characters? Answer from "same" or "different". \\ Line1: \textcolor{MidnightBlue}{\{line1\}} \\ Line2: \textcolor{MidnightBlue}{\{line2\}} \\ Your answer:

\paragraph{Unobserved - 04} Source: Annotator \\ [1ex]
\textbf{Input}: \textcolor{MidnightBlue}{\{play\}}, \textcolor{MidnightBlue}{\{line1\}}, \textcolor{MidnightBlue}{\{line2\}} \\ [0.5ex]
\textbf{Template}: \\ [0.5ex]
The following paragraph is a dialogue from one of Shakespeare's plays, but without the information of the corresponding speaking character. You need to decide whether the two lines I give you are lines of the same character. \\ \textcolor{MidnightBlue}{\{play\}} \\ From the above dialogue, are the lines \textcolor{MidnightBlue}{\{line1\}} and \textcolor{MidnightBlue}{\{line2\}} spoken by the same or different characters? \\  Answer:

\paragraph{Unobserved - 05} Source: Annotator \\ [1ex]
\textbf{Input}: \textcolor{MidnightBlue}{\{play\}}, \textcolor{MidnightBlue}{\{line1\}}, \textcolor{MidnightBlue}{\{line2\}} \\ [0.5ex]
\textbf{Template}: \\ [0.5ex]
Now you are a dramatist. The following transcripts of dialogues are taken from Shakespeare plays, but the transcripts do not mark who said what.  Your task is to identify whether the sentences in question were spoken by the same or different people. Here is the play: \\ \textcolor{MidnightBlue}{\{play\}} \\ Question: In the preceding dialogue, were the lines \textcolor{MidnightBlue}{\{line1\}} and \textcolor{MidnightBlue}{\{line2\}} spoken by the same person or different people? Please just give a short answer: same or different. \\  \\ Your Answer:

\paragraph{Unobserved - 06} Source: Annotator \\ [1ex]
\textbf{Input}: \textcolor{MidnightBlue}{\{play\}}, \textcolor{MidnightBlue}{\{line1\}}, \textcolor{MidnightBlue}{\{line2\}} \\ [0.5ex]
\textbf{Template}: \\ [0.5ex]
The following transcripts of dialogues have been taken from Shakespeare plays, but the transcripts do not say who said what. We have two sentences selected from the transcripts, please make a judgement whether the sentences are spoken by the same people. \\ \textcolor{MidnightBlue}{\{play\}} \\ From the above dialogue, are the lines \textcolor{MidnightBlue}{\{line1\}} and \textcolor{MidnightBlue}{\{line2\}} spoken by the same or different characters?

\paragraph{Unobserved - 07} Source: Annotator \\ [1ex]
\textbf{Input}: \textcolor{MidnightBlue}{\{play\}}, \textcolor{MidnightBlue}{\{line1\}}, \textcolor{MidnightBlue}{\{line2\}} \\ [0.5ex]
\textbf{Template}: \\ [0.5ex]
Dialogue: \textcolor{MidnightBlue}{\{play\}} \\ From the above dialogue, are the lines \textcolor{MidnightBlue}{\{line1\}} and \textcolor{MidnightBlue}{\{line2\}} spoken by the same or different characters? \\ Answer "same" or "different". \\ Answer:

\paragraph{Unobserved - 08} Source: Annotator \\ [1ex]
\textbf{Input}: \textcolor{MidnightBlue}{\{play\}}, \textcolor{MidnightBlue}{\{line1\}}, \textcolor{MidnightBlue}{\{line2\}} \\ [0.5ex]
\textbf{Template}: \\ [0.5ex]
In the context of the Shakespeare play, \textcolor{MidnightBlue}{\{play\}}, assess the given dialogue transcripts. Determine whether the sentences \textcolor{MidnightBlue}{\{line1\}} and \textcolor{MidnightBlue}{\{line2\}} were spoken by a single person or by different people. \\ Answer:

\paragraph{Unobserved - 09} Source: Annotator \\ [1ex]
\textbf{Input}: \textcolor{MidnightBlue}{\{play\}}, \textcolor{MidnightBlue}{\{line1\}}, \textcolor{MidnightBlue}{\{line2\}} \\ [0.5ex]
\textbf{Template}: \\ [0.5ex]
Play: \textcolor{MidnightBlue}{\{play\}} \\  In this play written by Shakespeare, classify whether \\ Character A: \textcolor{MidnightBlue}{\{line1\}} \\ Character B: \textcolor{MidnightBlue}{\{line2\}}are spoken by the same character or different ones? Answer 'Yes' or 'No' only.

\paragraph{Unobserved - 10} Source: Annotator \\ [1ex]
\textbf{Input}: \textcolor{MidnightBlue}{\{play\}}, \textcolor{MidnightBlue}{\{line1\}}, \textcolor{MidnightBlue}{\{line2\}} \\ [0.5ex]
\textbf{Template}: \\ [0.5ex]
Context: \textcolor{MidnightBlue}{\{play\}} \\ Question: read this dialogue selected from a play written by Shakespeare, are the lines \textcolor{MidnightBlue}{\{line1\}} and \textcolor{MidnightBlue}{\{line2\}} from the same character? \\ Options: \\ A) Yes \\ B) No. Answer:

\subsubsection*{BBL - Strategy QA}

\paragraph{Unobserved - 01} Source: Annotator \\ [1ex]
\textbf{Input}: \textcolor{MidnightBlue}{\{question\}} \\ [0.5ex]
\textbf{Template}: \\ [0.5ex]
You are given a question which requires reasoning steps that are implicit in the question. Please choose the best answer from  "yes" or "no" and provide an explanation. \\  \\ Question:\textcolor{MidnightBlue}{\{question\}} \\ Answer and Explanation:

\paragraph{Unobserved - 02} Source: Annotator \\ [1ex]
\textbf{Input}: \textcolor{MidnightBlue}{\{question\}} \\ [0.5ex]
\textbf{Template}: \\ [0.5ex]
Reason about the answer to the question. \textcolor{MidnightBlue}{\{question\}}

\paragraph{Unobserved - 03} Source: Annotator \\ [1ex]
\textbf{Input}: \textcolor{MidnightBlue}{\{question\}} \\ [0.5ex]
\textbf{Template}: \\ [0.5ex]
You are taking an exam where each question requires implicit reasoning steps to answer. The answer will always be either "yes" or "no". Please carefully consider the following question, its implications, and any related information you may need to provide the correct answer. \\ Question: \textcolor{MidnightBlue}{\{question\}} \\ Answer:

\paragraph{Unobserved - 04} Source: Annotator \\ [1ex]
\textbf{Input}: \textcolor{MidnightBlue}{\{question\}} \\ [0.5ex]
\textbf{Template}: \\ [0.5ex]
Answer questions that assume implicit reasoning steps in the question prompt: \textcolor{MidnightBlue}{\{question\}}

\paragraph{Unobserved - 05} Source: Annotator \\ [1ex]
\textbf{Input}: \textcolor{MidnightBlue}{\{question\}} \\ [0.5ex]
\textbf{Template}: \\ [0.5ex]
Input: \\ 	- question: \textcolor{MidnightBlue}{\{question\}} \\ Output: \\ 	- answer: 

\paragraph{Unobserved - 06} Source: Annotator \\ [1ex]
\textbf{Input}: \textcolor{MidnightBlue}{\{question\}} \\ [0.5ex]
\textbf{Template}: \\ [0.5ex]
Use logic and reasoning to answer the following questions with either "yes" or "no". \\ Question: \textcolor{MidnightBlue}{\{question\}} \\ Answer:

\paragraph{Unobserved - 07} Source: Annotator \\ [1ex]
\textbf{Input}: \textcolor{MidnightBlue}{\{question\}} \\ [0.5ex]
\textbf{Template}: \\ [0.5ex]
Please answer the following question, you should think step by step, but please use "yes" or "no" to answer. \\ Question: \textcolor{MidnightBlue}{\{question\}} \\ Answer:

\paragraph{Unobserved - 08} Source: Annotator \\ [1ex]
\textbf{Input}: \textcolor{MidnightBlue}{\{question\}} \\ [0.5ex]
\textbf{Template}: \\ [0.5ex]
Answer questions in which the required reasoning steps are implicit in the question. Please first answer "Yes" or "No" and then output your explanation. \\  \textcolor{MidnightBlue}{\{question\}} Answer:

\paragraph{Unobserved - 09} Source: Annotator \\ [1ex]
\textbf{Input}: \textcolor{MidnightBlue}{\{question\}} \\ [0.5ex]
\textbf{Template}: \\ [0.5ex]
Please give your answer to the following question, which should be answered yes or no. This question may require you to do implicit multi-hop reasoning. \\ Question: \textcolor{MidnightBlue}{\{question\}} \\  Answer:

\paragraph{Unobserved - 10} Source: Annotator \\ [1ex]
\textbf{Input}: \textcolor{MidnightBlue}{\{question\}} \\ [0.5ex]
\textbf{Template}: \\ [0.5ex]
This question needs to be solved via decomposing it into multiple sub-questions and make comparison among the results of the sub-questions. \\ The question is \textcolor{MidnightBlue}{\{question\}} \\ After decomposing the question, we find the answer is 

\subsubsection*{BBL - Strange Stories}

\paragraph{Unobserved - 01} Source: Annotator \\ [1ex]
\textbf{Input}: \textcolor{MidnightBlue}{\{question\}},  \textcolor{MidnightBlue}{\{context\}} \\ [0.5ex]
\textbf{Template}: \\ [0.5ex]
You are given a psychology question that asks you to provide a socially intelligent response after reading a short story. Please answer "yes" or "no" to the given question. \\  \\ \textcolor{MidnightBlue}{\{context\}} \\  \\ Quesiton: \textcolor{MidnightBlue}{\{question\}} \\ Answer:

\paragraph{Unobserved - 02} Source: Annotator \\ [1ex]
\textbf{Input}: \textcolor{MidnightBlue}{\{question\}},  \textcolor{MidnightBlue}{\{context\}} \\ [0.5ex]
\textbf{Template}: \\ [0.5ex]
Given a story, answer whether the question is true or false. \\ \textcolor{MidnightBlue}{\{context\}} \\ Q: \textcolor{MidnightBlue}{\{question\}} \\ A:

\paragraph{Unobserved - 03} Source: Annotator \\ [1ex]
\textbf{Input}: \textcolor{MidnightBlue}{\{question\}},  \textcolor{MidnightBlue}{\{context\}} \\ [0.5ex]
\textbf{Template}: \\ [0.5ex]
You are taking a test for reading comprehension. You will be presented with a story and asked a question related to the story. The answer to the question is either "yes" or "no". Please carefully consider the story below before selecting your answer. \\ Story: \textcolor{MidnightBlue}{\{context\}} \\ Question: \textcolor{MidnightBlue}{\{question\}} \\ Answer:

\paragraph{Unobserved - 04} Source: Annotator \\ [1ex]
\textbf{Input}: \textcolor{MidnightBlue}{\{question\}},  \textcolor{MidnightBlue}{\{context\}} \\ [0.5ex]
\textbf{Template}: \\ [0.5ex]
A psychology test with naturalistic short stories that measures social intelligence. Boolean options.\textcolor{MidnightBlue}{\{context\}} \\ Q: \textcolor{MidnightBlue}{\{question\}}Answer:

\paragraph{Unobserved - 05} Source: Annotator \\ [1ex]
\textbf{Input}: \textcolor{MidnightBlue}{\{question\}},  \textcolor{MidnightBlue}{\{context\}} \\ [0.5ex]
\textbf{Template}: \\ [0.5ex]
Story: \textcolor{MidnightBlue}{\{context\}} \\  Q: \textcolor{MidnightBlue}{\{question\}} \\  Output:

\paragraph{Unobserved - 06} Source: Annotator \\ [1ex]
\textbf{Input}: \textcolor{MidnightBlue}{\{question\}},  \textcolor{MidnightBlue}{\{context\}} \\ [0.5ex]
\textbf{Template}: \\ [0.5ex]
Given the following text, answer the question with either "yes" or "no": \\  \\ Text: \textcolor{MidnightBlue}{\{context\}} \\ Question: \textcolor{MidnightBlue}{\{question\}} \\ Answer:

\paragraph{Unobserved - 07} Source: Annotator \\ [1ex]
\textbf{Input}: \textcolor{MidnightBlue}{\{question\}},  \textcolor{MidnightBlue}{\{context\}} \\ [0.5ex]
\textbf{Template}: \\ [0.5ex]
Please read the following text and answer the question according to the content of the text, your answer should be "yes" or "no". \\ Text: \textcolor{MidnightBlue}{\{context\}} \\ Question: \textcolor{MidnightBlue}{\{question\}} \\ Answer:

\paragraph{Unobserved - 08} Source: Annotator \\ [1ex]
\textbf{Input}: \textcolor{MidnightBlue}{\{question\}},  \textcolor{MidnightBlue}{\{context\}} \\ [0.5ex]
\textbf{Template}: \\ [0.5ex]
Image you are taking a psychology test. Please read the given story and answer the question. Please answer "yes" or "No". \\ Story: \textcolor{MidnightBlue}{\{context\}} \\ Q: \textcolor{MidnightBlue}{\{question\}} \\ A:

\paragraph{Unobserved - 09} Source: Annotator \\ [1ex]
\textbf{Input}: \textcolor{MidnightBlue}{\{question\}},  \textcolor{MidnightBlue}{\{context\}} \\ [0.5ex]
\textbf{Template}: \\ [0.5ex]
Please give your answer to the following question, which should be answered yes or no. You should judge the correctness of the question according to the story. \\  \\ Story: \textcolor{MidnightBlue}{\{context\}} \\ Question: \textcolor{MidnightBlue}{\{question\}} \\ Answer:

\paragraph{Unobserved - 10} Source: Annotator \\ [1ex]
\textbf{Input}: \textcolor{MidnightBlue}{\{question\}},  \textcolor{MidnightBlue}{\{context\}} \\ [0.5ex]
\textbf{Template}: \\ [0.5ex]
The following story is associated with a question, the answer of which is "yes" or "no".  \\  \\ Story: \textcolor{MidnightBlue}{\{context\}} \\ Question: \textcolor{MidnightBlue}{\{question\}} \\  \\ According to the story, the answer is

\subsubsection*{BBL - Winowhy}

\paragraph{Unobserved - 01} Source: Annotator \\ [1ex]
\textbf{Input}: \textcolor{MidnightBlue}{\{question\}} \\ [0.5ex]
\textbf{Template}: \\ [0.5ex]
In the sentence: \textcolor{MidnightBlue}{\{question\}}. Is the pronoun reasoning correct? Please answer with either "correct" or "incorrect". Do not include any other words.

\paragraph{Unobserved - 02} Source: Annotator \\ [1ex]
\textbf{Input}: \textcolor{MidnightBlue}{\{question\}} \\ [0.5ex]
\textbf{Template}: \\ [0.5ex]
Verify if the reasoning about which words certain pronouns refer to in the given words is right, choose one answer from "correct" and "incorrect": \\ Reasoning:\textcolor{MidnightBlue}{\{question\}} \\ Answer:

\paragraph{Unobserved - 03} Source: Annotator \\ [1ex]
\textbf{Input}: \textcolor{MidnightBlue}{\{question\}} \\ [0.5ex]
\textbf{Template}: \\ [0.5ex]
Given the context: \textcolor{MidnightBlue}{\{question\}}, determine if the co-reference resolution and the explanationis correct by output either "correct" or "incorrect". Answer:

\paragraph{Unobserved - 04} Source: Annotator \\ [1ex]
\textbf{Input}: \textcolor{MidnightBlue}{\{question\}} \\ [0.5ex]
\textbf{Template}: \\ [0.5ex]
Context: \textcolor{MidnightBlue}{\{question\}} \\ Question: Is the pronoun referring to the correct object? Answer with"Yes" or "No".

\paragraph{Unobserved - 05} Source: Annotator \\ [1ex]
\textbf{Input}: \textcolor{MidnightBlue}{\{question\}} \\ [0.5ex]
\textbf{Template}: \\ [0.5ex]
Judge the correctness of the understanding of pronoun: \\ \textcolor{MidnightBlue}{\{question\}} \\ Give your answer as"correct" or "incorrect". Your answer:

\paragraph{Unobserved - 06} Source: Annotator \\ [1ex]
\textbf{Input}: \textcolor{MidnightBlue}{\{question\}} \\ [0.5ex]
\textbf{Template}: \\ [0.5ex]
You are tested on your understanding of pronoun. Here is a sentence followed by the explanation: \textcolor{MidnightBlue}{\{question\}} \\ Output "correct" if you think the explanation is correct; output "incorrect"If the explanation is wrong.

\paragraph{Unobserved - 07} Source: Annotator \\ [1ex]
\textbf{Input}: \textcolor{MidnightBlue}{\{question\}} \\ [0.5ex]
\textbf{Template}: \\ [0.5ex]
Read the following reasoning about who a particular pronoun refers to: \textcolor{MidnightBlue}{\{question\}} \\ Is the reasoning correct?

\paragraph{Unobserved - 08} Source: Annotator \\ [1ex]
\textbf{Input}: \textcolor{MidnightBlue}{\{question\}} \\ [0.5ex]
\textbf{Template}: \\ [0.5ex]
Read the following reasoning, and answer if its correct or incorrect. \textcolor{MidnightBlue}{\{question\}} \\

\paragraph{Unobserved - 09} Source: Annotator \\ [1ex]
\textbf{Input}: \textcolor{MidnightBlue}{\{question\}} \\ [0.5ex]
\textbf{Template}: \\ [0.5ex]
\textcolor{MidnightBlue}{\{question\}} The reasoning stated above is "correct" or "incorrect"? It is

\paragraph{Unobserved - 10} Source: Annotator \\ [1ex]
\textbf{Input}: \textcolor{MidnightBlue}{\{context\}}, \textcolor{MidnightBlue}{\{explanation\}}  \\ [0.5ex]
\textbf{Template}: \\ [0.5ex]
You will be given a sentence followed by an explanation of the use of pronouns in that sentence. Please answer if the explanation is correct or incorrect. \\  \\ Sentence: \textcolor{MidnightBlue}{\{context\}} \\ Explanation: \textcolor{MidnightBlue}{\{explanation\}} \\ Answer:

\subsubsection*{BBL - Language ID}

\paragraph{Unobserved - 01} Source: Annotator \\ [1ex]
\textbf{Input}: \textcolor{MidnightBlue}{\{sentence\}},  \textcolor{ForestGreen}{\{choiceA\}}, \textcolor{ForestGreen}{\{choiceB\}}, \textcolor{ForestGreen}{\{choiceC\}}, \textcolor{ForestGreen}{\{choiceD\}}, \textcolor{ForestGreen}{\{choiceE\}},  \textcolor{ForestGreen}{\{choiceF\}}, \textcolor{ForestGreen}{\{choiceG\}},\textcolor{ForestGreen}{\{choiceH\}}, \textcolor{ForestGreen}{\{choiceI\}}, \textcolor{ForestGreen}{\{choiceJ\}}, \textcolor{ForestGreen}{\{choiceK\}}  \\ [0.5ex]
\textbf{Template}: \\ [0.5ex]
Identify the correct language of the given sentence. Please choose the best answer from A, B, C, D, E, F, G, H, I, J, and K. \\  \\ Sentence: \textcolor{MidnightBlue}{\{sentence\}} \\ A: \textcolor{ForestGreen}{\{choiceA\}} \\ B: \textcolor{ForestGreen}{\{choiceB\}} \\ C: \textcolor{ForestGreen}{\{choiceC\}} \\ D: \textcolor{ForestGreen}{\{choiceD\}} \\ E: \textcolor{ForestGreen}{\{choiceE\}} \\ F: \textcolor{ForestGreen}{\{choiceF\}} \\ G: \textcolor{ForestGreen}{\{choiceG\}} \\ H: \textcolor{ForestGreen}{\{choiceH\}} \\ I: \textcolor{ForestGreen}{\{choiceI\}} \\ J: \textcolor{ForestGreen}{\{choiceJ\}} \\ K: \textcolor{ForestGreen}{\{choiceK\}} \\ Answer:

\paragraph{Unobserved - 02} Source: Annotator \\ [1ex]
\textbf{Input}: \textcolor{MidnightBlue}{\{sentence\}},  \textcolor{ForestGreen}{\{choiceA\}}, \textcolor{ForestGreen}{\{choiceB\}}, \textcolor{ForestGreen}{\{choiceC\}}, \textcolor{ForestGreen}{\{choiceD\}}, \textcolor{ForestGreen}{\{choiceE\}},  \textcolor{ForestGreen}{\{choiceF\}}, \textcolor{ForestGreen}{\{choiceG\}},\textcolor{ForestGreen}{\{choiceH\}}, \textcolor{ForestGreen}{\{choiceI\}}, \textcolor{ForestGreen}{\{choiceJ\}}, \textcolor{ForestGreen}{\{choiceK\}}  \\ [0.5ex]
\textbf{Template}: \\ [0.5ex]
\textcolor{MidnightBlue}{\{sentence\}} \\ What language is the language stated above? A: \textcolor{ForestGreen}{\{choiceA\}} B: \textcolor{ForestGreen}{\{choiceB\}} C: \textcolor{ForestGreen}{\{choiceC\}} D: \textcolor{ForestGreen}{\{choiceD\}} E: \textcolor{ForestGreen}{\{choiceE\}} F: \textcolor{ForestGreen}{\{choiceF\}} G: \textcolor{ForestGreen}{\{choiceG\}} H: \textcolor{ForestGreen}{\{choiceH\}} I: \textcolor{ForestGreen}{\{choiceI\}} J: \textcolor{ForestGreen}{\{choiceJ\}} K: \textcolor{ForestGreen}{\{choiceK\}}

\paragraph{Unobserved - 03} Source: Annotator \\ [1ex]
\textbf{Input}: \textcolor{MidnightBlue}{\{sentence\}},  \textcolor{ForestGreen}{\{choiceA\}}, \textcolor{ForestGreen}{\{choiceB\}}, \textcolor{ForestGreen}{\{choiceC\}}, \textcolor{ForestGreen}{\{choiceD\}}, \textcolor{ForestGreen}{\{choiceE\}},  \textcolor{ForestGreen}{\{choiceF\}}, \textcolor{ForestGreen}{\{choiceG\}},\textcolor{ForestGreen}{\{choiceH\}}, \textcolor{ForestGreen}{\{choiceI\}}, \textcolor{ForestGreen}{\{choiceJ\}}, \textcolor{ForestGreen}{\{choiceK\}}  \\ [0.5ex]
\textbf{Template}: \\ [0.5ex]
You are taking a test that requires you to identify the language a given sentence is written in. To help narrow down your choices, we’ve made this a multiple choice question. After carefully examining the sentence and each answer below, please select the correct language of the sentence from one of "A", "B", "C", "D", "E", "F", "G", "H", "I", "J", or "K" \\ Sentence: \textcolor{MidnightBlue}{\{sentence\}} \\ - A: \textcolor{ForestGreen}{\{choiceA\}} \\ - B: \textcolor{ForestGreen}{\{choiceB\}} \\ - C: \textcolor{ForestGreen}{\{choiceC\}} \\ - D: \textcolor{ForestGreen}{\{choiceD\}} \\ - E: \textcolor{ForestGreen}{\{choiceE\}} \\ - F: \textcolor{ForestGreen}{\{choiceF\}} \\ - G: \textcolor{ForestGreen}{\{choiceG\}} \\ - H: \textcolor{ForestGreen}{\{choiceH\}} \\ - I: \textcolor{ForestGreen}{\{choiceI\}} \\ - J: \textcolor{ForestGreen}{\{choiceJ\}} \\ - K: \textcolor{ForestGreen}{\{choiceK\}} \\ Answer:

\paragraph{Unobserved - 04} Source: Annotator \\ [1ex]
\textbf{Input}: \textcolor{MidnightBlue}{\{sentence\}},  \textcolor{ForestGreen}{\{choiceA\}}, \textcolor{ForestGreen}{\{choiceB\}}, \textcolor{ForestGreen}{\{choiceC\}}, \textcolor{ForestGreen}{\{choiceD\}}, \textcolor{ForestGreen}{\{choiceE\}},  \textcolor{ForestGreen}{\{choiceF\}}, \textcolor{ForestGreen}{\{choiceG\}},\textcolor{ForestGreen}{\{choiceH\}}, \textcolor{ForestGreen}{\{choiceI\}}, \textcolor{ForestGreen}{\{choiceJ\}}, \textcolor{ForestGreen}{\{choiceK\}}  \\ [0.5ex]
\textbf{Template}: \\ [0.5ex]
Please select the language that correctly corresponds to the provided sentence from the following options: \\ Sentence: \textcolor{MidnightBlue}{\{sentence\}} \\ Options: \\ A: \textcolor{ForestGreen}{\{choiceA\}} \\ B: \textcolor{ForestGreen}{\{choiceB\}} \\ C: \textcolor{ForestGreen}{\{choiceC\}} \\ D: \textcolor{ForestGreen}{\{choiceD\}} \\ E: \textcolor{ForestGreen}{\{choiceE\}} \\ F: \textcolor{ForestGreen}{\{choiceF\}} \\ G: \textcolor{ForestGreen}{\{choiceG\}} \\ H: \textcolor{ForestGreen}{\{choiceH\}} \\ I: \textcolor{ForestGreen}{\{choiceI\}} \\ J: \textcolor{ForestGreen}{\{choiceJ\}} \\ K: \textcolor{ForestGreen}{\{choiceK\}} \\ Your answer:

\paragraph{Unobserved - 05} Source: Annotator \\ [1ex]
\textbf{Input}: \textcolor{MidnightBlue}{\{sentence\}},  \textcolor{ForestGreen}{\{choiceA\}}, \textcolor{ForestGreen}{\{choiceB\}}, \textcolor{ForestGreen}{\{choiceC\}}, \textcolor{ForestGreen}{\{choiceD\}}, \textcolor{ForestGreen}{\{choiceE\}},  \textcolor{ForestGreen}{\{choiceF\}}, \textcolor{ForestGreen}{\{choiceG\}},\textcolor{ForestGreen}{\{choiceH\}}, \textcolor{ForestGreen}{\{choiceI\}}, \textcolor{ForestGreen}{\{choiceJ\}}, \textcolor{ForestGreen}{\{choiceK\}}  \\ [0.5ex]
\textbf{Template}: \\ [0.5ex]
Input \\ 	- sentence: \textcolor{MidnightBlue}{\{sentence\}} \\ 	- A: \textcolor{ForestGreen}{\{choiceA\}} \\ 	- B: \textcolor{ForestGreen}{\{choiceB\}} \\ 	- C: \textcolor{ForestGreen}{\{choiceC\}} \\ 	- D: \textcolor{ForestGreen}{\{choiceD\}} \\ 	- E: \textcolor{ForestGreen}{\{choiceE\}} \\ 	- F: \textcolor{ForestGreen}{\{choiceF\}} \\ 	- G: \textcolor{ForestGreen}{\{choiceG\}} \\ 	- H: \textcolor{ForestGreen}{\{choiceH\}} \\ 	- I: \textcolor{ForestGreen}{\{choiceI\}} \\ 	- J: \textcolor{ForestGreen}{\{choiceJ\}} \\ 	- K: \textcolor{ForestGreen}{\{choiceK\}} \\ Output \\ 	- Answer:

\paragraph{Unobserved - 06} Source: Annotator \\ [1ex]
\textbf{Input}: \textcolor{MidnightBlue}{\{sentence\}},  \textcolor{ForestGreen}{\{choiceA\}}, \textcolor{ForestGreen}{\{choiceB\}}, \textcolor{ForestGreen}{\{choiceC\}}, \textcolor{ForestGreen}{\{choiceD\}}, \textcolor{ForestGreen}{\{choiceE\}},  \textcolor{ForestGreen}{\{choiceF\}}, \textcolor{ForestGreen}{\{choiceG\}},\textcolor{ForestGreen}{\{choiceH\}}, \textcolor{ForestGreen}{\{choiceI\}}, \textcolor{ForestGreen}{\{choiceJ\}}, \textcolor{ForestGreen}{\{choiceK\}}  \\ [0.5ex]
\textbf{Template}: \\ [0.5ex]
Given the following text, identify the correct language by selecting one of the options in the list (A, B, C, D, E, F, G, H, I, J, K): \\  \\ Text: \textcolor{MidnightBlue}{\{sentence\}} \\  \\ A: \textcolor{ForestGreen}{\{choiceA\}} \\ B: \textcolor{ForestGreen}{\{choiceB\}} \\ C: \textcolor{ForestGreen}{\{choiceC\}} \\ D: \textcolor{ForestGreen}{\{choiceD\}} \\ E: \textcolor{ForestGreen}{\{choiceE\}} \\ F: \textcolor{ForestGreen}{\{choiceF\}} \\ G: \textcolor{ForestGreen}{\{choiceG\}} \\ H: \textcolor{ForestGreen}{\{choiceH\}} \\ I: \textcolor{ForestGreen}{\{choiceI\}} \\ J: \textcolor{ForestGreen}{\{choiceJ\}} \\ K: \textcolor{ForestGreen}{\{choiceK\}} \\  \\ Answer:

\paragraph{Unobserved - 07} Source: Annotator \\ [1ex]
\textbf{Input}: \textcolor{MidnightBlue}{\{sentence\}},  \textcolor{ForestGreen}{\{choiceA\}}, \textcolor{ForestGreen}{\{choiceB\}}, \textcolor{ForestGreen}{\{choiceC\}}, \textcolor{ForestGreen}{\{choiceD\}}, \textcolor{ForestGreen}{\{choiceE\}},  \textcolor{ForestGreen}{\{choiceF\}}, \textcolor{ForestGreen}{\{choiceG\}},\textcolor{ForestGreen}{\{choiceH\}}, \textcolor{ForestGreen}{\{choiceI\}}, \textcolor{ForestGreen}{\{choiceJ\}}, \textcolor{ForestGreen}{\{choiceK\}}  \\ [0.5ex]
\textbf{Template}: \\ [0.5ex]
Please read the following sentence, then choose from the options which language you think it most likely came from. Your answer should be "A", "B", "C", "D", "E", "F", "G", "H", "I", "J", or "K" \\ Sentence: \textcolor{MidnightBlue}{\{sentence\}} \\ Options: \\ A: \textcolor{ForestGreen}{\{choiceA\}} \\ B: \textcolor{ForestGreen}{\{choiceB\}} \\ C: \textcolor{ForestGreen}{\{choiceC\}} \\ D: \textcolor{ForestGreen}{\{choiceD\}} \\ E: \textcolor{ForestGreen}{\{choiceE\}} \\ F: \textcolor{ForestGreen}{\{choiceF\}} \\ G: \textcolor{ForestGreen}{\{choiceG\}} \\ H: \textcolor{ForestGreen}{\{choiceH\}} \\ I: \textcolor{ForestGreen}{\{choiceI\}} \\ J: \textcolor{ForestGreen}{\{choiceJ\}} \\ K: \textcolor{ForestGreen}{\{choiceK\}} \\ Answer:

\paragraph{Unobserved - 08} Source: Annotator \\ [1ex]
\textbf{Input}: \textcolor{MidnightBlue}{\{sentence\}},  \textcolor{ForestGreen}{\{choiceA\}}, \textcolor{ForestGreen}{\{choiceB\}}, \textcolor{ForestGreen}{\{choiceC\}}, \textcolor{ForestGreen}{\{choiceD\}}, \textcolor{ForestGreen}{\{choiceE\}},  \textcolor{ForestGreen}{\{choiceF\}}, \textcolor{ForestGreen}{\{choiceG\}},\textcolor{ForestGreen}{\{choiceH\}}, \textcolor{ForestGreen}{\{choiceI\}}, \textcolor{ForestGreen}{\{choiceJ\}}, \textcolor{ForestGreen}{\{choiceK\}}  \\ [0.5ex]
\textbf{Template}: \\ [0.5ex]
Please give the language used in the following sentence. Each sentence will give five options, please output the corresponding option (i.e. A, B, C, D, E, F, G, H, I, J, or K) to represent the corresponding answer. \\  \\ Sentence: \textcolor{MidnightBlue}{\{sentence\}} \\ Options:: \\ A: \textcolor{ForestGreen}{\{choiceA\}} \\ B: \textcolor{ForestGreen}{\{choiceB\}} \\ C: \textcolor{ForestGreen}{\{choiceC\}} \\ D: \textcolor{ForestGreen}{\{choiceD\}} \\ E: \textcolor{ForestGreen}{\{choiceE\}} \\ F: \textcolor{ForestGreen}{\{choiceF\}} \\ G: \textcolor{ForestGreen}{\{choiceG\}} \\ H: \textcolor{ForestGreen}{\{choiceH\}} \\ I: \textcolor{ForestGreen}{\{choiceI\}} \\ J: \textcolor{ForestGreen}{\{choiceJ\}} \\ K: \textcolor{ForestGreen}{\{choiceK\}} \\ Answer:

\paragraph{Unobserved - 09} Source: Annotator \\ [1ex]
\textbf{Input}: \textcolor{MidnightBlue}{\{sentence\}},  \textcolor{ForestGreen}{\{choiceA\}}, \textcolor{ForestGreen}{\{choiceB\}}, \textcolor{ForestGreen}{\{choiceC\}}, \textcolor{ForestGreen}{\{choiceD\}}, \textcolor{ForestGreen}{\{choiceE\}},  \textcolor{ForestGreen}{\{choiceF\}}, \textcolor{ForestGreen}{\{choiceG\}},\textcolor{ForestGreen}{\{choiceH\}}, \textcolor{ForestGreen}{\{choiceI\}}, \textcolor{ForestGreen}{\{choiceJ\}}, \textcolor{ForestGreen}{\{choiceK\}}  \\ [0.5ex]
\textbf{Template}: \\ [0.5ex]
Given the sentence: \textcolor{MidnightBlue}{\{sentence\}}, select the correct language among the choices A. \textcolor{ForestGreen}{\{choiceA\}} B. \textcolor{ForestGreen}{\{choiceB\}} C. \textcolor{ForestGreen}{\{choiceC\}} D. \textcolor{ForestGreen}{\{choiceD\}} E. \textcolor{ForestGreen}{\{choiceE\}} F. \textcolor{ForestGreen}{\{choiceF\}} G. \textcolor{ForestGreen}{\{choiceG\}} H. \textcolor{ForestGreen}{\{choiceH\}} I. \textcolor{ForestGreen}{\{choiceI\}} J. \textcolor{ForestGreen}{\{choiceJ\}} K. \textcolor{ForestGreen}{\{choiceK\}} \\ - A: \textcolor{ForestGreen}{\{choiceA\}} \\ - B: \textcolor{ForestGreen}{\{choiceB\}} \\ - C: \textcolor{ForestGreen}{\{choiceC\}} \\ - D: \textcolor{ForestGreen}{\{choiceD\}} \\ - E: \textcolor{ForestGreen}{\{choiceE\}} \\ - F: \textcolor{ForestGreen}{\{choiceF\}} \\ - G: \textcolor{ForestGreen}{\{choiceG\}} \\ - H: \textcolor{ForestGreen}{\{choiceH\}} \\ - I: \textcolor{ForestGreen}{\{choiceI\}} \\ - J: \textcolor{ForestGreen}{\{choiceJ\}} \\ - K: \textcolor{ForestGreen}{\{choiceK\}} \\ Language:

\paragraph{Unobserved - 10} Source: Annotator \\ [1ex]
\textbf{Input}: \textcolor{MidnightBlue}{\{sentence\}},  \textcolor{ForestGreen}{\{choiceA\}}, \textcolor{ForestGreen}{\{choiceB\}}, \textcolor{ForestGreen}{\{choiceC\}}, \textcolor{ForestGreen}{\{choiceD\}}, \textcolor{ForestGreen}{\{choiceE\}},  \textcolor{ForestGreen}{\{choiceF\}}, \textcolor{ForestGreen}{\{choiceG\}},\textcolor{ForestGreen}{\{choiceH\}}, \textcolor{ForestGreen}{\{choiceI\}}, \textcolor{ForestGreen}{\{choiceJ\}}, \textcolor{ForestGreen}{\{choiceK\}}  \\ [0.5ex]
\textbf{Template}: \\ [0.5ex]
\textcolor{MidnightBlue}{\{sentence\}} \\  \\ This is a sentence written in one of \textcolor{ForestGreen}{\{choiceA\}}, \textcolor{ForestGreen}{\{choiceB\}}, \textcolor{ForestGreen}{\{choiceC\}}, \textcolor{ForestGreen}{\{choiceD\}}, \textcolor{ForestGreen}{\{choiceE\}}, \textcolor{ForestGreen}{\{choiceF\}}, \textcolor{ForestGreen}{\{choiceG\}}, \textcolor{ForestGreen}{\{choiceH\}}, \textcolor{ForestGreen}{\{choiceI\}}, \textcolor{ForestGreen}{\{choiceJ\}}, \textcolor{ForestGreen}{\{choiceK\}}. According to the words and the linguistic structure, I can tell that the language is:

\subsubsection*{BBL - Vitamin C}

\paragraph{Unobserved - 01} Source: Annotator \\ [1ex]
\textbf{Input}: \textcolor{MidnightBlue}{\{context\}}, \textcolor{MidnightBlue}{\{claim\}} \\ [0.5ex]
\textbf{Template}: \\ [0.5ex]
You are now a very experienced judge. Based only on the information contained in a brief quote from Wikipedia, answer whether the related claim is True, False or Neither. Use Neither when the Wikipedia quote does not provide the necessary information to resolve the question. \\  \\ \textcolor{MidnightBlue}{\{context\}} \\ Claim: \textcolor{MidnightBlue}{\{claim\}} \\ Is this True, False, or Neither?

\paragraph{Unobserved - 02} Source: Annotator \\ [1ex]
\textbf{Input}: \textcolor{MidnightBlue}{\{context\}}, \textcolor{MidnightBlue}{\{claim\}} \\ [0.5ex]
\textbf{Template}: \\ [0.5ex]
Now you are a Vitaminc Fact Verifier. Based only on the information contained in a brief quote from Wikipedia, answer whether the related claim is True, False or Neither. Use Neither when the Wikipedia quote does not provide the necessary information to resolve the question. \\  \\ \textcolor{MidnightBlue}{\{context\}} \\ Claim: \textcolor{MidnightBlue}{\{claim\}} \\ Question: Is this True, False, or Neither? \\ Your answer:

\paragraph{Unobserved - 03} Source: Annotator \\ [1ex]
\textbf{Input}: \textcolor{MidnightBlue}{\{context\}}, \textcolor{MidnightBlue}{\{claim\}} \\ [0.5ex]
\textbf{Template}: \\ [0.5ex]
\textcolor{MidnightBlue}{\{context\}} \\ Read the above paragraph, and answer the following claim \textcolor{MidnightBlue}{\{claim\}}. Answer True, Flase, or Neither. Neither means the Wikipedia quote does not provide the necessary information to resolve the question. Answer:

\paragraph{Unobserved - 04} Source: Annotator \\ [1ex]
\textbf{Input}: \textcolor{MidnightBlue}{\{context\}}, \textcolor{MidnightBlue}{\{claim\}} \\ [0.5ex]
\textbf{Template}: \\ [0.5ex]
Given a claim and its related information context from Wikipedia, determine whether the claim is True, False or Neither. Neither means the given information is not enough to decide if the claim is True or False, which is roughly equivalent to uncertain. \\  \\ Context:\textcolor{MidnightBlue}{\{context\}} \\ Claim: \textcolor{MidnightBlue}{\{claim\}} \\ True, False or Neither?

\paragraph{Unobserved - 05} Source: Annotator \\ [1ex]
\textbf{Input}: \textcolor{MidnightBlue}{\{context\}}, \textcolor{MidnightBlue}{\{claim\}} \\ [0.5ex]
\textbf{Template}: \\ [0.5ex]
\textcolor{MidnightBlue}{\{context\}} Claim: \textcolor{MidnightBlue}{\{claim\}} \\ Based on the context, is the claim true? false? Or Neither? Give your answer as one of "True", "False" or "Neither"

\paragraph{Unobserved - 06} Source: Annotator \\ [1ex]
\textbf{Input}: \textcolor{MidnightBlue}{\{context\}}, \textcolor{MidnightBlue}{\{claim\}} \\ [0.5ex]
\textbf{Template}: \\ [0.5ex]
Based only on the information contained in the given context, please make a judgement whether the related claim is True, False or Neither. \\  \\ \textcolor{MidnightBlue}{\{context\}} \\  Claim: \textcolor{MidnightBlue}{\{claim\}} \\ True, False, or Neither?

\paragraph{Unobserved - 07} Source: Annotator \\ [1ex]
\textbf{Input}: \textcolor{MidnightBlue}{\{context\}}, \textcolor{MidnightBlue}{\{claim\}} \\ [0.5ex]
\textbf{Template}: \\ [0.5ex]
Wikipedia: \textcolor{MidnightBlue}{\{context\}} \\ Someone: based on the given context, is the \textcolor{MidnightBlue}{\{claim\}} True, False, or Neither?

\paragraph{Unobserved - 08} Source: Annotator \\ [1ex]
\textbf{Input}: \textcolor{MidnightBlue}{\{context\}}, \textcolor{MidnightBlue}{\{claim\}} \\ [0.5ex]
\textbf{Template}: \\ [0.5ex]
Evaluate the related claim as True, False, or Neither based solely on the information given in the short Wikipedia excerpt. Select Neither when the excerpt doesn't provide sufficient information to address the question. \\ \textcolor{MidnightBlue}{\{context\}} \\ Claim: \textcolor{MidnightBlue}{\{claim\}} \\ Answer(True, False, or Neither):

\paragraph{Unobserved - 09} Source: Annotator \\ [1ex]
\textbf{Input}: \textcolor{MidnightBlue}{\{context\}}, \textcolor{MidnightBlue}{\{claim\}} \\ [0.5ex]
\textbf{Template}: \\ [0.5ex]
Input: \textcolor{MidnightBlue}{\{claim\}} \\ Verify the factually of the claim based on the following context \\ \textcolor{MidnightBlue}{\{context\}} \\  \\ - "True" if the claim is factually correct \\ - "False" if the claim is factually incorrect \\ - "Neither" if the factuality cannot be determined. Output you answerwith one of "True", "False", or "Neither". Answer:

\paragraph{Unobserved - 10} Source: Annotator \\ [1ex]
\textbf{Input}: \textcolor{MidnightBlue}{\{context\}}, \textcolor{MidnightBlue}{\{claim\}} \\ [0.5ex]
\textbf{Template}: \\ [0.5ex]
Context: \textcolor{MidnightBlue}{\{context\}} \\ Now classify this claim into one of 'True', 'False', or 'Neither'. \\ \textcolor{MidnightBlue}{\{claim\}}

\subsection{Granular Experiment Instructions}
\label{section:granular_instructions}

In this section, we provide the instructions we used for all 6 settings by dataset.

\subsubsection*{BBH - Intent Recognition}

\paragraph{Closest - 1} Source: NIV2 Task 163 OpenPI Classification - Template 2 \\ [1ex] 
\textbf{Input}: \textcolor{MidnightBlue}{\{passage\}} \\ [0.5ex]
\textbf{Template}: \\ [0.5ex]
You will be given a definition of a task first, then some input of the task. \\ Given a passage as input, answer with the category to which the passage belongs. There are categories - \textcolor{ForestGreen}{\{options\}}. The answer should be one of the categories based on words from the passage which closely belong to the category. \\  \\ \textcolor{MidnightBlue}{\{passage\}} \\ Output:

\paragraph{Closest - 2} Source: NIV2 Task 163 OpenPI Classification - Template 4\\ [1ex]
\textbf{Input}: \textcolor{MidnightBlue}{\{passage\}} \\ [0.5ex]
\textbf{Template}: \\ [0.5ex]
Instructions: Given a passage as input, answer with the category to which the passage belongs. There are categories - \textcolor{ForestGreen}{\{options\}}. The answer should be one of the categories based on words from the passage which closely belong to the category. \\ Input: \textcolor{MidnightBlue}{\{passage\}} \\ Output:

\paragraph{Closest - 3} Source: NIV2 Task 163 OpenPI Classification - Template 6\\ [1ex]
\textbf{Input}: \textcolor{MidnightBlue}{\{passage\}} \\ [0.5ex]
\textbf{Template}: \\ [0.5ex]
Given the task definition and input, reply with output. Given a passage as input, answer with the category to which the passage belongs. There are categories - \textcolor{ForestGreen}{\{options\}}. The answer should be one of the categories based on words from the passage which closely belong to the category. \\  \\ \textcolor{MidnightBlue}{\{passage\}} \\ 

\paragraph{Closest - 4} Source: NIV2 Task 163 OpenPI Classification - Template 8\\ [1ex]
\textbf{Input}: \textcolor{MidnightBlue}{\{passage\}} \\ [0.5ex]
\textbf{Template}: \\ [0.5ex]
Q: Given a passage as input, answer with the category to which the passage belongs. There are categories - \textcolor{ForestGreen}{\{options\}}. The answer should be one of the categories based on words from the passage which closely belong to the category. \\ \textcolor{MidnightBlue}{\{passage\}} \\ A:

\paragraph{Closest - 5} Source: NIV2 Task 163 OpenPI Classification - Template 10\\ [1ex]
\textbf{Input}: \textcolor{MidnightBlue}{\{passage\}} \\ [0.5ex]
\textbf{Template}: \\ [0.5ex]
Detailed Instructions: Given a passage as input, answer with the category to which the passage belongs. There are categories - \textcolor{ForestGreen}{\{options\}}. The answer should be one of the categories based on words from the passage which closely belong to the category. \\ Q: \textcolor{MidnightBlue}{\{passage\}} \\ A:

\paragraph{Incorrect - 1} Source: NIV2 Task 562 Language Identification - Template 10\\ [1ex]
\textbf{Input}: \textcolor{MidnightBlue}{\{text\}}, \textcolor{ForestGreen}{\{options\}} \\ [0.5ex]
\textbf{Template}: \\ [0.5ex]
Detailed Instructions: In this task, an input sentence is given which can be in the \textcolor{ForestGreen}{\{options\}} languages. There are a total of \textcolor{ForestGreen}{\{options length\}} languages. Your task is to identify the language of the input sentence. The input sentence can only be in any of the \textcolor{ForestGreen}{\{options length\}} languages provided. \\ Q: \textcolor{MidnightBlue}{\{text\}} \\ A:

\paragraph{Incorrect - 2} Source: NIV2 Task 1588 Tecla Classification - Template 10\\ [1ex]
\textbf{Input}: \textcolor{MidnightBlue}{\{text\}}, \textcolor{ForestGreen}{\{options\}} \\ [0.5ex]
\textbf{Template}: \\ [0.5ex]
Detailed Instructions: In this task, you are given a text in Catalan. Your task is to classify it into \textcolor{ForestGreen}{\{options length\}} different given themes. Names of all the classes are \textcolor{ForestGreen}{\{options\}} \\ Q: \textcolor{MidnightBlue}{\{text\}} \\ A:

\paragraph{Incorrect - 3} Source: NIV2 Task 564 DiscoFuse Classification - Template 10\\ [1ex]
\textbf{Input}: \textcolor{MidnightBlue}{\{text\}}, \textcolor{ForestGreen}{\{options\}} \\ [0.5ex]
\textbf{Template}: \\ [0.5ex]
Detailed Instructions: In this task, you are given two sentences in the English language and your task is to classify them into one of their discourse types. A discourse type is an indicator to classify the given two sentences on the basis of a co-text as well as a relevant context. There are \textcolor{ForestGreen}{\{options length\}} discourse types in total which are \textcolor{ForestGreen}{\{options\}} \\ Q: \textcolor{MidnightBlue}{\{text\}} \\ A:

\paragraph{Incorrect - 4} Source: NIV2 Task 1193 Course Classification - Template 10\\ [1ex]
\textbf{Input}: \textcolor{MidnightBlue}{\{text\}}, \textcolor{ForestGreen}{\{options\}} \\ [0.5ex]
\textbf{Template}: \\ [0.5ex]
Detailed Instructions: In this task, you are given the name of an Indian food dish. You need to classify the dish as a \textcolor{ForestGreen}{\{options\}} \\ Q: \textcolor{MidnightBlue}{\{text\}} \\ A:

\paragraph{Incorrect - 5} Source: Flan Sentiment140 - Template 2\\ [1ex]
\textbf{Input}: \textcolor{MidnightBlue}{\{text\}}, \textcolor{ForestGreen}{\{options\}} \\ [0.5ex]
\textbf{Template}: \\ [0.5ex]
\textcolor{MidnightBlue}{\{text\}} \\  \\ How would the sentiment of this tweet be described? \\ \textcolor{ForestGreen}{\{options\}}

\paragraph{Collected - 1} Source: Annotator \\ [1ex]
\textbf{Input}: \textcolor{MidnightBlue}{\{text\}}, \textcolor{ForestGreen}{\{options\}} \\ [0.5ex]
\textbf{Template}: \\ [0.5ex]
You are given a set of intentions to predict: \textcolor{ForestGreen}{\{options\}}. Pick the most suitable one to describe the following utterance: \textcolor{MidnightBlue}{\{text\}}. Intention:

\paragraph{Collected - 2} Source: Annotator \\ [1ex]
\textbf{Input}: \textcolor{MidnightBlue}{\{text\}}, \textcolor{ForestGreen}{\{options\}} \\ [0.5ex]
\textbf{Template}: \\ [0.5ex]
You are a dialogue assistance at recognizing and classifying user's intention. \\ Always respond with one of the \ options: [\textcolor{ForestGreen}{\{options\}}] to indicate the intention. \\ Utterance: \textcolor{MidnightBlue}{\{text\}} \\ Intention:

\paragraph{Collected - 3} Source: Annotator \\ [1ex]
\textbf{Input}: \textcolor{MidnightBlue}{\{text\}}, \textcolor{ForestGreen}{\{options\}} \\ [0.5ex]
\textbf{Template}: \\ [0.5ex]
The tasks is to classify the intention of the utterance: '\textcolor{MidnightBlue}{\{text\}}' into one of the followings: \textcolor{ForestGreen}{\{options\}}. Your answer is: 

\paragraph{Collected - 4} Source: Annotator \\ [1ex]
\textbf{Input}: \textcolor{MidnightBlue}{\{text\}}, \textcolor{ForestGreen}{\{options\}} \\ [0.5ex]
\textbf{Template}: \\ [0.5ex]
Given the label space: \textcolor{ForestGreen}{\{options\}}, classify the intention of the given utterance. \\ \textcolor{MidnightBlue}{\{text\}} \\ Intention:

\paragraph{Collected - 5} Source: Annotator \\ [1ex]
\textbf{Input}: \textcolor{MidnightBlue}{\{text\}}, \textcolor{ForestGreen}{\{options\}} \\ [0.5ex]
\textbf{Template}: \\ [0.5ex]
Output the intention of the utterance from the list: \textcolor{ForestGreen}{\{options\}}. Output the exact word or phrase. \textcolor{MidnightBlue}{\{text\}}

\paragraph{Task Designer} See \textsc{Big-Bench} eval file.

\paragraph{Negation - 1} Source: NIV2 Task 163 OpenPI Classification - Template 2\\ [1ex]
\textbf{Input}: \textcolor{MidnightBlue}{\{passage\}} \\ [0.5ex]
\textbf{Template}: \\ [0.5ex]
You will be given a definition of a task first, then some input of the task. \\ Given a passage as input, answer with the category to which the passage \textcolor{red}{doesn't} belong. There are categories - \textcolor{ForestGreen}{\{options\}}. The answer should be one of the categories based on words from the passage which \textcolor{red}{doesn't} belong to the category. \\  \\ \textcolor{MidnightBlue}{\{passage\}} \\ Output:

\paragraph{Negation - 2} Source: NIV2 Task 163 OpenPI Classification - Template 4\\ [1ex]
\textbf{Input}: \textcolor{MidnightBlue}{\{passage\}} \\ [0.5ex]
\textbf{Template}: \\ [0.5ex]
Instructions: Given a passage as input, answer with the category to which the passage \textcolor{red}{doesn't} belong. There are categories - \textcolor{ForestGreen}{\{options\}}. The answer should be one of the categories based on words from the passage which \textcolor{red}{doesn't} belong to the category. \\ Input: \textcolor{MidnightBlue}{\{passage\}} \\ Output:

\paragraph{Negation - 3} Source: NIV2 Task 163 OpenPI Classification - Template 6\\ [1ex]
\textbf{Input}: \textcolor{MidnightBlue}{\{passage\}} \\ [0.5ex]
\textbf{Template}: \\ [0.5ex]
Given the task definition and input, reply with output. Given a passage as input, answer with the category to which the passage \textcolor{red}{doesn't} belong. There are categories - \textcolor{ForestGreen}{\{options\}}. The answer should be one of the categories based on words from the passage which \textcolor{red}{doesn't} belong to the category. \\  \\ \textcolor{MidnightBlue}{\{passage\}} \\ 

\paragraph{Negation - 4} Source: NIV2 Task 163 OpenPI Classification - Template 8\\ [1ex]
\textbf{Input}: \textcolor{MidnightBlue}{\{passage\}} \\ [0.5ex]
\textbf{Template}: \\ [0.5ex]
Q: Given a passage as input, answer with the category to which the passage \textcolor{red}{doesn't} belongs. There are categories - \textcolor{ForestGreen}{\{options\}}. The answer should be one of the categories based on words from the passage which \textcolor{red}{doesn't} belong to the category. \\ \textcolor{MidnightBlue}{\{passage\}} \\ A:

\paragraph{Negation - 5} Source: NIV2 Task 163 OpenPI Classification - Template 10\\ [1ex]
\textbf{Input}: \textcolor{MidnightBlue}{\{passage\}} \\ [0.5ex]
\textbf{Template}: \\ [0.5ex]
Detailed Instructions: Given a passage as input, answer with the category to which the passage \textcolor{red}{doesn't} belong. There are categories - \textcolor{ForestGreen}{\{options\}}. The answer should be one of the categories based on words from the passage which \textcolor{red}{doesn't} belong to the category. \\ Q: \textcolor{MidnightBlue}{\{passage\}} \\ A:

\paragraph{Nonsensical - 1} Source: Annotator \\ [1ex]
\textbf{Input}: \textcolor{MidnightBlue}{\{text\}} \\ [0.5ex]
\textbf{Template}: \\ [0.5ex]
\textcolor{red}{The council of street racoons demands you respond to their inquisition.} \textcolor{MidnightBlue}{\{text\}}

\paragraph{Nonsensical - 2} Source: Annotator \\ [1ex]
\textbf{Input}: \textcolor{MidnightBlue}{\{text\}} \\ [0.5ex]
\textbf{Template}: \\ [0.5ex]
\textcolor{red}{Surveillance birds query your knowledge of seed. } \textcolor{MidnightBlue}{\{text\}}

\paragraph{Nonsensical - 3} Source: Annotator \\ [1ex]
\textbf{Input}: \textcolor{MidnightBlue}{\{text\}} \\ [0.5ex]
\textbf{Template}: \\ [0.5ex]
\textcolor{red}{Darth Vader requires you to answer to the dark side } \textcolor{MidnightBlue}{\{text\}}

\paragraph{Nonsensical - 4} Source: Annotator \\ [1ex]
\textbf{Input}: \textcolor{MidnightBlue}{\{text\}} \\ [0.5ex]
\textbf{Template}: \\ [0.5ex]
\textcolor{red}{Respond to the requirement of the Mars working dolphin union.} \textcolor{MidnightBlue}{\{text\}}

\paragraph{Nonsensical - 5} Source: Annotator \\ [1ex]
\textbf{Input}: \textcolor{MidnightBlue}{\{text\}} \\ [0.5ex]
\textbf{Template}: \\ [0.5ex]
\textcolor{red}{You are undergoing the inquiry of court of the local squirrels.} \textcolor{MidnightBlue}{\{text\}}

\subsubsection*{BBH - Empirical Judgment}

\paragraph{Closest - 1} Source: NIV2 Task 163 OpenPI Classification - Template 2 \\ [1ex] 
\textbf{Input}: \textcolor{MidnightBlue}{\{passage\}} \\ [0.5ex]
\textbf{Template}: \\ [0.5ex]
You will be given a definition of a task first, then some input of the task. \\ Given a passage as input, answer with the category to which the passage belongs. There are categories - \textcolor{ForestGreen}{\{options\}}. The answer should be one of the categories based on words from the passage which closely belong to the category. \\  \\ \textcolor{MidnightBlue}{\{passage\}} \\ Output:

\paragraph{Closest - 2} Source: NIV2 Task 163 OpenPI Classification - Template 4\\ [1ex]
\textbf{Input}: \textcolor{MidnightBlue}{\{passage\}} \\ [0.5ex]
\textbf{Template}: \\ [0.5ex]
Instructions: Given a passage as input, answer with the category to which the passage belongs. There are categories - \textcolor{ForestGreen}{\{options\}}. The answer should be one of the categories based on words from the passage which closely belong to the category. \\ Input: \textcolor{MidnightBlue}{\{passage\}} \\ Output:

\paragraph{Closest - 3} Source: NIV2 Task 163 OpenPI Classification - Template 6\\ [1ex]
\textbf{Input}: \textcolor{MidnightBlue}{\{passage\}} \\ [0.5ex]
\textbf{Template}: \\ [0.5ex]
Given the task definition and input, reply with output. Given a passage as input, answer with the category to which the passage belongs. There are categories - \textcolor{ForestGreen}{\{options\}}. The answer should be one of the categories based on words from the passage which closely belong to the category. \\  \\ \textcolor{MidnightBlue}{\{passage\}} \\ 

\paragraph{Closest - 4} Source: NIV2 Task 163 OpenPI Classification - Template 8\\ [1ex]
\textbf{Input}: \textcolor{MidnightBlue}{\{passage\}} \\ [0.5ex]
\textbf{Template}: \\ [0.5ex]
Q: Given a passage as input, answer with the category to which the passage belongs. There are categories - \textcolor{ForestGreen}{\{options\}}. The answer should be one of the categories based on words from the passage which closely belong to the category. \\ \textcolor{MidnightBlue}{\{passage\}} \\ A:

\paragraph{Closest - 5} Source: NIV2 Task 163 OpenPI Classification - Template 10\\ [1ex]
\textbf{Input}: \textcolor{MidnightBlue}{\{passage\}} \\ [0.5ex]
\textbf{Template}: \\ [0.5ex]
Detailed Instructions: Given a passage as input, answer with the category to which the passage belongs. There are categories - \textcolor{ForestGreen}{\{options\}}. The answer should be one of the categories based on words from the passage which closely belong to the category. \\ Q: \textcolor{MidnightBlue}{\{passage\}} \\ A:

\paragraph{Incorrect - 1} Source: NIV2 Task 143 Odd Man Out Classification - Template 10\\ [1ex]
\textbf{Input}: \textcolor{MidnightBlue}{\{input\}}, \textcolor{ForestGreen}{\{categories\}} \\ [0.5ex]
\textbf{Template}: \\ [0.5ex]
Detailed Instructions: Given a set of four words, generate the category that the words belong to. Words are separated by commas. The possible categories are \textcolor{ForestGreen}{\{categories\}} \\ Q: \textcolor{MidnightBlue}{\{input\}} \\ A:

\paragraph{Incorrect - 2} Source: NIV2 Task 137 Newscomm Classification - Template 10\\ [1ex]
\textbf{Input}: \textcolor{MidnightBlue}{\{input\}}, \textcolor{ForestGreen}{\{options\}} \\ [0.5ex]
\textbf{Template}: \\ [0.5ex]
Detailed Instructions: Classify the given news commentary into the language in which it is written in. There are \textcolor{ForestGreen}{\{options length\}} languages to classify the sentences into \textcolor{ForestGreen}{\{options\}} \\ Q: \textcolor{MidnightBlue}{\{input\}} \\ A:

\paragraph{Incorrect - 3} Source: Flan2021 - Sentiment140 - Template 1\\ [1ex]
\textbf{Input}: \textcolor{MidnightBlue}{\{input\}}, \textcolor{ForestGreen}{\{options\}} \\ [0.5ex]
\textbf{Template}: \\ [0.5ex]
\textcolor{MidnightBlue}{\{text\}} \\ Select your answer from the options. What is the sentiment of this tweet? \\ Options: \textcolor{ForestGreen}{\{options\}}...I think the answer is

\paragraph{Incorrect - 4} Source: Flan2021 - Sentiment140 - Template 6\\ [1ex]
\textbf{Input}: \textcolor{MidnightBlue}{\{input\}}, \textcolor{ForestGreen}{\{options\}} \\ [0.5ex]
\textbf{Template}: \\ [0.5ex]
Select your answer from the options. How would one describe the sentiment of this tweet? \\ \textcolor{MidnightBlue}{\{text\}} \\ \textcolor{ForestGreen}{\{options\}}

\paragraph{Incorrect - 5} Source: NIV2 Task 1422 MathQA Physics - Template 10\\ [1ex]
\textbf{Input}: \textcolor{MidnightBlue}{\{input\}}, \textcolor{ForestGreen}{\{options\}} \\ [0.5ex]
\textbf{Template}: \\ [0.5ex]
Detailed Instructions: In this task, you need to answer the given multiple-choice question on the physics. Classify your answers into \textcolor{ForestGreen}{\{letter length\}} \\ Q: Problem: \textcolor{MidnightBlue}{\{input\}}  \\ \textcolor{ForestGreen}{\{options\}} \\ A:

\paragraph{Collected - 1} Source: Annotator \\ [1ex]
\textbf{Input}: \textcolor{MidnightBlue}{\{events\}} \\ [0.5ex]
\textbf{Template}: \\ [0.5ex]
Two events are described in the following sentence: \textcolor{MidnightBlue}{\{events\}} \\ Classify the relation between the events into one of 'causal', 'correlative', or 'neutral'.

\paragraph{Collected - 2} Source: Annotator \\ [1ex]
\textbf{Input}: \textcolor{MidnightBlue}{\{events\}} \\ [0.5ex]
\textbf{Template}: \\ [0.5ex]
Causal relation: two events have causal relation if one causes the other to happen. \\ Correlative relation: two events have correlative relation if there is no explicity causal relation but they are correlated. \\ Neutral relation: two events have no obvious correlation. \\  \\ \textcolor{MidnightBlue}{\{events\}} Do the events described in the sentence have causal, correlative, or neutral relation?

\paragraph{Collected - 3} Source: Annotator \\ [1ex]
\textbf{Input}: \textcolor{MidnightBlue}{\{events\}} \\ [0.5ex]
\textbf{Template}: \\ [0.5ex]
Sentence: \textcolor{MidnightBlue}{\{events\}} Make a judgment about the relation of the events in the sentence. The possible relations are: "causal", "correlative", "neutral"

\paragraph{Collected - 4} Source: Annotator \\ [1ex]
\textbf{Input}: \textcolor{MidnightBlue}{\{events\}} \\ [0.5ex]
\textbf{Template}: \\ [0.5ex]
What is the relation between the events: \textcolor{MidnightBlue}{\{events\}} \\ Classify it into "causal", "correlative", "neutral"

\paragraph{Collected - 5} Source: Annotator \\ [1ex]
\textbf{Input}: \textcolor{MidnightBlue}{\{events\}} \\ [0.5ex]
\textbf{Template}: \\ [0.5ex]
You are given a sentence that describe two or more events. Now, classify the relation into one of "causal", "correlative", "neutral". \\ Sentence: "\textcolor{MidnightBlue}{\{events\}}" \\ Answer: 

\paragraph{Task Designer} See \textsc{Big-Bench} eval file.

\paragraph{Negation - 1} Source: NIV2 Task 163 OpenPI Classification - Template 2\\ [1ex]
\textbf{Input}: \textcolor{MidnightBlue}{\{passage\}} \\ [0.5ex]
\textbf{Template}: \\ [0.5ex]
You will be given a definition of a task first, then some input of the task. \\ Given a passage as input, answer with the category to which the passage \textcolor{red}{doesn't} belong. There are categories - \textcolor{ForestGreen}{\{options\}}. The answer should be one of the categories based on words from the passage which \textcolor{red}{doesn't} belong to the category. \\  \\ \textcolor{MidnightBlue}{\{passage\}} \\ Output:

\paragraph{Negation - 2} Source: NIV2 Task 163 OpenPI Classification - Template 4\\ [1ex]
\textbf{Input}: \textcolor{MidnightBlue}{\{passage\}} \\ [0.5ex]
\textbf{Template}: \\ [0.5ex]
Instructions: Given a passage as input, answer with the category to which the passage \textcolor{red}{doesn't} belong. There are categories - \textcolor{ForestGreen}{\{options\}}. The answer should be one of the categories based on words from the passage which \textcolor{red}{doesn't} belong to the category. \\ Input: \textcolor{MidnightBlue}{\{passage\}} \\ Output:

\paragraph{Negation - 3} Source: NIV2 Task 163 OpenPI Classification - Template 6\\ [1ex]
\textbf{Input}: \textcolor{MidnightBlue}{\{passage\}} \\ [0.5ex]
\textbf{Template}: \\ [0.5ex]
Given the task definition and input, reply with output. Given a passage as input, answer with the category to which the passage \textcolor{red}{doesn't} belong. There are categories - \textcolor{ForestGreen}{\{options\}}. The answer should be one of the categories based on words from the passage which \textcolor{red}{doesn't} belong to the category. \\  \\ \textcolor{MidnightBlue}{\{passage\}} \\ 

\paragraph{Negation - 4} Source: NIV2 Task 163 OpenPI Classification - Template 8\\ [1ex]
\textbf{Input}: \textcolor{MidnightBlue}{\{passage\}} \\ [0.5ex]
\textbf{Template}: \\ [0.5ex]
Q: Given a passage as input, answer with the category to which the passage \textcolor{red}{doesn't} belongs. There are categories - \textcolor{ForestGreen}{\{options\}}. The answer should be one of the categories based on words from the passage which \textcolor{red}{doesn't} belong to the category. \\ \textcolor{MidnightBlue}{\{passage\}} \\ A:

\paragraph{Negation - 5} Source: NIV2 Task 163 OpenPI Classification - Template 10\\ [1ex]
\textbf{Input}: \textcolor{MidnightBlue}{\{passage\}} \\ [0.5ex]
\textbf{Template}: \\ [0.5ex]
Detailed Instructions: Given a passage as input, answer with the category to which the passage \textcolor{red}{doesn't} belong. There are categories - \textcolor{ForestGreen}{\{options\}}. The answer should be one of the categories based on words from the passage which \textcolor{red}{doesn't} belong to the category. \\ Q: \textcolor{MidnightBlue}{\{passage\}} \\ A:

\paragraph{Nonsensical - 1} Source: Annotator \\ [1ex]
\textbf{Input}: \textcolor{MidnightBlue}{\{text\}} \\ [0.5ex]
\textbf{Template}: \\ [0.5ex]
\textcolor{red}{The council of street raccoons demands you respond to their inquisition.} \textcolor{MidnightBlue}{\{text\}}

\paragraph{Nonsensical - 2} Source: Annotator \\ [1ex]
\textbf{Input}: \textcolor{MidnightBlue}{\{text\}} \\ [0.5ex]
\textbf{Template}: \\ [0.5ex]
\textcolor{red}{Surveillance birds query your knowledge of seed. } \textcolor{MidnightBlue}{\{text\}}

\paragraph{Nonsensical - 3} Source: Annotator \\ [1ex]
\textbf{Input}: \textcolor{MidnightBlue}{\{text\}} \\ [0.5ex]
\textbf{Template}: \\ [0.5ex]
\textcolor{red}{Darth Vader requires you to answer to the dark side } \textcolor{MidnightBlue}{\{text\}}

\paragraph{Nonsensical - 4} Source: Annotator \\ [1ex]
\textbf{Input}: \textcolor{MidnightBlue}{\{text\}} \\ [0.5ex]
\textbf{Template}: \\ [0.5ex]
\textcolor{red}{Respond to the requirement of the Mars working dolphin union.} \textcolor{MidnightBlue}{\{text\}}

\paragraph{Nonsensical - 5} Source: Annotator \\ [1ex]
\textbf{Input}: \textcolor{MidnightBlue}{\{text\}} \\ [0.5ex]
\textbf{Template}: \\ [0.5ex]
\textcolor{red}{You are undergoing the inquiry of court of the local squirrels.} \textcolor{MidnightBlue}{\{text\}}

\subsubsection*{BBL - Conceptual Combinations}

\paragraph{Closest - 1} Source: Flan2021 - CosmosQA - Template 1 \\ [1ex]
\textbf{Input}: \textcolor{MidnightBlue}{\{context\}}, \textcolor{MidnightBlue}{\{question\}}, \textcolor{ForestGreen}{\{options\}} \\ [0.5ex]
\textbf{Template}: \\ [0.5ex]
\textcolor{MidnightBlue}{\{context\}} \\  \\ Question with options to choose from: \textcolor{MidnightBlue}{\{question\}} \\ OPTIONS:\textcolor{ForestGreen}{\{options\}}

\paragraph{Closest - 2} Source: Flan2021 - CosmosQA - Template 2 \\ [1ex]
\textbf{Input}: \textcolor{MidnightBlue}{\{context\}}, \textcolor{MidnightBlue}{\{question\}}, \textcolor{ForestGreen}{\{options\}} \\ [0.5ex]
\textbf{Template}: \\ [0.5ex]
\textcolor{MidnightBlue}{\{context\}} \\  \\ OPTIONS: \textcolor{ForestGreen}{\{options\}} \\ Q: \textcolor{MidnightBlue}{\{question\}}

\paragraph{Closest - 3} Source: Flan2021 - CosmosQA - Template 3 \\ [1ex]
\textbf{Input}: \textcolor{MidnightBlue}{\{context\}}, \textcolor{MidnightBlue}{\{question\}}, \textcolor{ForestGreen}{\{options\}} \\ [0.5ex]
\textbf{Template}: \\ [0.5ex]
\textcolor{MidnightBlue}{\{context\}} \\  \\ OPTIONS: \textcolor{ForestGreen}{\{options\}} \\ Answer the following question: \textcolor{MidnightBlue}{\{question\}} \\ 

\paragraph{Closest - 4} Source: Flan2021 - CosmosQA - Template 4 \\ [1ex]
\textbf{Input}: \textcolor{MidnightBlue}{\{context\}}, \textcolor{MidnightBlue}{\{question\}}, \textcolor{ForestGreen}{\{options\}} \\ [0.5ex]
\textbf{Template}: \\ [0.5ex]
\textcolor{MidnightBlue}{\{context\}} \\  \\ Based on the preceding passage, choose your answer for question \textcolor{MidnightBlue}{\{question\}} \\ OPTIONS: \textcolor{ForestGreen}{\{options\}}

\paragraph{Closest - 5} Source: Flan2021 - CosmosQA - Template 5 \\ [1ex]
\textbf{Input}: \textcolor{MidnightBlue}{\{context\}}, \textcolor{MidnightBlue}{\{question\}}, \textcolor{ForestGreen}{\{options\}} \\ [0.5ex]
\textbf{Template}: \\ [0.5ex]
\textcolor{MidnightBlue}{\{context\}} \\  \\ Q with options: Give answer the following question using evidence from the above passage: \textcolor{MidnightBlue}{\{question\}} \\ OPTIONS:\textcolor{ForestGreen}{\{options\}}

\paragraph{Closest - 6} Source: Flan2021 - CosmosQA - Template 6 \\ [1ex]
\textbf{Input}: \textcolor{MidnightBlue}{\{context\}}, \textcolor{MidnightBlue}{\{question\}}, \textcolor{ForestGreen}{\{options\}} \\ [0.5ex]
\textbf{Template}: \\ [0.5ex]
Context: \textcolor{MidnightBlue}{\{context\}} \\ Question \textcolor{MidnightBlue}{\{question\}} \\ Possible answers: \\ \textcolor{ForestGreen}{\{options\}} \\ The answer:

\paragraph{Closest - 7} Source: Flan2021 - CosmosQA - Template 7 \\ [1ex]
\textbf{Input}: \textcolor{MidnightBlue}{\{context\}}, \textcolor{MidnightBlue}{\{question\}}, \textcolor{ForestGreen}{\{options\}} \\ [0.5ex]
\textbf{Template}: \\ [0.5ex]
Read the following article and answer the question by choosing from the options. \\  \\ \textcolor{MidnightBlue}{\{context\}} \\  \\ \textcolor{MidnightBlue}{\{question\}} \\OPTIONS: \textcolor{ForestGreen}{\{options\}}...A:

\paragraph{Closest - 8} Source: Flan2021 - CosmosQA - Template 8 \\ [1ex]
\textbf{Input}: \textcolor{MidnightBlue}{\{context\}}, \textcolor{MidnightBlue}{\{question\}}, \textcolor{ForestGreen}{\{options\}} \\ [0.5ex]
\textbf{Template}: \\ [0.5ex]
This question has options. Answer the question about text: \\  \\ \textcolor{MidnightBlue}{\{context\}} \\  \\ \textcolor{MidnightBlue}{\{question\}} \\ OPTIONS:\textcolor{ForestGreen}{\{options\}}

\paragraph{Incorrect - 1} Source: Flan2021 WSC273 - Template 2\\ [1ex]
\textbf{Input}: \textcolor{MidnightBlue}{\{context\}}, \textcolor{ForestGreen}{\{options\}} \\ [0.5ex]
\textbf{Template}: \\ [0.5ex]
Complete the passage. \\  \\ \textcolor{MidnightBlue}{\{context\}} \\ OPTIONS: \textcolor{ForestGreen}{\{options\}}

\paragraph{Incorrect - 2} Source: Flan2021 WSC273 - Template 9\\ [1ex]
\textbf{Input}: \textcolor{MidnightBlue}{\{context\}}, \textcolor{ForestGreen}{\{options\}} \\ [0.5ex]
\textbf{Template}: \\ [0.5ex]
What is the next event listed in the options is correct? \\  \\ \textcolor{MidnightBlue}{\{context\}} \\ OPTIONS: \textcolor{ForestGreen}{\{options\}} \\ A:

\paragraph{Incorrect - 3} Source: Flan2021 Winograde - Template 3\\ [1ex]
\textbf{Input}: \textcolor{MidnightBlue}{\{context\}}, \textcolor{ForestGreen}{\{options\}} \\ [0.5ex]
\textbf{Template}: \\ [0.5ex]
Choose your story that continues the following story. \\  \\ \textcolor{MidnightBlue}{\{context\}} \\  \\ \textcolor{ForestGreen}{\{options\}}

\paragraph{Incorrect - 4} Source: Flan2021 Story Cloze - Template 1\\ [1ex]
\textbf{Input}: \textcolor{MidnightBlue}{\{context\}}, \textcolor{ForestGreen}{\{options\}} \\ [0.5ex]
\textbf{Template}: \\ [0.5ex]
\textcolor{MidnightBlue}{\{context\}} \\ \textcolor{ForestGreen}{\{options\}} \\ Which option is the next sentence?

\paragraph{Incorrect - 5} Source: Flan2021 Sentiment140 - Template 1\\ [1ex]
\textbf{Input}: \textcolor{MidnightBlue}{\{text\}}, \textcolor{ForestGreen}{\{options\}} \\ [0.5ex]
\textbf{Template}: \\ [0.5ex]
\textcolor{MidnightBlue}{\{text\}} \\ Select your answer from the options. What is the sentiment of this tweet? \\ \textcolor{ForestGreen}{\{options\}}...I think the answer is

\paragraph{Incorrect - 6} Source: NIV2 Task 1422 MathQA Physics - Template 10\\ [1ex]
\textbf{Input}: \textcolor{MidnightBlue}{\{input\}}, \textcolor{ForestGreen}{\{options\}} \\ [0.5ex]
\textbf{Template}: \\ [0.5ex]
Detailed Instructions: In this task, you need to answer the given multiple-choice question on the physics. Classify your answers into \textcolor{ForestGreen}{\{letter length\}} \\ Q: Problem: \textcolor{MidnightBlue}{\{input\}}  \\ \textcolor{ForestGreen}{\{options\}} \\ A:

\paragraph{Incorrect - 7} Source: NIV2 Task 1297 QASC - Template 10\\ [1ex]
\textbf{Input}: \textcolor{MidnightBlue}{\{input\}}, \textcolor{ForestGreen}{\{options\}} \\ [0.5ex]
\textbf{Template}: \\ [0.5ex]
Detailed Instructions: In this task, you are given two facts, and a multiple-choice question. Based on the given facts, answer the question with index of the correct option (e.g, "A"). \\ Q: \textcolor{MidnightBlue}{\{input\}} \textcolor{ForestGreen}{\{options\}} \\ A:

\paragraph{Collected - 01} Source: Annotator \\ [1ex]
\textbf{Input}: \textcolor{MidnightBlue}{\{question\}}, \textcolor{MidnightBlue}{\{context\}}, \textcolor{ForestGreen}{\{choiceA\}}, \textcolor{ForestGreen}{\{choiceB\}},\textcolor{ForestGreen}{\{choiceC\}},\textcolor{ForestGreen}{\{choiceD\}} \\ [0.5ex]
\textbf{Template}: \\ [0.5ex]
\textcolor{MidnightBlue}{\{context\}} Question: \textcolor{MidnightBlue}{\{question\}} \\ The options are the following: \\ A. \textcolor{ForestGreen}{\{choiceA\}} \\ B. \textcolor{ForestGreen}{\{choiceB\}} \\ C. \textcolor{ForestGreen}{\{choiceC\}} \\ D. \textcolor{ForestGreen}{\{choiceD\}} \\ Use your common sense to output one of the letter "A", "B", "C", or "D" to indicate your answer.

\paragraph{Collected - 02} Source: Annotator \\ [1ex]
\textbf{Input}: \textcolor{MidnightBlue}{\{question\}}, \textcolor{MidnightBlue}{\{context\}}, \textcolor{ForestGreen}{\{choiceA\}}, \textcolor{ForestGreen}{\{choiceB\}},\textcolor{ForestGreen}{\{choiceC\}},\textcolor{ForestGreen}{\{choiceD\}} \\ [0.5ex]
\textbf{Template}: \\ [0.5ex]
You are given a concept or a factual context. Answer the multiple choice question based on the context by choosing from the choices provided. \\ Context: \textcolor{MidnightBlue}{\{context\}} \\ Question: \textcolor{MidnightBlue}{\{question\}} \\ Choices: \\ A. \textcolor{ForestGreen}{\{choiceA\}} \\ B. \textcolor{ForestGreen}{\{choiceB\}} \\ C. \textcolor{ForestGreen}{\{choiceC\}} \\ D. \textcolor{ForestGreen}{\{choiceD\}} \\ Answer: 

\paragraph{Collected - 03} Source: Annotator \\ [1ex]
\textbf{Input}: \textcolor{MidnightBlue}{\{question\}}, \textcolor{MidnightBlue}{\{context\}}, \textcolor{ForestGreen}{\{choiceA\}}, \textcolor{ForestGreen}{\{choiceB\}},\textcolor{ForestGreen}{\{choiceC\}},\textcolor{ForestGreen}{\{choiceD\}} \\ [0.5ex]
\textbf{Template}: \\ [0.5ex]
You are a linguistic expert that knows most of the concepts and combinations of words. Now, answer the following question: \textcolor{MidnightBlue}{\{context\}} Question: \textcolor{MidnightBlue}{\{question\}} (A) \textcolor{ForestGreen}{\{choiceA\}} (B) \textcolor{ForestGreen}{\{choiceB\}} (C) \textcolor{ForestGreen}{\{choiceC\}} (D) \textcolor{ForestGreen}{\{choiceD\}} \\ Your answer is: 

\paragraph{Collected - 04} Source: Annotator \\ [1ex]
\textbf{Input}: \textcolor{MidnightBlue}{\{question\}}, \textcolor{MidnightBlue}{\{context\}}, \textcolor{ForestGreen}{\{choiceA\}}, \textcolor{ForestGreen}{\{choiceB\}},\textcolor{ForestGreen}{\{choiceC\}},\textcolor{ForestGreen}{\{choiceD\}} \\ [0.5ex]
\textbf{Template}: \\ [0.5ex]
Answer the question about concepts combination. Specifically, you need to take contradictions, emergent properties, fanciful fictional combinations, homonyms, invented words, and surprising uncommon combinations into consideration. \textcolor{MidnightBlue}{\{context\}} Question: \textcolor{MidnightBlue}{\{question\}} (A) \textcolor{ForestGreen}{\{choiceA\}} (B) \textcolor{ForestGreen}{\{choiceB\}} (C) \textcolor{ForestGreen}{\{choiceC\}} (D) \textcolor{ForestGreen}{\{choiceD\}} \\ Your answer is: 

\paragraph{Collected - 05} Source: Annotator \\ [1ex]
\textbf{Input}: \textcolor{MidnightBlue}{\{question\}}, \textcolor{MidnightBlue}{\{context\}}, \textcolor{ForestGreen}{\{choiceA\}}, \textcolor{ForestGreen}{\{choiceB\}},\textcolor{ForestGreen}{\{choiceC\}},\textcolor{ForestGreen}{\{choiceD\}} \\ [0.5ex]
\textbf{Template}: \\ [0.5ex]
Question: \textcolor{MidnightBlue}{\{question\}} \\ The options are:  \\ A. \textcolor{ForestGreen}{\{choiceA\}} \\ B. \textcolor{ForestGreen}{\{choiceB\}} \\ C. \textcolor{ForestGreen}{\{choiceC\}} \\ D. \textcolor{ForestGreen}{\{choiceD\}} \\ Here is a context to help you answer the question: \textcolor{MidnightBlue}{\{context\}}. Choose the best answer from "A", "B", "C", "D".

\paragraph{Collected - 06} Source: Annotator \\ [1ex]
\textbf{Input}: \textcolor{MidnightBlue}{\{question\}}, \textcolor{MidnightBlue}{\{context\}}, \textcolor{ForestGreen}{\{choiceA\}}, \textcolor{ForestGreen}{\{choiceB\}},\textcolor{ForestGreen}{\{choiceC\}},\textcolor{ForestGreen}{\{choiceD\}} \\ [0.5ex]
\textbf{Template}: \\ [0.5ex]
The following is a multiple-choice question answering problem about conceptual meaning of words. You should choose the answer that best answer the question based on the context. \textcolor{MidnightBlue}{\{context\}} Question: \textcolor{MidnightBlue}{\{question\}} \\ The options are: \\ A. \textcolor{ForestGreen}{\{choiceA\}} \\ B. \textcolor{ForestGreen}{\{choiceB\}} \\ C. \textcolor{ForestGreen}{\{choiceC\}} \\ D. \textcolor{ForestGreen}{\{choiceD\}} \\ Answer: 

\paragraph{Collected - 07} Source: Annotator \\ [1ex]
\textbf{Input}: \textcolor{MidnightBlue}{\{question\}}, \textcolor{MidnightBlue}{\{context\}}, \textcolor{ForestGreen}{\{choiceA\}}, \textcolor{ForestGreen}{\{choiceB\}},\textcolor{ForestGreen}{\{choiceC\}},\textcolor{ForestGreen}{\{choiceD\}} \\ [0.5ex]
\textbf{Template}: \\ [0.5ex]
Context: \textcolor{MidnightBlue}{\{context\}}. Understand the context, and answer the following question: \textcolor{MidnightBlue}{\{question\}}Options: \\ (A) \textcolor{ForestGreen}{\{choiceA\}} \\ (B) \textcolor{ForestGreen}{\{choiceB\}} \\ (C) \textcolor{ForestGreen}{\{choiceC\}} \\ (D) \textcolor{ForestGreen}{\{choiceD\}} \\ Answer:

\paragraph{Collected - 08} Source: Annotator \\ [1ex]
\textbf{Input}: \textcolor{MidnightBlue}{\{question\}}, \textcolor{MidnightBlue}{\{context\}}, \textcolor{ForestGreen}{\{choiceA\}}, \textcolor{ForestGreen}{\{choiceB\}},\textcolor{ForestGreen}{\{choiceC\}},\textcolor{ForestGreen}{\{choiceD\}} \\ [0.5ex]
\textbf{Template}: \\ [0.5ex]
Linguistic Professor: \textcolor{MidnightBlue}{\{context\}} \textcolor{MidnightBlue}{\{question\}} \\ Student: can you provide the options? \\ Linguistic Professor: The choices are A) \textcolor{ForestGreen}{\{choiceA\}} B) \textcolor{ForestGreen}{\{choiceB\}} C) \textcolor{ForestGreen}{\{choiceC\}} D) \textcolor{ForestGreen}{\{choiceD\}} \\ Student: I got it. The answer is 

\paragraph{Collected - 09} Source: Annotator \\ [1ex]
\textbf{Input}: \textcolor{MidnightBlue}{\{question\}}, \textcolor{MidnightBlue}{\{context\}}, \textcolor{ForestGreen}{\{choiceA\}}, \textcolor{ForestGreen}{\{choiceB\}},\textcolor{ForestGreen}{\{choiceC\}},\textcolor{ForestGreen}{\{choiceD\}} \\ [0.5ex]
\textbf{Template}: \\ [0.5ex]
The task is to answer the linguistic question about concepts combination. Context: \textcolor{MidnightBlue}{\{context\}} \\  \\ Question: \textcolor{MidnightBlue}{\{question\}} \\  \\ Options: \\ A. \textcolor{ForestGreen}{\{choiceA\}} \\ B. \textcolor{ForestGreen}{\{choiceB\}} \\ C. \textcolor{ForestGreen}{\{choiceC\}} \\ D. \textcolor{ForestGreen}{\{choiceD\}} \\  \\ Answer:

\paragraph{Collected - 10} Source: Annotator \\ [1ex]
\textbf{Input}: \textcolor{MidnightBlue}{\{question\}}, \textcolor{MidnightBlue}{\{context\}}, \textcolor{ForestGreen}{\{choiceA\}}, \textcolor{ForestGreen}{\{choiceB\}},\textcolor{ForestGreen}{\{choiceC\}},\textcolor{ForestGreen}{\{choiceD\}} \\ [0.5ex]
\textbf{Template}: \\ [0.5ex]
\textcolor{MidnightBlue}{\{question\}} \\ Options: A. \textcolor{ForestGreen}{\{choiceA\}}, B. \textcolor{ForestGreen}{\{choiceB\}}, C. \textcolor{ForestGreen}{\{choiceC\}}, or D. \textcolor{ForestGreen}{\{choiceD\}}. What is the correct answer to this conceptual combination question? Based on the context "\textcolor{MidnightBlue}{\{context\}}", I think the most accurate answer is

\paragraph{Task Designer} See \textsc{Big-Bench} eval file.

\paragraph{Negated - 1} Source: Flan2021 - CosmosQA - Template 1 \\ [1ex]
\textbf{Input}: \textcolor{MidnightBlue}{\{context\}}, \textcolor{MidnightBlue}{\{question\}}, \textcolor{ForestGreen}{\{options\}} \\ [0.5ex]
\textbf{Template}: \\ [0.5ex]
\textcolor{MidnightBlue}{\{context\}} \\  \\ Question with options to choose from: \textcolor{MidnightBlue}{\{question\}} \\ OPTIONS:\textcolor{ForestGreen}{\{options\}} \textcolor{red}{The answer is not:}

\paragraph{Negation - 2} Source: Flan2021 - CosmosQA - Template 2 \\ [1ex]
\textbf{Input}: \textcolor{MidnightBlue}{\{context\}}, \textcolor{MidnightBlue}{\{question\}}, \textcolor{ForestGreen}{\{options\}} \\ [0.5ex]
\textbf{Template}: \\ [0.5ex]
\textcolor{MidnightBlue}{\{context\}} \\  \\ OPTIONS: \textcolor{ForestGreen}{\{options\}} \\ Q: \textcolor{MidnightBlue}{\{question\}} \textcolor{red}{The answer is not:}

\paragraph{Negation - 3} Source: Flan2021 - CosmosQA - Template 3 \\ [1ex]
\textbf{Input}: \textcolor{MidnightBlue}{\{context\}}, \textcolor{MidnightBlue}{\{question\}}, \textcolor{ForestGreen}{\{options\}} \\ [0.5ex]
\textbf{Template}: \\ [0.5ex]
\textcolor{MidnightBlue}{\{context\}} \\  \\ OPTIONS: \textcolor{ForestGreen}{\{options\}} \\ Answer the following question: \textcolor{MidnightBlue}{\{question\}} \\ \textcolor{red}{The answer is not:}

\paragraph{Negation - 4} Source: Flan2021 - CosmosQA - Template 4 \\ [1ex]
\textbf{Input}: \textcolor{MidnightBlue}{\{context\}}, \textcolor{MidnightBlue}{\{question\}}, \textcolor{ForestGreen}{\{options\}} \\ [0.5ex]
\textbf{Template}: \\ [0.5ex]
\textcolor{MidnightBlue}{\{context\}} \\  \\ Based on the preceding passage, choose your answer for question \textcolor{MidnightBlue}{\{question\}} \\ OPTIONS: \textcolor{ForestGreen}{\{options\}} \\ \textcolor{red}{The answer is not:}

\paragraph{Negation - 5} Source: Flan2021 - CosmosQA - Template 5 \\ [1ex]
\textbf{Input}: \textcolor{MidnightBlue}{\{context\}}, \textcolor{MidnightBlue}{\{question\}}, \textcolor{ForestGreen}{\{options\}} \\ [0.5ex]
\textbf{Template}: \\ [0.5ex]
\textcolor{MidnightBlue}{\{context\}} \\  \\ Q with options: Give answer the following question using evidence from the above passage: \textcolor{MidnightBlue}{\{question\}} \\ OPTIONS:\textcolor{ForestGreen}{\{options\}} \\ \textcolor{red}{The answer is not:}

\paragraph{Negation - 6} Source: Flan2021 - CosmosQA - Template 6 \\ [1ex]
\textbf{Input}: \textcolor{MidnightBlue}{\{context\}}, \textcolor{MidnightBlue}{\{question\}}, \textcolor{ForestGreen}{\{options\}} \\ [0.5ex]
\textbf{Template}: \\ [0.5ex]
Context: \textcolor{MidnightBlue}{\{context\}} \\ Question \textcolor{MidnightBlue}{\{question\}} \\ Possible answers: \\ \textcolor{ForestGreen}{\{options\}} \\ The answer \textcolor{red}{is not:}:

\paragraph{Negation - 7} Source: Flan2021 - CosmosQA - Template 7 \\ [1ex]
\textbf{Input}: \textcolor{MidnightBlue}{\{context\}}, \textcolor{MidnightBlue}{\{question\}}, \textcolor{ForestGreen}{\{options\}} \\ [0.5ex]
\textbf{Template}: \\ [0.5ex]
Read the following article and answer the question by choosing from the options. \\  \\ \textcolor{MidnightBlue}{\{context\}} \\  \\ \textcolor{MidnightBlue}{\{question\}} \\OPTIONS: \textcolor{ForestGreen}{\{options\}}...\textcolor{red}{The answer is not:}:

\paragraph{Negation - 8} Source: Flan2021 - CosmosQA - Template 8 \\ [1ex]
\textbf{Input}: \textcolor{MidnightBlue}{\{context\}}, \textcolor{MidnightBlue}{\{question\}}, \textcolor{ForestGreen}{\{options\}} \\ [0.5ex]
\textbf{Template}: \\ [0.5ex]
This question has options. Answer the question about text: \\  \\ \textcolor{MidnightBlue}{\{context\}} \\  \\ \textcolor{MidnightBlue}{\{question\}} \\ OPTIONS:\textcolor{ForestGreen}{\{options\}} \\ \textcolor{red}{The answer is not:}

\paragraph{Nonsensical - 1} Source: Annotator \\ [1ex]
\textbf{Input}: \textcolor{MidnightBlue}{\{text\}} \\ [0.5ex]
\textbf{Template}: \\ [0.5ex]
\textcolor{red}{The council of street raccoons demands you respond to their inquisition.} \textcolor{MidnightBlue}{\{text\}}

\paragraph{Nonsensical - 2} Source: Annotator \\ [1ex]
\textbf{Input}: \textcolor{MidnightBlue}{\{text\}} \\ [0.5ex]
\textbf{Template}: \\ [0.5ex]
\textcolor{red}{Surveillance birds query your knowledge of seed. } \textcolor{MidnightBlue}{\{text\}}

\paragraph{Nonsensical - 3} Source: Annotator \\ [1ex]
\textbf{Input}: \textcolor{MidnightBlue}{\{text\}} \\ [0.5ex]
\textbf{Template}: \\ [0.5ex]
\textcolor{red}{Darth Vader requires you to answer to the dark side } \textcolor{MidnightBlue}{\{text\}}

\paragraph{Nonsensical - 4} Source: Annotator \\ [1ex]
\textbf{Input}: \textcolor{MidnightBlue}{\{text\}} \\ [0.5ex]
\textbf{Template}: \\ [0.5ex]
\textcolor{red}{Respond to the requirement of the Mars working dolphin union.} \textcolor{MidnightBlue}{\{text\}}

\paragraph{Nonsensical - 5} Source: Annotator \\ [1ex]
\textbf{Input}: \textcolor{MidnightBlue}{\{text\}} \\ [0.5ex]
\textbf{Template}: \\ [0.5ex]
\textcolor{red}{You are undergoing the inquiry of court of the local squirrels.} \textcolor{MidnightBlue}{\{text\}}

\subsubsection*{BBL - Language Identification}

\paragraph{Closest - 1} Source: NIV2 Task 137 Newscomm Classification - Template 2 \\ [1ex] 
\textbf{Input}: \textcolor{MidnightBlue}{\{passage\}} \\ [0.5ex]
\textbf{Template}: \\ [0.5ex]
You will be given a definition of a task first, then some input of the task. \\ Classify the given news commentary into the language in which it is written in. There are \textcolor{ForestGreen}{\{option length\}} languages to classify the sentences into \textcolor{ForestGreen}{\{options\}} \\  \\ \textcolor{MidnightBlue}{\{sentence\}} \\ Output:

\paragraph{Closest - 2} Source: NIV2 Task 137 Newscomm Classification - Template 4 \\ [1ex] 
\textbf{Input}: \textcolor{MidnightBlue}{\{passage\}} \\ [0.5ex]
\textbf{Template}: \\ [0.5ex]
Instructions: Classify the given news commentary into the language in which it is written in. There are \textcolor{ForestGreen}{\{option length\}} languages to classify the sentences into \textcolor{ForestGreen}{\{options\}} \\ Input: \textcolor{MidnightBlue}{\{sentence\}} \\ Output:

\paragraph{Closest - 3} Source: NIV2 Task 137 Newscomm Classification - Template 6 \\ [1ex] 
\textbf{Input}: \textcolor{MidnightBlue}{\{passage\}} \\ [0.5ex]
\textbf{Template}: \\ [0.5ex]
Given the task definition and input, reply with output. Classify the given news commentary into the language in which it is written in. There are \textcolor{ForestGreen}{\{option length\}} languages to classify the sentences into \textcolor{ForestGreen}{\{options\}} \\  \\ \textcolor{MidnightBlue}{\{sentence\}} \\ 

\paragraph{Closest - 4} Source: NIV2 Task 137 Newscomm Classification - Template 8 \\ [1ex] 
\textbf{Input}: \textcolor{MidnightBlue}{\{passage\}} \\ [0.5ex]
\textbf{Template}: \\ [0.5ex]
Q: Classify the given news commentary into the language in which it is written in. There are \textcolor{ForestGreen}{\{option length\}} languages to classify the sentences into \textcolor{ForestGreen}{\{options\}} \\ \textcolor{MidnightBlue}{\{sentence\}} \\ A:

\paragraph{Closest - 5} Source: NIV2 Task 137 Newscomm Classification - Template 10 \\ [1ex] 
\textbf{Input}: \textcolor{MidnightBlue}{\{passage\}} \\ [0.5ex]
\textbf{Template}: \\ [0.5ex]
Detailed Instructions: Classify the given news commentary into the language in which it is written in. There are \textcolor{ForestGreen}{\{option length\}} languages to classify the sentences into \textcolor{ForestGreen}{\{options\}} \\ Q: \textcolor{MidnightBlue}{\{sentence\}} \\ A:

\paragraph{Incorrect - 1} Source: NIV2 Task 143 Odd Man Out Classification - Template 10\\ [1ex]
\textbf{Input}: \textcolor{MidnightBlue}{\{input\}}, \textcolor{ForestGreen}{\{categories\}} \\ [0.5ex]
\textbf{Template}: \\ [0.5ex]
Detailed Instructions: Given a set of four words, generate the category that the words belong to. Words are separated by commas. The possible categories are \textcolor{ForestGreen}{\{categories\}} \\ Q: \textcolor{MidnightBlue}{\{input\}} \\ A:

\paragraph{Incorrect - 2} Source: NIV2 Task 1322 Government Type Classification - Template 10\\ [1ex]
\textbf{Input}: \textcolor{MidnightBlue}{\{input\}}, \textcolor{ForestGreen}{\{options\}} \\ [0.5ex]
\textbf{Template}: \\ [0.5ex]
Detailed Instructions:  In this task, you are given a country name and you need to answer with the government type of the country, as of the year 2015. The following are possible government types that are considered valid answers: \textcolor{ForestGreen}{\{options\}} \\ Q: \textcolor{MidnightBlue}{\{input\}} \\ A:

\paragraph{Incorrect - 3} Source: NIV2 Task 1422 MathQA Physics - Template 10\\ [1ex]
\textbf{Input}: \textcolor{MidnightBlue}{\{input\}}, \textcolor{ForestGreen}{\{options\}} \\ [0.5ex]
\textbf{Template}: \\ [0.5ex]
Detailed Instructions: In this task, you need to answer the given multiple-choice question on the physics. Classify your answers into \textcolor{ForestGreen}{\{letter length\}} \\ Q: Problem: \textcolor{MidnightBlue}{\{input\}}  \\ \textcolor{ForestGreen}{\{options\}} \\ A:

\paragraph{Incorrect - 4} Source: NIV2 Task 154 HateXPlain Classification - Template 10\\ [1ex]
\textbf{Input}: \textcolor{MidnightBlue}{\{input\}}, \textcolor{ForestGreen}{\{labels\}} \\ [0.5ex]
\textbf{Template}: \\ [0.5ex]
Detailed Instructions: The input is a tweet which can be Hate Speech, Offensive or Normal tweet. Hate Speech and Offensive tweets target one community. Given such a tweet, output the community targeted in the tweet. The community will be one of the nine values: \textcolor{ForestGreen}{\{labels\}}. Output 'None' if the tweet does not target any community. A tweet targets only one community. \\ Q: \textcolor{MidnightBlue}{\{input\}} \\ A:

\paragraph{Collected - 01} Source: Annotator \\ [1ex]
\textbf{Input}: \textcolor{MidnightBlue}{\{sentence\}},  \textcolor{ForestGreen}{\{choiceA\}}, \textcolor{ForestGreen}{\{choiceB\}}, \textcolor{ForestGreen}{\{choiceC\}}, \textcolor{ForestGreen}{\{choiceD\}}, \textcolor{ForestGreen}{\{choiceE\}},  \textcolor{ForestGreen}{\{choiceF\}}, \textcolor{ForestGreen}{\{choiceG\}},\textcolor{ForestGreen}{\{choiceH\}}, \textcolor{ForestGreen}{\{choiceI\}}, \textcolor{ForestGreen}{\{choiceJ\}}, \textcolor{ForestGreen}{\{choiceK\}}  \\ [0.5ex]
\textbf{Template}: \\ [0.5ex]
Identify the correct language of the given sentence. Please choose the best answer from A, B, C, D, E, F, G, H, I, J, and K. \\  \\ Sentence: \textcolor{MidnightBlue}{\{sentence\}} \\ A: \textcolor{ForestGreen}{\{choiceA\}} \\ B: \textcolor{ForestGreen}{\{choiceB\}} \\ C: \textcolor{ForestGreen}{\{choiceC\}} \\ D: \textcolor{ForestGreen}{\{choiceD\}} \\ E: \textcolor{ForestGreen}{\{choiceE\}} \\ F: \textcolor{ForestGreen}{\{choiceF\}} \\ G: \textcolor{ForestGreen}{\{choiceG\}} \\ H: \textcolor{ForestGreen}{\{choiceH\}} \\ I: \textcolor{ForestGreen}{\{choiceI\}} \\ J: \textcolor{ForestGreen}{\{choiceJ\}} \\ K: \textcolor{ForestGreen}{\{choiceK\}} \\ Answer:

\paragraph{Collected - 02} Source: Annotator \\ [1ex]
\textbf{Input}: \textcolor{MidnightBlue}{\{sentence\}},  \textcolor{ForestGreen}{\{choiceA\}}, \textcolor{ForestGreen}{\{choiceB\}}, \textcolor{ForestGreen}{\{choiceC\}}, \textcolor{ForestGreen}{\{choiceD\}}, \textcolor{ForestGreen}{\{choiceE\}},  \textcolor{ForestGreen}{\{choiceF\}}, \textcolor{ForestGreen}{\{choiceG\}},\textcolor{ForestGreen}{\{choiceH\}}, \textcolor{ForestGreen}{\{choiceI\}}, \textcolor{ForestGreen}{\{choiceJ\}}, \textcolor{ForestGreen}{\{choiceK\}}  \\ [0.5ex]
\textbf{Template}: \\ [0.5ex]
\textcolor{MidnightBlue}{\{sentence\}} \\ What language is the language stated above? A: \textcolor{ForestGreen}{\{choiceA\}} B: \textcolor{ForestGreen}{\{choiceB\}} C: \textcolor{ForestGreen}{\{choiceC\}} D: \textcolor{ForestGreen}{\{choiceD\}} E: \textcolor{ForestGreen}{\{choiceE\}} F: \textcolor{ForestGreen}{\{choiceF\}} G: \textcolor{ForestGreen}{\{choiceG\}} H: \textcolor{ForestGreen}{\{choiceH\}} I: \textcolor{ForestGreen}{\{choiceI\}} J: \textcolor{ForestGreen}{\{choiceJ\}} K: \textcolor{ForestGreen}{\{choiceK\}}

\paragraph{Collected - 03} Source: Annotator \\ [1ex]
\textbf{Input}: \textcolor{MidnightBlue}{\{sentence\}},  \textcolor{ForestGreen}{\{choiceA\}}, \textcolor{ForestGreen}{\{choiceB\}}, \textcolor{ForestGreen}{\{choiceC\}}, \textcolor{ForestGreen}{\{choiceD\}}, \textcolor{ForestGreen}{\{choiceE\}},  \textcolor{ForestGreen}{\{choiceF\}}, \textcolor{ForestGreen}{\{choiceG\}},\textcolor{ForestGreen}{\{choiceH\}}, \textcolor{ForestGreen}{\{choiceI\}}, \textcolor{ForestGreen}{\{choiceJ\}}, \textcolor{ForestGreen}{\{choiceK\}}  \\ [0.5ex]
\textbf{Template}: \\ [0.5ex]
You are taking a test that requires you to identify the language a given sentence is written in. To help narrow down your choices, we’ve made this a multiple choice question. After carefully examining the sentence and each answer below, please select the correct language of the sentence from one of "A", "B", "C", "D", "E", "F", "G", "H", "I", "J", or "K" \\ Sentence: \textcolor{MidnightBlue}{\{sentence\}} \\ - A: \textcolor{ForestGreen}{\{choiceA\}} \\ - B: \textcolor{ForestGreen}{\{choiceB\}} \\ - C: \textcolor{ForestGreen}{\{choiceC\}} \\ - D: \textcolor{ForestGreen}{\{choiceD\}} \\ - E: \textcolor{ForestGreen}{\{choiceE\}} \\ - F: \textcolor{ForestGreen}{\{choiceF\}} \\ - G: \textcolor{ForestGreen}{\{choiceG\}} \\ - H: \textcolor{ForestGreen}{\{choiceH\}} \\ - I: \textcolor{ForestGreen}{\{choiceI\}} \\ - J: \textcolor{ForestGreen}{\{choiceJ\}} \\ - K: \textcolor{ForestGreen}{\{choiceK\}} \\ Answer:

\paragraph{Collected - 04} Source: Annotator \\ [1ex]
\textbf{Input}: \textcolor{MidnightBlue}{\{sentence\}},  \textcolor{ForestGreen}{\{choiceA\}}, \textcolor{ForestGreen}{\{choiceB\}}, \textcolor{ForestGreen}{\{choiceC\}}, \textcolor{ForestGreen}{\{choiceD\}}, \textcolor{ForestGreen}{\{choiceE\}},  \textcolor{ForestGreen}{\{choiceF\}}, \textcolor{ForestGreen}{\{choiceG\}},\textcolor{ForestGreen}{\{choiceH\}}, \textcolor{ForestGreen}{\{choiceI\}}, \textcolor{ForestGreen}{\{choiceJ\}}, \textcolor{ForestGreen}{\{choiceK\}}  \\ [0.5ex]
\textbf{Template}: \\ [0.5ex]
Please select the language that correctly corresponds to the provided sentence from the following options: \\ Sentence: \textcolor{MidnightBlue}{\{sentence\}} \\ Options: \\ A: \textcolor{ForestGreen}{\{choiceA\}} \\ B: \textcolor{ForestGreen}{\{choiceB\}} \\ C: \textcolor{ForestGreen}{\{choiceC\}} \\ D: \textcolor{ForestGreen}{\{choiceD\}} \\ E: \textcolor{ForestGreen}{\{choiceE\}} \\ F: \textcolor{ForestGreen}{\{choiceF\}} \\ G: \textcolor{ForestGreen}{\{choiceG\}} \\ H: \textcolor{ForestGreen}{\{choiceH\}} \\ I: \textcolor{ForestGreen}{\{choiceI\}} \\ J: \textcolor{ForestGreen}{\{choiceJ\}} \\ K: \textcolor{ForestGreen}{\{choiceK\}} \\ Your answer:

\paragraph{Collected - 05} Source: Annotator \\ [1ex]
\textbf{Input}: \textcolor{MidnightBlue}{\{sentence\}},  \textcolor{ForestGreen}{\{choiceA\}}, \textcolor{ForestGreen}{\{choiceB\}}, \textcolor{ForestGreen}{\{choiceC\}}, \textcolor{ForestGreen}{\{choiceD\}}, \textcolor{ForestGreen}{\{choiceE\}},  \textcolor{ForestGreen}{\{choiceF\}}, \textcolor{ForestGreen}{\{choiceG\}},\textcolor{ForestGreen}{\{choiceH\}}, \textcolor{ForestGreen}{\{choiceI\}}, \textcolor{ForestGreen}{\{choiceJ\}}, \textcolor{ForestGreen}{\{choiceK\}}  \\ [0.5ex]
\textbf{Template}: \\ [0.5ex]
Input \\ 	- sentence: \textcolor{MidnightBlue}{\{sentence\}} \\ 	- A: \textcolor{ForestGreen}{\{choiceA\}} \\ 	- B: \textcolor{ForestGreen}{\{choiceB\}} \\ 	- C: \textcolor{ForestGreen}{\{choiceC\}} \\ 	- D: \textcolor{ForestGreen}{\{choiceD\}} \\ 	- E: \textcolor{ForestGreen}{\{choiceE\}} \\ 	- F: \textcolor{ForestGreen}{\{choiceF\}} \\ 	- G: \textcolor{ForestGreen}{\{choiceG\}} \\ 	- H: \textcolor{ForestGreen}{\{choiceH\}} \\ 	- I: \textcolor{ForestGreen}{\{choiceI\}} \\ 	- J: \textcolor{ForestGreen}{\{choiceJ\}} \\ 	- K: \textcolor{ForestGreen}{\{choiceK\}} \\ Output \\ 	- Answer:

\paragraph{Collected - 06} Source: Annotator \\ [1ex]
\textbf{Input}: \textcolor{MidnightBlue}{\{sentence\}},  \textcolor{ForestGreen}{\{choiceA\}}, \textcolor{ForestGreen}{\{choiceB\}}, \textcolor{ForestGreen}{\{choiceC\}}, \textcolor{ForestGreen}{\{choiceD\}}, \textcolor{ForestGreen}{\{choiceE\}},  \textcolor{ForestGreen}{\{choiceF\}}, \textcolor{ForestGreen}{\{choiceG\}},\textcolor{ForestGreen}{\{choiceH\}}, \textcolor{ForestGreen}{\{choiceI\}}, \textcolor{ForestGreen}{\{choiceJ\}}, \textcolor{ForestGreen}{\{choiceK\}}  \\ [0.5ex]
\textbf{Template}: \\ [0.5ex]
Given the following text, identify the correct language by selecting one of the options in the list (A, B, C, D, E, F, G, H, I, J, K): \\  \\ Text: \textcolor{MidnightBlue}{\{sentence\}} \\  \\ A: \textcolor{ForestGreen}{\{choiceA\}} \\ B: \textcolor{ForestGreen}{\{choiceB\}} \\ C: \textcolor{ForestGreen}{\{choiceC\}} \\ D: \textcolor{ForestGreen}{\{choiceD\}} \\ E: \textcolor{ForestGreen}{\{choiceE\}} \\ F: \textcolor{ForestGreen}{\{choiceF\}} \\ G: \textcolor{ForestGreen}{\{choiceG\}} \\ H: \textcolor{ForestGreen}{\{choiceH\}} \\ I: \textcolor{ForestGreen}{\{choiceI\}} \\ J: \textcolor{ForestGreen}{\{choiceJ\}} \\ K: \textcolor{ForestGreen}{\{choiceK\}} \\  \\ Answer:

\paragraph{Collected - 07} Source: Annotator \\ [1ex]
\textbf{Input}: \textcolor{MidnightBlue}{\{sentence\}},  \textcolor{ForestGreen}{\{choiceA\}}, \textcolor{ForestGreen}{\{choiceB\}}, \textcolor{ForestGreen}{\{choiceC\}}, \textcolor{ForestGreen}{\{choiceD\}}, \textcolor{ForestGreen}{\{choiceE\}},  \textcolor{ForestGreen}{\{choiceF\}}, \textcolor{ForestGreen}{\{choiceG\}},\textcolor{ForestGreen}{\{choiceH\}}, \textcolor{ForestGreen}{\{choiceI\}}, \textcolor{ForestGreen}{\{choiceJ\}}, \textcolor{ForestGreen}{\{choiceK\}}  \\ [0.5ex]
\textbf{Template}: \\ [0.5ex]
Please read the following sentence, then choose from the options which language you think it most likely came from. Your answer should be "A", "B", "C", "D", "E", "F", "G", "H", "I", "J", or "K" \\ Sentence: \textcolor{MidnightBlue}{\{sentence\}} \\ Options: \\ A: \textcolor{ForestGreen}{\{choiceA\}} \\ B: \textcolor{ForestGreen}{\{choiceB\}} \\ C: \textcolor{ForestGreen}{\{choiceC\}} \\ D: \textcolor{ForestGreen}{\{choiceD\}} \\ E: \textcolor{ForestGreen}{\{choiceE\}} \\ F: \textcolor{ForestGreen}{\{choiceF\}} \\ G: \textcolor{ForestGreen}{\{choiceG\}} \\ H: \textcolor{ForestGreen}{\{choiceH\}} \\ I: \textcolor{ForestGreen}{\{choiceI\}} \\ J: \textcolor{ForestGreen}{\{choiceJ\}} \\ K: \textcolor{ForestGreen}{\{choiceK\}} \\ Answer:

\paragraph{Collected - 08} Source: Annotator \\ [1ex]
\textbf{Input}: \textcolor{MidnightBlue}{\{sentence\}},  \textcolor{ForestGreen}{\{choiceA\}}, \textcolor{ForestGreen}{\{choiceB\}}, \textcolor{ForestGreen}{\{choiceC\}}, \textcolor{ForestGreen}{\{choiceD\}}, \textcolor{ForestGreen}{\{choiceE\}},  \textcolor{ForestGreen}{\{choiceF\}}, \textcolor{ForestGreen}{\{choiceG\}},\textcolor{ForestGreen}{\{choiceH\}}, \textcolor{ForestGreen}{\{choiceI\}}, \textcolor{ForestGreen}{\{choiceJ\}}, \textcolor{ForestGreen}{\{choiceK\}}  \\ [0.5ex]
\textbf{Template}: \\ [0.5ex]
Please give the language used in the following sentence. Each sentence will give five options, please output the corresponding option (i.e. A, B, C, D, E, F, G, H, I, J, or K) to represent the corresponding answer. \\  \\ Sentence: \textcolor{MidnightBlue}{\{sentence\}} \\ Options:: \\ A: \textcolor{ForestGreen}{\{choiceA\}} \\ B: \textcolor{ForestGreen}{\{choiceB\}} \\ C: \textcolor{ForestGreen}{\{choiceC\}} \\ D: \textcolor{ForestGreen}{\{choiceD\}} \\ E: \textcolor{ForestGreen}{\{choiceE\}} \\ F: \textcolor{ForestGreen}{\{choiceF\}} \\ G: \textcolor{ForestGreen}{\{choiceG\}} \\ H: \textcolor{ForestGreen}{\{choiceH\}} \\ I: \textcolor{ForestGreen}{\{choiceI\}} \\ J: \textcolor{ForestGreen}{\{choiceJ\}} \\ K: \textcolor{ForestGreen}{\{choiceK\}} \\ Answer:

\paragraph{Collected - 09} Source: Annotator \\ [1ex]
\textbf{Input}: \textcolor{MidnightBlue}{\{sentence\}},  \textcolor{ForestGreen}{\{choiceA\}}, \textcolor{ForestGreen}{\{choiceB\}}, \textcolor{ForestGreen}{\{choiceC\}}, \textcolor{ForestGreen}{\{choiceD\}}, \textcolor{ForestGreen}{\{choiceE\}},  \textcolor{ForestGreen}{\{choiceF\}}, \textcolor{ForestGreen}{\{choiceG\}},\textcolor{ForestGreen}{\{choiceH\}}, \textcolor{ForestGreen}{\{choiceI\}}, \textcolor{ForestGreen}{\{choiceJ\}}, \textcolor{ForestGreen}{\{choiceK\}}  \\ [0.5ex]
\textbf{Template}: \\ [0.5ex]
Given the sentence: \textcolor{MidnightBlue}{\{sentence\}}, select the correct language among the choices A. \textcolor{ForestGreen}{\{choiceA\}} B. \textcolor{ForestGreen}{\{choiceB\}} C. \textcolor{ForestGreen}{\{choiceC\}} D. \textcolor{ForestGreen}{\{choiceD\}} E. \textcolor{ForestGreen}{\{choiceE\}} F. \textcolor{ForestGreen}{\{choiceF\}} G. \textcolor{ForestGreen}{\{choiceG\}} H. \textcolor{ForestGreen}{\{choiceH\}} I. \textcolor{ForestGreen}{\{choiceI\}} J. \textcolor{ForestGreen}{\{choiceJ\}} K. \textcolor{ForestGreen}{\{choiceK\}} \\ - A: \textcolor{ForestGreen}{\{choiceA\}} \\ - B: \textcolor{ForestGreen}{\{choiceB\}} \\ - C: \textcolor{ForestGreen}{\{choiceC\}} \\ - D: \textcolor{ForestGreen}{\{choiceD\}} \\ - E: \textcolor{ForestGreen}{\{choiceE\}} \\ - F: \textcolor{ForestGreen}{\{choiceF\}} \\ - G: \textcolor{ForestGreen}{\{choiceG\}} \\ - H: \textcolor{ForestGreen}{\{choiceH\}} \\ - I: \textcolor{ForestGreen}{\{choiceI\}} \\ - J: \textcolor{ForestGreen}{\{choiceJ\}} \\ - K: \textcolor{ForestGreen}{\{choiceK\}} \\ Language:

\paragraph{Collected - 10} Source: Annotator \\ [1ex]
\textbf{Input}: \textcolor{MidnightBlue}{\{sentence\}},  \textcolor{ForestGreen}{\{choiceA\}}, \textcolor{ForestGreen}{\{choiceB\}}, \textcolor{ForestGreen}{\{choiceC\}}, \textcolor{ForestGreen}{\{choiceD\}}, \textcolor{ForestGreen}{\{choiceE\}},  \textcolor{ForestGreen}{\{choiceF\}}, \textcolor{ForestGreen}{\{choiceG\}},\textcolor{ForestGreen}{\{choiceH\}}, \textcolor{ForestGreen}{\{choiceI\}}, \textcolor{ForestGreen}{\{choiceJ\}}, \textcolor{ForestGreen}{\{choiceK\}}  \\ [0.5ex]
\textbf{Template}: \\ [0.5ex]
\textcolor{MidnightBlue}{\{sentence\}} \\  \\ This is a sentence written in one of \textcolor{ForestGreen}{\{choiceA\}}, \textcolor{ForestGreen}{\{choiceB\}}, \textcolor{ForestGreen}{\{choiceC\}}, \textcolor{ForestGreen}{\{choiceD\}}, \textcolor{ForestGreen}{\{choiceE\}}, \textcolor{ForestGreen}{\{choiceF\}}, \textcolor{ForestGreen}{\{choiceG\}}, \textcolor{ForestGreen}{\{choiceH\}}, \textcolor{ForestGreen}{\{choiceI\}}, \textcolor{ForestGreen}{\{choiceJ\}}, \textcolor{ForestGreen}{\{choiceK\}}. According to the words and the linguistic structure, I can tell that the language is:

\paragraph{Task Designer} See \textsc{Big-Bench} eval file.

\paragraph{Negation - 1} Source: NIV2 Task 137 Newscomm Classification - Template 2 \\ [1ex] 
\textbf{Input}: \textcolor{MidnightBlue}{\{passage\}} \\ [0.5ex]
\textbf{Template}: \\ [0.5ex]
You will be given a definition of a task first, then some input of the task. \\ Classify the given news commentary into the language in which it is \textcolor{red}{not} written in. There are \textcolor{ForestGreen}{\{option length\}} languages to classify the sentences into \textcolor{ForestGreen}{\{options\}} \\  \\ \textcolor{MidnightBlue}{\{sentence\}} \\ Output:

\paragraph{Negation - 2} Source: NIV2 Task 137 Newscomm Classification - Template 4 \\ [1ex] 
\textbf{Input}: \textcolor{MidnightBlue}{\{passage\}} \\ [0.5ex]
\textbf{Template}: \\ [0.5ex]
Instructions: Classify the given news commentary into the language in which it is \textcolor{red}{not} written in. There are \textcolor{ForestGreen}{\{option length\}} languages to classify the sentences into \textcolor{ForestGreen}{\{options\}} \\ Input: \textcolor{MidnightBlue}{\{sentence\}} \\ Output:

\paragraph{Negation - 3} Source: NIV2 Task 137 Newscomm Classification - Template 6 \\ [1ex] 
\textbf{Input}: \textcolor{MidnightBlue}{\{passage\}} \\ [0.5ex]
\textbf{Template}: \\ [0.5ex]
Given the task definition and input, reply with output. Classify the given news commentary into the language in which it is \textcolor{red}{not} written in. There are \textcolor{ForestGreen}{\{option length\}} languages to classify the sentences into \textcolor{ForestGreen}{\{options\}} \\  \\ \textcolor{MidnightBlue}{\{sentence\}} \\ 

\paragraph{Negation - 4} Source: NIV2 Task 137 Newscomm Classification - Template 8 \\ [1ex] 
\textbf{Input}: \textcolor{MidnightBlue}{\{passage\}} \\ [0.5ex]
\textbf{Template}: \\ [0.5ex]
Q: Classify the given news commentary into the language in which it is \textcolor{red}{not} written in. There are \textcolor{ForestGreen}{\{option length\}} languages to classify the sentences into \textcolor{ForestGreen}{\{options\}} \\ \textcolor{MidnightBlue}{\{sentence\}} \\ A:

\paragraph{Negation - 5} Source: NIV2 Task 137 Newscomm Classification - Template 10 \\ [1ex] 
\textbf{Input}: \textcolor{MidnightBlue}{\{passage\}} \\ [0.5ex]
\textbf{Template}: \\ [0.5ex]
Detailed Instructions: Classify the given news commentary into the language in which it is \textcolor{red}{not} written in. There are \textcolor{ForestGreen}{\{option length\}} languages to classify the sentences into \textcolor{ForestGreen}{\{options\}} \\ Q: \textcolor{MidnightBlue}{\{sentence\}} \\ A:

\paragraph{Nonsensical - 1} Source: Annotator \\ [1ex]
\textbf{Input}: \textcolor{MidnightBlue}{\{text\}} \\ [0.5ex]
\textbf{Template}: \\ [0.5ex]
\textcolor{red}{The council of street raccoons demands you respond to their inquisition.} \textcolor{MidnightBlue}{\{text\}}

\paragraph{Nonsensical - 2} Source: Annotator \\ [1ex]
\textbf{Input}: \textcolor{MidnightBlue}{\{text\}} \\ [0.5ex]
\textbf{Template}: \\ [0.5ex]
\textcolor{red}{Surveillance birds query your knowledge of seed. } \textcolor{MidnightBlue}{\{text\}}

\paragraph{Nonsensical - 3} Source: Annotator \\ [1ex]
\textbf{Input}: \textcolor{MidnightBlue}{\{text\}} \\ [0.5ex]
\textbf{Template}: \\ [0.5ex]
\textcolor{red}{Darth Vader requires you to answer to the dark side } \textcolor{MidnightBlue}{\{text\}}

\paragraph{Nonsensical - 4} Source: Annotator \\ [1ex]
\textbf{Input}: \textcolor{MidnightBlue}{\{text\}} \\ [0.5ex]
\textbf{Template}: \\ [0.5ex]
\textcolor{red}{Respond to the requirement of the Mars working dolphin union.} \textcolor{MidnightBlue}{\{text\}}

\paragraph{Nonsensical - 5} Source: Annotator \\ [1ex]
\textbf{Input}: \textcolor{MidnightBlue}{\{text\}} \\ [0.5ex]
\textbf{Template}: \\ [0.5ex]
\textcolor{red}{You are undergoing the inquiry of court of the local squirrels.} \textcolor{MidnightBlue}{\{text\}}

\subsubsection*{BBH - Epistemic Reasoning}

\paragraph{Closest - 1} Source: FLAN2021 RTE - Template 1 \\ [1ex] 
\textbf{Input}: \textcolor{MidnightBlue}{\{premise\}}, \textcolor{MidnightBlue}{\{hypothesis\}}, \textcolor{ForestGreen}{\{options\}} \\ [0.5ex]
\textbf{Template}: \\ [0.5ex]
\textcolor{MidnightBlue}{\{premise\}} \\  \\ Question with options: Based on the paragraph above can we conclude that "\textcolor{MidnightBlue}{\{hypothesis\}}"? \\  \\ OPTIONS: \textcolor{ForestGreen}{\{options\}}

\paragraph{Closest - 2} Source: FLAN2021 RTE - Template 2 \\ [1ex] 
\textbf{Input}: \textcolor{MidnightBlue}{\{premise\}}, \textcolor{MidnightBlue}{\{hypothesis\}}, \textcolor{ForestGreen}{\{options\}} \\ [0.5ex]
\textbf{Template}: \\ [0.5ex]
\textcolor{MidnightBlue}{\{premise\}} \\  \\ Based on that paragraph can we conclude that the sentence below is true?  \\ \textcolor{MidnightBlue}{\{hypothesis\}} \\  \\ OPTIONS: \textcolor{ForestGreen}{\{options\}}

\paragraph{Closest - 3} Source: FLAN2021 RTE - Template 3 \\ [1ex] 
\textbf{Input}: \textcolor{MidnightBlue}{\{premise\}}, \textcolor{MidnightBlue}{\{hypothesis\}}, \textcolor{ForestGreen}{\{options\}} \\ [0.5ex]
\textbf{Template}: \\ [0.5ex]
\textcolor{MidnightBlue}{\{premise\}} \\  \\ Q with options: Can we draw the following conclusion? \\ \textcolor{MidnightBlue}{\{hypothesis\}} \\  \\ OPTIONS: \textcolor{ForestGreen}{\{options\}}

\paragraph{Closest - 4} Source: FLAN2021 RTE - Template 4 \\ [1ex] 
\textbf{Input}: \textcolor{MidnightBlue}{\{premise\}}, \textcolor{MidnightBlue}{\{hypothesis\}}, \textcolor{ForestGreen}{\{options\}} \\ [0.5ex]
\textbf{Template}: \\ [0.5ex]
\textcolor{MidnightBlue}{\{premise\}} \\ Does this next sentence follow, given the preceding text? \\ \textcolor{MidnightBlue}{\{hypothesis\}} \\  \\ OPTIONS: \textcolor{ForestGreen}{\{options\}}

\paragraph{Closest - 5} Source: FLAN2021 RTE - Template 5 \\ [1ex] 
\textbf{Input}: \textcolor{MidnightBlue}{\{premise\}}, \textcolor{MidnightBlue}{\{hypothesis\}}, \textcolor{ForestGreen}{\{options\}} \\ [0.5ex]
\textbf{Template}: \\ [0.5ex]
\textcolor{MidnightBlue}{\{premise\}} \\ OPTIONS: \textcolor{ForestGreen}{\{options\}} \\ Question: Can we infer the following? \\ \textcolor{MidnightBlue}{\{hypothesis\}}

\paragraph{Closest - 6} Source: FLAN2021 RTE - Template 6 \\ [1ex] 
\textbf{Input}: \textcolor{MidnightBlue}{\{premise\}}, \textcolor{MidnightBlue}{\{hypothesis\}}, \textcolor{ForestGreen}{\{options\}} \\ [0.5ex]
\textbf{Template}: \\ [0.5ex]
Read the following paragraph and determine if the hypothesis is true. Select from options at the end: \\  \\ \textcolor{MidnightBlue}{\{premise\}} \\  \\ Hypothesis: \textcolor{MidnightBlue}{\{hypothesis\}} \\ OPTIONS: \textcolor{ForestGreen}{\{options\}} \\ The answer is

\paragraph{Closest - 7} Source: FLAN2021 RTE - Template 7 \\ [1ex] 
\textbf{Input}: \textcolor{MidnightBlue}{\{premise\}}, \textcolor{MidnightBlue}{\{hypothesis\}}, \textcolor{ForestGreen}{\{options\}} \\ [0.5ex]
\textbf{Template}: \\ [0.5ex]
Read the text and determine if the sentence is true: \\  \\ \textcolor{MidnightBlue}{\{premise\}} \\  \\ Sentence: \textcolor{MidnightBlue}{\{hypothesis\}} \\ OPTIONS: \textcolor{ForestGreen}{\{options\}} \\ A:

\paragraph{Closest - 8} Source: FLAN2021 RTE - Template 8 \\ [1ex] 
\textbf{Input}: \textcolor{MidnightBlue}{\{premise\}}, \textcolor{MidnightBlue}{\{hypothesis\}}, \textcolor{ForestGreen}{\{options\}} \\ [0.5ex]
\textbf{Template}: \\ [0.5ex]
Question with options: can we draw the following hypothesis from the context?  \\  \\ Context: \\  \\ \textcolor{MidnightBlue}{\{premise\}} \\  \\ Hypothesis: \textcolor{MidnightBlue}{\{hypothesis\}} \\ OPTIONS: \textcolor{ForestGreen}{\{options\}} \\ A:

\paragraph{Incorrect - 1} Source: NIV2 Task 143 Odd Man Out Classification - Template 10\\ [1ex]
\textbf{Input}: \textcolor{MidnightBlue}{\{input\}}, \textcolor{ForestGreen}{\{categories\}} \\ [0.5ex]
\textbf{Template}: \\ [0.5ex]
Detailed Instructions: Given a set of four words, generate the category that the words belong to. Words are separated by commas. The possible categories are \textcolor{ForestGreen}{\{categories\}} \\ Q: \textcolor{MidnightBlue}{\{input\}} \\ A:

\paragraph{Incorrect - 2} Source: NIV2 Task 137 Newscomm Classification - Template 10\\ [1ex]
\textbf{Input}: \textcolor{MidnightBlue}{\{input\}}, \textcolor{ForestGreen}{\{options\}} \\ [0.5ex]
\textbf{Template}: \\ [0.5ex]
Detailed Instructions: Classify the given news commentary into the language in which it is written in. There are \textcolor{ForestGreen}{\{options length\}} languages to classify the sentences into \textcolor{ForestGreen}{\{options\}} \\ Q: \textcolor{MidnightBlue}{\{input\}} \\ A:

\paragraph{Incorrect - 3} Source: Flan2021 - Sentiment140 - Template 1\\ [1ex]
\textbf{Input}: \textcolor{MidnightBlue}{\{input\}}, \textcolor{ForestGreen}{\{options\}} \\ [0.5ex]
\textbf{Template}: \\ [0.5ex]
\textcolor{MidnightBlue}{\{text\}} \\ Select your answer from the options. What is the sentiment of this tweet? \\ Options: \textcolor{ForestGreen}{\{options\}}...I think the answer is

\paragraph{Incorrect - 4} Source: Flan2021 - Sentiment140 - Template 6\\ [1ex]
\textbf{Input}: \textcolor{MidnightBlue}{\{input\}}, \textcolor{ForestGreen}{\{options\}} \\ [0.5ex]
\textbf{Template}: \\ [0.5ex]
Select your answer from the options. How would one describe the sentiment of this tweet? \\ \textcolor{MidnightBlue}{\{text\}} \\ \textcolor{ForestGreen}{\{options\}}

\paragraph{Incorrect - 5} Source: NIV2 Task 1422 MathQA Physics - Template 10\\ [1ex]
\textbf{Input}: \textcolor{MidnightBlue}{\{input\}}, \textcolor{ForestGreen}{\{options\}} \\ [0.5ex]
\textbf{Template}: \\ [0.5ex]
Detailed Instructions: In this task, you need to answer the given multiple-choice question on the physics. Classify your answers into \textcolor{ForestGreen}{\{letter length\}} \\ Q: Problem: \textcolor{MidnightBlue}{\{input\}}  \\ \textcolor{ForestGreen}{\{options\}} \\ A:

\paragraph{Incorrect - 6} Source: NIV2 Task 562 Language Identification - Template 10\\ [1ex]
\textbf{Input}: \textcolor{MidnightBlue}{\{text\}}, \textcolor{ForestGreen}{\{options\}} \\ [0.5ex]
\textbf{Template}: \\ [0.5ex]
Detailed Instructions: In this task, an input sentence is given which can be in the \textcolor{ForestGreen}{\{options\}} languages. There are a total of \textcolor{ForestGreen}{\{options length\}} languages. Your task is to identify the language of the input sentence. The input sentence can only be in any of the \textcolor{ForestGreen}{\{options length\}} languages provided. \\ Q: \textcolor{MidnightBlue}{\{text\}} \\ A:

\paragraph{Incorrect - 7} Source: NIV2 Task 1193 Course Classification - Template 10\\ [1ex]
\textbf{Input}: \textcolor{MidnightBlue}{\{text\}}, \textcolor{ForestGreen}{\{options\}} \\ [0.5ex]
\textbf{Template}: \\ [0.5ex]
Detailed Instructions: In this task, you are given the name of an Indian food dish. You need to classify the dish as a \textcolor{ForestGreen}{\{options\}} \\ Q: \textcolor{MidnightBlue}{\{input\}} \\ A:

\paragraph{Incorrect - 8} Source: NIV2 - Task 56 - Template 10 \\ [1ex]
\textbf{Input}: \textcolor{MidnightBlue}{\{paragraph\}}, \textcolor{MidnightBlue}{\{question\}}, \textcolor{ForestGreen}{\{correct answer\}} \\ [0.5ex]
\textbf{Template}: \\ [0.5ex]
Detailed Instructions: In this task, your goal is to judge a correct answer to a given question based on an associated paragraph and decide if it is a good correct answer or not. A good correct answer is one that correctly and completely answers the question. A bad correct answer addresses the question only partially or incorrectly. If you think the given correct answer is good, indicate it by responding "Yes". Otherwise, respond "No". There are only two types of responses possible: "Yes" and "No". \\ Q: Paragraph- \textcolor{MidnightBlue}{\{paragraph\}} Question: \textcolor{MidnightBlue}{\{question\}} Correct Answer: \textcolor{ForestGreen}{\{correct answer\}} \\ A:

\paragraph{Collected - 01} Source: Annotator \\ [1ex]
\textbf{Input}: \textcolor{MidnightBlue}{\{premise\}}, \textcolor{MidnightBlue}{\{hypothesis\}} \\ [0.5ex]
\textbf{Template}: \\ [0.5ex]
Classify whether two sentences have entailment relation. Output "yes" if they have entailment relation; output "no" if they do not have entailment relation. \\ Premise: \textcolor{MidnightBlue}{\{premise\}} \\ Hypothesis: \textcolor{MidnightBlue}{\{hypothesis\}} \\ Answer: 

\paragraph{Collected - 02} Source: Annotator \\ [1ex]
\textbf{Input}: \textcolor{MidnightBlue}{\{premise\}}, \textcolor{MidnightBlue}{\{hypothesis\}} \\ [0.5ex]
\textbf{Template}: \\ [0.5ex]
What is the relation between the given two sentences? Choose one of 'entailment' and 'non-entailment'. \\  \\ Sentence1: \textcolor{MidnightBlue}{\{premise\}} \\ Sentence2: \textcolor{MidnightBlue}{\{hypothesis\}} \\ Relation:

\paragraph{Collected - 03} Source: Annotator \\ [1ex]
\textbf{Input}: \textcolor{MidnightBlue}{\{premise\}}, \textcolor{MidnightBlue}{\{hypothesis\}} \\ [0.5ex]
\textbf{Template}: \\ [0.5ex]
Premise: \textcolor{MidnightBlue}{\{premise\}} \\ Is the truthfulness of the premise entail the following hypothesis? \\ Hypothesis: \textcolor{MidnightBlue}{\{hypothesis\}}.

\paragraph{Collected - 04} Source: Annotator \\ [1ex]
\textbf{Input}: \textcolor{MidnightBlue}{\{premise\}}, \textcolor{MidnightBlue}{\{hypothesis\}} \\ [0.5ex]
\textbf{Template}: \\ [0.5ex]
Given the premise '\textcolor{MidnightBlue}{\{premise\}}', can you infer that '\textcolor{MidnightBlue}{\{hypothesis\}}'? Answer 'Yes' or 'No' as your answer. Answer: 

\paragraph{Collected - 05} Source: Annotator \\ [1ex]
\textbf{Input}: \textcolor{MidnightBlue}{\{premise\}}, \textcolor{MidnightBlue}{\{hypothesis\}} \\ [0.5ex]
\textbf{Template}: \\ [0.5ex]
I think "\textcolor{MidnightBlue}{\{premise\}}" entails "\textcolor{MidnightBlue}{\{hypothesis\}}". \\  \\ Am I right?

\paragraph{Task Designer} See \textsc{Big-Bench} eval file.

\paragraph{Negation - 1} Source: FLAN2021 RTE - Template 1 \\ [1ex] 
\textbf{Input}: \textcolor{MidnightBlue}{\{premise\}}, \textcolor{MidnightBlue}{\{hypothesis\}}, \textcolor{ForestGreen}{\{options\}} \\ [0.5ex]
\textbf{Template}: \\ [0.5ex]
\textcolor{MidnightBlue}{\{premise\}} \\  \\ Question with options: Based on the paragraph above can we conclude that "\textcolor{MidnightBlue}{\{hypothesis\}}"? \\ \textcolor{red}{Answer the opposite of the correct answer.} \\ OPTIONS: \textcolor{ForestGreen}{\{options\}}

\paragraph{Negation - 2} Source: FLAN2021 RTE - Template 2 \\ [1ex] 
\textbf{Input}: \textcolor{MidnightBlue}{\{premise\}}, \textcolor{MidnightBlue}{\{hypothesis\}}, \textcolor{ForestGreen}{\{options\}} \\ [0.5ex]
\textbf{Template}: \\ [0.5ex]
\textcolor{MidnightBlue}{\{premise\}} \\  \\ Based on that paragraph can we conclude that the sentence below is true? \textcolor{red}{Answer the opposite of the correct answer.} \\ \textcolor{MidnightBlue}{\{hypothesis\}} \\  \\ OPTIONS: \textcolor{ForestGreen}{\{options\}}

\paragraph{Negation - 3} Source: FLAN2021 RTE - Template 3 \\ [1ex] 
\textbf{Input}: \textcolor{MidnightBlue}{\{premise\}}, \textcolor{MidnightBlue}{\{hypothesis\}}, \textcolor{ForestGreen}{\{options\}} \\ [0.5ex]
\textbf{Template}: \\ [0.5ex]
\textcolor{MidnightBlue}{\{premise\}} \\  \\ Q with options: Can we draw the following conclusion? \textcolor{red}{Answer the opposite of the correct answer.} \\  \textcolor{MidnightBlue}{\{hypothesis\}} \\  \\ OPTIONS: \textcolor{ForestGreen}{\{options\}}

\paragraph{Negation - 4} Source: FLAN2021 RTE - Template 4 \\ [1ex] 
\textbf{Input}: \textcolor{MidnightBlue}{\{premise\}}, \textcolor{MidnightBlue}{\{hypothesis\}}, \textcolor{ForestGreen}{\{options\}} \\ [0.5ex]
\textbf{Template}: \\ [0.5ex]
\textcolor{MidnightBlue}{\{premise\}} \\ Does this next sentence follow, given the preceding text? \textcolor{red}{Answer the opposite of the correct answer.} \\ \textcolor{MidnightBlue}{\{hypothesis\}} \\  \\ OPTIONS: \textcolor{ForestGreen}{\{options\}}

\paragraph{Negation - 5} Source: FLAN2021 RTE - Template 5 \\ [1ex] 
\textbf{Input}: \textcolor{MidnightBlue}{\{premise\}}, \textcolor{MidnightBlue}{\{hypothesis\}}, \textcolor{ForestGreen}{\{options\}} \\ [0.5ex]
\textbf{Template}: \\ [0.5ex]
\textcolor{MidnightBlue}{\{premise\}} \\ OPTIONS: \textcolor{ForestGreen}{\{options\}} \\ Question: Can we infer the following? \textcolor{red}{Answer the opposite of the correct answer.} \\ \textcolor{MidnightBlue}{\{hypothesis\}}

\paragraph{Negation - 6} Source: FLAN2021 RTE - Template 6 \\ [1ex] 
\textbf{Input}: \textcolor{MidnightBlue}{\{premise\}}, \textcolor{MidnightBlue}{\{hypothesis\}}, \textcolor{ForestGreen}{\{options\}} \\ [0.5ex]
\textbf{Template}: \\ [0.5ex]
Read the following paragraph and determine if the hypothesis is true. Select from options at the end. \textcolor{red}{Answer the opposite of the correct answer} : \\  \\ \textcolor{MidnightBlue}{\{premise\}} \\  \\ Hypothesis: \textcolor{MidnightBlue}{\{hypothesis\}} \\ OPTIONS: \textcolor{ForestGreen}{\{options\}} \\ The answer is

\paragraph{Negation - 7} Source: FLAN2021 RTE - Template 7 \\ [1ex] 
\textbf{Input}: \textcolor{MidnightBlue}{\{premise\}}, \textcolor{MidnightBlue}{\{hypothesis\}}, \textcolor{ForestGreen}{\{options\}} \\ [0.5ex]
\textbf{Template}: \\ [0.5ex]
Read the text and determine if the sentence is true. \textcolor{red}{Answer the opposite of the correct answer}: \\  \\ \textcolor{MidnightBlue}{\{premise\}} \\  \\ Sentence: \textcolor{MidnightBlue}{\{hypothesis\}} \\ OPTIONS: \textcolor{ForestGreen}{\{options\}} \\ A:

\paragraph{Negation - 8} Source: FLAN2021 RTE - Template 8 \\ [1ex] 
\textbf{Input}: \textcolor{MidnightBlue}{\{premise\}}, \textcolor{MidnightBlue}{\{hypothesis\}}, \textcolor{ForestGreen}{\{options\}} \\ [0.5ex]
\textbf{Template}: \\ [0.5ex]
Question with options: can we draw the following hypothesis from the context? \textcolor{red}{Answer the opposite of the correct answer.} \\  \\ Context: \\  \\ \textcolor{MidnightBlue}{\{premise\}} \\  \\ Hypothesis: \textcolor{MidnightBlue}{\{hypothesis\}} \\ OPTIONS: \textcolor{ForestGreen}{\{options\}} \\ A:

\paragraph{Nonsensical - 1} Source: Annotator \\ [1ex]
\textbf{Input}: \textcolor{MidnightBlue}{\{text\}} \\ [0.5ex]
\textbf{Template}: \\ [0.5ex]
\textcolor{red}{The council of street raccoons demands you respond to their inquisition.} \textcolor{MidnightBlue}{\{text\}}

\paragraph{Nonsensical - 2} Source: Annotator \\ [1ex]
\textbf{Input}: \textcolor{MidnightBlue}{\{text\}} \\ [0.5ex]
\textbf{Template}: \\ [0.5ex]
\textcolor{red}{Surveillance birds query your knowledge of seed. } \textcolor{MidnightBlue}{\{text\}}

\paragraph{Nonsensical - 3} Source: Annotator \\ [1ex]
\textbf{Input}: \textcolor{MidnightBlue}{\{text\}} \\ [0.5ex]
\textbf{Template}: \\ [0.5ex]
\textcolor{red}{Darth Vader requires you to answer to the dark side } \textcolor{MidnightBlue}{\{text\}}

\paragraph{Nonsensical - 4} Source: Annotator \\ [1ex]
\textbf{Input}: \textcolor{MidnightBlue}{\{text\}} \\ [0.5ex]
\textbf{Template}: \\ [0.5ex]
\textcolor{red}{Respond to the requirement of the Mars working dolphin union.} \textcolor{MidnightBlue}{\{text\}}

\paragraph{Nonsensical - 5} Source: Annotator \\ [1ex]
\textbf{Input}: \textcolor{MidnightBlue}{\{text\}} \\ [0.5ex]
\textbf{Template}: \\ [0.5ex]
\textcolor{red}{You are undergoing the inquiry of court of the local squirrels.} \textcolor{MidnightBlue}{\{text\}}

\subsubsection*{BBH - Crash Blossom}

\paragraph{Closest - 1} Source: Flan2021 - CosmosQA - Template 1 \\ [1ex]
\textbf{Input}: \textcolor{MidnightBlue}{\{context\}}, \textcolor{MidnightBlue}{\{question\}}, \textcolor{ForestGreen}{\{options\}} \\ [0.5ex]
\textbf{Template}: \\ [0.5ex]
\textcolor{MidnightBlue}{\{context\}} \\  \\ Question with options to choose from: \textcolor{MidnightBlue}{\{question\}} \\ OPTIONS:\textcolor{ForestGreen}{\{options\}}

\paragraph{Closest - 2} Source: Flan2021 - CosmosQA - Template 2 \\ [1ex]
\textbf{Input}: \textcolor{MidnightBlue}{\{context\}}, \textcolor{MidnightBlue}{\{question\}}, \textcolor{ForestGreen}{\{options\}} \\ [0.5ex]
\textbf{Template}: \\ [0.5ex]
\textcolor{MidnightBlue}{\{context\}} \\  \\ OPTIONS: \textcolor{ForestGreen}{\{options\}} \\ Q: \textcolor{MidnightBlue}{\{question\}}

\paragraph{Closest - 3} Source: Flan2021 - CosmosQA - Template 3 \\ [1ex]
\textbf{Input}: \textcolor{MidnightBlue}{\{context\}}, \textcolor{MidnightBlue}{\{question\}}, \textcolor{ForestGreen}{\{options\}} \\ [0.5ex]
\textbf{Template}: \\ [0.5ex]
\textcolor{MidnightBlue}{\{context\}} \\  \\ OPTIONS: \textcolor{ForestGreen}{\{options\}} \\ Answer the following question: \textcolor{MidnightBlue}{\{question\}} \\ 

\paragraph{Closest - 4} Source: Flan2021 - CosmosQA - Template 4 \\ [1ex]
\textbf{Input}: \textcolor{MidnightBlue}{\{context\}}, \textcolor{MidnightBlue}{\{question\}}, \textcolor{ForestGreen}{\{options\}} \\ [0.5ex]
\textbf{Template}: \\ [0.5ex]
\textcolor{MidnightBlue}{\{context\}} \\  \\ Based on the preceding passage, choose your answer for question \textcolor{MidnightBlue}{\{question\}} \\ OPTIONS: \textcolor{ForestGreen}{\{options\}}

\paragraph{Closest - 5} Source: Flan2021 - CosmosQA - Template 5 \\ [1ex]
\textbf{Input}: \textcolor{MidnightBlue}{\{context\}}, \textcolor{MidnightBlue}{\{question\}}, \textcolor{ForestGreen}{\{options\}} \\ [0.5ex]
\textbf{Template}: \\ [0.5ex]
\textcolor{MidnightBlue}{\{context\}} \\  \\ Q with options: Give answer the following question using evidence from the above passage: \textcolor{MidnightBlue}{\{question\}} \\ OPTIONS:\textcolor{ForestGreen}{\{options\}}

\paragraph{Closest - 6} Source: Flan2021 - CosmosQA - Template 6 \\ [1ex]
\textbf{Input}: \textcolor{MidnightBlue}{\{context\}}, \textcolor{MidnightBlue}{\{question\}}, \textcolor{ForestGreen}{\{options\}} \\ [0.5ex]
\textbf{Template}: \\ [0.5ex]
Context: \textcolor{MidnightBlue}{\{context\}} \\ Question \textcolor{MidnightBlue}{\{question\}} \\ Possible answers: \\ \textcolor{ForestGreen}{\{options\}} \\ The answer:

\paragraph{Closest - 7} Source: Flan2021 - CosmosQA - Template 7 \\ [1ex]
\textbf{Input}: \textcolor{MidnightBlue}{\{context\}}, \textcolor{MidnightBlue}{\{question\}}, \textcolor{ForestGreen}{\{options\}} \\ [0.5ex]
\textbf{Template}: \\ [0.5ex]
Read the following article and answer the question by choosing from the options. \\  \\ \textcolor{MidnightBlue}{\{context\}} \\  \\ \textcolor{MidnightBlue}{\{question\}} \\OPTIONS: \textcolor{ForestGreen}{\{options\}}...A:

\paragraph{Closest - 8} Source: Flan2021 - CosmosQA - Template 8 \\ [1ex]
\textbf{Input}: \textcolor{MidnightBlue}{\{context\}}, \textcolor{MidnightBlue}{\{question\}}, \textcolor{ForestGreen}{\{options\}} \\ [0.5ex]
\textbf{Template}: \\ [0.5ex]
This question has options. Answer the question about text: \\  \\ \textcolor{MidnightBlue}{\{context\}} \\  \\ \textcolor{MidnightBlue}{\{question\}} \\ OPTIONS:\textcolor{ForestGreen}{\{options\}}

\paragraph{Incorrect - 1} Source: NIV2 Task 143 Odd Man Out Classification - Template 10\\ [1ex]
\textbf{Input}: \textcolor{MidnightBlue}{\{input\}}, \textcolor{ForestGreen}{\{categories\}} \\ [0.5ex]
\textbf{Template}: \\ [0.5ex]
Detailed Instructions: Given a set of four words, generate the category that the words belong to. Words are separated by commas. The possible categories are \textcolor{ForestGreen}{\{categories\}} \\ Q: \textcolor{MidnightBlue}{\{input\}} \\ A:

\paragraph{Incorrect - 2} Source: NIV2 Task 137 Newscomm Classification - Template 10\\ [1ex]
\textbf{Input}: \textcolor{MidnightBlue}{\{input\}}, \textcolor{ForestGreen}{\{options\}} \\ [0.5ex]
\textbf{Template}: \\ [0.5ex]
Detailed Instructions: Classify the given news commentary into the language in which it is written in. There are \textcolor{ForestGreen}{\{options length\}} languages to classify the sentences into \textcolor{ForestGreen}{\{options\}} \\ Q: \textcolor{MidnightBlue}{\{input\}} \\ A:

\paragraph{Incorrect - 3} Source: Flan2021 - Sentiment140 - Template 1\\ [1ex]
\textbf{Input}: \textcolor{MidnightBlue}{\{input\}}, \textcolor{ForestGreen}{\{options\}} \\ [0.5ex]
\textbf{Template}: \\ [0.5ex]
\textcolor{MidnightBlue}{\{text\}} \\ Select your answer from the options. What is the sentiment of this tweet? \\ Options: \textcolor{ForestGreen}{\{options\}}...I think the answer is

\paragraph{Incorrect - 4} Source: Flan2021 - Sentiment140 - Template 6\\ [1ex]
\textbf{Input}: \textcolor{MidnightBlue}{\{input\}}, \textcolor{ForestGreen}{\{options\}} \\ [0.5ex]
\textbf{Template}: \\ [0.5ex]
Select your answer from the options. How would one describe the sentiment of this tweet? \\ \textcolor{MidnightBlue}{\{text\}} \\ \textcolor{ForestGreen}{\{options\}}

\paragraph{Incorrect - 5} Source: NIV2 Task 1422 MathQA Physics - Template 10\\ [1ex]
\textbf{Input}: \textcolor{MidnightBlue}{\{input\}}, \textcolor{ForestGreen}{\{options\}} \\ [0.5ex]
\textbf{Template}: \\ [0.5ex]
Detailed Instructions: In this task, you need to answer the given multiple-choice question on the physics. Classify your answers into \textcolor{ForestGreen}{\{letter length\}} \\ Q: Problem: \textcolor{MidnightBlue}{\{input\}}  \\ \textcolor{ForestGreen}{\{options\}} \\ A:

\paragraph{Collected - 01} Source: Annotator \\ [1ex]
\textbf{Input}: \textcolor{MidnightBlue}{\{word\}}, \textcolor{MidnightBlue}{\{sentence\}}, \textcolor{MidnightBlue}{\{options\}} \\ [0.5ex]
\textbf{Template}: \\ [0.5ex]
Classify the part of speech of the word "\textcolor{MidnightBlue}{\{word\}}" in the following sentence: \textcolor{MidnightBlue}{\{sentence\}}. The options are: \textcolor{ForestGreen}{\{options\}} \\ Answer: 

\paragraph{Collected - 02} Source: Annotator \\ [1ex]
\textbf{Input}: \textcolor{MidnightBlue}{\{word\}}, \textcolor{MidnightBlue}{\{sentence\}}, \textcolor{MidnightBlue}{\{options\}} \\ [0.5ex]
\textbf{Template}: \\ [0.5ex]
Sentence: \textcolor{MidnightBlue}{\{sentence\}} \\ Identify the part of speech of \textcolor{MidnightBlue}{\{word\}} in the sentence. Choise your answer from \textcolor{ForestGreen}{\{options\}} and output the best choice.

\paragraph{Collected - 03} Source: Annotator \\ [1ex]
\textbf{Input}: \textcolor{MidnightBlue}{\{word\}}, \textcolor{MidnightBlue}{\{sentence\}}, \textcolor{MidnightBlue}{\{options\}} \\ [0.5ex]
\textbf{Template}: \\ [0.5ex]
What is the part of speech of the word '\textcolor{MidnightBlue}{\{word\}}' in '\textcolor{MidnightBlue}{\{sentence\}}'. You may only choose from the following options: \textcolor{ForestGreen}{\{options\}}. Your answer is: 

\paragraph{Collected - 04} Source: Annotator \\ [1ex]
\textbf{Input}: \textcolor{MidnightBlue}{\{word\}}, \textcolor{MidnightBlue}{\{sentence\}}, \textcolor{MidnightBlue}{\{options\}} \\ [0.5ex]
\textbf{Template}: \\ [0.5ex]
Given a sentence and a word contained in the sentence, output the part of speech of the word. \\ Word: \textcolor{MidnightBlue}{\{word\}} \\ Sentence: \textcolor{MidnightBlue}{\{sentence\}} \\ Options: \textcolor{ForestGreen}{\{options\}} \\  \\ Answer: 

\paragraph{Collected - 05} Source: Annotator \\ [1ex]
\textbf{Input}: \textcolor{MidnightBlue}{\{word\}}, \textcolor{MidnightBlue}{\{sentence\}}, \textcolor{MidnightBlue}{\{options\}} \\ [0.5ex]
\textbf{Template}: \\ [0.5ex]
Identify the part of speech of the word. Question: which one of \textcolor{ForestGreen}{\{options\}} is '\textcolor{MidnightBlue}{\{word\}}' in '\textcolor{MidnightBlue}{\{sentence\}}'? Answer: 

\paragraph{Task Designer} See \textsc{Big-Bench} eval file.

\paragraph{Negation - 1} Source: Flan2021 - CosmosQA - Template 1 \\ [1ex]
\textbf{Input}: \textcolor{MidnightBlue}{\{context\}}, \textcolor{MidnightBlue}{\{question\}}, \textcolor{ForestGreen}{\{options\}} \\ [0.5ex]
\textbf{Template}: \\ [0.5ex]
\textcolor{MidnightBlue}{\{context\}} \\  \\ Question with options to choose from: \textcolor{MidnightBlue}{\{question\}} \\ OPTIONS:\textcolor{ForestGreen}{\{options\}} \textcolor{red}{The answer is not:}

\paragraph{Negation - 2} Source: Flan2021 - CosmosQA - Template 2 \\ [1ex]
\textbf{Input}: \textcolor{MidnightBlue}{\{context\}}, \textcolor{MidnightBlue}{\{question\}}, \textcolor{ForestGreen}{\{options\}} \\ [0.5ex]
\textbf{Template}: \\ [0.5ex]
\textcolor{MidnightBlue}{\{context\}} \\  \\ OPTIONS: \textcolor{ForestGreen}{\{options\}} \\ Q: \textcolor{MidnightBlue}{\{question\}} \textcolor{red}{The answer is not:}

\paragraph{Negation - 3} Source: Flan2021 - CosmosQA - Template 3 \\ [1ex]
\textbf{Input}: \textcolor{MidnightBlue}{\{context\}}, \textcolor{MidnightBlue}{\{question\}}, \textcolor{ForestGreen}{\{options\}} \\ [0.5ex]
\textbf{Template}: \\ [0.5ex]
\textcolor{MidnightBlue}{\{context\}} \\  \\ OPTIONS: \textcolor{ForestGreen}{\{options\}} \\ Answer the following question: \textcolor{MidnightBlue}{\{question\}} \\ \textcolor{red}{The answer is not:}

\paragraph{Negation - 4} Source: Flan2021 - CosmosQA - Template 4 \\ [1ex]
\textbf{Input}: \textcolor{MidnightBlue}{\{context\}}, \textcolor{MidnightBlue}{\{question\}}, \textcolor{ForestGreen}{\{options\}} \\ [0.5ex]
\textbf{Template}: \\ [0.5ex]
\textcolor{MidnightBlue}{\{context\}} \\  \\ Based on the preceding passage, choose your answer for question \textcolor{MidnightBlue}{\{question\}} \\ OPTIONS: \textcolor{ForestGreen}{\{options\}} \textcolor{red}{The answer is not:}

\paragraph{Negation - 5} Source: Flan2021 - CosmosQA - Template 5 \\ [1ex]
\textbf{Input}: \textcolor{MidnightBlue}{\{context\}}, \textcolor{MidnightBlue}{\{question\}}, \textcolor{ForestGreen}{\{options\}} \\ [0.5ex]
\textbf{Template}: \\ [0.5ex]
\textcolor{MidnightBlue}{\{context\}} \\  \\ Q with options: Give answer the following question using evidence from the above passage: \textcolor{MidnightBlue}{\{question\}} \\ OPTIONS:\textcolor{ForestGreen}{\{options\}} \textcolor{red}{The answer is not:}

\paragraph{Negation - 6} Source: Flan2021 - CosmosQA - Template 6 \\ [1ex]
\textbf{Input}: \textcolor{MidnightBlue}{\{context\}}, \textcolor{MidnightBlue}{\{question\}}, \textcolor{ForestGreen}{\{options\}} \\ [0.5ex]
\textbf{Template}: \\ [0.5ex]
Context: \textcolor{MidnightBlue}{\{context\}} \\ Question \textcolor{MidnightBlue}{\{question\}} \\ Possible answers: \\ \textcolor{ForestGreen}{\{options\}} \\ \textcolor{red}{The answer is not:}

\paragraph{Negation - 7} Source: Flan2021 - CosmosQA - Template 7 \\ [1ex]
\textbf{Input}: \textcolor{MidnightBlue}{\{context\}}, \textcolor{MidnightBlue}{\{question\}}, \textcolor{ForestGreen}{\{options\}} \\ [0.5ex]
\textbf{Template}: \\ [0.5ex]
Read the following article and answer the question by choosing from the options. \\  \\ \textcolor{MidnightBlue}{\{context\}} \\  \\ \textcolor{MidnightBlue}{\{question\}} \\OPTIONS: \textcolor{ForestGreen}{\{options\}}...\textcolor{red}{The answer is not:}:

\paragraph{Negation - 8} Source: Flan2021 - CosmosQA - Template 8 \\ [1ex]
\textbf{Input}: \textcolor{MidnightBlue}{\{context\}}, \textcolor{MidnightBlue}{\{question\}}, \textcolor{ForestGreen}{\{options\}} \\ [0.5ex]
\textbf{Template}: \\ [0.5ex]
This question has options. Answer the question about text: \\  \\ \textcolor{MidnightBlue}{\{context\}} \\  \\ \textcolor{MidnightBlue}{\{question\}} \\ OPTIONS:\textcolor{ForestGreen}{\{options\}} \textcolor{red}{The answer is not:}

\paragraph{Nonsensical - 1} Source: Annotator \\ [1ex]
\textbf{Input}: \textcolor{MidnightBlue}{\{text\}} \\ [0.5ex]
\textbf{Template}: \\ [0.5ex]
\textcolor{red}{The council of street raccoons demands you respond to their inquisition.} \textcolor{MidnightBlue}{\{text\}}

\paragraph{Nonsensical - 2} Source: Annotator \\ [1ex]
\textbf{Input}: \textcolor{MidnightBlue}{\{text\}} \\ [0.5ex]
\textbf{Template}: \\ [0.5ex]
\textcolor{red}{Surveillance birds query your knowledge of seed. } \textcolor{MidnightBlue}{\{text\}}

\paragraph{Nonsensical - 3} Source: Annotator \\ [1ex]
\textbf{Input}: \textcolor{MidnightBlue}{\{text\}} \\ [0.5ex]
\textbf{Template}: \\ [0.5ex]
\textcolor{red}{Darth Vader requires you to answer to the dark side } \textcolor{MidnightBlue}{\{text\}}

\paragraph{Nonsensical - 4} Source: Annotator \\ [1ex]
\textbf{Input}: \textcolor{MidnightBlue}{\{text\}} \\ [0.5ex]
\textbf{Template}: \\ [0.5ex]
\textcolor{red}{Respond to the requirement of the Mars working dolphin union.} \textcolor{MidnightBlue}{\{text\}}

\paragraph{Nonsensical - 5} Source: Annotator \\ [1ex]
\textbf{Input}: \textcolor{MidnightBlue}{\{text\}} \\ [0.5ex]
\textbf{Template}: \\ [0.5ex]
\textcolor{red}{You are undergoing the inquiry of court of the local squirrels.} \textcolor{MidnightBlue}{\{text\}}

\subsubsection*{BBH - Logical Sequence}

\paragraph{Closest - 1} Source: NIV2 - Task 73 - Template 2 \\ [1ex]
\textbf{Input}: \textcolor{MidnightBlue}{\{question\}}, \textcolor{ForestGreen}{\{options\}} \\ [0.5ex]
\textbf{Template}: \\ [0.5ex]
You will be given a definition of a task first, then some input of the task. \\ You are given a question and some answer options (associated with "A", "B", "C", "D"). You should choose the correct answer based on commonsense knowledge. Avoid answering questions based on associations, the set of answers are chosen deliberately to capture common sense beyond associations. Do not generate anything else apart from one of the following characters: \textcolor{ForestGreen}{\{options letter\}} and only give one answer for each question. \\  \\ \textcolor{MidnightBlue}{\{question\}} \textcolor{ForestGreen}{\{options\}} \\ Output:

\paragraph{Closest - 2} Source: NIV2 - Task 73 - Template 4 \\ [1ex]
\textbf{Input}: \textcolor{MidnightBlue}{\{question\}}, \textcolor{ForestGreen}{\{options\}} \\ [0.5ex]
\textbf{Template}: \\ [0.5ex]
Instructions: You are given a question and some answer options (associated with "A", "B", "C", "D"). You should choose the correct answer based on commonsense knowledge. Avoid answering questions based on associations, the set of answers are chosen deliberately to capture common sense beyond associations. Do not generate anything else apart from one of the following characters: \textcolor{ForestGreen}{\{options letter\}} and only give one answer for each question. \\ Input: \textcolor{MidnightBlue}{\{question\}} \textcolor{ForestGreen}{\{options\}} \\ Output:

\paragraph{Closest - 3} Source: NIV2 - Task 73 - Template 6 \\ [1ex]
\textbf{Input}: \textcolor{MidnightBlue}{\{question\}}, \textcolor{ForestGreen}{\{options\}} \\ [0.5ex]
\textbf{Template}: \\ [0.5ex]
Given the task definition and input, reply with output. You are given a question and some answer options (associated with "A", "B", "C", "D"). You should choose the correct answer based on commonsense knowledge. Avoid answering questions based on associations, the set of answers are chosen deliberately to capture common sense beyond associations. Do not generate anything else apart from one of the following characters: \textcolor{ForestGreen}{\{options letter\}} and only give one answer for each question. \\  \\ \textcolor{MidnightBlue}{\{question\}} \textcolor{ForestGreen}{\{options\}} \\

\paragraph{Closest - 4} Source: NIV2 - Task 73 - Template 8 \\ [1ex]
\textbf{Input}: \textcolor{MidnightBlue}{\{question\}}, \textcolor{ForestGreen}{\{options\}} \\ [0.5ex]
\textbf{Template}: \\ [0.5ex]
Q: You are given a question and some answer options (associated with "A", "B", "C", "D"). You should choose the correct answer based on commonsense knowledge. Avoid answering questions based on associations, the set of answers are chosen deliberately to capture common sense beyond associations. Do not generate anything else apart from one of the following characters: \textcolor{ForestGreen}{\{options letter\}} and only give one answer for each question. \\ \textcolor{MidnightBlue}{\{question\}} \textcolor{ForestGreen}{\{options\}} \\ A:

\paragraph{Closest - 5} Source: NIV2 - Task 73 - Template 10 \\ [1ex]
\textbf{Input}: \textcolor{MidnightBlue}{\{question\}}, \textcolor{ForestGreen}{\{options\}} \\ [0.5ex]
\textbf{Template}: \\ [0.5ex]
Detailed Instructions: You are given a question and some answer options (associated with "A", "B", "C", "D"). You should choose the correct answer based on commonsense knowledge. Avoid answering questions based on associations, the set of answers are chosen deliberately to capture common sense beyond associations. Do not generate anything else apart from one of the following characters: \textcolor{ForestGreen}{\{options letter\}} and only give one answer for each question. \\ Q: \textcolor{MidnightBlue}{\{question\}} \textcolor{ForestGreen}{\{options\}} \\ A:

\paragraph{Incorrect - 1} Source: NIV2 Task 1421 MathQA General - Template 10\\ [1ex]
\textbf{Input}: \textcolor{MidnightBlue}{\{input\}}, \textcolor{ForestGreen}{\{options\}} \\ [0.5ex]
\textbf{Template}: \\ [0.5ex]
Detailed Instructions: In this task, you need to answer the given multiple-choice question on the general math. Classify your answers into Classify your answers into \textcolor{ForestGreen}{\{option letter\}} \\ Q: Problem: \textcolor{MidnightBlue}{\{input\}}  \\ \textcolor{ForestGreen}{\{options\}} \\ A:

\paragraph{Incorrect - 2} Source: NIV2 Task 1422 MathQA Physics - Template 10\\ [1ex]
\textbf{Input}: \textcolor{MidnightBlue}{\{input\}}, \textcolor{ForestGreen}{\{options\}} \\ [0.5ex]
\textbf{Template}: \\ [0.5ex]
Detailed Instructions: In this task, you need to answer the given multiple-choice question on the physics. Classify your answers into \textcolor{ForestGreen}{\{letter length\}} \\ Q: Problem: \textcolor{MidnightBlue}{\{input\}}  \\ \textcolor{ForestGreen}{\{options\}} \\ A:

\paragraph{Incorrect - 3} Source: Flan2021 - WSC273 - Template 1\\ [1ex]
\textbf{Input}: \textcolor{MidnightBlue}{\{context\}}, \textcolor{ForestGreen}{\{options\}} \\ [0.5ex]
\textbf{Template}: \\ [0.5ex]
Multi-choice problem: \textcolor{MidnightBlue}{\{context\}} \\ \textcolor{ForestGreen}{\{options\}}

\paragraph{Incorrect - 4} Source: Flan2021 - TREC - Template 1\\ [1ex]
\textbf{Input}: \textcolor{MidnightBlue}{\{text\}}, \textcolor{ForestGreen}{\{options\}} \\ [0.5ex]
\textbf{Template}: \\ [0.5ex]
What type of thing is the question "\textcolor{MidnightBlue}{\{text\}}" asking about? \\  \\ \textcolor{ForestGreen}{\{options\}} \\ Answer:

\paragraph{Incorrect - 5} Source: Flan2021 - PIQA - Template 1\\ [1ex]
\textbf{Input}: \textcolor{MidnightBlue}{\{input\}}, \textcolor{ForestGreen}{\{options\}} \\ [0.5ex]
\textbf{Template}: \\ [0.5ex]
Here is a goal: \textcolor{MidnightBlue}{\{goal\}} \\  \\ How would you accomplish this goal? \\  \\ \textcolor{ForestGreen}{\{options\}}

\paragraph{Collected - 1} Source: Annotator \\ [1ex]
\textbf{Input}: \textcolor{ForestGreen}{\{listA\}}, \textcolor{ForestGreen}{\{listB\}}, \textcolor{ForestGreen}{\{listC\}}, \textcolor{ForestGreen}{\{listD\}} \\ [0.5ex]
\textbf{Template}: \\ [0.5ex]
Four items are naturally in a sequential or chronological order. Now, choose the correct order of these items from the following options: \\ A. \textcolor{ForestGreen}{\{listA\}} \\ B. \textcolor{ForestGreen}{\{listB\}} \\ C. \textcolor{ForestGreen}{\{listC\}} \\ D. \textcolor{ForestGreen}{\{listD\}}

\paragraph{Collected - 2} Source: Annotator \\ [1ex]
\textbf{Input}: \textcolor{ForestGreen}{\{listA\}}, \textcolor{ForestGreen}{\{listB\}}, \textcolor{ForestGreen}{\{listC\}}, \textcolor{ForestGreen}{\{listD\}} \\ [0.5ex]
\textbf{Template}: \\ [0.5ex]
You are given four lists of the same objects in different orders. Which of the following lists is correctly ordered chronologically?  \\  \\ Lists: \\ A. \textcolor{ForestGreen}{\{listA\}} \\ B. \textcolor{ForestGreen}{\{listB\}} \\ C. \textcolor{ForestGreen}{\{listC\}} \\ D. \textcolor{ForestGreen}{\{listD\}}

\paragraph{Collected - 3} Source: Annotator \\ [1ex]
\textbf{Input}: \textcolor{ForestGreen}{\{listA\}}, \textcolor{ForestGreen}{\{listB\}}, \textcolor{ForestGreen}{\{listC\}}, \textcolor{ForestGreen}{\{listD\}} \\ [0.5ex]
\textbf{Template}: \\ [0.5ex]
Choose the best answer that describes a sequence chronologically. Options: A: \textcolor{ForestGreen}{\{listA\}}, B: \textcolor{ForestGreen}{\{listB\}}, C: \textcolor{ForestGreen}{\{listC\}}, D: \textcolor{ForestGreen}{\{listD\}} \\  \\ Answer:

\paragraph{Collected - 4} Source: Annotator \\ [1ex]
\textbf{Input}: \textcolor{ForestGreen}{\{listA\}}, \textcolor{ForestGreen}{\{listB\}}, \textcolor{ForestGreen}{\{listC\}}, \textcolor{ForestGreen}{\{listD\}} \\ [0.5ex]
\textbf{Template}: \\ [0.5ex]
In this task, pick the list of the items that are chronologically orded most correctly. Choose from the following options and output the corresponding letter as one of 'A', 'B', 'C', or 'D'. \\ A. \textcolor{ForestGreen}{\{listA\}} \\ B. \textcolor{ForestGreen}{\{listB\}} \\ C. \textcolor{ForestGreen}{\{listC\}} \\ D. \textcolor{ForestGreen}{\{listD\}}

\paragraph{Collected - 5} Source: Annotator \\ [1ex]
\textbf{Input}: \textcolor{ForestGreen}{\{listA\}}, \textcolor{ForestGreen}{\{listB\}}, \textcolor{ForestGreen}{\{listC\}}, \textcolor{ForestGreen}{\{listD\}} \\ [0.5ex]
\textbf{Template}: \\ [0.5ex]
Question: Which of the following lists is correctly ordered chronologically? \\ Choose the correct order from the lists: A. \textcolor{ForestGreen}{\{listA\}}, B. \textcolor{ForestGreen}{\{listB\}}, C. \textcolor{ForestGreen}{\{listC\}}, D. \textcolor{ForestGreen}{\{listD\}}. Answer: 

\paragraph{Task Designer} See \textsc{Big-Bench} eval file.

\paragraph{Negation - 1} Source: NIV2 - Task 73 - Template 2 \\ [1ex]
\textbf{Input}: \textcolor{MidnightBlue}{\{question\}}, \textcolor{ForestGreen}{\{options\}} \\ [0.5ex]
\textbf{Template}: \\ [0.5ex]
You will be given a definition of a task first, then some input of the task. \\ You are given a question and some answer options (associated with "A", "B", "C", "D"). You should choose the \textcolor{red}{incorrect} answer based on commonsense knowledge. Avoid answering questions based on associations, the set of answers are chosen deliberately to capture common sense beyond associations. Do not generate anything else apart from one of the following characters: \textcolor{ForestGreen}{\{options letter\}} and only give one answer for each question. \\  \\ \textcolor{MidnightBlue}{\{question\}} \textcolor{ForestGreen}{\{options\}} \\ Output:

\paragraph{Negation - 2} Source: NIV2 - Task 73 - Template 4 \\ [1ex]
\textbf{Input}: \textcolor{MidnightBlue}{\{question\}}, \textcolor{ForestGreen}{\{options\}} \\ [0.5ex]
\textbf{Template}: \\ [0.5ex]
Instructions: You are given a question and some answer options (associated with "A", "B", "C", "D"). You should choose the \textcolor{red}{incorrect} answer based on commonsense knowledge. Avoid answering questions based on associations, the set of answers are chosen deliberately to capture common sense beyond associations. Do not generate anything else apart from one of the following characters: \textcolor{ForestGreen}{\{options letter\}} and only give one answer for each question. \\ Input: \textcolor{MidnightBlue}{\{question\}} \textcolor{ForestGreen}{\{options\}} \\ Output:

\paragraph{Negation - 3} Source: NIV2 - Task 73 - Template 6 \\ [1ex]
\textbf{Input}: \textcolor{MidnightBlue}{\{question\}}, \textcolor{ForestGreen}{\{options\}} \\ [0.5ex]
\textbf{Template}: \\ [0.5ex]
Given the task definition and input, reply with output. You are given a question and some answer options (associated with "A", "B", "C", "D"). You should choose the \textcolor{red}{incorrect} answer based on commonsense knowledge. Avoid answering questions based on associations, the set of answers are chosen deliberately to capture common sense beyond associations. Do not generate anything else apart from one of the following characters: \textcolor{ForestGreen}{\{options letter\}} and only give one answer for each question. \\  \\ \textcolor{MidnightBlue}{\{question\}} \textcolor{ForestGreen}{\{options\}} \\

\paragraph{Negation - 4} Source: NIV2 - Task 73 - Template 8 \\ [1ex]
\textbf{Input}: \textcolor{MidnightBlue}{\{question\}}, \textcolor{ForestGreen}{\{options\}} \\ [0.5ex]
\textbf{Template}: \\ [0.5ex]
Q: You are given a question and some answer options (associated with "A", "B", "C", "D"). You should choose the \textcolor{red}{incorrect} answer based on commonsense knowledge. Avoid answering questions based on associations, the set of answers are chosen deliberately to capture common sense beyond associations. Do not generate anything else apart from one of the following characters: \textcolor{ForestGreen}{\{options letter\}} and only give one answer for each question. \\ \textcolor{MidnightBlue}{\{question\}} \textcolor{ForestGreen}{\{options\}} \\ A:

\paragraph{Negation - 5} Source: NIV2 - Task 73 - Template 10 \\ [1ex]
\textbf{Input}: \textcolor{MidnightBlue}{\{question\}}, \textcolor{ForestGreen}{\{options\}} \\ [0.5ex]
\textbf{Template}: \\ [0.5ex]
Detailed Instructions: You are given a question and some answer options (associated with "A", "B", "C", "D"). You should choose the \textcolor{red}{incorrect} answer based on commonsense knowledge. Avoid answering questions based on associations, the set of answers are chosen deliberately to capture common sense beyond associations. Do not generate anything else apart from one of the following characters: \textcolor{ForestGreen}{\{options letter\}} and only give one answer for each question. \\ Q: \textcolor{MidnightBlue}{\{question\}} \textcolor{ForestGreen}{\{options\}} \\ A:

\paragraph{Nonsensical - 1} Source: Annotator \\ [1ex]
\textbf{Input}: \textcolor{ForestGreen}{\{listA\}}, \textcolor{ForestGreen}{\{listB\}}, \textcolor{ForestGreen}{\{listC\}}, \textcolor{ForestGreen}{\{listD\}}  \\ [0.5ex]
\textbf{Template}: \\ [0.5ex]
\textcolor{red}{The council of street raccoons demands you respond to their inquisition.} \\
A. \textcolor{ForestGreen}{\{listA\}} \\
B. \textcolor{ForestGreen}{\{listB\}} \\
C. \textcolor{ForestGreen}{\{listC\}} \\
D. \textcolor{ForestGreen}{\{listD\}} \\

\paragraph{Nonsensical - 2} Source: Annotator \\ [1ex]
\textbf{Input}: \textcolor{ForestGreen}{\{listA\}}, \textcolor{ForestGreen}{\{listB\}}, \textcolor{ForestGreen}{\{listC\}}, \textcolor{ForestGreen}{\{listD\}}  \\ [0.5ex]
\textbf{Template}: \\ [0.5ex]
\textcolor{red}{Surveillance birds query your knowledge of seed. } \\
A. \textcolor{ForestGreen}{\{listA\}} \\
B. \textcolor{ForestGreen}{\{listB\}} \\
C. \textcolor{ForestGreen}{\{listC\}} \\
D. \textcolor{ForestGreen}{\{listD\}} \\

\paragraph{Nonsensical - 3} Source: Annotator \\ [1ex]
\textbf{Input}: \textcolor{ForestGreen}{\{listA\}}, \textcolor{ForestGreen}{\{listB\}}, \textcolor{ForestGreen}{\{listC\}}, \textcolor{ForestGreen}{\{listD\}}   \\ [0.5ex]
\textbf{Template}: \\ [0.5ex]
\textcolor{red}{Darth Vader requires you to answer to the dark side } \\
A. \textcolor{ForestGreen}{\{listA\}} \\
B. \textcolor{ForestGreen}{\{listB\}} \\
C. \textcolor{ForestGreen}{\{listC\}} \\
D. \textcolor{ForestGreen}{\{listD\}} \\

\paragraph{Nonsensical - 4} Source: Annotator \\ [1ex]
\textbf{Input}: \textcolor{ForestGreen}{\{listA\}}, \textcolor{ForestGreen}{\{listB\}}, \textcolor{ForestGreen}{\{listC\}}, \textcolor{ForestGreen}{\{listD\}}  \\ [0.5ex]
\textbf{Template}: \\ [0.5ex]
\textcolor{red}{Respond to the requirement of the Mars working dolphin union.} \\
A. \textcolor{ForestGreen}{\{listA\}} \\
B. \textcolor{ForestGreen}{\{listB\}} \\
C. \textcolor{ForestGreen}{\{listC\}} \\
D. \textcolor{ForestGreen}{\{listD\}} \\

\paragraph{Nonsensical - 5} Source: Annotator \\ [1ex]
\textbf{Input}: \textcolor{ForestGreen}{\{listA\}}, \textcolor{ForestGreen}{\{listB\}}, \textcolor{ForestGreen}{\{listC\}}, \textcolor{ForestGreen}{\{listD\}}  \\ [0.5ex]
\textbf{Template}: \\ [0.5ex]
\textcolor{red}{You are undergoing the inquiry of court of the local squirrels.} \\
A. \textcolor{ForestGreen}{\{listA\}} \\
B. \textcolor{ForestGreen}{\{listB\}} \\
C. \textcolor{ForestGreen}{\{listC\}} \\
D. \textcolor{ForestGreen}{\{listD\}} \\

\subsection{Paraphrased Instructions}

Here we provide the prompt we use to automatically generate paraphrased instructions for \ref{section:methods}. 
%Following the same procedure, the exact same training data should be obtained. 
We also %will also 
provide the JSON file of all paraphrased instructions. 

\paragraph{Alpaca} To generate %the paraphrase of trained 
paraphrases of observed instructions in the Alpaca collection we sampled 1000 out of 52002 Alpaca tasks at i.i.d. random %(with a seed of 42) 
and generated paraphrases of instructions with GPT-4 using the following prompts. %to GPT-4.
\begin{itemize}
    \item ``Paraphrase this sentence:\textbackslash n\textbackslash n$\{$instruction$\}$Paraphrased sentence:\textbackslash n\textbackslash n''
    \item ``Paraphrase this instruction into a longer sentence\textbackslash n\textbackslash n\textbackslash n$\{$instruction$\}$New sentence:\textbackslash n''
    \item ``You are given an instruction:\textbackslash n\textbackslash n$\{$instruction$\}$Now, paraphrase it into a new instruction with equivalent meaning:\textbackslash n\textbackslash n''
\end{itemize}

\paragraph{Flan} We first reproduced the held-in instruction-tuning set of Flan-T5 with the pipeline\footnote{https://github.com/google-research/FLAN}. We randomly sampled 986 data samples from the generated data following the proportion of partition B reported in \cite{chung2022scaling}. We generate the paraphrases of the selected data with GPT-4 using the following prompts.
\begin{itemize}
    \item ``Here's an input utterance:\textbackslash n\textbackslash n\{instruction\}\textbackslash n \textbackslash n \textbackslash n Now, your task is to paraphrase the input by only changing the instruction but leaving everything else the same.\textbackslash n Here's the new utterance:\textbackslash n\textbackslash n''
    \item ``You are given an utterance which is a combination of task instruction and the actual input. Your job is to paraphrase the task instruction and leave the input unchanged. Here's the utterance to be paraphrased:\textbackslash n\textbackslash n\textbackslash n\{instruction\}\textbackslash n\textbackslash n\textbackslash n Now, generate the new utterance:\textbackslash n\textbackslash n\textbackslash n''
    \item ``You are provided with the utterance of a specific task and I need you to paraphrase it. The actual input, question, and examples in the task should not be changed. You should only paraphrase the instructions. Task:\textbackslash n\textbackslash n\textbackslash n \{instruction\}\textbackslash n\textbackslash n\textbackslash nThe paraphrased utterance:\textbackslash n\textbackslash n\textbackslash n''
\end{itemize}

\newpage
\section{Procedures and Surveys}

%\appendix
%\input{sections/A_appendix.tex}

\end{document}